\def\isarxiv{1} 

\ifdefined\isarxiv
\documentclass[11pt]{article}

\usepackage[numbers]{natbib}

\usepackage{booktabs} 

\else
\documentclass[twoside]{article}

\usepackage{booktabs} 

\usepackage[accepted]{aistats2026}
\usepackage[round]{natbib}

\fi

\usepackage{amsmath}
\usepackage{amsthm}
\usepackage{amssymb}
\usepackage{algorithm}
\usepackage{subfig}
\usepackage{algpseudocode}
\usepackage{graphicx}
\usepackage{grffile}
\usepackage{wrapfig,epsfig}
\usepackage{url}
\usepackage{xcolor}
\usepackage{epstopdf}
\usepackage{tikz}

\usepackage{bbm}
\usepackage{dsfont}

\allowdisplaybreaks

\ifdefined\isarxiv

\usepackage{tikz}
\usepackage{hyperref}  
\hypersetup{colorlinks=true,citecolor=blue,linkcolor=blue} 
\usetikzlibrary{arrows}
\usepackage[margin=1in]{geometry}

\else

\usepackage{microtype}
\usepackage{hyperref}
\definecolor{mydarkblue}{rgb}{0,0.08,0.45}
\hypersetup{colorlinks=true, citecolor=mydarkblue,linkcolor=mydarkblue}

\fi
\graphicspath{{./figs/}}
\graphicspath{{./Figures/}}

\usepackage[most]{tcolorbox}

\tcbset{
  aibox/.style={
    width=474.18663pt,
    top=10pt,
    colback=white,
    colframe=black,
    colbacktitle=black,
    enhanced,
    center,
    attach boxed title to top left={yshift=-0.1in,xshift=0.15in},
    boxed title style={boxrule=0pt,colframe=white,},
  }
}
\newtcolorbox{AIbox}[2][]{aibox,title=#2,#1}

\theoremstyle{plain}
\newtheorem{theorem}{Theorem}[section]
\newtheorem{lemma}[theorem]{Lemma}
\newtheorem{definition}[theorem]{Definition}

\newtheorem{fact}[theorem]{Fact}
\newtheorem{remark}[theorem]{Remark}

\newtheorem{example}[theorem]{Example}

\newcommand{\wh}{\widehat}
\newcommand{\wt}{\widetilde}
\newcommand{\ov}{\overline}

\newcommand{\R}{\mathbb{R}}

\renewcommand{\d}{\mathrm{d}}

\renewcommand{\tilde}{\wt}

\newcommand{\Tmat}{{\cal T}_{\mathrm{mat}}}
\newcommand{\paren}[1]{\left(#1\right)}

\DeclareMathOperator*{\E}{{\mathbb{E}}}

\DeclareMathOperator*{\Z}{\mathbb{Z}}

\DeclareMathOperator{\supp}{supp}
\DeclareMathOperator{\poly}{poly}

\DeclareMathOperator{\diag}{diag}

\makeatletter
\newcommand*{\RN}[1]{\expandafter\@slowromancap\romannumeral #1@}
\makeatother

\newcommand{\ourmethod}[1]{{\color{blue}{ourmethod}}}

\usepackage{lineno}

\begin{document}

\ifdefined\isarxiv

\date{}

\title{Support Basis: Fast Attention Beyond Bounded Entries} 
\author{
Maryam Aliakbarpour\thanks{\texttt{maryama@rice.edu}. Rice University.}
\and
Vladimir Braverman\thanks{\texttt{vova@cs.jhu.edu}. Johns Hopkins University.}
\and
Junze Yin\thanks{\texttt{jy158@rice.edu}. Rice University.}
\and
Haochen Zhang\thanks{\texttt{hz112@rice.edu}. Rice University.}
}

\else

%

%

\twocolumn[

\aistatstitle{Support Basis: Fast Attention Beyond Bounded Entries}

\aistatsauthor{ Maryam Aliakbarpour \And Vladimir Braverman \And Junze Yin \And Haochen Zhang}

\aistatsaddress{ Rice University \And Johns Hopkins University
 \And  Rice University \And Rice University } ]

\fi

\ifdefined\isarxiv
\begin{titlepage}
  \maketitle
  \begin{abstract}

Large language models (LLMs) have demonstrated remarkable performance across a wide range of tasks. However, the quadratic complexity of softmax attention remains a central bottleneck that limits their scalability. Alman and Song (NeurIPS 2023a; NeurIPS 2024a) proposed sub-quadratic time algorithms for attention inference and training, respectively, but they rely on the restrictive \emph{bounded-entry assumption}. We show that this assumption rarely holds in practice, which significantly limits their applicability to modern LLMs.

In this paper, we introduce \emph{support-basis decomposition}, a new technique for accurate and efficient attention inference and training \emph{without} the bounded-entry assumption. We empirically show that the entries of the query and key matrices exhibit sub-Gaussian behavior. Leveraging this widely observed property, we perform exact computation on sparse components and polynomial approximation on dense components. Without relying on restrictive assumptions, we theoretically show that our algorithm achieves sub-quadratic runtime while matching the approximation error of prior work, and we empirically validate its computational efficiency and downstream task performance\footnote{Our code can be found at: \url{https://github.com/yinj66/support_basis}.}. We further generalize our method to a multi-threshold setting that eliminates all distributional assumptions, providing the first theoretical justification for the empirical success of polynomial attention. Moreover, we show that softmax attention can be closely approximated by multiple polynomial attentions with significantly smaller $\ell_p$ error.

  \end{abstract}
  \thispagestyle{empty}
\end{titlepage}

{\hypersetup{linkcolor=black}
\tableofcontents
}
\newpage

\else

\begin{abstract}

Large language models (LLMs) have demonstrated remarkable performance across a wide range of tasks. However, the quadratic complexity of softmax attention remains a central bottleneck that limits their scalability. Alman and Song (NeurIPS 2023a; NeurIPS 2024a) proposed sub-quadratic time algorithms for attention inference and training, respectively, but they rely on the restrictive \emph{bounded-entry assumption}. We show that this assumption rarely holds in practice, which significantly limits their applicability to modern LLMs.

In this paper, we introduce \emph{support-basis decomposition}, a new technique for accurate and efficient attention inference and training \emph{without} the bounded-entry assumption. We empirically show that the entries of the query and key matrices exhibit sub-Gaussian behavior. Leveraging this widely observed property, we perform exact computation on sparse components and polynomial approximation on dense components. Without relying on restrictive assumptions, we theoretically show that our algorithm achieves sub-quadratic runtime while matching the approximation error of prior work, and we empirically validate its computational efficiency and downstream task performance\footnote{Our code can be found at: \url{https://github.com/yinj66/support_basis}.}. We further generalize our method to a multi-threshold setting that eliminates all distributional assumptions, providing the first theoretical justification for the empirical success of polynomial attention. Moreover, we show that softmax attention can be closely approximated by multiple polynomial attentions with significantly smaller $\ell_p$ error.
\end{abstract}

\fi

\ifdefined\isarxiv

\section{Introduction}

\else
\section{INTRODUCTION}

\fi

Large language models (LLMs) are powerful AI models for understanding and generating human-like text. Prominent examples include BERT \citep{dclt18}, OPT \citep{zrg+22}, GPT series \citep{bmr+20,rns+18,rwc+19,o23}, and Llama series \citep{tli+23,tms+23,dja+24}. These models are built upon the Transformer architecture \citep{vsp+17}, which utilizes the \emph{softmax attention computation} as a core mechanism. Using this design, LLMs can handle a wide range of language tasks, like language modeling \citep{mms+19}, sentiment analysis \citep{uas20}, creative writing \citep{cha22,o23}, and natural language translation \citep{hwl21}.

The exact softmax attention computation enables models to emphasize the most influential parts of the input when generating outputs, rather than weighting all words uniformly. The attention weights, computed via a softmax function over the inputs, determine the relative importance assigned to each word. By selectively focusing on highly relevant inputs, attention layers allow models to handle longer sequences more effectively.

However, a crucial drawback of softmax attention is its quadratic time complexity in the input length $n$. Given query, key, and value matrices $Q, K, V \in \R^{n \times d}$ and entrywise exponential function $\exp$, where $d \ll n$ is the hidden dimension, computing the $\paren{Q, K}$-softmax attention matrix $A = \exp\paren{QK^\top / d} \in \R^{n \times n}$ and the subsequent product $AV$ are both computationally expensive, requiring $O\paren{n^2 d}$ time\footnote{In the standard attention literature \citep{vsp+17}, $A := \exp\paren{QK^\top / \sqrt{d}}$. We use $\exp\paren{QK^\top / d}$ solely for the simplicity in our theoretical analysis. We use the standard definition $\exp\paren{QK^\top / \sqrt{d}}$ when running our experiments.}. This quadratic complexity makes it challenging for LLMs to process long input sequences efficiently. Since $A$ contains $n^2$ entries, it is impossible to improve the asymptotic time complexity when explicitly computing all entries of $A$. Thus, prior works on attention optimization that aim to improve quadratic complexity attempt to implicitly utilize $A$ \citep{as23,bpc20,kkl20,zgd+20,kvpf20,kmz24,wlk+20,gswy23} to approximate it. 
Among these, \cite{as23} 
is the only work that provides a time-optimal algorithm for approximating softmax attention without reducing it to a simpler problem. Nevertheless, their algorithm relies on strong assumptions, which makes it very difficult to implement in modern transformers.  
To address this issue, we design an efficient algorithm for approximating attention via a novel mathematical technique, support-basis decomposition, eliminating the strong assumptions in \cite{as23}. Additionally, we combine the polynomial method with sketching techniques to theoretically justify the competitive performance of polynomial attention \citep{kmz24}, where the entrywise exponential is replaced with an entrywise polynomial $p(x) = x^\beta$ for $\beta \geq 2$.

\paragraph{Organization} In Section~\ref{sub:intro:prob}, we present the problem formulation and limitations of prior work. In Section~\ref{sub:intro:our}, we address these challenges and go beyond.

\subsection{Problem Setup and Prior Approaches}
\label{sub:intro:prob}

To justify the quadratic time complexity, we present the mathematical formulation of the exact attention computation. The $(Q, K)$-softmax-attention matrix $A = \exp\paren{QK^\top / d} \in \R^{n \times n}$ captures relationships between each row of the query and the key by the inner product $A_{i, j} := \exp\paren{\frac{1}{d}\left \langle Q_{i, *}, K_{j, *}\right \rangle}$, where $i, j \in [n] := \{1, 2, \dots, n\}$, and $Q_{i, *}, K_{j, *} \in \R^{d}$ are the $i$-th and $j$-th rows of $Q$ and $K$, respectively. The inner product $\left \langle Q_{i, *}, K_{j, *}\right \rangle = \|Q_{i, *}\|_2  \|K_{j, *}\|_2 \cos\theta$, computed in $O\paren{d}$ time, quantifies the alignment of $Q_{i, *}$ and $K_{j, *}$, while the entrywise exponential function further accentuates these relationships by magnifying large similarities. 

We further make each row of $A$ form a probability distribution by dividing each entry by the corresponding row sum. Finally, contextual information is aggregated by multiplying the normalized attention matrix with the value matrix $V \in \R^{n \times d}$. The exact softmax attention computation problem is defined as follows:
\begin{definition}[The exact attention computation]\label{def:exact_attention_computation} 
    Given $Q, K, V \in \R^{n \times d}$, we let $A = \exp\paren{QK^\top / d} \in \R^{n \times n}$ be the $\paren{Q, K}$-softmax-attention matrix. For all $i, j \in [n]$, we let $\diag : \R^n \to \R^{n \times n}$ as $\diag\paren{x}_{i, j} := x_i$ if $i = j$, and $\diag\paren{x}_{i, j} := 0$ otherwise.
    Let $D := \diag\paren{A \mathbf{1}_n} \in \R^{n \times n}$ and $\mathbf{1}_n \in \R^n$ be the all-ones vector. The exact attention computation $\mathrm{Att}\paren{Q, K, V } \in \R^{n \times d}$ is defined as: 
    \begin{align*}
        \mathrm{Att}\paren{Q, K, V } := D^{-1} A V.
    \end{align*}
\end{definition}

To reduce the quadratic time complexity $O\paren{n^2 d}$, \cite{as23} studies approximate attention computation to avoid fully constructing $A \in \R^{n \times n}$:
\begin{definition}[The approximate attention computation]\label{def:approximate_attention_computation}
Let $\epsilon > 0$ be the accuracy parameter. For all matrix $M$, we let $\left \| M \right \|_{\infty}:=\max_{i,j} | M_{i,j} |$. Given $Q, K, V \in \R^{n \times d}$, the goal of the approximate attention computation is to output a matrix $P \in \R^{n \times d}$ with: 
\begin{align*}
    \left \| P - D^{-1} A V \right \|_{\infty} \leq \epsilon \cdot \|V\|_\infty.
\end{align*}
\end{definition}

Under the bounded entry assumption, \cite{as23} presents a time-optimal algorithm that approximates attention computation with small error:
\begin{itemize}
\item {\bf Upper bound of \cite{as23}}: if the bounded entry assumption holds, that is with $B = o\paren{\sqrt{\log n}}$, $\left \| Q \right \|_{\infty}, \left \| K \right \|_{\infty}, \left \| V \right \|_{\infty} \leq B$, then there exists a sub-quadratic time algorithm (Algorithm~\ref{alg:main_as23}) that solves the approximate attention computation problem with $\epsilon = 1 / \poly\paren{n}$;
\item {\bf Lower bound of \cite{as23}}: if the bounded entry assumption fails, there does not exist any sub-quadratic time algorithm approximating attention computation under the Strong Exponential Time Hypothesis (SETH, Definition~\ref{def:seth}) for $\epsilon = 1 / \poly\paren{n}$.
\end{itemize}
    Building upon \cite{as23}, \cite{as24b} further shows that the gradient of the attention loss $\min_{X \in \R^{d \times d}} L(X)$:
    \begin{align}\label{eq:loss}
        \min_{X \in \R^{d \times d}} \| D(X)^{-1} \exp(X_\ell X X_\ell^\top) X_\ell W_V - E \|_F^2
    \end{align}
    can be approximated in sub-quadratic time with the bounded entry assumption. $X_{\ell} \in \R^{n \times d}$ denotes the $\ell$-th layer input, $D(X) = \diag\paren{\exp(X_{\ell} X X_{\ell}^\top / d)\mathbf{1}_n}$, $X = W_Q W_K^\top$ is the variable, $E \in \R^{n \times d}$ denotes the desired output, and $W_Q, W_K, W_V \in \R^{d \times d}$ are the weights for query, key, and value, respectively. This is equivalent to the attention definition (Definition~\ref{def:exact_attention_computation}), with the weights written explicitly, since attention backpropagation requires gradients with respect to the weights. When the bounded-entry assumption is violated, \cite{as24b} shows that no sub-quadratic-time algorithm exists for approximating the gradient of the attention loss to accuracy $\epsilon = 1 / \poly(n)$.

\ifdefined\isarxiv

\else

\begin{figure}[!ht]
    \centering
    \includegraphics[width=\linewidth]{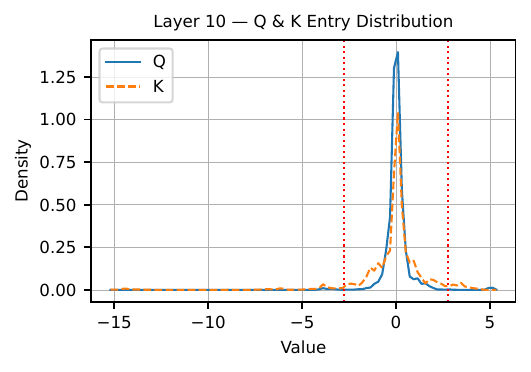}
    \caption{Distribution of entries in $Q$ and $K$ of Layer~10 in TinyLlama-1.1B \citep{zzwl24}.
    The red dashed lines mark the thresholds $\pm\sqrt{\log n}$.}
    \label{fig:entry_distribution}
\end{figure}

\fi

To obtain an accurate and efficient algorithm for approximating attention computation and its gradient (for attention inference and training, respectively), all entries in the matrices $Q, K, V$ must be sufficiently small, i.e., $\max \left \{\left \| Q \right \|_{\infty}, \left \| K \right \|_{\infty}, \left \| V \right \|_{\infty}\right \} \leq o\paren{\sqrt{\log n}}$. However, this bounded-entry assumption rarely holds in modern Transformer architectures (see Figure~\ref{fig:entry_distribution}). Therefore, it is natural to ask:
\begin{center}
    {\it Can we accurately and efficiently approximate the attention computation with a practical assumption?}
\end{center}

\subsection{Our Results}
\label{sub:intro:our}

\ifdefined\isarxiv

\begin{wrapfigure}{r}{0.55\textwidth}
    \centering
    \includegraphics[width=\linewidth]{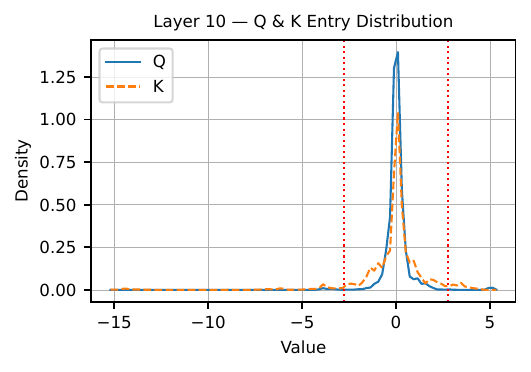}
    \caption{Distribution of entries in the query and key matrices of Layer~10 in TinyLlama-1.1B \citep{zzwl24}.
    The red dashed lines mark the thresholds $\pm\sqrt{\log n}$.}
    \label{fig:entry_distribution}
\end{wrapfigure}

\else

\fi

\paragraph{Contribution 1---An Accurate, Efficient, and Implementable Algorithm} We provide a positive answer to this question. Our “practical assumption” stems from the fact that the entries of the query, key, and value matrices resemble a sub-Gaussian distribution (Figure~\ref{fig:entry_distribution}). These entries cluster near the mean, and the number of extreme entries—those whose absolute values exceed $o\paren{\sqrt{\log n}}$—is small (see figures in Appendix~\ref{sec:distribution} for more details beyond Figure~\ref{fig:entry_distribution}: we provide empirical evidence of sub-Gaussian-like behavior in widely used modern Transformer architectures, such as TinyLlama-1.1B \citep{zzwl24}, LLaDA-8B-Base \citep{nie2025large}, OPT-1.3B \citep{zrg+22}, and Phi-2 \citep{jba23}, across \emph{multiple} layers). 

Under this practical assumption, the number of large entries is small, so we decompose the query $Q$ into the sum of a sparse matrix, $Q^{(L)}$, whose rows contain large entries, and a dense matrix, $Q^{(s)}$, whose rows contain only small entries. We apply the same decomposition to the key $K$. We then perform the exact computation for the sparse components and use the polynomial method of \cite{as23} to approximate the small and dense component. We call this \emph{single-threshold support basis decomposition}, where “single-threshold” denotes $T = o\paren{\sqrt{\log n}}$ and “support basis” (Definition~\ref{def:support_basis}) refers to the disjoint matrices $Q^{\paren{s}} \paren{K^{\paren{s}}}^\top$ and $Q^{\paren{L}} \paren{K^{\paren{L}}}^\top + Q^{\paren{s}} \paren{K^{\paren{L}}}^\top + Q^{\paren{L}} \paren{K^{\paren{s}}}^\top$.
\begin{theorem}[Informal version of Theorem~\ref{thm:subgaussian-main}]
\label{thm:subgaussian-main-informal}
Let \( Q, K, V \in \mathbb{R}^{n \times d} \) be the query, key, and value matrices. Suppose entries of $Q, K$ are independent and sub-Gaussian with variance proxies \( \sigma_Q^2 \) and \( \sigma_K^2 \), respectively. 
Let $\epsilon, \delta \in (0, 0.1)$ respectively be the accuracy parameter and failure probability. Then, with probability at least $1 - \delta$, there exists a sub-quadratic time algorithm (Algorithm~\ref{alg:approx_attention}) that outputs $P \in \mathbb{R}^{n \times d}$ satisfying
\begin{align*}
    \left\| P - D^{-1} A V \right\|_\infty \leq \epsilon \cdot \|V\|_\infty,
\end{align*}
where $A = \exp(QK^\top / d)$ and $D = \mathrm{diag}(A \cdot \mathbf{1}_n)$. 
\end{theorem}
Using our single-threshold support basis, we can further approximate the gradient of the attention loss in sub-quadratic time if the entries of $Q$ and $K$ are sub-Gaussian. Unlike \cite{as24b}, our result enable efficient backward propagation without the bounded entry assumption (see Section~\ref{sub:technique:single} for detail).

Our empirical results on the distributions of the $Q$ and $K$ entries across multiple transformer models in multiple layers (e.g., Figure~\ref{fig:entry_distribution}) imply that our sub-quadratic-time randomized algorithm can be applied to all of these models throughout the entire network. Since we show that both the forward and backward propagation of attention can be approximated in sub-quadratic time whenever sub-Gaussianity holds, we naturally extend our efficient algorithm to \emph{multi-layer attention}, aligning it more closely with modern Transformer architectures.

However, it is also natural to ask:
\begin{center}
    {\it Can we generalize our support basis to any arbitrary $Q, K \in \R^{n \times d}$ without any assumption?}
\end{center}

\paragraph{Contribution 2---Efficient Attention Approximation With No Assumption} We provide a positive answer to this question. Without any assumptions, entries of $Q$ and $K$ may not cluster near the mean, so we cannot infer that the number of large entries is small. Thus, we develop \emph{multiple-threshold support basis} to approximate the attention computation in sub-quadratic time. However, due to the \textbf{lower bound} of \cite{as23}, this unavoidably leads to a weaker accuracy guarantee $\epsilon > 1/\poly(n)$ in $\ell_\infty$ norm.
\begin{theorem}[Informal version of Theorem~\ref{thm:attention_approximation_bucketing}]\label{thm:attention_approximation_bucketing_informal}
    Let \( Q, K, V \in \mathbb{R}^{n \times d} \) be the query, key, and value matrices. Let $\epsilon, \delta \in (0, 0.1)$ respectively be the accuracy parameter and failure probability. Let $B = \max \{\|Q\|_\infty, \|K\|_\infty\}$ and $b = \min_{i \in [n], j \in [d]} \{|Q_{i, j}|, |K_{i, j}|\}$.
    Then, with probability $1 - \delta$, the approximate attention computation (Definition~\ref{def:approximate_attention_computation}) can be solved in sub-quadratic time by outputting $P \in \R^{n \times d}$ that satisfies
    \begin{align*}
        \left \|P - D^{-1} A V\right \|_\infty \lesssim \epsilon \exp(3B^2) \cdot \left \| V \right \|_\infty.
    \end{align*}
\end{theorem}
Beyond the contribution in attention approximation, our result also provides a theoretical justification for why polynomial attention performs well in practice. While \cite{kmz24} proposes a fast algorithm to approximate a polynomial attention and empirically demonstrates its effectiveness on downstream tasks, a theoretical explanation for this success remains lacking.
\begin{center}
    {\it Why does the polynomial attention have comparable performance with the softmax attention?}
\end{center}

\paragraph{Contribution 3---A Theoretical Justification for the Empirical Performance of Polynomial Attention and Beyond} 

In our analysis of Theorem~\ref{thm:attention_approximation_bucketing_informal}, we bridge this gap by showing that softmax attention is $\epsilon_1$-close to a polynomial attention, and this polynomial attention can be approximated via the sketching methods of \cite{kmz24} with accuracy $\epsilon_2$. It implies the resulting approximation is $(\epsilon_1 + \epsilon_2)$-close to softmax attention, which explains why \cite{kmz24} obtains experimental results for polynomial attention that are comparable to those of softmax attention.

Moreover, we note that the attention approximation method of \cite{as23} does not perform well when the entries of $Q$ and $K$ are far apart. This motivates us to design a \emph{multiple-threshold support basis}, which partitions $Q$ and $K$ into several disjoint components. For the partitions with larger entries, instead of performing exact computation as in the \emph{single-threshold support basis}, whose running time becomes large without the sub-Gaussian assumption, we approximate them by a sum of polynomial attentions. 
It supports the empirical results of \cite{kmz24} and goes further: we can significantly reduce the $\ell_p$ error when approximating softmax attention using multiple polynomial attentions compared to a single polynomial attention, although the $\ell_\infty$ error {\bf lower bound} of \cite{as23} remains impossible to bypass.
Thus, combining multiple polynomial attentions better captures the behavior of softmax attention than a single polynomial attention in \cite{kmz24}.

\paragraph{Contribution 4---Experimental Results} Besides empirically validating our assumption that the entries of query and key matrices resemble sub-Gaussian distributions, we also present runtime experiments showing that our method is faster than exact attention computation and achieves significantly lower error than \cite{as23}. Additionally, we evaluate downstream tasks with both \cite{as23} and our method across various benchmarks. Since the assumptions of \cite{as23} rarely hold in practice, it yields 0\% accuracy, whereas our method aligns with practical settings and achieves performance comparable to exact attention (see Section~\ref{sec:exp}).

\paragraph{Notation.} Let $\Z_+$ be the set of positive integers. For all sets $Y$, we use $|Y|$ to denote its cardinality. We define the combination $\binom{n}{r} := \frac{n!}{(n - r)! r!}$.
For all $i \in [d]$ and $x, y \in \R^d$, we define $x \circ y \in \R^d$ as $(x \circ y)_i := x_i \cdot y_i$; for all $A, B \in \R^{n \times d}$, for all $i \in [n]$ and $j \in [d]$, $A \circ B \in \R^{n \times d}$ is defined as $(A \circ B)_{i, j} := A_{i, j} \cdot B_{i, j}$. 
For all $p \in \Z_+$, we define the $\ell_p$ norm of $x$ as $\|x\|_p : = \paren{\sum_{i \in [d]} |x_i|^p}^{1/p}$ and $A^{\circ p} \in \R^{n \times d}$ as $(A^{\circ p})_{i, j} := A_{i, j}^p$.
We define $\supp(A) := \{(i, j) \in [n] \times [d] \mid A_{i, j} \neq 0\}$.
${\bf 1}_{n \times d}$ is the $n \times d$ matrix whose entries are all $1$'s.
$\Pr[\cdot]$ denotes the probability. The expectation of the discrete random variable $X \in \R$ is $\E[X] := \sum_i x_i \cdot \Pr[X = x_i]$.

\paragraph{Roadmap.} In Section~\ref{sec:related_work}, we present the related work. In Section~\ref{sec:technique}, we give an overview of techniques we use to prove our main results (Theorem~\ref{thm:subgaussian-main-informal} and Theorem~\ref{thm:attention_approximation_bucketing_informal}). In Section~\ref{sec:exp}, we present our experimental results. 

\ifdefined\isarxiv

\section{Related Work}

\else
\section{RELATED WORK}

\fi

\label{sec:related_work}

\paragraph{Polynomial Methods for Attention Approximation} 

As the attention approximation algorithm of \cite{as23} is time optimal, it has inspired numerous follow-up works \citep{as24a, as24b, as25, chl+24, lssz24}. They generalize the polynomial method from \cite{as23} to establish upper and lower bounds for various attention-approximation–related problems. \cite{as24a} extends this method to study tensor attention inference, showing that the $p$-th order tensor attention can only be approximated in almost linear time under an even stronger assumption: $\max \left \{\left \| Q \right \|_{\infty}, \left \| K \right \|_{\infty}, \left \| V \right \|_{\infty}\right \} \leq o\paren{ \sqrt[p]{\log n}}$. Similarly, \cite{as24b} proves that the gradient of attention can be approximated in almost linear time, and \cite{lssz24} shows the same for the gradient of tensor attention. Moreover, \cite{as25, chl+24} apply the method of \cite{as23} to Rotary
Position Embedding (RoPE) 
attention (proposed by \cite{sal+24}) inference and training, respectively. All of these works rely on the strict bounded-entry assumption and contain no experimental results. Although our method is specifically built upon \cite{as23}, it can be generalized to all these works to eliminate the bounded-entry assumption, since they all rely on similar techniques.

A recent work by~\cite{ghs+25} develops upon \cite{as23} by finding the relationships between the entry bound $B > 0$ and different hidden dimensions $d$. They show that when $d = O(1)$ and $B = \poly(n)$, the approximate attention computation problem admits a truly subquadratic-time algorithm with runtime $\widetilde{O}(n^{2 - 1/d} \cdot \poly\log(B/\varepsilon))$. In contrast, when $d = 2^{\Theta(\log^* n)}$, they establish a conditional lower bound under the SETH, proving that any algorithm computing the approximate attention computation problem must take $n^{2 - o(1)}$ time. Finally, in the regime where $d = \poly(n)$, the standard algorithm with runtime $O(n^2 d)$ is conditionally optimal. Although \cite{ghs+25} gives a more flexible choice of $d$, it is still unclear how we can design a fast algorithm approximating the attention computation problem if there are entries in $Q, K, V$ larger than $B$.

\paragraph{Other Attention Optimization Works} 

Other attention approximation works also rely on different types of assumptions that are incomparable to our work. Reformer \citep{kkl20} introduces Locality-Sensitive Hashing (LSH) Attention, which reduces the computational complexity from $O\paren{n^2}$ to $O\paren{ n \log n}$ by assuming that, for each query vector, only a small subset of the nearest key vectors substantially contributes to the softmax output. Longformer \citep{bpc20} and Big Bird \citep{zgd+20} build on the sparse attention matrix assumption. 
Linear Transformer \citep{kvpf20} replaces the softmax operation with a kernel-based formulation, and \cite{kmz24} replaces the entrywise exponential with an entrywise polynomial. Linformer \citep{wlk+20} assumes the self-attention matrix is low-rank. 

The most similar work to our \emph{multiple-threshold support basis} result is \cite{lls+24}, which is also motivated by alleviating strict assumptions in prior works and develops a sub-quadratic time algorithm for attention approximation with a causal mask. However, their approach is limited to the case where the causal mask is a lower-triangular matrix in which all lower-triangular entries are equal to 1. Consequently, when applied with this mask, the attention matrix also becomes lower triangular. \cite{lls+24} decomposes the masked attention into a set of $k$ “conv bases” with $k \leq n$ and analyzes each basis in $O(n \log n)$ time via Fast Fourier Transform to achieve sub-quadratic runtime. While this is similar in spirit to our approach, our task is significantly more challenging: we decompose a dense $n \times n$ attention matrix with no zero entries into a sum of disjoint components and analyze each using the polynomial method and sketching. Both works share the advantage of tuning the number of bases, but our method is more promising because our bucketing scheme generates at most a logarithmic number of bases. In contrast, the number of bases in \cite{lls+24} can be as large as $n$ in the worst case, which yields no improvement in runtime.

\ifdefined\isarxiv

\section{Technique Overview}

\else
\section{TECHNIQUE OVERVIEW}

\fi

\label{sec:technique}
Our theoretical proofs are deferred to the appendix. Below, we first discuss the origin and limitations of the bounded-entry assumption in attention optimization (Section~\ref{sub:technique:limitation}). We then summarize the techniques used in our proofs to remove this bounded-entry assumption and support our main results. In Section~\ref{sub:technique:single}, we show that a single-threshold support basis suffices to eliminate this assumption, proving Theorem~\ref{thm:subgaussian-main-informal}. In Section~\ref{sub:technique:multiple}, we introduce multiple-threshold support bases to prove Theorem~\ref{thm:attention_approximation_bucketing_informal}.

\subsection{Bounded Entry Assumption Limitations}
\label{sub:technique:limitation}

In Lemma 3.4 of \cite{as23} (Lemma~\ref{lem:wt_A_small_rank}), with
$g = O\left(\max\left\{ \frac{\log(1/\epsilon)}{\log(\log(1/\epsilon)/B)}, B^2 \right\}\right)$ being the degree of the Chebyshev polynomial, if the bounded entry assumption hold ($B = o\paren{\sqrt{\log n}}$), then
there exists an algorithm that returns $U_1, U_2 \in \mathbb{R}^{n \times r}$ such that the attention matrix can be low-rank approximated $A \approx U_1 U_2^\top$, with the rank
$r = \binom{2(d+g)}{2g}$.
The construction of the low-rank approximation is by using a degree-$g$ entrywise Chebyshev polynomial $U_1 U_2^\top = p\paren{QK^\top / d}$ (see details in Appendix~\ref{sub:preli:poly}) to approximate the attention matrix $\exp\paren{QK^\top / d}$ so that computing $U_1 \paren{U_2^\top V}$ ($O\paren{nrd}$ time) takes less time than $AV$ ($O\paren{n^2 d}$ time) if $r < n$.
If the bounded entry assumption is not satisfied, i.e. $B \ge \Omega(\sqrt{\log n})$, then $g \ge c_1 \log n$ for some $c_1 > 1$, regardless of $\epsilon$.
Therefore, with $d = c_2 \log n$ for some $c_2 > 1$, we get
\[
\binom{2(d+g)}{2g}
\ge
\left(\frac{d + g}{d}\right)^{2d}
\ge
2^{\Omega(d)}
\ge
2^{\Omega(\log n)}
\ge
n^{\Omega(1)}.
\]
It implies that the attention approximation using $U_1, U_2 \in \mathbb{R}^{n \times n^{\Omega(1)}}$
requires more than $n^2$ time, regardless of the error $\epsilon$.
This is a significant drawback: if $B \ge \Omega(\sqrt{\log n})$, then the Chebyshev polynomial
cannot yield a sub-quadratic time algorithm.
If we force $g < o\paren{\log n}$ to ensure $r < n$, the error guarantee breaks.
Thus, directly using the sub-quadratic algorithm from~\cite{as23} when
$B \ge \Omega(\sqrt{\log n})$ may fail to provide any meaningful accuracy guarantee.

\subsection{Single-Threshold Support Basis}
\label{sub:technique:single}

To eliminate the bounded entry assumption, the most straightforward approach is to scan all entries of $Q, K \in \R^{n \times d}$ and move those larger than the threshold $T = o\paren{\sqrt{\log n}}$ into $Q^{\paren{L}}$ and $K^{\paren{L}}$, respectively. The matrices containing the remaining smaller entries are denoted $Q^{\paren{s}}$ and $K^{\paren{s}}$. We denote this process as $\textsc{Split}(Q, T)$ and $\textsc{Split}(K, T)$. We perform exact computation for the terms related to $Q^{\paren{L}}$ and $K^{\paren{L}}$, and use the polynomial method \citep{as23} to approximate $Q^{\paren{s}} \paren{K^{\paren{s}}}^\top$. However, we notice that since $\exp$ is applied entry-wisely, $\paren{\exp\paren{A + B}} \cdot V = \paren{\exp\paren{A} \circ \exp\paren{B}} \cdot V$, which cannot be further simplified as multiplication $\cdot$ is not distributive over Hadamard product $\circ$.
Therefore, we cannot divide and conquer it using this naive approach.

\paragraph{Divide: Design a Valid Support Basis}
Our technique is to design a novel decomposition method that breaks down $QK^\top / d$ into a sum of \emph{disjoint matrices}---For a set of matrices $\{A_i\}_{i \in [m]} \subset \R^{n \times d}$ where $m \in \Z_+$, $A_i$ is said to be a disjoint matrix if for all $\paren{j, k} \in [n] \times [d]$, for all $i' \neq i$, if $\paren{A_i}_{j, k} \neq 0$, then $\paren{A_{i'}}_{j, k} = 0$ (Definition~\ref{def:disjoint}). With two disjoint matrices $A^{\paren{s}}, A^{\paren{L}}$ satisfying $QK^\top = Q^{(L)} K^\top + Q^{(s)} \paren{K^{(L)}}^\top + Q^{(s)} \paren{K^{(s)}}^\top = A^{\paren{s}} + A^{\paren{L}}$, we define $\{A^{\paren{s}}, A^{\paren{L}}\}$ as a \emph{support basis} of $QK^\top$. 
In our work, we construct $A^{\paren{s}}$ as a dense matrix, and all of its entries are bounded by $o\paren{\sqrt{\log n}}$; we construct $A^{\paren{L}}$ as a sparse matrix, and its entries can be large.
Therefore, $A^{\paren{s}}$ is suitable for approximation using the polynomial method, whereas $A^{\paren{L}}$ is suitable to be computed exactly since its sparsity ensures the fast matrix multiplication. By Fact~\ref{fac:exp_split}, this decomposition has a desirable mathematical property, namely $\exp\paren{QK^\top / d}$ equals:
\begin{align}\label{eq:disjoint}
    \exp\paren{A^{\paren{s}} / d} + \exp\paren{A^{\paren{L}} / d} - {\bf 1}_{n \times n}.
\end{align}
This additive form is essential for the divide-and-conquer strategy, as multiplication $\cdot$ is distributive over addition $+$. This can break the original attention computation problem $D^{-1} \exp\paren{QK^\top / d} V$ into a sum of two simpler problems $D^{-1} \exp\paren{A^{\paren{s}} / d} V$ and $D^{-1} \paren{\exp\paren{A^{\paren{L}} / d} - {\bf 1}_{n \times n}} V$. Before presenting how we conquer each of them, we first present how to construct such a valid support basis $\{A^{\paren{s}}, A^{\paren{L}}\}$. All of the entries of $Q^{\paren{L}}K^\top + Q^{\paren{s}}\paren{K^{\paren{L}}}^\top$ are ``potentially large'', so they should be included in $A^{\paren{L}}$. To further ensure that our current $A^{\paren{L}}$ is disjoint from the remaining dense component $Q^{\paren{s}}\paren{K^{\paren{s}}}^\top$, for all $\paren{i, j} \in [n] \times [n]$ satisfying $\paren{Q^{\paren{L}}K^\top + Q^{\paren{s}}\paren{K^{\paren{L}}}^\top}_{i, j} \neq 0$, we extract the corresponding entry $\paren{Q^{\paren{s}}\paren{K^{\paren{s}}}^\top}_{i, j}$ and add it into $\paren{A^{\paren{L}}}_{i, j}$. Regarding $A^{\paren{s}}$, we define $\paren{A^{\paren{s}}}_{i, j}$ to be $\paren{Q^{\paren{s}}\paren{K^{\paren{s}}}^\top}_{i, j}$ if $\paren{A^{\paren{L}}}_{i, j} = 0$ and $0$ otherwise. Then, this forms a valid support basis. 

\ifdefined\isarxiv

\else

\begin{figure}[!ht]
    \centering
    \includegraphics[width=\linewidth]{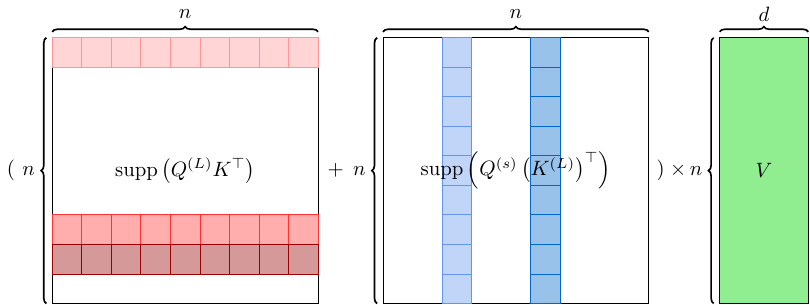}
    \caption{A visualization of $A^{\paren{L}} V$. By the definition of $A^{\paren{L}}$, $\supp\paren{A^{\paren{L}}} = \supp\paren{Q^{\paren{L}}K^\top + Q^{\paren{s}}\paren{K^{\paren{L}}}^\top}$. Therefore, visualizing $A^{\paren{L}} V$ is equivalent to visualize $\paren{Q^{\paren{L}}K^\top + Q^{\paren{s}}\paren{K^{\paren{L}}}} V$. By definition, $Q^{\paren{L}}, K^{\paren{L}}$ are sparse matrices with $\left |\supp\paren{Q^{\paren{L}}} \right| = \left |\supp\paren{K^{\paren{L}}} \right| = O\paren{n^\alpha}$ for $\alpha \in \paren{0, 1}$, so there are at most $O\paren{n^\alpha}$ non-zero rows (red blocks) in $Q^{\paren{L}}K^\top$ and $O\paren{n^\alpha}$ non-zero columns (blue blocks) in $Q^{\paren{s}}\paren{K^{\paren{L}}}^\top$.}
    \label{fig:AL}
\end{figure}
\fi

\ifdefined\isarxiv

\begin{figure}[!ht]
    \centering
    \includegraphics[width=\linewidth]{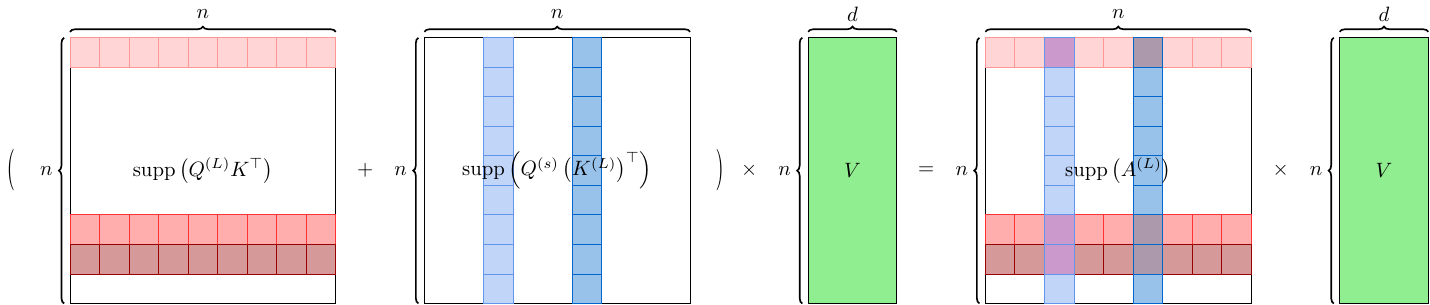}
    \caption{A visualization of $A^{\paren{L}} V$. By the definition of $A^{\paren{L}}$, $\supp\paren{A^{\paren{L}}} = \supp\paren{Q^{\paren{L}}K^\top + Q^{\paren{s}}\paren{K^{\paren{L}}}^\top}$. Therefore, visualizing $A^{\paren{L}} V$ is equivalent to visualize $\paren{Q^{\paren{L}}K^\top + Q^{\paren{s}}\paren{K^{\paren{L}}}} V$. By definition, $Q^{\paren{L}}, K^{\paren{L}}$ are sparse matrices with $\left |\supp\paren{Q^{\paren{L}}} \right| = \left |\supp\paren{K^{\paren{L}}} \right| = O\paren{n^\alpha}$ for $\alpha \in \paren{0, 1}$, so there are at most $O\paren{n^\alpha}$ non-zero rows (red blocks) in $Q^{\paren{L}}K^\top$ and $O\paren{n^\alpha}$ non-zero columns (blue blocks) in $Q^{\paren{s}}\paren{K^{\paren{L}}}^\top$.}
    \label{fig:AL}
\end{figure}

\else

\fi

\paragraph{Conquer: Approximate the Attention in $A^{\paren{L}}$ Sparsity Time} Now, we have shown that we can construct a support basis $\{A^{ \paren{s}}, A^{\paren{L}}\}$ to make attention computation be divided into the sum of two smaller problems: $D^{-1} \exp\paren{A^{\paren{s}} / d} V$ and $D^{-1} \paren{\exp\paren{A^{\paren{L}} / d} - {\bf 1}_{n \times n}} V$. Below, we justify the statement that if the number of large entries (those greater than $o\paren{\sqrt{\log n}}$) of $Q$ and $K$ is upper bounded by $O\paren{n^{\alpha}}$ with $\alpha \in \paren{0, 1}$, then the attention computation can be approximated in sub-quadratic time.

To conquer each of these smaller problems, we efficiently approximate the former using polynomial method from \cite{as23} and exactly compute the latter. To ensure that our attention approximation algorithm runs in sub-quadratic time, it suffices to make the exact computation part sub-quadratic, since the polynomial method always takes $n^{1 + o\paren{1}}$ time. Recall that we define $A^{\paren{L}} \in \R^{n \times n}$ as $\paren{A^{\paren{L}}}_{i, j} := \paren{Q K^\top}_{i, j}$ if 
     $\paren{Q^{\paren{L}}K^\top + Q^{\paren{s}}\paren{K^{\paren{L}}}^\top}_{i,j} \neq 0$
and $0$ otherwise. Since $Q, K \in \R^{n \times d}$, computing an arbitrary entry of $Q K^\top$ only takes $d$ time. Thus, we need to know how many of the entries of $A^{\paren{L}}$ are non-zero. 
As the number of large entries of $Q$ and $K$ is at most $O\paren{n^{\alpha}}$ with $\alpha \in \paren{0, 1}$, we consider $Q^{\paren{s}}$ and $K$ to be dense and $Q^{\paren{L}}, K^{\paren{L}}$ to be sparse. Thus, the support of $A^{\paren{L}}$ is represented as in Figure~\ref{fig:AL}. We get $\mathrm{supp}\paren{\exp\paren{A^{\paren{L}} / d} - {\bf 1}_{n \times n}} = \mathrm{supp}\paren{A^{\paren{L}}}$ as $\exp\paren{0} - 1 = 0$. When computing $\paren{\exp\paren{A^{\paren{L}} / d} - {\bf 1}_{n \times n}} V$, it suffices to consider:

\begin{itemize}
\item {\bf Case 1}: if a matrix $A_1 \in \R^{n \times n}$ is sparse with $n^{\alpha}$ rows (red blocks in Figure~\ref{fig:AL}) that are non-zero and the rest of the rows are all 0, then we only need to compute $n^{\alpha} d$ numbers of inner products when computing $A_1 V \in \mathbb{R}^{n \times d}$. As each inner product takes $O(n)$ time, it takes $O(n^{1 + \alpha} d)$ time in total to compute $A_1 V$.

\item {\bf Case 2}: if a matrix $A_2 \in \R^{n \times n}$ is sparse with $n^{\alpha}$ columns (blue blocks in Figure~\ref{fig:AL}) that are non-zero and the rest of the rows are all 0, computing $A_2 V \in \mathbb{R}^{n \times d}$ requires $nd$ numbers of inner product, but since each inner product takes only $O(n^{\alpha})$ time, then in total, it still takes $O(n^{1 + \alpha} d)$ time.
\end{itemize}
    
Combining both cases together, with $d = O\paren{\log n}$, it takes $O\paren{n^{1 + \alpha}}$ time to compute 
\ifdefined\isarxiv
\begin{align*}
    \paren{\exp\paren{A^{\paren{L}} / d} - {\bf 1}_{n \times n}} V.
\end{align*}
\else
$\paren{\exp\paren{A^{\paren{L}} / d} - {\bf 1}_{n \times n}} V$.
\fi
As $O\paren{n^{1 + \alpha}}$ is the sparsity of $A^{\paren{L}}$ (Lemma~\ref{lem:sparsity_AL}), we call it $A^{\paren{L}}$ sparsity time. 

\begin{algorithm}[!ht]\caption{We approximate the attention computation problem using the support basis $\{A^{\paren{s}}, A^{\paren{L}}\}$.}\label{alg:approx_attention}
\begin{algorithmic}[1]

\Procedure{\textsc{ApproxAttention}}{$Q \in \R^{n \times d}, K \in \R^{n \times d}, V \in \R^{n \times d}, n, d, \epsilon, T$}
    \State $Q^{\paren{L}}, Q^{\paren{s}} \in \R^{n \times d} \gets \textsc{Split}(Q, T)$ 
    \State $K^{\paren{L}}, K^{\paren{s}} \in \R^{n \times d} \gets \textsc{Split}(K, T)$
    \State Explicitly compute $A^{\paren{L}}$. \label{line:atten_al}
    \State $U_1, U_2  \in \R^{n \times r} \gets \textsc{Polynomial}(A^{\paren{s}}, \epsilon)$. \Comment{To approximate $\exp\paren{A^{\paren{s}} / d}$.} \label{line:atten_poly}
    \State $d_1 \gets U_1 \paren{U_2^\top {\bf 1}_n} \in \R^n$. \label{line:atten_d1}
    \State $d_2 \gets \paren{\exp\paren{A^{\paren{L}} / d} - {\bf 1}_{n \times n}} {\bf 1}_n \in \R^n$. \label{line:atten_d2}
    \State $D^{-1} \gets \diag\paren{d_1 + d_2}^{-1} \in \R^{n \times n}$. \label{line:atten_d-1} 
    \State $C_1 \gets U_1 \paren{U_2^\top V} \in \R^{n \times d}$. \label{line:compute_C1_main}
    \State $C_2 \gets \paren{\exp\paren{A^{\paren{L}} / d} - {\bf 1}_{n \times n}} V \in \R^{n \times d}$. \label{line:compute_C2_main}
    \State \Return $D^{-1} (C_1 + C_2)$. \label{line:atten_return}
\EndProcedure
\end{algorithmic}
\end{algorithm}

Constructing $D = \diag\paren{\exp\paren{QK^\top / d} {\bf 1}_n}$ also takes $A^{\paren{L}}$ sparsity time: because of the linearity property of $\diag : \R^n \to \R^{n \times n}$, the time complexity is dominated by computing 
\ifdefined\isarxiv
\begin{align*}
    \paren{\exp\paren{A^{\paren{L}} / d} - {\bf 1}_{n \times n}} {\bf 1}_n,
\end{align*}
\else
$\paren{\exp\paren{A^{\paren{L}} / d} - {\bf 1}_{n \times n}} {\bf 1}_n$,
\fi
which, similarly, takes $O\paren{n^{1 + \alpha}}$ time. Finally, as $D^{-1}$ is diagonal, the attention computation problem can be approximated in $A^{\paren{L}}$ sparsity time.

\paragraph{Sub-Gaussianity of Query and Key} We have now justified the statement that if the number of large entries (those greater than $o\paren{\sqrt{\log n}}$) of $Q$ and $K$ is upper bounded by $O\paren{n^{\alpha}}$ with $\alpha \in \paren{0, 1}$, then the attention computation can be approximated in sub-quadratic time. What remains is to justify why this “if” condition holds. This is where our assumption—that the entries of the query and key matrices follow a sub-Gaussian distribution—becomes crucial. Suppose $ Q, K \in \mathbb{R}^{n \times d} $ are random matrices with independent sub-Gaussian entries having variances $ \sigma_Q^2 $ and $ \sigma_K^2 $, respectively, we have
    $\Pr\left[ \left |Q_{i,j} \right| \geq t \right] \leq 2 \exp\left( -\frac{t^2}{\sigma_Q^2} \right)$ and $\Pr\left[ \left |K_{i,j} \right| \geq t \right] \leq 2 \exp\left( -\frac{t^2}{\sigma_K^2} \right)$.
This implies that the expected number of large entries in $M^{\paren{L}} \in \{Q^{\paren{L}}, K^{\paren{L}}\}$ is bounded as: $\mathbb{E}\left[\left | \mathrm{supp}\paren{M^{\paren{L}}}\right | \right]
\leq 2nd \cdot \exp\left( -\frac{T^2}{\sigma_M^2} \right)$. Applying the multiplicative Chernoff bound, we obtain that with high probability, our “if” condition holds:
\begin{align*}
    \Pr\left[ \left |\mathrm{supp}\paren{M^{\paren{L}}} \right | > n^\alpha \right] \leq \exp\left( -\Omega\paren{n^\alpha} \right).
\end{align*}
Putting everything together, we conclude the proof sketch of Theorem~\ref{thm:subgaussian-main-informal} by Modus Ponens.

\paragraph{Generalize to Multi-Layer Attention}
Generalizing our method to multi-layer attention requires efficient algorithms for both inference and training. Inference can be achieved using Algorithm~\ref{alg:approx_attention}, while training involves efficiently approximating the gradient of the loss function (Eq.~\eqref{eq:loss}). As noted in \cite{as24b}, the dominant term in the time complexity of computing $\frac{\d L}{\d X}$ still arises from the attention matrix $A \in \R^{n \times n}$. Thus, replacing $A$ with the low-rank matrix $U_1 U_2^\top$ like \cite{as23}, \cite{as24b} can approximate $\frac{\d L}{\d X}$ in sub-quadratic time.

On the other hand, our method expresses $A$ as $\exp\paren{A^{\paren{s}} / d} + \exp\paren{A^{\paren{L}} / d} - {\bf 1}_{n \times n}$ (Eq.~\eqref{eq:disjoint}). Since the entries of $\exp\paren{A^{\paren{s}} / d}$ are small and satisfy the bounded entry assumption, we can approximate it by $U_1 U_2^\top$ with little sacrifice in accuracy. However, although $\mathcal{B} = \exp\paren{A^{\paren{L}} / d} - {\bf 1}_{n \times n}$ is sparse, its product with other dense matrices is dense, and thus further computation may still require quadratic time $\Omega\paren{n^2}$.
To address this issue, we use the approximate SVD from \cite{cw17}, allowing us to approximate the best rank-$k$ low-rank approximation of $\mathcal{B}$ in $A^{\paren{L}}$ sparsity time (Theorem~\ref{thm:approximate_svd}). Choosing $k = n^{o(1)}$, we approximate $\frac{\d L}{\d X}$ in $A^{\paren{L}}$ sparsity time, which is also sub-quadratic by sub-Gaussianity, thereby generalizing our method to multi-layer attention.

\subsection{Multiple-Threshold Support Basis}
\label{sub:technique:multiple}

The single-threshold support basis works well when $Q$ and $K$ only have a small number of ``large'' entries. However, if we do \emph{not} make any distributional assumptions, we cannot bound the number of large entries in $Q$ and $K$ with high probability. Thus, the exact computation may become too expensive.

\paragraph{Divide: Design a Multiple-Threshold Support Basis}
To address it, we develop the \emph{multiple-threshold support basis} to decompose $Q$ and $K$ into several disjoint components by a sequence of thresholds $0 < T_1 < T_2 < \cdots < T_m$.
These thresholds partition the range of entry magnitudes into $m$ disjoint intervals. For example, entries within $(T_{\ell - 1}, T_{\ell}]$ are assigned to the $\ell$-th ``bucket''. We write $Q = Q^{(T_1)} + Q^{(T_2)} + \cdots + Q^{(T_m)}$ and $K = K^{(T_1)} + K^{(T_2)} + \cdots + K^{(T_m)}$. To ensure the disjointness of each $A^{\paren{T_\ell, T_{\ell'}}} := Q^{\paren{T_\ell}} \paren{K^{\paren{T_{\ell'}}}}^\top$, if we find a large entry from $Q$ or $K$ in the interval $(T_{\ell - 1}, T_{\ell}]$, we take the entire row containing this entry and include it in $Q^{(T_\ell)}$ or $K^{(T_\ell)}$.
We then expand the product $QK^\top$ as
    $QK^\top = \sum_{\ell=1}^m \sum_{\ell'=1}^m A^{\paren{T_\ell, T_{\ell'}}}$.
Since there are $m^2$ such matrices $A^{(T_\ell, T_{\ell'})}$, we choose $T_\ell$ to grow exponentially with respect to $\ell$ to keep $m^2$ small.  
Thus, we only need to handle a small number of $A^{(T_\ell, T_{\ell'})}$ to achieve the desired running time.
Due to its disjointness, the support basis $\{ A^{\paren{T_l, T_{l'}}}  \}_{T_l, T_{l'} \in  \{T_\ell \}_{\ell = 1}^{m}}$ admits a desirable additive decomposition (by induction) analogous to Eq.~\eqref{eq:disjoint} for the single-threshold support basis.

\paragraph{Conquer: Reduce Gaussian Kernels to Polynomial Kernels and Approximate Them by Sketching}
The largest entries in each of $Q^{(T_\ell)} (K^{(T_{\ell'})})^\top$ is at most $d \cdot T_\ell \cdot T_{\ell'}$. Therefore, to make each of them satisfy the bounded entry assumption, we need to take out a scalar from this matrix $Q^{(T_\ell)} (K^{(T_{\ell'})})^\top = \frac{T_\ell \cdot T_{\ell'}}{\log n} \cdot  Q^{(\ell)} (K^{(\ell')})^\top$ so that the $\ell_\infty$ norm of $Q^{(\ell)} (K^{(\ell')})^\top$ is small. As $\exp$ is applied entry-wisely, we can get 
    $\exp(Q^{(T_\ell)} (K^{(T_{\ell'})})^\top) = \exp(Q^{(\ell)} (K^{(\ell')})^\top)^{\circ \frac{T_\ell \cdot T_{\ell'}}{\log n}}$,
where the power $\frac{T_\ell \cdot T_{\ell'}}{\log n}$ is also applied entry-wisely on the matrix $\exp(Q^{(\ell)} (K^{(\ell')})^\top)$. Then, we apply the polynomial method \citep{as23} to approximate $\exp\paren{Q^{(\ell)} (K^{(\ell')})^\top} \approx U_1^{\paren{ \ell, \ell'}} \paren{ U_2^{\paren{ \ell, \ell'}}}^\top$ in almost linear time. We can finally approximate the polynomial attention 
    $D^{-1} \paren{U_1^{\paren{ \ell, \ell'}} \paren{ U_2^{\paren{ \ell, \ell'}}}^\top}^{\circ \frac{T_\ell \cdot T_{\ell'}}{\log n}} V$
in linear time in $n$ via oblivious sketching \citep{kmz24, akk+20}, which provides a randomized low-dimensional embedding so that 
    $\left \langle \paren{U_1^{\paren{ \ell, \ell'}}}_{i, *}, \paren{U_2^{\paren{ \ell, \ell'}}}_{j, *} \right \rangle^{\frac{T_\ell \cdot T_{\ell'}}{\log n}} \approx \left \langle \phi'\paren{\paren{U_1^{\paren{ \ell, \ell'}}}_{i, *}}, \phi'\paren{\paren{U_2^{\paren{ \ell, \ell'}}}_{j, *}} \right\rangle$
without explicitly computing the high dimensional inner product.

\paragraph{Error Analysis}
The reason to use bucketing is to control the range of values over which we approximate $\exp(x)$. Without bucketing, we would need to approximate $\exp(x)$ over the entire range of entries in $QK^\top / d$, which may contain both very small and very large values. If this range is wide, then we must take out a large scalar, which leads to a very high-degree polynomial. This will greatly increase the error, as by the mean value theorem, we can show
    $\left \|\mathcal{F}^{\circ p} - \mathcal{G}^{\circ p}\right \|_\infty \leq p \cdot \beta^{p-1} \cdot \left \|\mathcal{F} - \mathcal{G}\right \|_\infty$,
where $\left \|\mathcal{F} - \mathcal{G}\right \|_\infty$ is the error of the Chebyshev polynomial \citep{as23} and $\beta$ is the largest entry in $\mathcal{F}$ and $\mathcal{G}$.
Bucketing addresses this issue by splitting the entries into groups (buckets) according to their magnitude. In each bucket, the maximum absolute value $B_{\text{bucket}}$ is much smaller than the global maximum $B$. Thus, we can approximate $\exp(x)$ over $[-B_{\text{bucket}}, B_{\text{bucket}}]$ using a much lower-degree polynomial, reducing all $\ell_p$ error.

In attention computation, bucketing corresponds to 
decomposing the $(Q, K)$-softmax attention matrix into a sum of disjoint components, each of which can be approximated by a polynomial attention. This yields
\[
    \exp\paren{QK^\top / d} \approx 
    \sum_{\text{buckets } b} p_b\paren{U_1^{\paren{ \ell, \ell'}} \paren{ U_2^{\paren{ \ell, \ell'}}}^\top}.
\]
Each $p_b$ is a polynomial tailored to the value range of its bucket. Because each polynomial is specialized to a narrower range, it 
matches the exponential function more closely within that range. Summing these accurate, range-specific approximations produces a result that is closer to the original softmax attention than using a single polynomial approximation over the full range. This concludes our proof sketch for Theorem~\ref{thm:attention_approximation_bucketing_informal}.

\ifdefined\isarxiv

\section{Experimental Results}

\else
\section{EXPERIMENTAL RESULTS}

\fi

\label{sec:exp}

\paragraph{Computational Efficiency}

\begin{figure*}[!ht]
    \centering
    \subfloat[$n = 8,192$]
    {
    \includegraphics[width=0.24\linewidth]{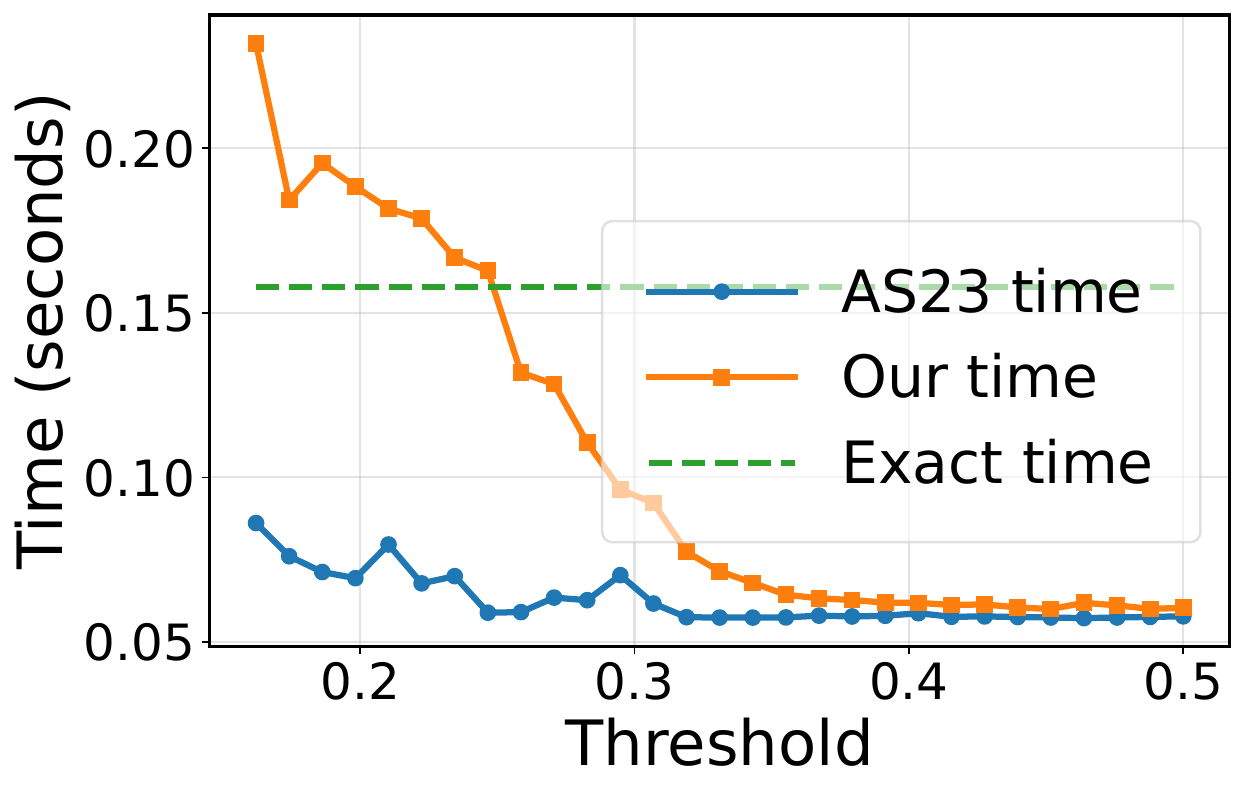}
    \includegraphics[width=0.24\linewidth]{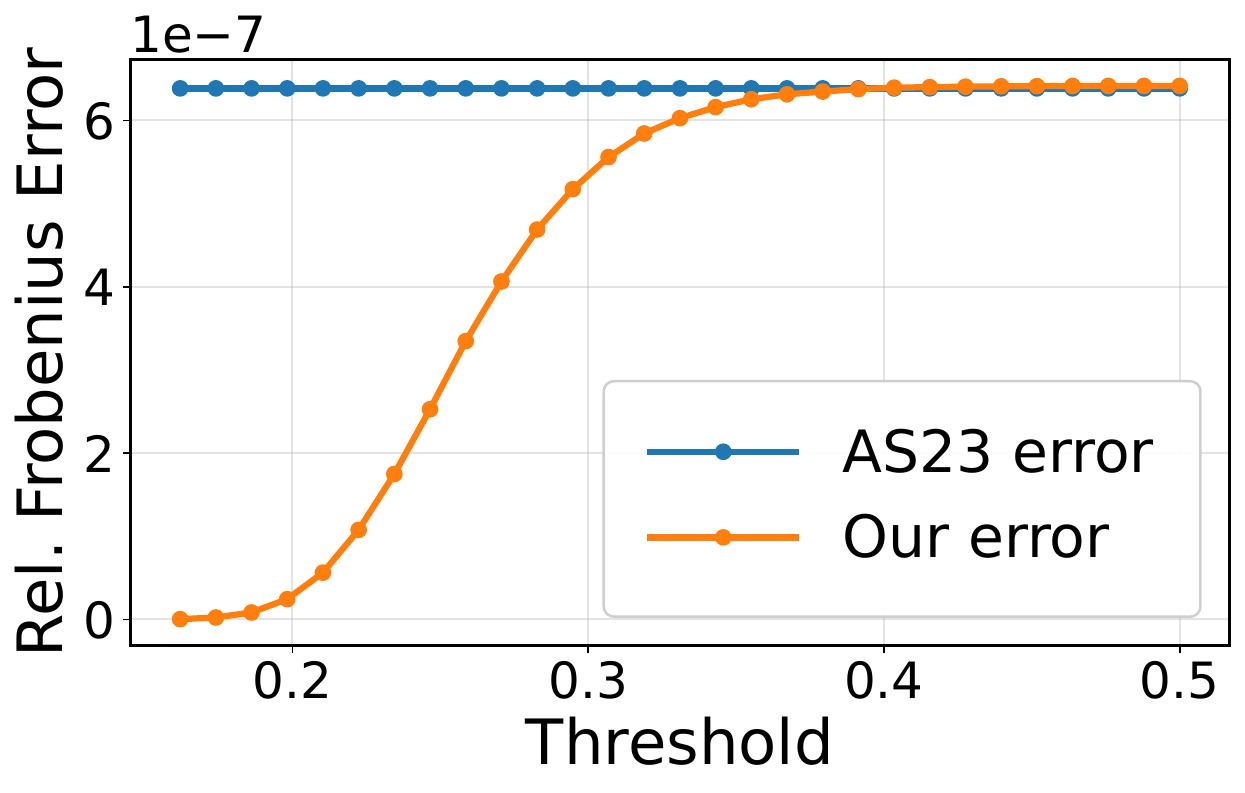}
    \label{fig:5000}
    }
    \subfloat[$n = 16,384$]
    {\includegraphics[width=0.24\linewidth]{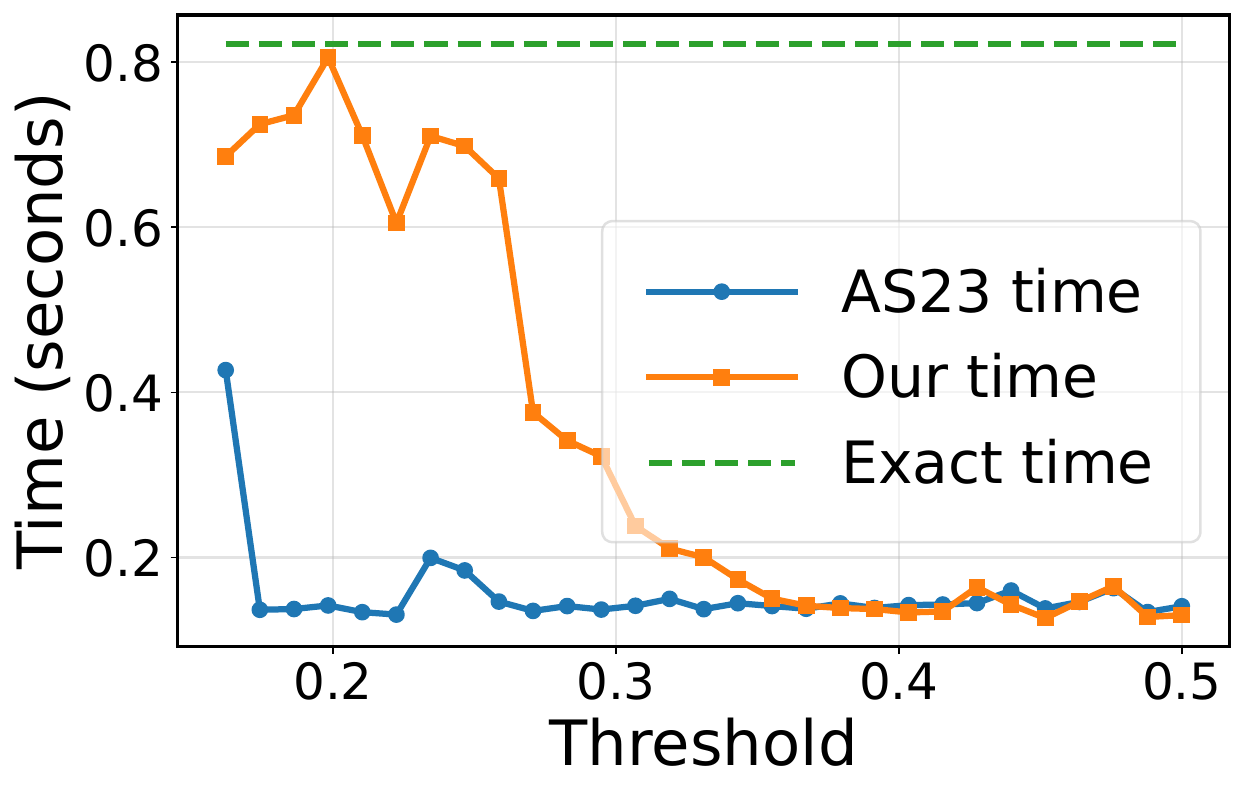}
    \includegraphics[width=0.24\linewidth]{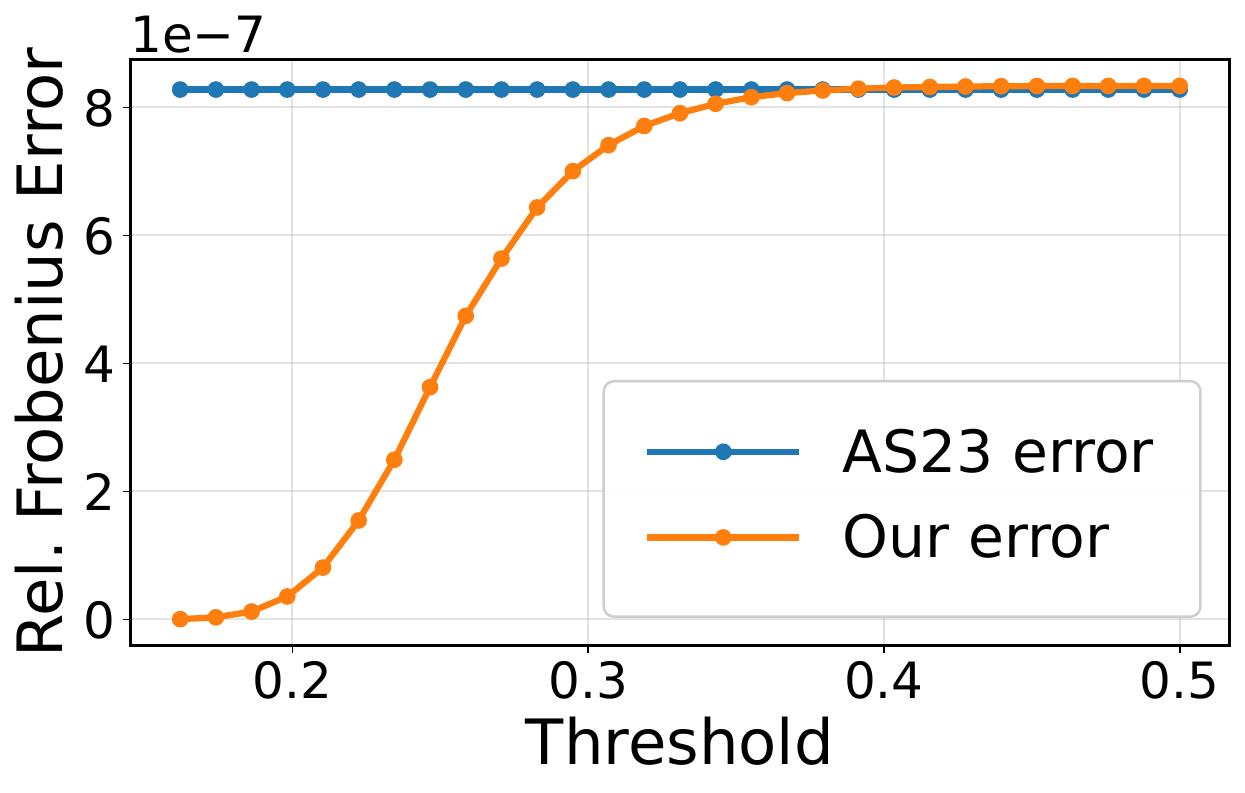}
    \label{fig:8192}
    }\\
    \subfloat[$n = 24,576$]
    {
    \includegraphics[width=0.24\linewidth]{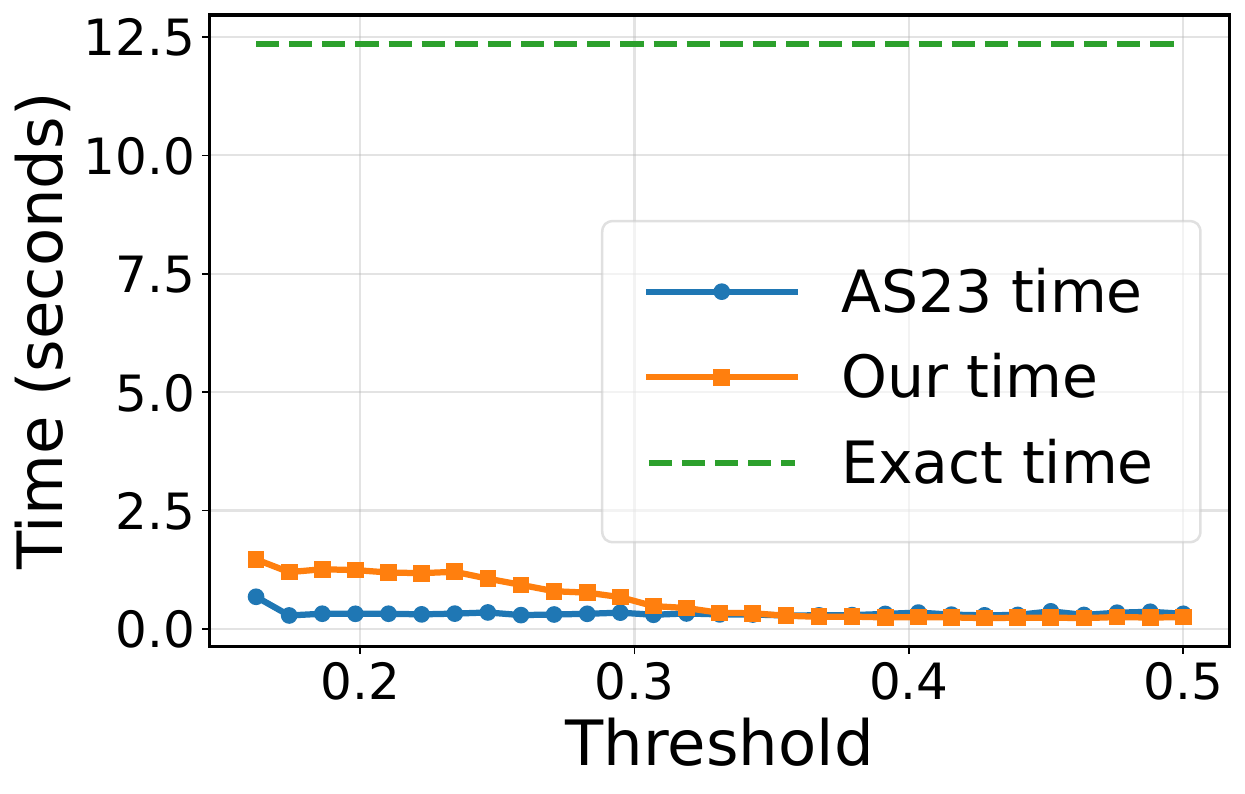}
    \includegraphics[width=0.24\linewidth]{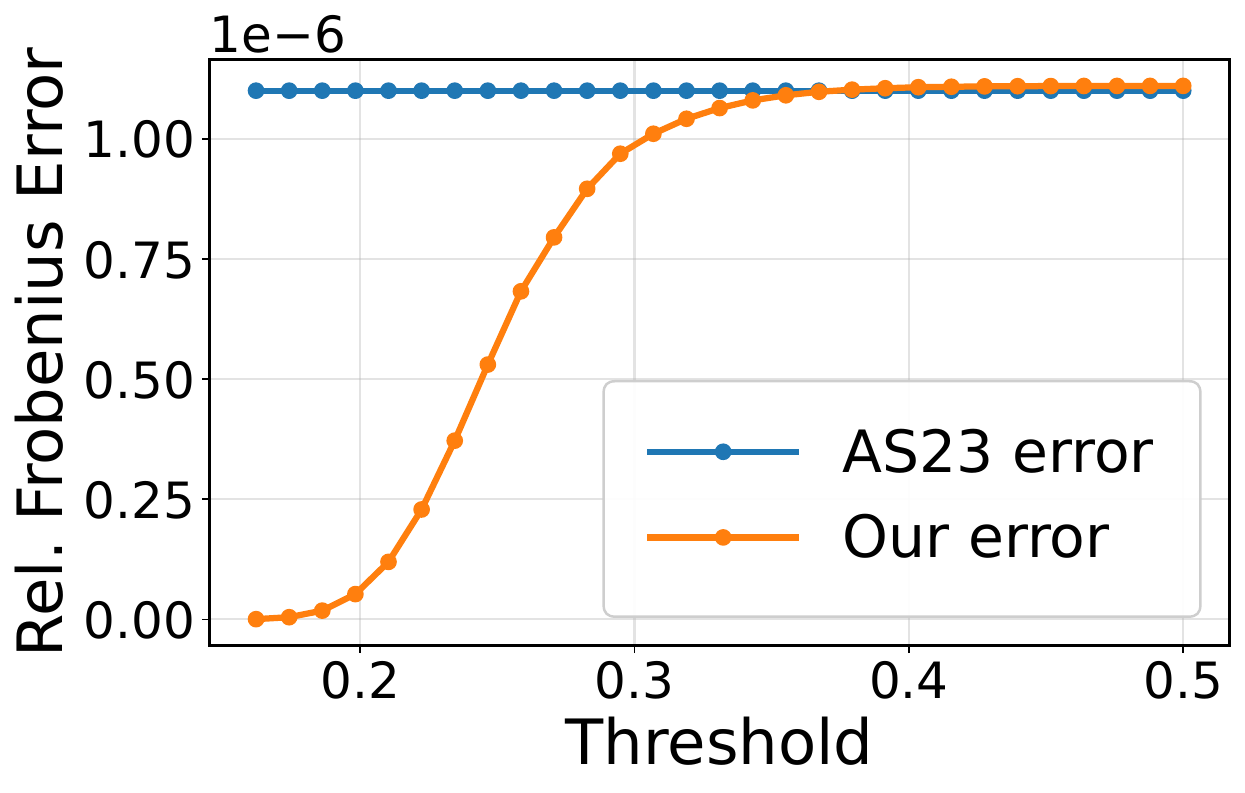}
    \label{fig:16000}
    }
    \subfloat[$n = 32,768$]
    {\includegraphics[width=0.24\linewidth]{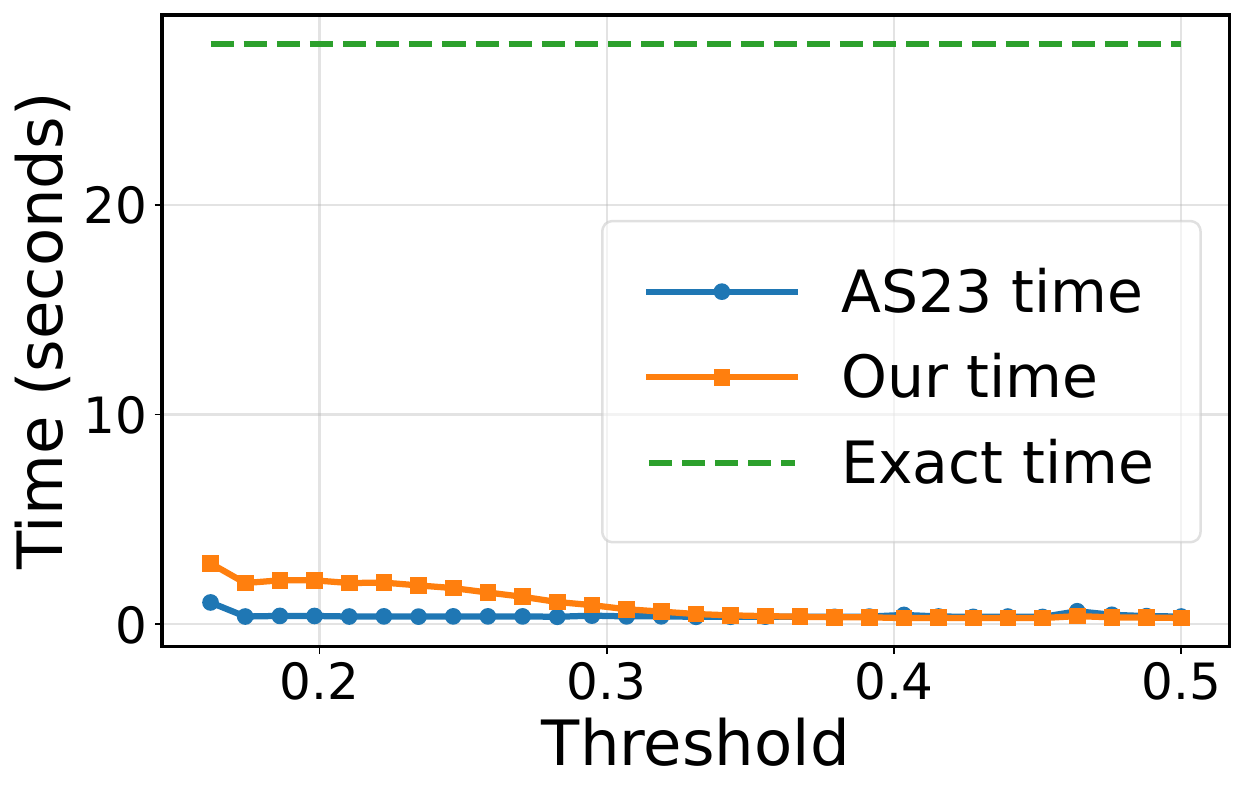}
    \includegraphics[width=0.24\linewidth]{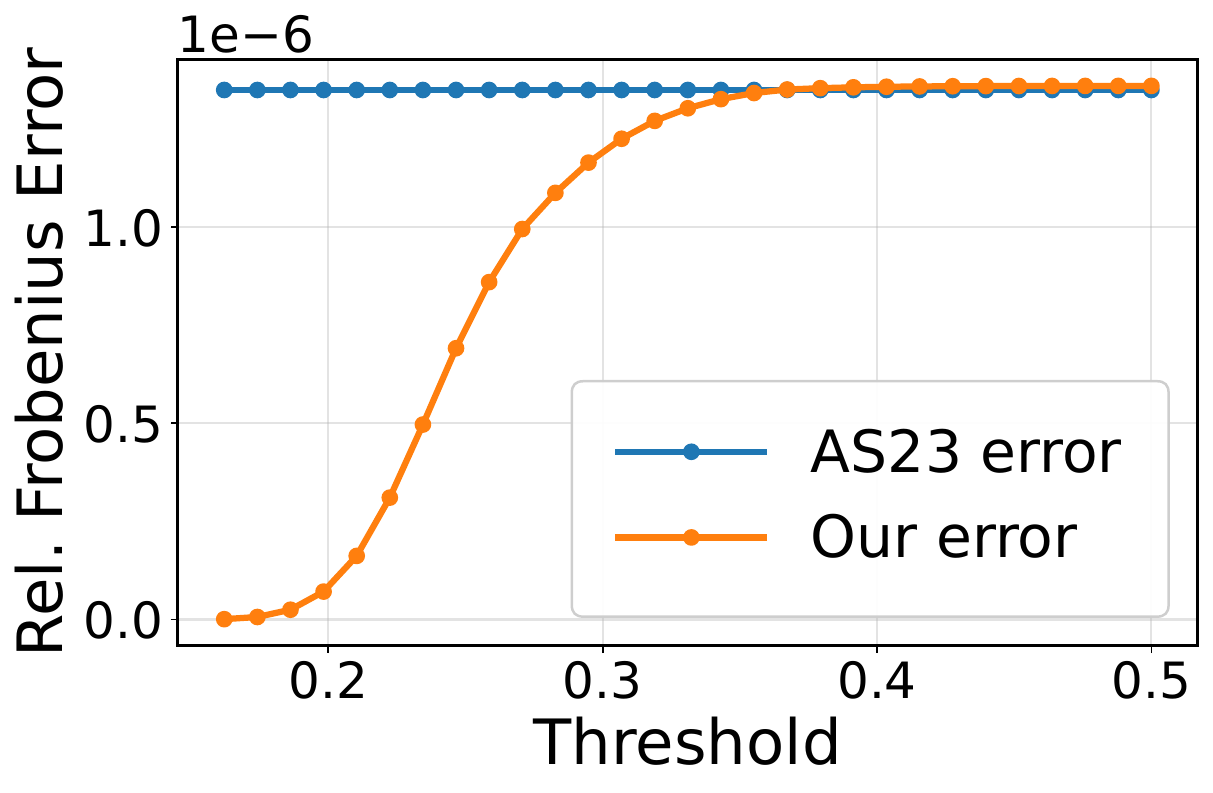}
    \label{fig:32000}
    }
    \caption{For each of Figures~\ref{fig:5000}, \ref{fig:8192}, \ref{fig:16000}, and \ref{fig:32000}, the left plot shows the change in running time with respect to the threshold~$T$, while the right plot shows the change in error with respect to~$T$. The blue curves correspond to \cite{as23}, the orange curves correspond to our single-threshold support basis method (Algorithm~\ref{alg:approx_attention}), and the dotted green lines represent the exact attention computation (Definition~\ref{def:exact_attention_computation}).} 
    \label{fig:attention_time_exp}
\end{figure*}

We first compare the computational efficiency of the exact attention computation, the method of \cite{as23}, and our proposed approach. We used an Apple M3 CPU to run this experiment.
With $Q, K, V \in \R^{n \times d}$, we set $d = 64$ and $n \in \{8192, 16384, 24576, 32768\}$ consistent with real-world Transformer deployments. 
To satisfy our sub-Gaussian assumption, each entry of $Q$ and $K$ is drawn independently from a Gaussian distribution with mean $0$ and standard deviation $0.1$, i.e., for all $(i,j) \in [n] \times [d]$, 
    $Q_{i,j} \sim \mathcal{N}(0, 0.1^2), K_{i,j} \sim \mathcal{N}(0, 0.1^2)$.

\begin{table*}[!ht]
\centering
\caption{Performance of LLaDA-8B-Instruct with different attention approximation methods across various benchmarks. We evaluate the accuracy on these benchmarks. (4) and (6) are degrees of Chebyshev polynomial.}
\begin{tabular}{lcccccc}
\toprule
 & \textbf{GSM8K} & \textbf{MMLU} & \textbf{Hellaswag} & \textbf{ARC-easy} & \textbf{ARC-challenge} & \textbf{Average} \\
\midrule
Exact computation          & 60.57\% & 62.56\% & 75.97\% & 91.62\% & 84.64\% & 75.47\% \\
Our method (4)    & 54.81\% & 61.17\% & 74.66\% & 89.90\% & 82.42\% & 72.59\% \\
Our method (6)    & 60.65\% & 62.62\% & 75.75\% & 91.58\% & 84.98\% & 75.11\% \\
\cite{as23} (4)    & 0\%     & 0\%     & 0\%     & 0\%     & 0\% & 0\%     \\
\cite{as23} (6)    & 0\%     & 0\%     & 0\%     & 0\%     & 0\% & 0\%    \\
\bottomrule
\label{tab:downstream_task}
\end{tabular}
\end{table*}

\cite{as23} uses polynomial approximation to all entries, so it suffers a significant drop in accuracy due to the bounded-entry assumption. In contrast, our method allows for more flexible control of the accuracy–efficiency trade-off by partitioning large and small entries using a threshold $T > 0$. Small entries satisfies the bounded-entry assumption, so we apply polynomial approximation to gain efficiency with little loss in accuracy; for large entries, when the assumption is violated, we use exact computation.
We evaluate our method with $T$ ranging from 0.15 to 0.5. As $T$ increases, our method performs more approximation and less exact computation.
For all $n \in \{8192, 16384, 24576, 32768\}$, we can approximate the attention computation both efficiently and accurately by tuning $T$.

In addition, the approximation error $\epsilon$ is \emph{not} driven by the sequence length $n$ but instead is determined by the entry bound $B$ in $Q$ and $K$. Our experiment (right plot of each Figures~\ref{fig:5000}--\ref{fig:32000}) is consistent with the theoretical result in the Chebyshev approximation bound. When $\epsilon = 1/\mathrm{poly}(n)$ and $d = O(\log n)$, the required polynomial degree satisfies $g = O\paren{\max\left\{\frac{\log(1/\epsilon)}{\log(\log(1/\epsilon)/B)}, B^2\right\}}$ (Lemma 3.4 of \cite{as23}). Thus, when $g$ and $n$ are fixed, increasing $B$ reduces the denominator $\log(\log(1/\epsilon)/B)$, which forces $\log(1/\epsilon)$ to shrink. Equivalently, $\epsilon$ must increase when $B$ grows, which is exactly what we observe in our experiment (Figure~\ref{fig:attention_time_exp}).

In contrast, the runtime heavily depends on $n$. Since our method reduces the attention computation from $O(n^2 d)$ to sub-quadratic time, the absolute runtime improvement becomes more pronounced as $n$ increases. As shown in Figure~\ref{fig:attention_time_exp}, larger sequence lengths yield substantially larger speedups \emph{without} introducing additional drawbacks in approximation error, making our method more suitable for larger transformer models with very long contexts, which aligns with the development of modern transformer architectures.

\paragraph{Performance on Downstream Tasks}

We evaluate the performance of LLaDA-8B-Instruct \citep{nie2025large} on a range of standard benchmarks, including GSM8K \citep{gsm8k}, MMLU \citep{mmlu}, Hellaswag \citep{zellers2019hellaswag}, ARC-easy \citep{arc}, and ARC-challenge \citep{arc}. 
We perform the exact computation for the largest and smallest 10\% of entries in $Q, K$ and applies the polynomial approximation to the remaining 80\%. 
In contrast, \cite{as23} approximates all entries of the attention matrix. 
Table~\ref{tab:downstream_task} shows the results. Our method demonstrates a clear advantage over \cite{as23} in practical settings. With a degree-4 Chebyshev approximation, our approach incurs less than a 3\% drop in accuracy compared to exact attention. With a degree-6 Chebyshev approximation, the performance is nearly indistinguishable from the original computation. By contrast, \cite{as23} fails to produce meaningful results, yielding zero accuracy across all benchmarks. 
The superiority of our method over \cite{as23} stems from two key factors. 
First, in multi-layer transformers, approximation errors in the attention computation accumulate across blocks; our hybrid strategy is substantially more robust to such error propagation. 
Second, large entries in $Q$ and $K$ are particularly sensitive to approximation error. By computing these entries exactly, our method avoids the large errors that \cite{as23} cannot, since it treats all entries uniformly.

\section*{Acknowledgment}

Maryam Aliakbarpour is affiliated with the Ken Kennedy Institute at Rice University. Vladimir Braverman is supported in part by NSF Grant CNS 2528780.

\ifdefined\isarxiv

\else
\bibliography{ref}
\bibliographystyle{plainnat}
\section*{Checklist}

\begin{enumerate}

  \item For all models and algorithms presented, check if you include:
  \begin{enumerate}
    \item A clear description of the mathematical setting, assumptions, algorithm, and/or model. [Yes]
    \item An analysis of the properties and complexity (time, space, sample size) of any algorithm. [Yes]
    \item (Optional) Anonymized source code, with specification of all dependencies, including external libraries. [Not Applicable]

    Justification: We will release our source code if our paper is accepted.

  \end{enumerate}

  \item For any theoretical claim, check if you include:
  \begin{enumerate}
    \item Statements of the full set of assumptions of all theoretical results. [Yes]
    \item Complete proofs of all theoretical results. [Yes]
    \item Clear explanations of any assumptions. [Yes]     
  \end{enumerate}

  \item For all figures and tables that present empirical results, check if you include:

  \begin{enumerate}
    \item The code, data, and instructions needed to reproduce the main experimental results (either in the supplemental material or as a URL). [No]
    
    Justification: We will release our code, data, and instructions to reproduce figures and tables if our paper is accepted.
    \item All the training details (e.g., data splits, hyperparameters, how they were chosen). [Not Applicable]

    Justification: We did not train any models. However, for our experiment in which we run inference, we have provided training details, such as hyperparameters, in our paper.
    \item A clear definition of the specific measure or statistics and error bars (e.g., with respect to the random seed after running experiments multiple times). [Not Applicable]

    Justification: In our paper, we do not use error bars.
    
    \item A description of the computing infrastructure used. (e.g., type of GPUs, internal cluster, or cloud provider). [Yes]
  \end{enumerate}

  \item If you are using existing assets (e.g., code, data, models) or curating/releasing new assets, check if you include:

  \begin{enumerate}
    \item Citations of the creator If your work uses existing assets. [Yes]
    \item The license information of the assets, if applicable. [Yes]
    \item New assets either in the supplemental material or as a URL, if applicable. [Not Applicable]
    \item Information about consent from data providers/curators. [Not Applicable]
    \item Discussion of sensible content if applicable, e.g., personally identifiable information or offensive content. [Not Applicable]
  \end{enumerate}

  \item If you used crowdsourcing or conducted research with human subjects, check if you include:
  \begin{enumerate}
    \item The full text of instructions given to participants and screenshots. [Not Applicable]
    \item Descriptions of potential participant risks, with links to Institutional Review Board (IRB) approvals if applicable. [Not Applicable]
    \item The estimated hourly wage paid to participants and the total amount spent on participant compensation. [Not Applicable]
  \end{enumerate}

\end{enumerate}

\fi

\newpage
\onecolumn
\appendix

\ifdefined\isarxiv

\else
\aistatstitle{Support Basis: Fast Attention Beyond Bounded Entries: \\
Supplementary Materials}

\tableofcontents
\fi

\ifdefined\isarxiv

{\LARGE  
\begin{center}
    \textbf{Appendix}
\end{center}

}
\else

\fi

\paragraph{Roadmap.}

Our paper presents two main results. First, we introduce the use of a \emph{single threshold support basis} to separate the large and small entries of the query and key matrices $Q, K \in \R^{n \times d}$. Under the sub-Gaussian assumption, this allows us to design a sub-quadratic time algorithm for approximating the attention computation, without requiring the bounded-entry condition $\|Q\|_\infty, \|K\|_\infty < o\paren{\sqrt{\log n}}$ used in \cite{as23}. We first use Appendix~\ref{sec:preli} to present the background of \cite{as23,kmz24}. Then, we use Appendix~\ref{sec:single_basis}, Appendix~\ref{sec:attention_single_basis}, and Appendix~\ref{sec:attention_single_basis_subgaussian} to support our result.

Specifically, in Appendix~\ref{sec:single_basis}, we formally define the support basis and show how to split the large and small entries of the query and key matrices $Q, K \in \R^{n \times d}$ in the attention computation. In Appendix~\ref{sec:attention_single_basis}, we use our theoretical result on the \emph{single threshold support basis} to approximate the attention computation in $A^{\paren{L}}$ sparsity time. In Appendix~\ref{sec:attention_single_basis_subgaussian}, we introduce our sub-Gaussian assumption and show that the number of large entries in $Q$ and $K$ is bounded, so that with high probability, the running time of our attention approximation algorithm, namely the $A^{\paren{L}}$ sparsity time, is sub-quadratic. In Appendix~\ref{sec:multi_layer}, we show how we can generalize our algorithm to multi-layer attention.

Second, we propose the use of a \emph{multiple thresholds support basis} to decompose $Q, K \in \R^{n \times d}$ into the sum of several matrices according to these thresholds. For each resulting component, we apply the polynomial approximation method from \cite{as23} and the sketching technique from \cite{akk+20, kmz24} to approximate and solve all subproblems in sub-quadratic time—without making any distributional assumptions, though at the cost of reduced accuracy. We use Appendix~\ref{sec:multi_basis}, Appendix~\ref{sec:multi_basis_threshold}, Appendix~\ref{sec:multi_basis_sketching}, and Appendix~\ref{sec:multi_basis_attention} to support this result. 

Specifically, in Appendix~\ref{sec:multi_basis}, we generalize the theoretical results of the \emph{single threshold support basis} to the setting with multiple thresholds and explain how to decompose $Q, K \in \R^{n \times d}$ accordingly. In Appendix~\ref{sec:multi_basis_threshold}, we specify how to choose the thresholds to ensure that the number of decomposed components remains bounded. In Appendix~\ref{sec:multi_basis_sketching}, we adapt the sketching techniques from \cite{akk+20, kmz24} to our setting. In Appendix~\ref{sec:multi_basis_attention}, we combine the sketching techniques with our \emph{multiple thresholds support basis} to approximate the attention computation in sub-quadratic time, without assuming bounded entries or any distributional conditions.

Finally, in Appendix~\ref{sec:facts}, we present the basic mathematical facts used throughout the paper to support our proofs. In Appendix~\ref{sec:more_related}, we present additional related works. In Appendix~\ref{sec:distribution}, we show the entry distributions of the query and key matrices on different transformer architectures across different layers.

\paragraph{Notation.}

For all positive integer $n, d$, we denote $\R, \R^n, \R^{n \times d}$ as sets containing all real numbers, all $n$-dimensional vectors, and $n \times d$ matrices, whose entries are all in $\R$. We define $[n] := \{1, 2, \dots, n\}$. For all set $X$, we use $|X|$ to denote its cardinality, namely the number of elements in this set. For all sets $X$ and $Y$, we define $X \times Y := \{\paren{x, y} \mid x \in X, y \in Y\}$.

For all $a \in \R$, we define $\lfloor a \rfloor$ as the largest integer satisfying $\lfloor a \rfloor \leq a$. For all $x, y \in \R^d$, their inner product is $\langle x, y \rangle := \sum_{i = 1}^d x_i y_i$. With any arbitrary $i \in [d]$, the Hadamard product $\circ$ is a binary operation: $x \circ y \in \R^d$ is defined as $(x \circ y)_i := x_i \cdot y_i$. For all $p$ being a positive integer or $\infty$, we define the $\ell_p$ norm of $x$ as $\|x\|_p : = \paren{\sum_{i \in [d]} |x_i|^p}^{1/p}$. We let ${\bf 1}_d$ and ${\bf 0}_d$ respectively denote the $d$-dimensional vectors whose entries are all $1$'s and $0$'s. Similarly, we respectively define ${\bf 1}_{n \times d}$ and ${\bf 0}_{n \times d}$ as the $n \times d$ matrix whose entries are all $1$'s and $0$'s. We define $\diag(x) \in \R^{d \times d}$ to be the diagonal matrix with $\diag(x)_{i, i} = x_i$.

Let $A \in \R^{n \times d}$. We let $A_{i, j} \in \R$ be the $(i, j)$-th entry of $A$, $A_{i, *} \in \R^d$ be the $i$-th row, and $A_{*, j} \in \R^n$ be the $j$-th column. We let $A^\top \in \R^{d \times n}$ be the transpose. $\| A \|_F, \|A\|_\infty \in \R$ respectively are the Frobenius norm and $\ell_\infty$ norm, where $\|A\|_F := \sqrt{\sum_{i \in [n]} \sum_{j \in [d]} |A_{i, j}|^2}$ and $\| A \|_\infty := \max_{i \in [n], j \in [d]} |A_{i, j}|$. We define $\supp(A) := \{(i, j) \in [n] \times [d] \mid A_{i, j} \neq 0\}$. Let $n_1, n_2, d_1, d_2, p$ be positive integers. We define $\exp(A) \in \R^{n \times d}$ and $A^{\circ p} \in \R^{n \times d}$ as the entry-wise exponential and power, namely for all $i \in [n]$ and $j \in [d]$, we have $\exp(A)_{i, j} := \exp\paren{A_{i, j}}$ and $\paren{A^{\circ p}}_{i, j} := \paren{A_{i, j}}^p$. We define $\|A\|_p := \sqrt[p]{\sum_{i, j} |A_{i, j}|^p}$. Let $A \in \R^{n_1 \times d_1}$. 
We let $A^{\otimes p}$ be $\underbrace{A \otimes A \otimes \cdots \otimes A}_{p} \in \R^{n_1 \times d_1^p}$ be the row-wise Kronecker product. 

For a differentiable function $f$, we use $f'$ to denote its derivative. 
For a probability space $(\Omega, \mathcal{F}, \Pr)$, where $\Omega$ is the sample space, $\mathcal{F}$ is the $\sigma$-algebra of events, and $\Pr: \mathcal{F} \to [0, 1]$ is the probability measure, we define the discrete random variable $X : \Omega \to \R$ so that the expectation of $X$ is $\E[X] := \sum_i x_i \cdot \Pr[X = x_i]$, and for all $A \in \mathcal{F}$, we define the indicator function $\mathbb{I}[A]: \Omega \to \{0, 1\}$ as $\mathbb{I}[A](\omega) := 1$ if $\omega \in A$ and $\mathbb{I}[A](\omega) := 0$ if $\omega \notin A$, for all $\omega \in \Omega$.

\ifdefined\isarxiv

\section{Preliminaries}

\else
\section{PRELIMINARIES}

\fi

\label{sec:preli}

In this section, we provide the theoretical foundation necessary to understand and contextualize our contributions in attention approximation. We give a summary of the prior work~\citep{as23,aa22}, which established sub-quadratic time algorithms for attention approximation under strong boundedness assumptions. We also present the important mathematical properties from \cite{kmz24,akk+20}.

In Appendix~\ref{sub:preli:poly}, we give an overview of techniques in \cite{as23,aa22}. In Appendix~\ref{sub:multi_basis_sketching:background}, we present the background related to the sketching technique and polynomial attention \citep{kmz24,akk+20}. 

\subsection{Background: Technique Overview of the Polynomial Method}
\label{sub:preli:poly}

\paragraph{Accuracy-Efficiency trade-off.}
Before delving into our method, we first provide an overview of the techniques used in \cite{aa22, as23} to explain the origin of the bounded entry assumption. The attention approximation (Definition~\ref{def:approximate_attention_computation}) can be computed in sub-quadratic time by replacing the entry-wise exponential function $\exp: \mathbb{R} \to \mathbb{R}$ in the softmax unit with a degree-$g$ Chebyshev polynomial $p: \mathbb{R} \to \mathbb{R}$, thereby reducing the time complexity. This polynomial is also applied entry-wisely to $QK^\top / d \in \R^{n \times n}$, where each entry $\langle Q_{i, *}, K_{j, *} \rangle / d$ is the inner product between a query and a key vector. When this polynomial is expanded, it produces a collection of monomials that are products of certain entries from \( Q \) and \( K \). These terms can be reorganized into a low-rank (rank-$r$) factorization of the form \( U_1 U_2^\top \), where \( U_1, U_2 \in \mathbb{R}^{n \times r} \) and \( r = \binom{2d + 2g}{2d} \). Similar as \cite{as23_video}, we give the following example to approximate the $(i, j)$-th entry of $\exp\paren{QK^\top / d}$: 
\begin{align}
        & ~ p\paren{\langle Q_{i, *}, K_{j, *} \rangle / d } \notag\\
        = & ~ p\paren{\frac{1}{d} \sum_{\ell = 1}^d Q_{i, \ell} \cdot  K_{j, \ell}} \notag\\
        = & ~ 3 Q_{i, 1} K_{j, 1} Q_{i, 3}  K_{j, 3} - 4 Q_{i, 7}  K_{j, 7} Q_{i, 16}^3  K_{j, 16}^3 + \dots \label{eq:poly_expansion}\\
        = & ~ \left \langle \underbrace{\begin{bmatrix}
            3 Q_{i, 1} Q_{i, 3}\\ -4 Q_{i, 7} Q_{i, 16}^3\\  \vdots\end{bmatrix}}_{\paren{U_1}_{i, *}}, \underbrace{\begin{bmatrix} K_{j, 1} K_{j, 3}\\ K_{j, 7} K_{j, 16}^3\\ \vdots \end{bmatrix}}_{\paren{U_2}_{j, *}} \right\rangle \label{eq:poly_expansion_inner_product}\\
        = & ~ \paren{U_1}_{i, *}^\top \cdot \paren{U_2}_{j, *}. \label{eq:poly_expansion_U1_U2}
    \end{align}

\begin{figure*}[!ht]
    \centering
    \includegraphics[width=\linewidth]{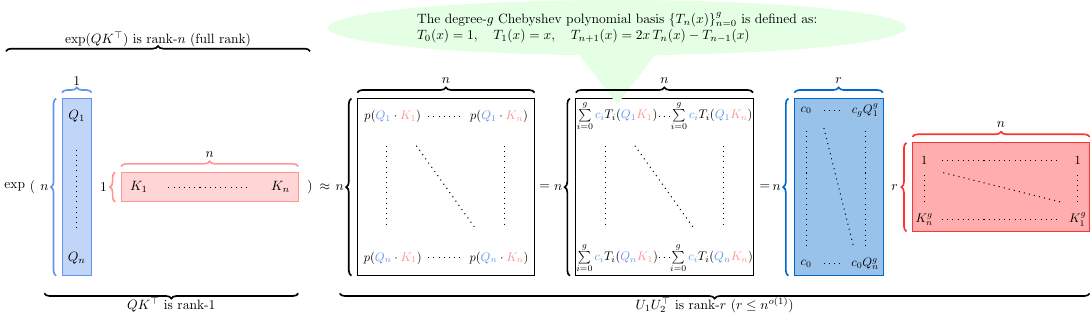} 
    \caption{A visualization of the polynomial method. For simplicity, we set $d = 1$. The matrix $QK^\top$ has rank 1, but the entrywise exponential function $\exp$ may map it to a full-rank matrix. \cite{as23} shows that replacing $\exp$ with a degree-$g$ Chebyshev polynomial $p$ can produce a matrix of rank $r$.  
    }
    \label{fig:poly_method}
\end{figure*}

For all $B > 0$, to accurately approximate \(\exp(x)\) on the range \([ -B, B]\), the degree \(g\) of the polynomial must grow with \(B\) (see Lemma~\ref{lem:wt_A_small_rank} for details). This leads to an accuracy–efficiency trade-off: increasing \(g\) improves the approximation accuracy, but also increases the rank \( r = \binom{2d + 2g}{2d} \), thereby slowing down the computation.

\paragraph{Time complexity for constructing \( U_1, U_2 \in \mathbb{R}^{n \times r} \) and approximating the attention computation.} We now explain why constructing $U_1, U_2 \in \mathbb{R}^{n \times r}$ takes $O(n r g)$ time and using them to approximate the attention computation takes $O(n r d)$ time. As shown in Eq.~\eqref{eq:poly_expansion_U1_U2}, each degree-$g$ Chebyshev polynomial may give one row of $U_1$, denoted $(U_1)_{i,*}$, and one row of $U_2$, denoted $(U_2)_{j,*}$. As $i, j \in [n]$, we need $n$ numbers of Chebyshev polynomials to fully construct $U_1, U_2 \in \R^{n \times r}$. Each polynomial expansion contains $r$ monomials, so each row vector $(U_1)_{i,*}$ or $(U_2)_{j,*}$ has $r$ entries to be computed (see from Eq.~\eqref{eq:poly_expansion} to Eq.~\eqref{eq:poly_expansion_inner_product} as an example).
Additionally, if the polynomial is degree-$g$, computing the value of each entry of $(U_1)_{i,*}$ or $(U_2)_{j,*}$ involves multiplying up to $O(g)$ scalar factors, i.e., entries from $Q$ or $K$ (see entries in Eq.~\eqref{eq:poly_expansion_inner_product} as an example). Thus, each monomial evaluation costs $O(g)$ scalar multiplications. Since we compute $n$ rows, each with $r$ monomials, and each monomial costs $O(g)$ operations, constructing $U_1$ and $U_2$ requires $O(n r g)$ time.
Now, we do not need to explicitly form the $n \times n$ attention matrix $A = \exp\paren{QK^\top / d}$ but instead work with the much smaller factors $U_1$ and $U_2$ satisfying $A \approx U_1 U_2^\top$.
Finally, using the associativity of matrix multiplication, the attention computation (as in Definition~\ref{def:exact_attention_computation})
\begin{align*}
    \diag\paren{\underbrace{U_1}_{n \times r} \paren{\underbrace{U_2^\top}_{r \times n} \underbrace{{\bf 1}_n}_{n \times 1}}}^{-1} \underbrace{U_1}_{n \times r} \paren{\underbrace{U_2^\top}_{r \times n} \underbrace{V}_{n \times d}}
\end{align*}
can be approximated in $O\paren{nrd}$ time.

\paragraph{Limitations.}
To approximate the attention in almost linear time, we need to make $O(nrg + nrd) = O(n^{1 + o(1)})$, which is equivalent to making $d, g, r \leq n^{o(1)}$. However, keeping $g$ and $r$ small requires $B$ (the maximum absolute entry of $Q$ or $K$) to be small. The analysis in~\cite{as23} shows that achieving both a fast runtime and a small approximation error requires $d = O(\log n)$ and $B = o\left(\sqrt{\log n}\right)$. This condition on $B$ is known as the bounded entry assumption.

\subsubsection{Background: Low Rank Approximation}\label{sec:low_rank:definition}

In this section, we review concepts of matrix low-rank approximation. A key idea is to approximate the softmax attention matrix, which is dense and expensive to compute, using a low-rank decomposition constructed from polynomial expansions. This technique is central to the method proposed in~\cite{as23}, where the exponential function is approximated by a Chebyshev polynomial, and the resulting polynomial matrix is factorized into a product of two low-rank matrices. This allows for efficient computation of the attention matrix without explicitly forming it. We formally define the notion of an \((\epsilon, r)\)-approximation and present a key lemma from~\cite{as23} that shows how to construct such a factorization efficiently.

\begin{definition}\label{def:epsilon_g_approx}
Let $r \geq 1$ denote a positive integer. Let $\epsilon \in \paren{ 0,0.1}$ denote an accuracy parameter. 
Given a matrix $A \in \R^{n \times n}_{\geq 0}$, we say $\wt{A} \in \R^{n \times n}_{\geq 0}$ is an $\paren{ \epsilon,r}$-approximation of $A$ if 
\begin{itemize}
    \item $\wt{A} = U_1 \cdot U_2^\top$ for some matrices $U_1, U_2 \in \R^{n \times r}$ (i.e., $\wt{A}$ has rank at most $r$), and
    \item $| \wt{A}_{i,j} - A_{i,j} | \leq \epsilon \cdot A_{i,j}$ for all $\paren{ i,j} \in [n]^2$.
\end{itemize}
\end{definition}

\begin{lemma}[Lemma 3.2 in \cite{as23}]\label{lem:rank_is_small}
Let $M = X Y^\top \in \R^{n \times n}$ denote a matrix with 
$X, Y \in \R^{n \times d}$. Let $P\paren{ x}$ denote a degree-$g$ polynomial, and define $r = \binom{2\paren{ g+d}}{2g}$.

There is an algorithm that runs in $O\paren{ n r g}$ time and, given as input the matrix $X,Y$, constructs matrices $U_1, U_2 \in \R^{n \times r}$ such that $P\paren{ M} = U_1 U_2^\top$. (Here, $P\paren{ M}$ denotes the entry-wise application of $P$ to $M$.)
\end{lemma}

\subsubsection{Background: Approximating the Attention Computation Assuming \texorpdfstring{$d = O\paren{ \log n}$}{} and Bounded Entries}
\label{sub:preli:approx}

In this section, we describe the algorithm introduced in \cite{as23} for approximating the attention computation.

\begin{algorithm}[!ht]\caption{Algorithm 1 of \cite{as23}. It takes $Q \in \R^{n \times d},K \in \R^{n \times d},V \in \R^{n \times d}$ as input and approximates the attention computation (Definition~\ref{def:approximate_attention_computation}) under the assumptions $d = O\paren{ \log n}$, $\left \| Q \right \|_{\infty} \leq B$, $\left \| K \right \|_{\infty} \leq B$, and $\left \| V \right \|_{\infty} \leq B$, where $B = o\paren{ \sqrt{\log n}}$.}\label{alg:main_as23}
\begin{algorithmic}[1]
\Procedure{PolyAttention}{$Q \in \R^{n \times d},K \in \R^{n \times d},V \in \R^{n \times d}, n, d,  B, \epsilon$} \Comment{Theorem~\ref{thm:formal_main_upper_bound}}
    \State \Comment{$\epsilon$ is the accuracy output}
    \State \Comment{$\left \|Q\right \|_{\infty}, \left \| K \right \|_{\infty}, \left \| V \right \|_{\infty} \leq B$}
    \State $g \gets O\paren{  \max\left \{ \frac{\log\paren{ 1/\epsilon}}{\log\paren{ \log\paren{ 1/\epsilon} / B}} , B^2\right \} }$
    \State $r \gets r = \binom{2\paren{ g+d}}{2g}$
    \State Construct $U_1, U_2 \in \R^{n \times r}$ via Lemma~\ref{lem:wt_A_small_rank} \Comment{$O\paren{ nrg}$ time}
    \State $\wt{w} \gets U_1 \cdot \paren{ U_2^\top {\bf 1}_n}$ \Comment{$O\paren{ n r}$ time}
    \State $\wt{D}^{-1} = \diag\paren{  \wt{w}^{-1} }$ \Comment{$O\paren{ n}$ time}
    \State Compute $U_2^\top V \in \R^{r \times d}$ \Comment{Takes $\Tmat\paren{ r,n,d}$ time}
    \State Compute $U_1 \cdot \paren{ U_2^\top V}$ \Comment{ $\Tmat\paren{ n,r,d}$ time}
    \State $P \gets \wt{D}^{-1} \cdot \paren{ U_1 \cdot \paren{ U_2^\top V}} $ \Comment{$O\paren{ nd}$ time}
    \State \Return $P$ \Comment{$P \in \R^{n \times d}$}
\EndProcedure
\end{algorithmic}
\end{algorithm}

\begin{lemma}[Corollary 2.2 in \cite{as23}]\label{cor:aa22_from_-B_to_B}
Let $B > 1$ and let $\epsilon \in \paren{ 0,0.1}$.
There is a polynomial $P: \R \rightarrow \R$ of degree $g := \Theta\paren{  \max\left \{ \frac{\log\paren{ 1/\epsilon}}{ \log\paren{  \log\paren{ 1/\epsilon} / B  } } , B \right \} }$ such that for all $x \in [-B,B]$, we have
\begin{align*}
|P\paren{ x} - \exp\paren{ x}| < \epsilon \cdot \exp\paren{ x}.
\end{align*}
\end{lemma}

\begin{lemma}[An improved version of Lemma 3.4 in \cite{as23}]\label{lem:wt_A_small_rank}
Let $B = o\paren{ \sqrt{\log n}}$. Suppose $Q, K \in \R^{n \times d}$, with $\left \| Q K^\top \right \|_{\infty} \leq dB^2$. Let $A:=\exp\paren{ QK^\top /d} \in \R^{n \times n}$. For accuracy parameter $\epsilon \in \paren{ 0,1}$, there is a positive integer $g$ bounded above by 
\begin{align*}
g = O \paren{  \max \left \{ \frac{\log\paren{ 1/\epsilon}}{ \log\paren{ \log\paren{ 1/\epsilon}/B} }, B^2 \right \} },
\end{align*}
and a positive integer $r$ bounded above by
\begin{align*}
r \leq \binom{2\paren{ g+d}}{2g}
\end{align*}
such that: 
There is a matrix $\wt{A} \in \R^{n \times n}$ that is an $\paren{ \epsilon,r}$-approximation (Definition~\ref{def:epsilon_g_approx}) of $A \in \R^{n \times n}$.
Furthermore, the matrices $U_1$ and $U_2$ defining $\wt{A}$ can be computed in $O\paren{ n \cdot r}$ time.
\end{lemma}
\begin{proof}
    By the assumption that $\left \| Q K^\top \right \|_{\infty} \leq dB^2$, we can get
    \begin{align*}
        \left \| Q K^\top/d \right \|_{\infty} \leq B^2.
    \end{align*}

    Thus, applying Lemma~\ref{cor:aa22_from_-B_to_B} (with bound $B^2$ on its entries), there is a degree-$g$ polynomial $P$ such that the matrix $\wt{A} = P\paren{ M}$ is an $\paren{ \epsilon,r}$-approximation to $A$. We can then compute $U_1, U_2$ using Lemma~\ref{lem:rank_is_small}, which gives the bound  
\begin{align*}
    r = \binom{2\paren{ g+d}}{2g}.
\end{align*}
This completes the proof.
\end{proof}

\subsubsection{Background: Main Results of \texorpdfstring{\cite{as23}}{}}
\label{sub:preli:main}

In this section, we formally present the core theoretical results from \cite{as23}, which establish both upper and lower bounds for the approximate attention computation problem. 
The upper bound shows that when the input matrices $Q, K, V \in \mathbb{R}^{n \times d}$ have entries bounded in $\ell_\infty$ norm by $B = o(\sqrt{\log n})$, and the hidden dimension satisfies $d = O(\log n)$, then the attention computation can be approximated to within error $\epsilon = 1/\mathrm{poly}(n)$ in almost linear time $n^{1 + o(1)}$. 

\begin{theorem}[Upper bound, Theorem 3.8 of \cite{as23}]\label{thm:formal_main_upper_bound}
With $d = O(\log n),B = o(\sqrt{\log n}),\epsilon_a = 1/\poly(n))$, the approximate attention computation problem (see Definition~\ref{def:approximate_attention_computation}) can be solved in time $\Tmat(n,n^{o(1)},d) = n^{1+o(1)}$.
\end{theorem}

In contrast, the lower bound establishes a fundamental computational limitation: under the SETH, any algorithm that approximates attention within accuracy $\epsilon = 1/\mathrm{poly}(n)$ must take at least $n^{2 - o(1)}$ time when $B = \Theta(\sqrt{\log n})$.

\begin{definition}[Strong Exponential Time Hypothesis ({\sf SETH}), Hypothesis 4.1 in \cite{as23}]\label{def:seth}
For every $\epsilon > 0$ there is a positive integer $k \geq 3$ such that $k$-$\mathsf{SAT}$ on formulas with $n$ variables cannot be solved in $O(2^{(1-\epsilon )n})$ time, even by a randomized algorithm.
\end{definition}

\begin{theorem}[Lower bound, Theorem 4.6 of \cite{as23}]\label{thm:formal_main_lower_bound}
Assuming {\sf SETH}, for every sufficiently small $q >0$, there are constants  
 $C > 0$ and $C_{\alpha}>0$ and $C_{\beta} > 1$ such that Approximate Attention Computation (Definition~\ref{def:approximate_attention_computation}) for parameters $d = O(\log n)$, $B = C_{\beta} \sqrt{\log n}$, and $\epsilon_a = n^{- C_{\alpha}}$ requires $\Omega(n^{2-q})$ time.
\end{theorem}

\subsection{Background: Sketching and Polynomial Attention}
\label{sub:multi_basis_sketching:background}

Below, we define a version of polynomial attention from~\cite{kmz24}.

\begin{definition}[Polynomial Attention \citep{kmz24}]\label{def:attention_poly_unit}
Let $g: \R \to \R$ be defined as $g(z) = z^{\beta}$ for $\beta \geq 2$, where $g(W)_{i, j} = g(W_{i, j})$ if $W$ is a matrix and $g(x)_{i} = g(x_{i})$ if $x$ is a vector. Given the input sequence $X \in \R^{n \times d}$ and the query, key, and value weights $W_Q, W_K, W_V \in \R^{d \times d}$, the polynomial attention computation is defined as:
\begin{align*}
    D^{-1} A V,
\end{align*}
where $A = g(X W_Q W_K^\top X^\top / \sqrt{d})$ and $D:= \diag( A {\bf 1}_n )$. 
\end{definition}

As demonstrated in Theorem 1.1 of \cite{kmz24}, one can construct a randomized feature map $\phi'$ for the degree-$p$ polynomial kernel that ensures the resulting approximate attention weights remain non-negative, satisfy provable error bounds, and can be computed in time proportional to the sequence length $n$.

\begin{theorem}[Theorem 1.1 in \cite{kmz24}]\label{thm:poly_attention}
Let $p \geq 2$ be an even integer, $\epsilon \in \paren{ 0, 0.5}$ be an error parameter. Let $d$ be the dimension of the vectors to be mapped. There exists a randomized feature mapping 
\[
\phi': \mathbb{R}^d \to \mathbb{R}^{z^2}, \quad \text{for } z = \Theta\paren{ \frac{p}{\epsilon^2} \log \frac{1}{\delta}}
\]
defined as $\phi'(x) := \paren{Sx^{\otimes \paren{p / 2}}}^{\otimes 2}$ where $S \in \R^{z \times d^{p / 2}}$ is a sketching matrix, such that for all set of vectors $\left \{q_i \in \mathbb{R}^d\right \}_{i \in [n]}$, $\left \{k_j \in \mathbb{R}^d\right \}_{j \in [n]}$, the following hold with probability at least $1 - \delta$:
\begin{enumerate}
    \item $\left \langle \phi'\paren{ q_i}, \phi'\paren{ k_j} \right \rangle \geq 0$ for all $i, j \in [n]$;
    \item 
    \[
    \sum_{i,j} | \left \langle \phi'\paren{ q_i}, \phi'\paren{ k_j} \right \rangle - \left \langle q_i, k_j \right \rangle^p |^2 
    \leq \epsilon^2 \sum_{i,j} \left\|q_i\right \|_2^{2p} \left\|k_j\right \|_2^{2p};
    \]
    \item Computing $\phi'\paren{ x}$ for $x \in \mathbb{R}^d$ requires:
    \begin{itemize}
        \item $\frac{p}{2}$ matrix-vector multiplications with matrices of size $d \times z$,
        \item $\frac{p}{2} - 2$ matrix-vector multiplications with matrices of size $z \times z$,
        \item $\frac{p}{2} - 1$ Hadamard products of $z$-dimensional vectors,
        \item and $1$ self-Kronecker product of an $z$-dimensional vector.
    \end{itemize}
\end{enumerate}
\end{theorem}

To analyze the concentration and tail behavior of random variables—particularly in the context of sketching and randomized linear algebra—it is useful to work with their moment norms. The following definition introduces the $L_t$ norm of a real-valued random variable, which captures its $t$-th moment in a normalized form. These norms play a key role in bounding approximation errors and analyzing stability under random projections. An important property of $L_t$ norms is that they satisfy the triangle inequality, as guaranteed by the Minkowski inequality \citep{akk+20}.

\begin{definition}[Definition 4.1 in \cite{akk+20}]
For every integer \( t \geq 1 \) and any random variable \( X \in \mathbb{R} \), we write
\[
\|X\|_{L_t} = \left( \mathbb{E} \left[ |X|^t \right] \right)^{1/t}.
\]
\textit{Note that} \( \|X + Y\|_{L_t} \leq \|X\|_{L_t} + \|Y\|_{L_t} \) \textit{for all random variables \( X, Y \) by the Minkowski Inequality}.
\end{definition}

Below, we present the JL moment property: it provides a probabilistic guarantee on how well a random matrix preserves the norm of any fixed unit vector—not just with high probability, but in expectation and in $L_t$ norm. This property strengthens the classical JL lemma by quantifying how tightly concentrated the squared norm $\|Sx\|_2^2$ is around its mean.

\begin{definition}[JL Moment Property, Definition 4.2 in \cite{akk+20}]\label{def:jl_moment}
For every positive integer $t$ and every $\delta, \epsilon \geq 0$, we say a distribution over random matrices $S \in \mathbb{R}^{m \times d}$ has the $(\epsilon, \delta, t)$-\textit{JL-moment property}, when
\[
\left\| \|Sx\|_2^2 - 1 \right\|_{L_t} \leq \epsilon \delta^{1/t} \quad \text{and} \quad \mathbb{E}\left[\|Sx\|_2^2\right] = 1
\]
for all $x \in \mathbb{R}^d$ such that $\|x\| = 1$.
\end{definition}

By \cite{akk+20}, the sketching matrix satisfying the JL Moment Property has the following property:

\begin{lemma}[Two vector JL Moment Property, Lemma 4.1 in \cite{akk+20}]\label{lem:jl_moment}
For all $x, y \in \mathbb{R}^d$, if $S$ has the $(\epsilon, \delta, t)$-\textit{JL Moment Property}, then
\[
\left\| (Sx)^\top (Sy) - x^\top y \right\|_{L_t} \leq \epsilon \delta^{1/t} \|x\|_2 \|y\|_2.
\]
\end{lemma}

\ifdefined\isarxiv

\section{(Batch) Gaussian Kernel Density Estimation via Single Threshold Support Basis}

\else
\section{(BATCH) GAUSSIAN KERNEL DENSITY ESTIMATION VIA SINGLE THRESHOLD SUPPORT BASIS}

\fi

\label{sec:single_basis}

In Appendix~\ref{sub:single_basis:disjoint}, we present the definition of disjoint matrices and analyze their mathematical properties. In Appendix~\ref{sub:single_basis:supp}, we give the formal definition of support basis and construct a single threshold support basis for $QK^\top$. In Appendix~\ref{sub:single_basis:bounded}, we generalize the bounded entry lemma \citep{as23} from the threshold $B = o\paren{\sqrt{\log n}}$ to any arbitrary threshold being a positive real number. In Appendix~\ref{sub:single_basis:sparsity}, we show that if the number of large entries in $Q$ and $K$ is small, then we can compute certain matrix multiplications in $A^{\paren{L}}$ sparsity time. In Appendix~\ref{sub:single_basis:batchkde}, we use our constructed single threshold support basis to design an $A^{\paren{L}}$ sparsity time algorithm to approximate the (batch) Gaussian kernel density estimation $AV$.  

\subsection{Disjoint Matrices and Their Properties}
\label{sub:single_basis:disjoint}

In order to decompose the $\paren{Q, K}$-softmax-attention matrix into simpler components, we begin by introducing the notion of disjoint matrices. This disjointness condition ensures that matrix components do not overlap in their non-zero entries, which is a crucial property that enables additive decompositions of matrix exponentials. We formalize this idea in the following definition.

\begin{definition}[Disjoint matrices]\label{def:disjoint}
    Let $\left \{A_k\right \}_{k \in [l]}$ be a set of $n$ by $n$ matrices, where $l \geq 2$ is an arbitrary positive integer. We say $A_k$'s are disjoint matrices if for all $i, j \in [n]$, for all $k \in [l]$, for all $\ov{k} \in [l] \setminus \left \{k\right \}$, if $\paren{A_k}_{i, j} \neq 0$, then $\paren{A_{\ov{k}}}_{i, j} = 0$.
\end{definition}

Now, we show how the exponential of a sum of disjoint matrices can be simplified. Since disjoint matrices do not share any overlapping non-zero entries, the exponential of their sum equals the sum of their exponentials minus a constant matrix.

\begin{fact}\label{fac:exp_split}
    Let $B, C \in \R^{n \times n}$ be disjoint matrices (see Definition~\ref{def:disjoint}). 

    Then, we have
    \begin{align*}
        \exp\paren{B + C} = \exp\paren{B} + \exp\paren{C} - {\bf 1}_{n \times n}.
    \end{align*}
\end{fact}
\begin{proof}
    Without loss of generality, suppose $C_{i, j} \neq 0$ and $B_{i, j} = 0$.
    Then, for all $i, j \in [n]$, we have
    \begin{align*}
        \exp\paren{B + C}_{i, j} 
        = & ~ \exp\paren{B_{i, j} + C_{i, j}}\\
        = & ~ \exp\paren{C}_{i, j} + 1 - 1\\
        = & ~ \exp\paren{C}_{i, j} + \exp\paren{B}_{i, j} - \paren{{\bf 1}_{n \times n}}_{i, j},
    \end{align*}
    where the first step follows from the fact that $\exp$ is applied entry-wisely, the second step follows from $B_{i, j} = 0$, and the last step follows from $\exp\paren{0} = 1$.
\end{proof}

\subsection{Support Basis}
\label{sub:single_basis:supp}

To construct a support basis for the matrix $QK^\top$, we must first define how to partition the matrix into ``large'' and ``small'' components based on the entries in $Q$ and $K$. 

\begin{definition}\label{def:Q_K} 
    Let $T > 0$ denote a threshold, $\alpha \in (0, 1)$, and $C > 1$ denote a fixed constant.  
    Let $Q, K \in \R^{n \times d}$ be the query and key matrices, respectively. We decompose these matrices as $Q = Q^{\paren{L}} + Q^{\paren{s}}$ and $K = K^{\paren{L}} + K^{\paren{s}}$, where:  
    \begin{itemize}
        \item $\left |\mathrm{supp}\paren{Q^{\paren{L}}} \right |, \left |\mathrm{supp}\paren{K^{\paren{L}}}\right | \leq C n^{\alpha}$, and for all $i \in [n]$ and $j \in [d]$, if $|Q_{i,j}| > T$, then $Q^{\paren{L}}_{i,j} := Q_{i,j}$; otherwise, $Q^{\paren{L}}_{i,j} := 0$. We define $K^{\paren{L}}$ in a similar manner.
        \item For all $i \in [n]$ and $j \in [d]$, if $|Q_{i,j}| \leq T$, then $Q^{\paren{s}}_{i,j} := Q_{i,j}$; otherwise, $Q^{\paren{s}}_{i,j} := 0$. We define $K^{\paren{s}}$ in a similar manner.
    \end{itemize}
\end{definition}

We define $A^{(L)}$ and $A^{(s)}$ as follows: 
\begin{definition}\label{def:split_Q_K}
    Given the query and the key matrices $Q, K \in \R^{n \times d}$, we let $Q^{\paren{L}}, Q^{\paren{s}}, K^{\paren{L}}, K^{\paren{s}} \in \R^{n \times d}$ be defined as in Definition~\ref{def:Q_K}. Let $i, j \in [n]$. 
    We define matrices $A^{\paren{L}}$ and $A^{\paren{s}}$ as follows: 
    \begin{align*}
        A^{\paren{L}}_{i, j} := \begin{cases}
            \paren{Q K^\top}_{i, j} &\quad\quad \mathrm{if}~\paren{Q^{\paren{L}}\paren{K^{\paren{L}}}^\top + Q^{\paren{s}}\paren{K^{\paren{L}}}^\top + Q^{\paren{L}}\paren{K^{\paren{s}}}^\top}_{i,j} \not = 0 \\0 &\quad \quad  \mathrm{otherwise.}
        \end{cases}
    \end{align*}
    \begin{align*}
        A^{\paren{s}}_{i, j} := \begin{cases}
            0 & \quad\quad \mathrm{if}~\paren{Q^{\paren{L}}\paren{K^{\paren{L}}}^\top + Q^{\paren{s}}\paren{K^{\paren{L}}}^\top + Q^{\paren{L}}\paren{K^{\paren{s}}}^\top}_{i,j} \not = 0 \\
            \paren{Q^{\paren{s}}\paren{K^{\paren{s}}}^\top}_{i, j} &\quad\quad \mathrm{otherwise.}
        \end{cases}
    \end{align*}
\end{definition}

Now, we provide the formal definition of the support basis. For a given matrix $A$, it requires all matrices in this basis to be disjoint (see Definition~\ref{def:disjoint}) to ensure the additive decomposition (see Fact~\ref{fac:exp_split}) and the sum of all matrices in this basis to be equal to the given matrix $A$.

\begin{definition}[Support basis]\label{def:support_basis}
Let $\left \{A_k\right \}_{k \in [l]}$ be a set of $n$ by $n$ matrices, where $l \geq 2$ is an arbitrary positive integer. 
    Given a matrix $A \in \R^{n \times n}$, we say that $\left \{A_k\right \}_{k \in [l]}$ is a support basis of $A$ if
    \begin{enumerate}
        \item $\sum_{k \in [l]} A_k = A$ and
        \item $A_k$'s are disjoint matrices (see Definition~\ref{def:disjoint}).
    \end{enumerate}
\end{definition}

Below, we formally prove that the resulting matrices $A^{(s)}$ and $A^{(L)}$ (as defined in Definition~\ref{def:split_Q_K}) are disjoint and together sum to $QK^\top$, forming a valid support basis of $QK^\top$.

\begin{lemma}\label{lem:query_key_split}
    Given the query and the key matrices $Q, K \in \R^{n \times d}$, we let $Q^{\paren{L}}, Q^{\paren{s}}, K^{\paren{L}}, K^{\paren{s}} \in \R^{n \times d}$ be defined as in Definition~\ref{def:Q_K} and $A^{\paren{L}}, A^{\paren{s}} \in \R^{n \times n}$ be defined as in Definition~\ref{def:split_Q_K}.

    Then, $\left \{A^{\paren{L}}, A^{\paren{s}}\right \}$ is a support basis of the matrix $QK^\top \in \R^{n \times n}$.
\end{lemma}

\begin{proof}

    We first show the first condition of the support basis, namely $A^{\paren{s}} + A^{\paren{L}} = QK^\top$.

By definition~\ref{def:split_Q_K}, we can see that for all arbitrary $i, j \in [n]$, if 
\begin{align*}
    \paren{Q^{\paren{L}}\paren{K^{\paren{L}}}^\top + Q^{\paren{s}}\paren{K^{\paren{L}}}^\top + Q^{\paren{L}}\paren{K^{\paren{s}}}^\top}_{i,j}
\end{align*}
is not zero then:
\begin{align*}
   A^{\paren{s}}_{i, j} + A^{\paren{L}}_{i, j}  = \paren{Q K^\top}_{i, j} + 0 = \paren{QK^\top}_{i,j}
   .
\end{align*}
On the other hand, if $\paren{Q^{\paren{L}}\paren{K^{\paren{L}}}^\top + Q^{\paren{s}}\paren{K^{\paren{L}}}^\top + Q^{\paren{L}}\paren{K^{\paren{s}}}^\top}_{i,j}$ is zero then we have: 
\begin{align*}
    A^{\paren{s}}_{i, j} + A^{\paren{L}}_{i, j} 
    = & ~ 0 + \paren{Q^{\paren{s}}\paren{K^{\paren{s}}}^\top}_{i, j} \\
    = & ~ \paren{QK^\top}_{i,j} - \paren{Q^{\paren{L}}\paren{K^{\paren{L}}}^\top + Q^{\paren{s}}\paren{K^{\paren{L}}}^\top + Q^{\paren{L}}\paren{K^{\paren{s}}}^\top}_{i,j}
    \\ 
    = & ~ \paren{QK^\top}_{ij} + 0 \\
    = & ~ \paren{QK^\top}_{ij},
\end{align*}
where in the second equality, we use $ QK^\top
        = \paren{Q^{\paren{L}} + Q^{\paren{s}}}\paren{K^{\paren{L}} + K^{\paren{s}}}^\top$.

Now, we show the second condition of the support basis that $A^{\paren{s}}$ and $A^{\paren{L}}$ are disjoint matrices.

This follows directly from Definition~\ref{def:split_Q_K}: if $\paren{Q^{\paren{L}}\paren{K^{\paren{L}}}^\top + Q^{\paren{s}}\paren{K^{\paren{L}}}^\top + Q^{\paren{L}}\paren{K^{\paren{s}}}^\top}_{i,j} \neq 0$, then $A^{\paren{L}}_{i, j} = \paren{QK^\top}_{i, j}$ and $A^{\paren{s}}_{i, j} = 0$; otherwise $A^{\paren{L}}_{i, j} = 0$ and $A^{\paren{s}}_{i, j} = \paren{Q^{\paren{s}}\paren{K^{\paren{s}}}^\top}_{i, j}$.
\end{proof}

Now, we provide a concrete example of the support basis to better illustrate this decomposition.

\begin{example}[The support basis $\left \{A^{\paren{L}}, A^{\paren{s}}\right \}$ of $QK^\top$]

Let $Q = \begin{bmatrix}
a & b \\
c & d \\
e & f
\end{bmatrix} \in \R^{3 \times 2}$ and $K^\top = \begin{bmatrix}
g & i & k \\
h & j & l
\end{bmatrix} \in \R^{2 \times 3}$. With a given threshold $T > 0$, we have $c, k > T$, and all other entries of $Q$ and $K$ are smaller than $T$. Therefore, by Definition~\ref{def:Q_K}, we have
\begin{align*}
    \underbrace{\begin{bmatrix}
0 & 0 \\
c & 0 \\
0 & 0
\end{bmatrix}}_{Q^{\paren{L}}} + \underbrace{\begin{bmatrix}
a & b \\
0 & d \\
e & f
\end{bmatrix}}_{Q^{\paren{s}}} = \underbrace{\begin{bmatrix}
a & b \\
c & d \\
e & f
\end{bmatrix}}_{Q} ~~\text{and}~~ 
\underbrace{\begin{bmatrix}
0 & 0 & k \\
0 & 0 & 0
\end{bmatrix}}_{\paren{K^{\paren{L}}}^\top} + \underbrace{\begin{bmatrix}
g & i & 0 \\
h & j & l
\end{bmatrix}}_{\paren{K^{\paren{s}}}^\top} = \underbrace{\begin{bmatrix}
g & i & k \\
h & j & l
\end{bmatrix}}_{K^\top}
\end{align*}

This setup allows us to decompose the matrix product $QK^\top$ into a sum of disjoint components:
    \begin{align*}
    & ~ \begin{bmatrix}
a & b \\
c & d \\
e & f
\end{bmatrix}
\begin{bmatrix}
g & i & k \\
h & j & l
\end{bmatrix}\\
= & ~
\begin{bmatrix}
ag + bh & ai + bj & ak + bl \\
cg + dh & ci + dj & ck + dl \\
eg + hf & ei + fj & ek + fl
\end{bmatrix}\\
= & ~
\underbrace{\begin{bmatrix}
0 & 0 & ak + bl \\
cg + dh & ci + dj & ck + dl \\
0 & 0 & ek + fl
\end{bmatrix}}_{A^{\paren{L}}}
+
\underbrace{\begin{bmatrix}
ag + bh & ai + bj & 0 \\
0 & 0 & 0 \\
eg + hf & ei + fj & 0
\end{bmatrix}}_{A^{\paren{s}}}\\
= & ~
\underbrace{\begin{bmatrix}
0 & 0 & 0 \\
0 & 0 & ck \\
0 & 0 & 0
\end{bmatrix}}_{Q^{\paren{L}}\paren{K^{\paren{L}}}^\top}
+
\underbrace{\begin{bmatrix}
0 & 0 & 0 \\
cg & ci & 0 \\
0 & 0 & 0
\end{bmatrix}}_{Q^{\paren{L}}\paren{K^{\paren{s}}}^\top}
+
\underbrace{\begin{bmatrix}
0 & 0 & ak \\
0 & 0 & 0 \\
0 & 0 & ek
\end{bmatrix}}_{Q^{\paren{s}}\paren{K^{\paren{L}}}^\top}
+
\underbrace{\begin{bmatrix}
ag + bh & ai + bj & bl \\
dh & dj & dl \\
eg + hf & ei + fj & fl
\end{bmatrix}}_{Q^{\paren{s}}\paren{K^{\paren{s}}}^\top}\\
= & ~
\underbrace{\begin{bmatrix}
0 & 0 \\
c & 0 \\
0 & 0
\end{bmatrix}}_{Q^{\paren{L}}}
\underbrace{\begin{bmatrix}
0 & 0 & k \\
0 & 0 & 0
\end{bmatrix}}_{\paren{K^{\paren{L}}}^\top}
+
\underbrace{\begin{bmatrix}
0 & 0 \\
c & 0 \\
0 & 0
\end{bmatrix}}_{Q^{\paren{L}}}
\underbrace{\begin{bmatrix}
g & i & 0 \\
h & j & l
\end{bmatrix}}_{\paren{K^{\paren{s}}}^\top}
+
\underbrace{\begin{bmatrix}
a & b \\
0 & d \\
e & f
\end{bmatrix}}_{Q^{\paren{s}}}
\underbrace{\begin{bmatrix}
0 & 0 & k \\
0 & 0 & 0
\end{bmatrix}}_{\paren{K^{\paren{L}}}^\top}
+
\underbrace{\begin{bmatrix}
a & b \\
0 & d \\
e & f
\end{bmatrix}}_{Q^{\paren{s}}}
\underbrace{\begin{bmatrix}
g & i & 0 \\
h & j & l
\end{bmatrix}}_{\paren{K^{\paren{s}}}^\top}.
\end{align*}
This example illustrates how the product $QK^\top$ can be partitioned into disjoint matrix components, forming a support basis that enables structured approximation.
\end{example}

\subsection{Bounded Entry}
\label{sub:single_basis:bounded}

We now establish an upper bound on the entries of the matrix $A^{(s)}$, which corresponds to the contribution from the small entries of $Q$ and $K$. This allows us to apply polynomial approximation techniques with controlled error. We generalize the bounded-entry result from \cite{as23} to arbitrary thresholds.

\begin{lemma}[Bounded entry, an improved version of Lemma 3.3 in \cite{as23}]\label{lem:bounded_entry}
Given $Q, K \in \R^{n \times d}$, we let $Q^{\paren{s}}, K^{\paren{s}}$ be defined as in Definition~\ref{def:Q_K}. Let $A^{\paren{s}} \in \mathbb{R}^{n \times n}$ be defined as in Definition~\ref{def:split_Q_K}, where 
\begin{align*}
        A^{\paren{s}}_{i, j} := \begin{cases}
            0 & \quad\quad \mathrm{if}~\paren{Q^{\paren{L}}\paren{K^{\paren{L}}}^\top + Q^{\paren{s}}\paren{K^{\paren{L}}}^\top + Q^{\paren{L}}\paren{K^{\paren{s}}}^\top}_{i,j} \not = 0 \\
            \paren{Q^{\paren{s}}\paren{K^{\paren{s}}}^\top}_{i, j} &\quad\quad \mathrm{otherwise.}
        \end{cases}
    \end{align*}

Then, for a given threshold $T > 0$, we have
\begin{align*}
    \left \|A^{\paren{s}} / d \right \|_\infty \leq T^2.
\end{align*}
\end{lemma}

\begin{proof}
We have
\begin{align*}
    \left \|A^{\paren{s}}\right \|_\infty
    = & ~ \max_{\paren{i, j} \in [n] \times [n]} \left |A^{\paren{s}}_{i, j} \right|\\
    = & ~ \max_{\paren{i, j} \in [n] \times [n]} \left |\sum_{l = 1}^d Q^{\paren{s}}_{i, l} K^{\paren{s}}_{j, l}\right|\\
    \leq & ~ \sum_{l = 1}^d \left |\max_{i \in [n]} Q^{\paren{s}}_{i, l} \right| \cdot \left |\max_{j \in [n]} K^{\paren{s}}_{j, l}\right|\\
    \leq & ~ \sum_{l = 1}^d T^2\\
    = & ~ d T^2,
\end{align*}
where the first step follows from the definition of the $\ell_\infty$ norm, the second step follows from the definition of $A^{\paren{s}}$ (see Definition~\ref{def:split_Q_K}), the third step follows from the triangle inequality (see Fact~\ref{fac:vector_norm}), and the fourth step follows from Definition~\ref{def:Q_K}.
\end{proof}

\subsection{Running Time and Sparsity Analysis}
\label{sub:single_basis:sparsity}

Next, we analyze the sparsity and computational cost associated with the matrix $A^{(L)}$, which captures the contribution from the large entries of $Q$ and $K$. Our goal is to exactly compute the part of the attention computation involving $A^{(L)}$ and approximate the part involving $A^{(s)}$. Therefore, it is important to know the time complexity for constructing $A^{(L)}$ and its sparsity.

\begin{lemma}\label{lem:sparsity_AL}
    Let $\alpha \in (0, 1)$. Given the query and the key matrices $Q, K \in \R^{n \times d}$, we let $Q^{\paren{L}}, Q^{\paren{s}}, K^{\paren{L}}, K^{\paren{s}} \in \R^{n \times d}$ be defined as in Definition~\ref{def:Q_K} 
    and $A^{\paren{L}}, A^{\paren{s}} \in \R^{n \times d}$ be defined as in Definition~\ref{def:split_Q_K}.

    Then, we have
    \begin{enumerate}
        \item $\left |\mathrm{supp}\paren{A^{\paren{L}}} \right | = O\paren{n^{1 + \alpha}}$, and
        \item it takes $O\paren{n^{1 + \alpha}d}$ time to compute $A^{\paren{L}}$.
    \end{enumerate}
\end{lemma}
\begin{proof}
    {\bf Proof of Part 1.}

    Note that by Definition~\ref{def:split_Q_K}, we have
    \begin{align*}
        A^{\paren{L}}_{i, j} := \begin{cases}
            \paren{Q K^\top}_{i, j} &\quad\quad \mathrm{if}~\paren{Q^{\paren{L}}\paren{K^{\paren{L}}}^\top + Q^{\paren{s}}\paren{K^{\paren{L}}}^\top + Q^{\paren{L}}\paren{K^{\paren{s}}}^\top}_{i,j} \not = 0 \\0 &\quad \quad  \mathrm{otherwise.}
        \end{cases}
    \end{align*}

    Suppose $\paren{i, j} \notin \mathrm{supp}\paren{Q^{\paren{L}}\paren{K^{\paren{L}}}^\top + Q^{\paren{s}}\paren{K^{\paren{L}}}^\top + Q^{\paren{L}}\paren{K^{\paren{s}}}^\top}$.

    Then, we have
    \begin{align*}
        A^{\paren{L}}_{i, j}
        = & ~ \paren{Q^{\paren{L}}\paren{K^{\paren{L}}}^\top + Q^{\paren{s}}\paren{K^{\paren{L}}}^\top + Q^{\paren{L}}\paren{K^{\paren{s}}}^\top}_{i, j}\\
        = & ~ 0.
    \end{align*}

    Therefore, it suffices to consider the case when 
    \begin{align*}
        \paren{i, j} \in \mathrm{supp}\paren{Q^{\paren{L}}\paren{K^{\paren{L}}}^\top + Q^{\paren{s}}\paren{K^{\paren{L}}}^\top + Q^{\paren{L}}\paren{K^{\paren{s}}}^\top}.
    \end{align*}

    Note that by Definition~\ref{def:Q_K}, we have
    \begin{align}\label{eq:nnz_assumption}
        \left | \mathrm{supp}\paren{Q^{\paren{L}}} \right |, \left | \mathrm{supp}\paren{K^{\paren{L}}} \right | \leq C n^{\alpha},
    \end{align}
    for a fixed constant $C > 1$.

    Therefore, we have
    \begin{align}\label{eq:support_of_QK}
        & ~ \left |\mathrm{supp}\paren{Q^{\paren{L}}\paren{K^{\paren{L}}}^\top + Q^{\paren{s}}\paren{K^{\paren{L}}}^\top + Q^{\paren{L}}\paren{K^{\paren{s}}}^\top} \right|\notag\\
        \leq & ~ \left|\mathrm{supp}\paren{Q^{\paren{L}}\paren{K^{\paren{L}}}^\top} \right| + \left|\mathrm{supp}\paren{Q^{\paren{s}}\paren{K^{\paren{L}}}^\top} \right| + \left|\mathrm{supp}\paren{Q^{\paren{L}}\paren{K^{\paren{s}}}^\top} \right| \notag\\
        \leq & ~ O\paren{n^{2 \alpha}} + \left|\mathrm{supp}\paren{Q^{\paren{s}}\paren{K^{\paren{L}}}^\top} \right| + \left|\mathrm{supp}\paren{Q^{\paren{L}}\paren{K^{\paren{s}}}^\top} \right| \notag\\
        \leq & ~ O\paren{n^{2 \alpha}} + O\paren{n^{1 + \alpha}} + O\paren{n^{1 + \alpha}} \notag\\
        = & ~ O\paren{n^{1 + \alpha}},
    \end{align}
    where the first step follows from Fact~\ref{fac:support_inequality}, the second step follows from Eq.~\eqref{eq:nnz_assumption}, the third step follows from Eq.~\eqref{eq:nnz_assumption}, and the last step follows from $\alpha \in (0, 1)$.

    {\bf Proof of Part 2.}

    The proof of this part is similar to that of {\bf Part 1}.

    It also suffices to only consider the case when 
    \begin{align*}
        \paren{i, j} \in \mathrm{supp}\paren{Q^{\paren{L}}\paren{K^{\paren{L}}}^\top + Q^{\paren{s}}\paren{K^{\paren{L}}}^\top + Q^{\paren{L}}\paren{K^{\paren{s}}}^\top}.
    \end{align*}

    Because of Eq.~\eqref{eq:nnz_assumption}, we have that 
    \begin{itemize}
        \item computing $Q^{\paren{L}}\paren{K^{\paren{L}}}^\top$ takes $O\paren{n^{2 \alpha}}$ time,
        \item computing $Q^{\paren{L}}\paren{K^{\paren{s}}}^\top$ takes $O\paren{n^{1 + \alpha}}$ time, and
        \item computing $Q^{\paren{s}}\paren{K^{\paren{L}}}^\top$ takes $O\paren{n^{1 + \alpha}}$ time.
    \end{itemize}

    Moreover, because of Eq.~\eqref{eq:support_of_QK}, we need to compute $O\paren{n^{1 + \alpha}}$ numbers of entries of $Q^{\paren{s}}\paren{K^{\paren{s}}}^\top$. 
    
    Since $Q^{\paren{s}}, K^{\paren{s}} \in \R^{n \times d}$, for each $i, j \in [n]$, we have
    \begin{align*}
       \paren{Q^{\paren{s}} \paren{K^{\paren{s}}}^\top}_{i, j} = \sum_{k \in [d]} \paren{Q^{\paren{s}}}_{i, k} \paren{K^{\paren{s}}}^\top_{k, j},
    \end{align*}
    which takes $O\paren{d}$ time.

    Therefore, in total, it takes $O\paren{n^{1 + \alpha}d}$ times to compute $\paren{Q^{\paren{s}}\paren{K^{\paren{s}}}^\top}_{i, j}$ for all 
    \begin{align*}
        \paren{i, j} \in \mathrm{supp}\paren{Q^{\paren{L}}\paren{K^{\paren{L}}}^\top + Q^{\paren{s}}\paren{K^{\paren{L}}}^\top + Q^{\paren{L}}\paren{K^{\paren{s}}}^\top}.
    \end{align*}

    Therefore, the time complexity of computing
    \begin{align*}
        \paren{QK^\top}_{i, j} = \paren{Q^{\paren{L}}\paren{K^{\paren{L}}}^\top + Q^{\paren{s}}\paren{K^{\paren{L}}}^\top + Q^{\paren{L}}\paren{K^{\paren{s}}}^\top + Q^{\paren{s}}\paren{K^{\paren{s}}}^\top}_{i, j}
    \end{align*}
    for all $\paren{i, j} \in \mathrm{supp}\paren{Q^{\paren{L}}\paren{K^{\paren{L}}}^\top + Q^{\paren{s}}\paren{K^{\paren{L}}}^\top + Q^{\paren{L}}\paren{K^{\paren{s}}}^\top}$ is
    \begin{align*}
        O\paren{n^{2\alpha}} + O\paren{n^{1 + \alpha}} + O\paren{n^{1 + \alpha}} + O\paren{n^{1 + \alpha}d} = O\paren{n^{1 + \alpha} d}.
    \end{align*}
\end{proof}

\subsection{(Batch) Gaussian Kernel Density Estimation}
\label{sub:single_basis:batchkde}

We now formalize the task of computing $AV$, where $A = \exp(QK^\top / d)$ is the $\paren{Q, K}$-softmax-attention matrix. This operation is closely related to a classical technique in statistics known as \emph{Gaussian Kernel Density Estimation (KDE)}. In Gaussian KDE, the goal is to estimate a density function by averaging Gaussian kernels centered at given data points. Given data points $\{x_1, \dots, x_n\} \subset \mathbb{R}^d$ and a query point $q \in \mathbb{R}^d$, the classical Gaussian KDE estimate is:
\begin{align*}
    \wh{f}(q) = \frac{1}{n} \sum_{j=1}^n \exp\left(-\frac{\|q - x_j\|_2^2}{2h^2} \right),
\end{align*}
where the Gaussian kernel can be expressed as:
\begin{align*}
    \exp\left(-\frac{\|q - k\|_2^2}{2h^2} \right)
    = & ~ \exp\left(-\frac{\|q\|_2^2 + \|k\|_2^2 - 2 \langle q, k \rangle}{2h^2} \right),
\end{align*}
and the attention computation problem focus primarily on
\begin{align*}
    \exp\left(\frac{\langle q, k \rangle}{h^2} \right).
\end{align*}

The matrix product $AV$ can therefore be interpreted as a \emph{batch} version of Gaussian KDE, where each row of $AV$ aggregates value vectors $V_{*, k}$ weighted by the similarity between query $Q_{i, *}$ and key $K_{j, *}$. This motivates the name \emph{(Batch) Gaussian Kernel Density Estimation}, as it generalizes Gaussian KDE to the matrix setting. We now formally define the problem.

\begin{definition}[(Batch) Gaussian kernel density estimation]\label{def:gaussian_kde}
    Given the query $Q \in \R^{n \times d}$, key $K \in \R^{n \times d}$, and value $V \in \R^{n \times d}$, we define the $\paren{Q, K}$-softmax-attention matrix as $A = \exp\paren{QK^\top / d}$. The goal of the (batch) Gaussian kernel density estimation is to output $\mathsf S \in \R^{n \times d}$ satisfying, for all $\epsilon > 0$,
    \begin{align*}
        \left \|\mathsf S - A V\right \|_\infty < \epsilon.
    \end{align*}
\end{definition}

Having defined the (Batch) Gaussian Kernel Density Estimation problem, we now show how our algorithm can approximate $AV$ efficiently by leveraging the support basis decomposition. Specifically, we use polynomial approximation for the component $A^{(s)}$, which contains only small entries and is thus suitable for Chebyshev expansion, and we compute the remaining part $A^{(L)}$ explicitly, exploiting its sparsity. The following lemma guarantees the correctness and efficiency of this approach by bounding the approximation error and analyzing the overall runtime.

\begin{algorithm}[!ht]\caption{In this algorithm, we approximate the (batch) Gaussian kernel density estimation: given the query, key, and value matrices $Q \in \R^{n \times d}, K \in \R^{n \times d}, V \in \R^{n \times d}$, the goal is to output $\wt AV$, where $\wt A \in \R^{n \times n}$ is the $\paren{\epsilon, r}$-approximation of the $\paren{Q, K}$-softmax-attention matrix $A$.}\label{alg:gaussian_kde}
\begin{algorithmic}[1]

\Procedure{\textsc{GaussianKDE}}{$Q \in \R^{n \times d}, K \in \R^{n \times d}, V \in \R^{n \times d}, n, d, \epsilon, T = o\paren{\sqrt{\log n}}$}
    \State $Q^{\paren{L}}, Q^{\paren{s}}, K^{\paren{L}}, K^{\paren{s}} \in \R^{n \times d} \gets \textsc{Split}(Q, T), \textsc{Split}(K, T)$ \label{line:split} 
    \State Explicitly compute $A^{\paren{L}}$. \label{line:compute_AL} 
    \State $U_1, U_2 \in \R^{n \times r} \gets \textsc{Polynomial}(Q^{\paren{s}}, K^{\paren{s}}, \epsilon_0)$.\Comment{To approximate $\exp\paren{A^{\paren{s}} / d}$.} \label{line:polymethod} 
    \State Get $C_1 \gets U_1 \paren{U_2^\top V} \in \R^{n \times d}$. \label{line:compute_C1}
    \State Get $C_2 \gets \paren{\exp\paren{A^{\paren{L}} / d} - {\bf 1}_{n \times n}} V \in \R^{n \times d}$. \label{line:compute_C2}
    \State \Return $C_1 + C_2 \in \R^{n \times d}$. \label{line:GaussianKDE_output}
\EndProcedure
\end{algorithmic}
\end{algorithm}

\begin{lemma}\label{lem:gaussian_kde}
    Given the query $Q \in \R^{n \times d}$, key $K \in \R^{n \times d}$, and value $V \in \R^{n \times d}$, we define the $\paren{Q, K}$-softmax-attention matrix as $A = \exp\paren{QK^\top / d}$ and $Q^{\paren{L}}, Q^{\paren{s}}, K^{\paren{L}}, K^{\paren{s}} \in \R^{n \times d}$ as in Definition~\ref{def:Q_K}. Suppose by Lemma~\ref{lem:wt_A_small_rank}, there exists $U_1, U_2 \in \R^{n \times r}$ such that $U_1 U_2^\top$ is the $\paren{\epsilon_0, r}$-approximation of $\exp\paren{Q^{\paren{s}}\paren{K^{\paren{s}}}^\top / d}$ for all arbitrary $\epsilon_0 \in \paren{0, 0.1}$.

    Then, we can solve the (Batch) Gaussian kernel density estimation (Definition~\ref{def:gaussian_kde}) by outputting $\mathsf{S} \in \R^{n \times d}$ (Algorithm~\ref{alg:gaussian_kde}) satisfying
    \begin{align*}
        \left \|\mathsf{S} - A V\right \|_\infty < n \epsilon_0 \left \|V \right \|_\infty
    \end{align*}
    in $O\paren{n^{1 + \alpha} d}$ time.
\end{lemma}
\begin{proof}

{\bf Proof of correctness.}

    We have
    \begin{align}\label{eq:A}
        AV
        = & ~ \exp\paren{QK^\top / d}V \notag\\
        = & ~ \exp\paren{\paren{A^{\paren{s}} + A^{\paren{L}}}  / d}V \notag\\
        = & ~ \exp\paren{A^{\paren{s}} / d}V + \exp\paren{A^{\paren{L}} / d}V - {\bf 1}_{n \times n}V,
    \end{align}
where the first step follows from the definition of the $\paren{Q, K}$-softmax-attention matrix, the second step follows from Lemma~\ref{lem:query_key_split}, and the third step follows from Fact~\ref{fac:exp_split}.

By the Lemma~\ref{lem:bounded_entry}, we have
\begin{align*}
    \left \|A^{\paren{s}} / d\right \|_\infty \leq T^2,
\end{align*}
for some threshold $T > 0$.

By setting the threshold $T$ to be equal to $B = o\paren{\sqrt{\log n}}$, we can see that this satisfies the assumption of using Lemma~\ref{lem:wt_A_small_rank} and Algorithm~\ref{alg:main_as23}.
Therefore, using Lemma~\ref{lem:wt_A_small_rank}, there exists $U_1, U_2 \in \R^{n \times r}$ such that $U_1 U_2^\top$ is the $\paren{\epsilon_0, r}$-approximation of $\exp\paren{Q^{\paren{s}}\paren{K^{\paren{s}}}^\top / d}$, satisfying
    \begin{align}\label{eq:low_rank_A2}
        \left \|\exp\paren{A^{\paren{s}} / d} - U_1U_2^\top\right \|_\infty < \epsilon_0,
    \end{align}
    for all arbitrary $\epsilon_0 \in \paren{0, 0.1}$.

In Algorithm~\ref{alg:gaussian_kde}, we output
\begin{align}\label{eq:wt_A_V}
    \mathsf{S}
    = & ~ C_1 + C_2 \notag\\
    = & ~ U_1 \paren{U_2^\top V} + \paren{\exp\paren{A^{\paren{L}} / d} - {\bf 1}_{n \times n}} V \notag\\
    = & ~ U_1 \paren{U_2^\top V} + \exp\paren{A^{\paren{L}} / d} V - {\bf 1}_{n \times n} V,
\end{align}
where the first step follows from the output (Line~\ref{line:GaussianKDE_output}) of Algorithm~\ref{alg:gaussian_kde} and the second step follows from updates $C_1 \gets U_1 \paren{U_2^\top V} \in \R^{n \times d}$ (Line~\ref{line:compute_C1}) and $C_2 \gets \paren{\exp\paren{A^{\paren{L}} / d} - {\bf 1}_{n \times n}} V \in \R^{n \times d}$ (Line~\ref{line:compute_C2}).

Therefore, combining Eq.~\eqref{eq:A} and Eq.~\eqref{eq:wt_A_V}, we have
\begin{align*}
    \left \|\mathsf{S} - AV\right \|_\infty
    = & ~ \left \|U_1 U_2^\top V - \exp\paren{A^{\paren{s}} / d}V \right \|_\infty\\
    = & ~ \left \|\paren{U_1 U_2^\top - \exp\paren{A^{\paren{s}} / d}} V \right \|_\infty\\
    \leq & ~ n \left \|U_1 U_2^\top - \exp\paren{A^{\paren{s}} / d}\right \|_\infty  \left \|V \right \|_\infty\\
    \leq & ~ n \epsilon_0 \left \|V \right \|_\infty,
\end{align*}
where the first step follows from combining Eq.~\eqref{eq:wt_A_V} and Eq.~\eqref{eq:A}, the second step follows from the distributive law, the third step follows from the definition of the $\ell_\infty$ norm, and the fourth step follows from Eq.~\eqref{eq:low_rank_A2}.

{\bf Proof of the running time.}

In Line~\ref{line:split}, we need to check each entry of $Q, K \in \R^{n \times d}$ finding the ones greater than the threshold $T$, which takes $O(nd)$ time.

In Line~\ref{line:compute_AL}, we compute $A^{\paren{L}} \in \R^{n \times n}$. By {\bf Part 2} of Lemma~\ref{lem:sparsity_AL}, we know that it takes $O\paren{n^{1 + \alpha}d}$ time to compute $A^{\paren{L}}$.

In Line~\ref{line:polymethod}, we construct $U_1, U_2 \in \R^{n \times r}$, which by Lemma~\ref{lem:wt_A_small_rank} takes $O(nr)$ time.

In Line~\ref{line:compute_C1}, it takes $O(nrd)$ time to compute $U_2^\top V \in \R^{r \times d}$ and takes $O(nrd)$ time to compute $U_1 \paren{U_2^\top V} \in \R^{n \times d}$.

In Line~\ref{line:compute_C2}, it takes $O(n^{1 + \alpha}d)$ time to compute $\paren{\exp\paren{A^{\paren{L}} / d} - {\bf 1}_{n \times n}} V$ as by {\bf Part 1} of Lemma~\ref{lem:sparsity_AL}, we know that $\left | \mathrm{supp}\paren{A^{\paren{L}}} \right | = O\paren{n^{1 + \alpha}}$ which implies $\left |\mathrm{supp}\paren{\exp\paren{A^{\paren{L}} / d} - {\bf 1}_{n \times n}} \right | = O\paren{n^{1 + \alpha}}$.

In total, it takes $O(n^{1 + \alpha}d + nrd)$ time. By Lemma~\ref{lem:wt_A_small_rank}, we know that $O(nr) = n^{1 + o(1)}$. Therefore, we have
\begin{align*}
    O(n^{1 + \alpha}d + nrd) = O(n^{1 + \alpha}d)
\end{align*}
\end{proof}

\ifdefined\isarxiv

\section{Attention Optimization via Single Threshold Support Basis}

\else
\section{ATTENTION OPTIMIZATION VIA SINGLE THRESHOLD SUPPORT BASIS}

\fi

\label{sec:attention_single_basis}

In Appendix~\ref{sub:attention_single_basis:p_implies_q}, we cite a lemma from \cite{as23} stating that if we can have the relative error between the $\paren{Q, K}$-softmax-attention matrix $A$ and the approximate $\paren{Q, K}$-softmax-attention matrix $\wt A$, then we can find the relative error between $\diag(A {\bf 1}_n)$ and $\diag(\wt A {\bf 1}_n)$. In Appendix~\ref{sub:attention_single_basis:p}, we get the relative error between $\diag(A {\bf 1}_n)$ and $\diag(\wt A {\bf 1}_n)$ by finding the relative error between $A$ and $\wt A$. In Appendix~\ref{sub:attention_single_basis:main}, we combine these relative errors with our result on the (batch) Gaussian kernel density estimation (Lemma~\ref{lem:gaussian_kde}) to construct an $A^{\paren{L}}$ sparsity time algorithm to approximate the attention computation.

\subsection{Relationship Between the Attention Matrix and the Normalization Matrix}

\label{sub:attention_single_basis:p_implies_q}

We cite Lemma 3.5 from \cite{as23} which states that if we can get $\left |\widetilde{A}_{i,j} - A_{i,j} \right| \leq \epsilon_A \cdot A_{i,j}$, then we have $\left | \widetilde{D}_{i,i} - D_{i,i} \right | \leq \epsilon_A \cdot D_{i,i}$.

\begin{lemma}[Lemma 3.5 in \cite{as23}]\label{lem:relationship}
Let $A \in \mathbb{R}^{n \times n}$ be any matrix whose entries are all positive and $\epsilon_A \in (0, 0.1)$ be any parameter. Let $\widetilde{A} \in \mathbb{R}^{n \times n}$ be any matrix such that, for all $(i, j) \in [n] \times [n]$, we have
\[
\left |\widetilde{A}_{i,j} - A_{i,j} \right| \leq \epsilon_A \cdot A_{i,j}.
\]
Define the matrices $D, \widetilde{D} \in \mathbb{R}^{n \times n}$ by $D = \mathrm{diag}(A \mathbf{1}_n)$ and $\widetilde{D} = \mathrm{diag}(\widetilde{A} \mathbf{1}_n)$. Then, for all $i \in [n]$ we have
\[
\left | \widetilde{D}_{i,i} - D_{i,i} \right | \leq \epsilon_A \cdot D_{i,i}.
\]
\end{lemma}

\begin{proof}
We have
\begin{align*}
    \left |\widetilde{D}_{i,i} - D_{i,i} \right| 
    = & ~ \left| \sum_{j=1}^n \widetilde{A}_{i,j} - \sum_{j=1}^n A_{i,j} \right| \\
    \leq & ~ \sum_{j=1}^n \left|\widetilde{A}_{i,j} - A_{i,j} \right| \\
    \leq & ~ \sum_{j=1}^n \epsilon_A A_{i,j} \\
    = & ~ \epsilon_A \cdot D_{i,i},
\end{align*}
where the first step follows from $D = \mathrm{diag}(A \mathbf{1}_n)$ and $\widetilde{D} = \mathrm{diag}(\widetilde{A} \mathbf{1}_n)$, the second step follows from the triangle inequality (Fact~\ref{fac:vector_norm}), the third step follows from the assumption in the lemma statement $\left |\widetilde{A}_{i,j} - A_{i,j} \right| \leq \epsilon_A \cdot A_{i,j}$, and the last step follows from $D = \mathrm{diag}(A \mathbf{1}_n)$.
\end{proof}

\subsection{The Error Bound for the Normalization Matrix}
\label{sub:attention_single_basis:p}

Now, we prove $\left |\widetilde{A}_{i,j} - A_{i,j} \right| \leq \epsilon_A \cdot A_{i,j}$ so that by the logic of Lemma~\ref{lem:relationship}, we can get $\left |\widetilde{D}_{i,i} - D_{i,i} \right| \leq \epsilon \cdot D_{i,i}$.

\begin{lemma}\label{lem:relative_D}
    Given the query $Q \in \R^{n \times d}$ and the key $K \in \R^{n \times d}$, we define the $\paren{Q, K}$-softmax-attention matrix as $A = \exp\paren{QK^\top / d}$ and $Q^{\paren{L}}, Q^{\paren{s}}, K^{\paren{L}}, K^{\paren{s}} \in \R^{n \times d}$ as in Definition~\ref{def:Q_K}. Suppose by Lemma~\ref{lem:wt_A_small_rank}, there exists $U_1, U_2 \in \R^{n \times r}$ such that $U_1 U_2^\top$ is the $\paren{\epsilon_0, r}$-approximation of $\exp\paren{Q^{\paren{s}}\paren{K^{\paren{s}}}^\top / d}$ for all arbitrary $\epsilon_0 \in \paren{0, 0.1}$. With $\wt A = U_1 U_2^\top + \exp\paren{A^{\paren{L}} / d} - {\bf 1}_{n \times n}$, we define $\wt D := \diag\paren{\wt A {\bf 1}_n}$ and $D := \diag\paren{A {\bf 1}_n}$.

    Then, we can show that for all $\epsilon \in \paren{0, 0.1}$ and $i \in [n]$,
    \[
    \left |\widetilde{D}_{i,i} - D_{i,i} \right| \leq \epsilon \cdot D_{i,i}.
    \]
\end{lemma}
\begin{proof}

First, for all $i, j \in [n]$, we have
\begin{align*}
    \left |\widetilde{A}_{i,j} - A_{i,j}\right |
    = & ~ \left |\paren{\exp\paren{A^{\paren{s}} / d} + \exp\paren{A^{\paren{L}} / d} - {\bf 1}_{n \times n}}_{i,j} - \paren{U_1U_2^\top + \exp\paren{A^{\paren{L}} / d} - {\bf 1}_{n \times n}}_{i,j}\right | \\
    = & ~ \left |\paren{\exp\paren{A^{\paren{s}} / d} + \exp\paren{A^{\paren{L}} / d} - {\bf 1}_{n \times n} - U_1U_2^\top - \exp\paren{A^{\paren{L}} / d} + {\bf 1}_{n \times n}}_{i,j}\right | \\
    = & ~ \left |\paren{\exp\paren{A^{\paren{s}} / d} - U_1U_2^\top}_{i, j}\right | \\
    \leq & ~ \epsilon \paren{\exp\paren{A^{\paren{s}} / d}}_{i, j},
\end{align*}
where the first step follows from $A = \exp\paren{QK^\top / d} = \exp\paren{A^{\paren{s}} / d} + \exp\paren{A^{\paren{L}} / d} - {\bf 1}_{n \times n}$ (see Eq.~\eqref{eq:A} for detail) and $\wt A = U_1 U_2^\top + \exp\paren{A^{\paren{L}} / d} - {\bf 1}_{n \times n}$ (see from the lemma statement) and the last step follows from Lemma~\ref{lem:wt_A_small_rank}.

Therefore, by Lemma~\ref{lem:relationship}, we have for all $i \in [n]$, 
\begin{align*}
    \left | \widetilde{D}_{i,i} - D_{i,i} \right | \leq \epsilon \cdot D_{i,i}.
\end{align*}
\end{proof}

\subsection{Main Result}
\label{sub:attention_single_basis:main}

We now state our main theoretical result for the single-threshold support basis framework. Building on Lemma~\ref{lem:gaussian_kde} and Lemma~\ref{lem:relative_D}, we show that the entire attention computation $D^{-1}AV$ can be approximated in $A^{\paren{L}}$ sparsity time with a relative error guarantee. Our approach combines two key components: a polynomial approximation for the bounded portion of the $\paren{Q, K}$-softmax-attention matrix, and an explicit computation for the sparse, large entries. Together, these enable us to construct an approximate attention output $P$ such that
\begin{align*}
    \| P - D^{-1}AV \|_\infty < \epsilon \cdot \|V\|_\infty,
\end{align*}
for all accuracy $\epsilon > 0$. The following theorem formalizes this guarantee and provides the corresponding runtime bound.

\begin{theorem}\label{thm:attention_approximation}
    Given the query $Q \in \R^{n \times d}$, key $K \in \R^{n \times d}$, and value $V \in \R^{n \times d}$, we define the $\paren{Q, K}$-softmax-attention matrix as $A = \exp\paren{QK^\top / d}$ and $Q^{\paren{L}}, Q^{\paren{s}}, K^{\paren{L}}, K^{\paren{s}} \in \R^{n \times d}$ as in Definition~\ref{def:Q_K}. 
    Let $\epsilon \in (0, 0.1)$.

    Then, we can solve the approximate attention computation (Definition~\ref{def:approximate_attention_computation}) by outputting $P \in \R^{n \times d}$ (Algorithm~\ref{alg:approx_attention}) satisfying
    \begin{align*}
        \left \|P - D^{-1} A V\right \|_\infty < \epsilon \left \|V \right \|_\infty
    \end{align*}
    in $O(n^{1 + \alpha}d)$ time.
\end{theorem}

\begin{proof}
    {\bf Proof of correctness.}

    By the triangle inequality (see Fact~\ref{fac:vector_norm}), we have
    \begin{align}\label{eq:D-1AV}
        \left \| D^{-1} AV - \wt D^{-1} \wt AV \right \|_\infty
        = & ~ \left \| D^{-1} AV - D^{-1} \wt AV + D^{-1} \wt AV -  \wt D^{-1} \wt AV \right \|_\infty \notag\\
        \leq & ~ \left \| D^{-1} AV - D^{-1} \wt AV \right \|_\infty + \left \| D^{-1} \wt AV -  \wt D^{-1} \wt AV \right \|_\infty.
    \end{align}

    Similar as \cite{as23}, we have for each $(i,j) \in [n] \times [d]$, 
\begin{align*}
\left|( \widetilde{D}^{-1} \widetilde{A} V - D^{-1} \wt A V )_{i,j}\right| 
= & ~ \left| \sum_{l=1}^n \left( \widetilde{D}^{-1}_{i,i} - D^{-1}_{i,i} \right) \cdot \widetilde{A}_{i,l} \cdot V_{l,j} \right| \\
\leq & ~ \sum_{l=1}^n \left| \widetilde{D}^{-1}_{i,i} - D^{-1}_{i,i} \right| \cdot |\widetilde{A}_{i,l}| \cdot \|V\|_\infty \\
= & ~ \sum_{l=1}^n \left| \frac{D_{i,i} - \widetilde{D}_{i,i}}{D_{i,i} \widetilde{D}_{i,i}} \right| \cdot |\widetilde{A}_{i,l}| \cdot \|V\|_\infty \\
\leq & ~ \epsilon \cdot \sum_{l=1}^n \left| \widetilde{D}_{i,i}^{-1} \widetilde{A}_{i,l} \right| \cdot \|V\|_\infty \\
= & ~ \epsilon \cdot \left| \sum_{l=1}^n \widetilde{D}_{i,i}^{-1} \widetilde{A}_{i,l} \right| \cdot \|V\|_\infty \\
= & ~ \epsilon \cdot \|V\|_\infty,
\end{align*}
where the second step follows from the triangle inequality (see Fact~\ref{fac:vector_norm}), the fourth step follows from Lemma~\ref{lem:relative_D} that
\[
\left| \frac{D_{i,i} - \widetilde{D}_{i,i}}{D_{i,i}} \right| \leq \epsilon,
\]
the fifth step follows from $\widetilde{D}_{i,i}^{-1} > 0$ and $\widetilde{A}_{i,l} > 0$, and the last step follows from $\left| \sum_{l=1}^n \widetilde{D}_{i,i}^{-1} \widetilde{A}_{i,l} \right| = 1$ as $D = \diag(A {\bf 1}_n)$.

Also, for each $(i,j) \in [n] \times [d]$,
\begin{align*}
\left|( D^{-1} \widetilde{A} V - D^{-1} A V )_{i,j}\right| 
= & ~ \left| \sum_{l=1}^n D^{-1}_{i,i} (\widetilde{A}_{i,l} - A_{i,l}) \cdot V_{l,j} \right| \\
\leq & ~ \sum_{l=1}^n \left| D^{-1}_{i,i} \right| \cdot |\widetilde{A}_{i,l} - A_{i,l}| \cdot \|V\|_\infty \\
= & ~ \sum_{l=1}^n D^{-1}_{i,i} \cdot |\widetilde{A}_{i,l} - A_{i,l}| \cdot \|V\|_\infty \\
\leq & ~ \sum_{l=1}^n D^{-1}_{i,i} \cdot \epsilon A_{i,l} \cdot \|V\|_\infty \\
= & ~ \epsilon \cdot \|V\|_\infty,
\end{align*}
where the second step follows from triangle inequality (see Fact~\ref{fac:vector_norm}), the third step follows from $D^{-1}_{i,i} > 0$, the fourth step follows from $|\widetilde{A}_{i,l} - A_{i,l}| \leq \epsilon \cdot A_{i,l}$ (see Lemma~\ref{lem:relative_D}), and the last step follows from definition of $D_{i,i}$.

{\bf Proof of Running time.}

Similar as the proof of Lemma~\ref{lem:gaussian_kde}, Line~\ref{line:atten_al} to Line~\ref{line:atten_d-1} takes $O(n^{1 + \alpha}d)$ time, Line~\ref{line:compute_C1_main} to Line~\ref{line:compute_C2_main} take $O(n^{1 + \alpha}d)$ time, and computing $D^{-1} (C_1 + C_2)$ takes $O(nd)$ time (as $D^{-1}$ is a diagonal matrix).
\end{proof}

\ifdefined\isarxiv

\section{Attention Optimization Under the Sub-Gaussian Distribution}

\else
\section{ATTENTION OPTIMIZATION UNDER THE SUB-GAUSSIAN DISTRIBUTION}

\fi

\label{sec:attention_single_basis_subgaussian}

Now, we justify the sparsity assumption used in our single-threshold support basis decomposition by introducing a natural probabilistic model: the sub-Gaussian distribution. Empirically, the entries of the query and key matrices in large language models tend to concentrate near zero and exhibit light tails—properties well captured by sub-Gaussian random variables. Under this assumption, we show that the number of ``large'' entries in $Q$ and $K$ (those exceeding a chosen threshold) is small with high probability. This validates our decomposition strategy and ensures that the matrix $A^{(L)}$ remains sparse. We also analyze the expected number of large entries and their contribution to the $\paren{Q, K}$-softmax-attention matrix, supporting the theoretical guarantees presented in Appendices~\ref{sec:single_basis} and~\ref{sec:attention_single_basis}.

Specifically, in Appendix~\ref{sub:attention_single_basis_subgaussian:def}, we present the formal definition of a sub-Gaussian distribution. In Appendix~\ref{sub:attention_single_basis_subgaussian:expectation}, we assume that the entries of the query matrix $Q$ and the key matrix $K$ follow sub-Gaussian distributions and show that the expected number of entries exceeding any threshold $T > 0$ is bounded. In Appendix~\ref{sub:attention_single_basis_subgaussian:nnz}, we prove that the number of non-zero entries in $Q^{(L)}$ and $K^{(L)}$ is at most $n^{\alpha}$ for some $\alpha \in (0, 1)$, with high probability. Finally, in Appendix~\ref{sub:attention_single_basis_subgaussian:attention}, we combine this sparsity result with our main result on attention optimization (Theorem~\ref{thm:attention_approximation}) to show that, with high probability, the attention computation can be approximated in sub-quadratic time. Thus, we finally can formally justify that the $A^{\paren{L}}$ sparsity time $O\paren{n^{1 + \alpha}}$ in Theorem~\ref{thm:attention_approximation} is truly sub-quadratic in $n$.

\subsection{The Definition of Sub-Gaussian Distribution}
\label{sub:attention_single_basis_subgaussian:def}

To formally analyze the statistical properties of the query and key matrices, we begin by introducing the notion of a sub-Gaussian distribution. Sub-Gaussian random variables exhibit tail behavior similar to or lighter than that of a Gaussian, making them a natural model for the entries of $Q$ and $K$ in practice. This assumption enables us to rigorously bound the probability and expected number of large entries, which is essential for proving sparsity and deriving sub-quadratic algorithms.

\begin{definition}[Proposition 2.5.2 in \cite{v18}]
A random variable \( X \) is said to be sub-Gaussian with variance \( \sigma^2 \), denoted as $X \in \mathrm{subG}(\sigma^2)$, if
\[
\Pr\left[ |X| \geq t \right] \leq 2 \exp\left( - \frac{t^2}{\sigma^2} \right) \quad \text{for all } t \geq 0.
\]
\end{definition}

\subsection{The Expected Number of Large Entries}

\label{sub:attention_single_basis_subgaussian:expectation}

Under the sub-Gaussian assumption, we can now quantify how many entries in the query and key matrices exceed a given threshold. Specifically, the following lemma shows that the expected number of such large entries decays exponentially with the threshold. This result supports our sparsity assumption for $Q^{(L)}$ and $K^{(L)}$, and plays a key role in bounding the size of the matrix $A^{(L)}$ used in our support basis decomposition.

\begin{lemma}[Expected Number of Large Entries Under Sub-Gaussianity]
\label{lem:expected-large-entries}
Let \( Q, K, V \in \mathbb{R}^{n \times d} \) be random matrices whose entries are independent and sub-Gaussian with variance proxies \( \sigma_Q^2 \), \( \sigma_K^2 \), and \( \sigma_V^2 \), respectively, which implies that for all \( i \in [n] \), \( j \in [d] \),
\[
\Pr\left[ |Q_{i,j}| \geq t \right] \leq 2 \exp\left( -\frac{t^2}{\sigma_Q^2} \right),
\]
and similarly for \( K_{i,j} \) and \( V_{i,j} \). Let \( T > 0 \) be a threshold. Define $Q^{(L)} \in \R^{n \times d}$ as
\[
Q^{(L)}_{i,j} := \begin{cases}
Q_{i,j} & \text{if } |Q_{i,j}| > T, \\
0 & \text{otherwise},
\end{cases}
\]
and similarly for \( K^{(L)} , V^{(L)} \in \R^{n \times d}\). Then, for all $M^{(L)} \in \{Q^{(L)}, K^{(L)}, V^{(L)}\}$ and $\sigma_M \in \{\sigma_Q, \sigma_K, \sigma_V\}$, we have
\[
\mathbb{E}\left[\left | \mathrm{supp}(M^{(L)})\right | \right]
\leq 2nd \cdot \exp\left( -\frac{T^2}{\sigma_M^2} \right).
\]
\end{lemma}

\begin{proof}

Fix a matrix \( M \in \{Q, K, V\} \). For each \( i \in [n], j \in [d] \), define the indicator variable
\[
X_{i,j} := \mathbb{I}\left[ |M_{i,j}| > T \right].
\]
Then, \( \left |\mathrm{supp}(M^{(L)}) \right | = \sum_{i=1}^{n} \sum_{j=1}^{d} X_{i,j} \). 

Therefore, taking expectation and using linearity, we have
\begin{align*}
    \mathbb{E}\left[\left |\mathrm{supp}\paren{M^{(L)}} \right | \right] 
    = & ~ \sum_{i=1}^{n} \sum_{j=1}^{d} \Pr\left[ |M_{i,j}| > T \right] \\
    \leq & ~ 2nd \cdot \exp\left( -\frac{T^2}{\sigma_M^2} \right),
\end{align*}
where the last step follows from the sub-Gaussian tail bound.
\end{proof}

\subsection{The Number of Non-Zero Entries}
\label{sub:attention_single_basis_subgaussian:nnz}

Next, we strengthen the result of Lemma~\ref{lem:expected-large-entries} by moving from an expectation bound to a high-probability guarantee. Specifically, we show that with high probability, the number of large entries in the query and key matrices is at most $n^\alpha$ for some $\alpha \in (0,1)$.

\begin{lemma}[The bound of the number of non-zero entries of $Q^{(L)}$ and $K^{(L)}$]
\label{lem:joint-sparsity}
Let \( Q, K \in \mathbb{R}^{n \times d} \) be random matrices with independent entries such that each \( Q_{i,j}, K_{i,j} \) is sub-Gaussian with variance proxies \( \sigma_Q^2 \) and \( \sigma_K^2 \), respectively. For a constant $c \in (0, 1)$ and a threshold \( T = \sqrt{c \log n} > 0 \), we define the large-entry matrices:
\[
Q^{(L)}_{i,j} := \begin{cases}
Q_{i,j} & \text{if } |Q_{i,j}| > T, \\
0 & \text{otherwise},
\end{cases}
\quad
K^{(L)}_{i,j} := \begin{cases}
K_{i,j} & \text{if } |K_{i,j}| > T, \\
0 & \text{otherwise}.
\end{cases}
\]
Then, for all $M^{\paren{L}} \in \{Q^{\paren{L}}, K^{\paren{L}}\}$, there exists \( 0 < \alpha < 1 \), such that
\[
\Pr\left[ \left |\mathrm{supp}\paren{M^{(L)}} \right | > n^\alpha \right]
\leq \exp\left( -\Omega(n^\alpha) \right).
\]
\end{lemma}

\begin{proof}
Since each \( Q_{i,j} \in \mathrm{subG}(\sigma_Q^2) \), we have
\[
\Pr\left[ |Q_{i,j}| > T \right] \leq 2 \exp\left( -\frac{T^2}{\sigma_Q^2} \right),
\]
and similarly for \( K_{i,j} \). Let \( X = \left |\mathrm{supp}\paren{Q^{(L)}} \right | = \sum_{i,j} \mathbb{I}\left[ |Q_{i,j}| > T \right] \).
We note that $X$ is a sum of independent entries.

Then, by Lemma~\ref{lem:expected-large-entries}, we have
\begin{align*}
    \mathbb{E}[X] 
    \leq & ~ 2nd \cdot \exp\left( -\frac{T^2}{\sigma_Q^2} \right)\\
    = & ~ 2nd \cdot \exp\left( -\frac{c \log n}{\sigma_Q^2} \right)\\
    = & ~ 2nd \cdot \paren{e^{\log n}}^{- \frac{c}{\sigma_Q^2}}\\
    = & ~ 2nd \cdot n^{-c/\sigma_Q^2}\\
    = & ~ o(n^\alpha),
\end{align*}
where the last step follows from $c, \sigma_Q^2 > 0$ and $d = O(\log n)$.

Therefore, by a multiplicative Chernoff bound (see Fact~\ref{fac:chernoff}), for all \( \delta > 0 \),
\[
\Pr\left[ X > (1+\delta) \mathbb{E}[X] \right] \leq \exp\left( - \frac{\delta^2 \cdot \mathbb{E}[X]}{3} \right).
\]
In particular, for large enough \( n \), \( n^\alpha > (1+\delta)\mathbb{E}[X] \), so
\[
\Pr\left[ \left |\mathrm{supp}\paren{Q^{(L)}} \right | > n^\alpha \right] \leq \exp\left( -\Omega(n^\alpha) \right),
\]
and likewise for \( \left |\mathrm{supp}\paren{K^{(L)}} \right | \).
\end{proof}

\subsection{Approximating the Attention Computation Under the Sub-Gaussian Assumption}
\label{sub:attention_single_basis_subgaussian:attention}

Finally, we combine the sparsity guarantees established in Lemma~\ref{lem:joint-sparsity} with our main result on attention approximation (Theorem~\ref{thm:attention_approximation}). The following theorem shows that, under the sub-Gaussian assumption, our algorithm approximates the attention computation and runs in sub-quadratic time with high probability. This provides a rigorous justification for the practical efficiency of our method and demonstrates that strong runtime guarantees can be achieved without requiring the bounded entry assumption, $B < o\paren{\sqrt{\log n}}$.

\begin{theorem}[Main Theorem Under Sub-Gaussian Assumption]
\label{thm:subgaussian-main}
Let \( Q, K, V \in \mathbb{R}^{n \times d} \) be the query, key, and value matrices. Suppose entries of $Q, K$ are independent and sub-Gaussian with variance proxies \( \sigma_Q^2 \) and \( \sigma_K^2 \), respectively. It implies that for all \( i \in [n] \), \( j \in [d] \),
\[
\Pr\left[ |Q_{i,j}| \geq t \right] \leq 2 \exp\left( -\frac{t^2}{\sigma_Q^2} \right),
\]
and similarly for \( K_{i,j} \). 
Fix a threshold $T = \sqrt{c \log n}$ for some constant $c \in (0, 1)$, and let $\epsilon \in (0, 0.1)$ be a target accuracy parameter. Then, with probability at least $1 - \exp(-\Omega(n^\alpha))$, for some $0 < \alpha < 1$, the output $P \in \mathbb{R}^{n \times d}$ from Algorithm~\ref{alg:approx_attention} satisfies
\[
\left\| P - D^{-1} A V \right\|_\infty \leq \epsilon \cdot \|V\|_\infty,
\]
where $A := \exp(QK^\top / d)$ and $D := \mathrm{diag}(A \cdot \mathbf{1}_n)$. Furthermore, the runtime of the algorithm is
\[
O(n^{1 + \alpha}d).
\]
\end{theorem}

\begin{proof}
    By Lemma~\ref{lem:joint-sparsity}, we have with probability $1 - \exp\left( -\Omega(n^\alpha) \right)$,
    \begin{align*}
        \left |\mathrm{supp}\paren{M^{(L)}} \right | \leq n^\alpha,
    \end{align*}
    with $\alpha \in (0, 1)$.

    Therefore, by Lemma~\ref{lem:sparsity_AL}, we have
    \begin{enumerate}
        \item $\left |\mathrm{supp}\paren{A^{(L)}} \right | = O\paren{n^{1 + \alpha}}$, and
        \item it takes $O\paren{n^{1 + \alpha}d}$ time to compute $A^{\paren{L}}$.
    \end{enumerate}

    Since $\left |\mathrm{supp}\paren{A^{(L)}} \right | = O\paren{n^{1 + \alpha}}$, computing $\paren{\exp\paren{A^{\paren{L}} / d} - {\bf 1}_{n \times n}} V$ (Line~\ref{line:compute_C2} Algorithm~\ref{alg:gaussian_kde}) and $\paren{\exp\paren{A^{\paren{L}}} - {\bf 1}_{n \times n}} {\bf 1}_n$ (Line~\ref{line:atten_d2} in Algorithm~\ref{alg:approx_attention}) take $O\paren{n^{1 + \alpha}d}$ time.
\end{proof}

\ifdefined\isarxiv

\section{Generalization to Multi-Layer Transformer Architectures}

\else
\section{GENERALIZATION TO MULTI-LAYER TRANSFORMER ARCHITECTURES}

\fi

\label{sec:multi_layer}

Recall that from Definition~\ref{def:exact_attention_computation} we define the single-layer attention computation as 
\begin{align*}
    D^{-1} A V = D^{-1} \exp\paren{QK^\top / d} V.
\end{align*}

In transformer architectures, this only denotes one single layer of the forward propagation. Without loss of generality, we can suppose this is the $\beta$-th layer: the query matrix is defined as $Q = X_\beta W_Q$, the key matrix is defined as $K = X_\beta W_K$, and the value matrix is defined as $V = X_\beta W_V$, where $X_\beta \in \R^{n \times d}$ is the input for the $\beta$-th layer and $W_Q, W_K, W_V \in \R^{d \times d}$ are the weight matrices for the query, key, and value respectively. When attempting to optimize the $\beta$-th layer forward propagation, we can exactly compute the matrices $Q, K, V$ in $O(nd^2)$ time, which is already sub-quadratic in $n$. The $\beta$-th layer can be formulated as the following recursive relation:
\begin{align*}
 X_{\beta + 1} \gets D^{-1} \exp(X_{\beta} W_Q W_K^\top X_{\beta}^\top) X_{\beta} W_V
\end{align*} 

To generalize our result to multi-layer transformer architectures, we also need to efficiently approximate the gradients of the attention computation with respect to these weight matrices. We first define the attention computation loss function as follows:
\begin{definition}[Attention loss]\label{def:attention_loss}
    Let $B \in \R^{n \times d}$ and $X, Y \in \R^{d \times d}$ be the weights where $X = W_Q W_K^\top$ and $Y = W_V$. Given the input $X_\ell \in \R^{n \times d}$, we define the attention loss as:
    \begin{align*}
        \min_{X,Y \in \R^{d \times d}} L(X, Y) := \min_{X,Y \in \R^{d \times d}}  \| D(X)^{-1} \exp(X_\ell X X_\ell^\top) X_\ell Y - B \|_F^2,
    \end{align*}
    where the diagonal matrix $D(X) \in \R^{n \times n}$ is defined as $D(X) := \diag( \exp(X_\ell X X_\ell^\top ) {\bf 1}_n )$.
\end{definition}

Since the dependence of the attention loss $L$ on $Y = W_V$ is linear, it is very straightforward to compute the gradient with respect to the value weight $\frac{\partial L(X, Y)}{\partial Y}$. Therefore, prior works such as \cite{as24b} focus on the optimization of $\frac{\partial L(X, Y)}{\partial X}$.

\begin{definition}[Approximate attention loss gradient computation]
Let the attention loss $L(X, Y)$ be defined as in Definition~\ref{def:attention_loss}.
For all $\epsilon > 0$, the goal is to output a vector $\wt{g}$ such that
\begin{align*}
    \left\| \wt{g} - \frac{\partial L(X, Y)}{\partial \mathrm{vec}\paren{X}} \right\|_\infty \leq \epsilon.
\end{align*}
\end{definition}

With $x = \mathrm{vec}\paren{X}$, $A\paren{x}$ be the $\paren{Q, K}$-softmax-attention matrix, and $\mathcal{Q}\paren{x} := VV^\top A\paren{x} - VE^\top$, \cite{as24b,dsxy23} show that the gradient of the attention can be expressed as 
\begin{align*}
    \frac{\partial L}{\partial x} = \mathrm{vec}\paren{X_\ell \paren{\mathcal{P}_1\paren{x} - \mathcal{P}_2\paren{x}} X_\ell^\top},
\end{align*}
where $\mathcal{P}_1\paren{x} = A\paren{x} \circ \mathcal{Q}\paren{x}$ and $\mathcal{P}_2\paren{x} := A\paren{x} \diag\paren{r\paren{x}}$, with $r\paren{x} \in \R^n$ being defined as $r\paren{x}_i := \langle A\paren{x}_{i, *}, \mathcal{Q}\paren{x}_{i, *} \rangle$. \cite{as24b} note that similar to the inference, the dominating term of the time complexity for computing $\frac{\d L}{\d x}$ is still the terms related to $A\paren{x}$. Therefore, by replacing $A\paren{x}$ with $U_1U_2^\top$ from \cite{as23}, \cite{as24b} shows that $\frac{\d L}{\d x}$ can be approximated in almost linear time.

Similarly, our method expresses $A\paren{x}$ as $\exp\paren{A^{\paren{s}}\paren{x} / d} + \exp\paren{A^{\paren{L}}\paren{x} / d} - {\bf 1}_{n \times n}$ (see Eq.~\eqref{eq:disjoint}). Entries of $\exp\paren{A^{\paren{s}} / d}$ are small, so we can use the same technique as \cite{as24b} to approximate it by $U_1U_2^\top$. $\exp\paren{A^{\paren{L}}\paren{x} / d} - {\bf 1}_{n \times n}$ is a sparse matrix, but the product with other dense matrices is dense. Thus, further computation may still require quadratic time. 

To address this issue, we use the approximate singular value decomposition from \cite{cw17}. We can approximate its best rank-$k$ low-rank approximation in $A^{\paren{L}}\paren{x}$ sparsity time (Theorem~\ref{thm:approximate_svd}). Choosing $k = n^{o(1)}$, we can approximate $\frac{\d L}{\d x}$ in $A^{\paren{L}}\paren{x}$ sparsity time, and by sub-Gaussianity, we can generalize our method to multi-layer attention.

\begin{theorem}[Theorem 8.1 in \cite{cw17}]\label{thm:approximate_svd}
For $A \in \R^{n \times n}$, there is an algorithm that, with failure probability $1/10$, 
finds matrices $L, W \in \R^{n \times k}$ with orthonormal columns, and diagonal 
$D \in \R^{k \times k}$, so that
\[
\|A - LDW^\top\|_F \leq (1+\varepsilon)\Delta_k.
\]
The algorithm runs in time
\[
O(\mathrm{nnz}(A)) + \tilde{O}(nk^2 \varepsilon^{-4} + k^3 \varepsilon^{-5}).
\]
\end{theorem}

\ifdefined\isarxiv

\section{Multiple Thresholds Support Basis Decomposition}

\else
\section{MULTIPLE THRESHOLDS SUPPORT BASIS DECOMPOSITION}

\fi

\label{sec:multi_basis}

We have now completed the presentation of the proofs for our first result, which uses the single-threshold support basis to approximate the attention computation without relying on the bounded-entry assumption required in \cite{as23}. Instead, we introduce the sub-Gaussian assumption, which, to the best of our knowledge, is satisfied by all existing transformer architectures (see Appendix~\ref{sec:distribution}). 

To extend our theoretical results to settings where the sub-Gaussian assumption may not hold, we next present a framework based on multiple-threshold support basis decomposition. Unlike the single-threshold approach, this method does not require any distributional assumptions on the entries of $Q$ and $K$. However, due to the lower bound established in \cite{as23} (see Theorem~\ref{thm:formal_main_lower_bound})\footnote{Assuming \textsf{SETH}, with $d = O(\log n)$ and the entries of $Q$ and $K$ not bounded above by $o\left(\sqrt{\log n}\right)$, no sub-quadratic time algorithm can approximate the attention computation to error $\epsilon_a = n^{-C_\alpha}$.}, we must sacrifice the approximation error in order to achieve sub-quadratic runtime.

Beyond the optimization perspective offered by our multiple-threshold support basis decomposition, we also establish a theoretical connection between polynomial attention and the Chebyshev polynomial approximation of softmax attention (Definition~\ref{def:exact_attention_computation}). This provides a theoretical justification for the strong empirical performance of high-degree polynomial attention observed in \cite{kmz24}. 

Our theoretical results go further than merely explaining existing empirical findings. Specifically, we show that the sum of multiple polynomial attentions achieves a significantly smaller $\ell_p$ error with respect to softmax attention, even though the $\ell_\infty$ error remains the same as that of a single polynomial approximation. This suggests that aggregating multiple polynomial approximations more accurately captures the behavior of softmax attention. We hope this insight will inspire future empirical research to investigate the benefits of summing multiple polynomial attentions, rather than relying solely on a single polynomial approximation as in \cite{kmz24}.

In Appendix~\ref{sub:multi_basis:definition}, we introduce some basic definitions for multiple thresholds support basis decomposition for decomposing the query $Q$ and key $K$ matrices. In Appendix~\ref{sub:multi_basis:decompose}, we prove that the decomposition satisfies the definition of support basis. In Appendix~\ref{sub:multi_basis:additive}, we use the mathematical property of the support basis developed earlier to get the additive decomposition of the $\paren{Q, K}$-softmax-attention matrix $\exp\paren{QK^\top / d}$. 

\subsection{Basic Definitions}

\label{sub:multi_basis:definition}

Now, we present the definitions for our multiple-threshold support basis decomposition. We aim to partition the query $Q$ and key $K$ matrices according to multiple value ranges, enabling a finer-grained approximation of the $\paren{Q, K}$-softmax-attention matrix. 
\begin{definition}[Multi-threshold decomposition of query and key matrices]\label{def:general_Q_K}
    Let $n, m, d$ be positive integers where $2 \leq m \leq n$. Let $Q, K \in \R^{n \times d}$ be the query and key matrices, respectively. Let $\min_{i \in [n], j \in [d]} \{|Q_{i, j}|, |K_{i, j}| \} = T_0 < T_1 < \dots < T_{m - 1} < T_m = \infty$ be a sequence of thresholds.  
    For all $i \in [n]$ and $\ell \in [m]$, we define
    \[
    Q^{\paren{T_\ell}}_{i,*} := 
    \begin{cases}
        Q_{i,*} & \text{if } T_{\ell - 1} \leq \max_{j \in [d]} |Q_{i,j}| < T_\ell \\ 
        {\bf 0}_{1 \times d} & \text{otherwise},
    \end{cases}
    \]
    and similarly for $K^{\paren{T_\ell}}_{i,j}$.
\end{definition}

We now formally define how to decompose the matrix $QK^\top$ into a sum of components based on multiple thresholds. By ensuring that the resulting matrices are disjoint, we preserve the additive structure necessary for efficient and accurate approximation. The following definition specifies the multi-threshold disjoint decomposition of $QK^\top$.

\begin{definition}[Multi-threshold disjoint decomposition of $QK^\top$]\label{def:multi_threshold_disjoint}
Let $n, m, d$ be positive integers where $2 \leq m \leq n$. Let $Q, K \in \R^{n \times d}$ be the query and key matrices, respectively. Let $\min_{i \in [n], j \in [d]} \{|Q_{i, j}|, |K_{i, j}| \} = T_0 < T_1 < \dots < T_{m - 1} < T_m = \infty$ be a sequence of thresholds.   

For all $T_l, T_{l'} \in  \{T_\ell \}_{\ell = 1}^{m}$, we let $Q^{\paren{T_l}}, K^{\paren{T_{l'}}}$ be defined as in Definition~\ref{def:general_Q_K}, and we define
\begin{align*}
    A^{\paren{T_l, T_{l'}}} := Q^{\paren{T_l}} \paren{K^{\paren{T_{l'}}}}^\top.
\end{align*}

\end{definition}

We now provide an example to better illustrate the definitions above.

\begin{example}[An example of 2-threshold disjoint decomposition of $QK^\top$]

Let $T_1 > 0$ be a positive real number and $T_2 = \infty$. Let $\begin{bmatrix}
a & b \\
c & d \\
e & f
\end{bmatrix} \in \R^{3 \times 2}$ and $\begin{bmatrix}
g & i & k \\
h & j & l
\end{bmatrix} \in \R^{2 \times 3}$, where $|c|, |k| \geq T_1$ and the absolute value of other entries are smaller than $T_1$.

This setup allows us to decompose the matrix product $QK^\top$ into a sum of disjoint components based on the threshold values $T_1$ and $T_2$ as follows:
    \begin{align*}
    & ~ \begin{bmatrix}
a & b \\
c & d \\
e & f
\end{bmatrix}
\begin{bmatrix}
g & i & k \\
h & j & l
\end{bmatrix}\\
= & ~
\begin{bmatrix}
ag + bh & ai + bj & ak + bl \\
cg + dh & ci + dj & ck + dl \\
eg + hf & ei + fj & ek + fl
\end{bmatrix}\\
= & ~
\underbrace{\begin{bmatrix}
0 & 0 & 0 \\
0 & 0 & ck + dl \\
0 & 0 & 0
\end{bmatrix}}_{A^{\paren{T_2, T_2}}}
+
\underbrace{\begin{bmatrix}
0 & 0 & 0 \\
cg + dh & ci + dj & 0 \\
0 & 0 & 0
\end{bmatrix}}_{A^{\paren{T_2, T_1}}}
+
\underbrace{\begin{bmatrix}
0 & 0 & ak + bl \\
0 & 0 & 0 \\
0 & 0 & ek + fl
\end{bmatrix}}_{A^{\paren{T_1, T_2}}}
+
\underbrace{\begin{bmatrix}
ag + bh & ai + bj & 0 \\
0 & 0 & 0 \\
eg + hf & ei + fj & 0
\end{bmatrix}}_{A^{\paren{T_1, T_1}}}\\
= & ~
\underbrace{\begin{bmatrix}
0 & 0\\
c & d\\
0 & 0\\
\end{bmatrix}}_{Q^{\paren{T_2}}}
\underbrace{\begin{bmatrix}
0 & 0 & k \\
0 & 0 & l
\end{bmatrix}}_{\paren{K^{\paren{T_2}}}^\top}
+
\underbrace{\begin{bmatrix}
0 & 0\\
c & d\\
0 & 0
\end{bmatrix}}_{Q^{\paren{T_2}}}
\underbrace{\begin{bmatrix}
g & i & 0 \\
h & j & 0
\end{bmatrix}}_{\paren{K^{\paren{T_1}}}^\top}
+
\underbrace{\begin{bmatrix}
a & b \\
0 & 0 \\
e & f
\end{bmatrix}}_{Q^{\paren{T_1}}}
\underbrace{\begin{bmatrix}
0 & 0 & k \\
0 & 0 & l
\end{bmatrix}}_{\paren{K^{\paren{T_2}}}^\top}
+
\underbrace{\begin{bmatrix}
a & b \\
0 & 0 \\
e & f
\end{bmatrix}}_{Q^{\paren{T_1}}}
\underbrace{\begin{bmatrix}
g & i & 0\\
h & j & 0
\end{bmatrix}}_{\paren{K^{\paren{T_1}}}^\top}.
\end{align*}
This example illustrates how the product $QK^\top$ can be partitioned into disjoint matrix components, forming a support basis that enables structured approximation.
\end{example}

\subsection{Multi-Threshold Support Basis Decomposition of \texorpdfstring{$QK^\top$}{}}
\label{sub:multi_basis:decompose}

We now show that the multi-threshold decomposition of $QK^\top$ forms a valid support basis. This helps with applying polynomial approximations to each component individually.

\begin{lemma}\label{lem:support_basis}
Let $n, m, d$ be positive integers where $2 \leq m \leq n$. Let $Q, K \in \R^{n \times d}$ be the query and key matrices, respectively. Let $\min_{i \in [n], j \in [d]} \{|Q_{i, j}|, |K_{i, j}| \} = T_0 < T_1 < \dots < T_{m - 1} < T_m = \infty$ be a sequence of thresholds.   

Then, for all $T_l, T_{l'} \in  \{T_\ell \}_{\ell = 1}^{m}$, the collection $\{ A^{\paren{T_l, T_{l'}}}  \}_{T_l, T_{l'} \in  \{T_\ell \}_{\ell = 1}^{m}}$ forms a support basis (as defined in Definition~\ref{def:support_basis}) of $QK^\top$.
\end{lemma}

\begin{proof}

By Definition~\ref{def:support_basis}, it suffices to show:
\begin{enumerate}
    \item $\sum_{l = 1}^m \sum_{l' = 1}^m A^{\paren{T_l, T_{l'}}} = QK^\top$ and
    \item $A^{\paren{T_l, T_{l'}}}$'s are disjoint matrices (see Definition~\ref{def:disjoint}).
\end{enumerate}

{\bf Proof of Part 1.}

By Definition~\ref{def:general_Q_K}, we can see that
    \[
    Q = \sum_{\ell = 1}^{m} Q^{\paren{T_\ell}} \quad \text{and} \quad K = \sum_{\ell = 1}^{m} K^{\paren{T_\ell}},
    \]

We have
\begin{align*}
    QK^\top 
    = & ~ \paren{ \sum_{\ell=1}^m Q^{\paren{T_\ell}} } \paren{ \sum_{\ell'=1}^m K^{\paren{T_{\ell'}}} }^\top \\
    = & ~ \sum_{\ell=1}^m \sum_{\ell'=1}^m Q^{\paren{T_\ell}} \paren{K^{\paren{T_{\ell'}}}}^\top \\
    = & ~ \sum_{\ell=1}^m \sum_{\ell'=1}^m A^{\paren{T_\ell, T_{\ell'}}}.
\end{align*}

{\bf Proof of Part 2.}

     Recall from Definition~\ref{def:disjoint}, we say $A^{\paren{T_l, T_{l'}}}$'s are disjoint matrices if for all $i, j \in [n]$, for all $T_l, T_{l'} \in  \{T_\ell \}_{\ell = 1}^{m}$, for all $\paren{\ov{T}_l, \ov{T}_{l'}} \in  \{T_\ell \}_{\ell = 1}^{m} \times  \{T_\ell \}_{\ell = 1}^{m} \setminus  \{\paren{T_l, T_{l'}} \}$, if $\paren{A^{\paren{T_l, T_{l'}}}}_{i, j} \neq 0$, then $\paren{A^{\paren{\ov{T}_l, \ov{T}_{l'}}}}_{i, j} = 0$.

     Suppose we have $\paren{A^{\paren{T_l, T_{l'}}}}_{i, j} \neq 0$, which implies
     \begin{align*}
         \paren{A^{\paren{T_l, T_{l'}}}}_{i, j}
         = & ~ \paren{Q^{\paren{T_l}} \paren{K^{\paren{T_{l'}}}}^\top}_{i, j}\\
         = & ~ Q^{\paren{T_l}}_{i, *} \paren{K^{\paren{T_{l'}}}_{j, *}}^\top\\
         \neq & ~ 0.
     \end{align*}

     This implies that $Q^{\paren{T_l}}_{i, *} = Q_{i, *}$ and $K^{\paren{T_l}}_{j, *} = K_{j, *}$.

     Therefore, by Definition~\ref{def:general_Q_K}, we can see that for all $\ov{T}_l \in  \{T_\ell \}_{\ell = 1}^{m} \setminus  \{T_l \}$, for all $\ov{T}_{l'} \in  \{T_\ell \}_{\ell = 1}^{m} \setminus  \{T_{l'} \}$, both $Q^{\paren{\ov{T}_{l}}}_{i, *} = {\bf 0}_{1 \times d}$ and $K^{\paren{\ov{T}_{l'}}}_{k, *} = {\bf 0}_{1 \times d}$.

     Finally, we can get for all $\paren{\ov{T}_l, \ov{T}_{l'}} \in  \{T_\ell \}_{\ell = 1}^{m} \times  \{T_\ell \}_{\ell = 1}^{m} \setminus  \{\paren{T_l, T_{l'}} \}$, $\paren{A^{\paren{\ov{T}_l, \ov{T}_{l'}}}}_{i, j} = 0$.
\end{proof}

\subsection{Additive Decomposition of \texorpdfstring{$\exp\paren{QK^\top / d}$}{}}

\label{sub:multi_basis:additive}

Applying the entry-wise exponential function to disjoint matrices allows us to express the exponential of their sum as the sum of their exponentials. We now generalize Fact~\ref{fac:exp_split}, originally stated for the single-threshold support basis, to the multi-threshold support basis setting as follows:
\begin{lemma}\label{lem:disjoint_decomposition_entrywise_exponential}
Let $n, m, d$ be positive integers where $2 \leq m \leq n$. 
Let $Q = \sum_{\ell = 1}^{m} Q^{\paren{T_\ell}} \in \mathbb{R}^{n \times d} $ and $K = \sum_{\ell' = 1}^{m} K^{\paren{T_{\ell'}}} \in \mathbb{R}^{n \times d}$ be defined as in Definition~\ref{def:general_Q_K}.
Let $A^{\paren{T_\ell, T_{\ell'}}} = Q^{\paren{T_\ell}} \paren{K^{\paren{T_{\ell'}}}}^\top \in \R^{n \times n}$ be defined as in Definition~\ref{def:multi_threshold_disjoint}.

Then, we have
\[
\exp\paren{QK^\top / d} = \sum_{\ell = 1}^{m} \sum_{\ell' = 1}^{m} \exp\paren{A^{\paren{T_\ell, T_{\ell'}}} / d} - \paren{m^2 - 1}\cdot \mathbf{1}_{n \times n}.
\]
\end{lemma}
\begin{proof}
    We prove this lemma using mathematical induction.

    {\bf Base case (when $m = 2$).}

    We have
    \begin{align*}
        & ~ \exp\paren{QK^\top / d} \\
        = & ~ \exp\paren{\paren{Q^{\paren{T_1}} + Q^{\paren{T_2}}} \paren{K^{\paren{T_1}} + K^{\paren{T_2}}}^\top / d}\\
        = & ~ \exp\paren{\paren{Q^{\paren{T_1}}\paren{K^{\paren{T_1}}}^\top + Q^{\paren{T_2}}\paren{K^{\paren{T_1}}}^\top + Q^{\paren{T_1}}\paren{K^{\paren{T_2}}}^\top + Q^{\paren{T_2}}\paren{K^{\paren{T_2}}}^\top} / d}\\
        = & ~ \exp\paren{\paren{A^{\paren{T_1, T_{1}}} + A^{\paren{T_2, T_{1}}} + A^{\paren{T_1, T_{2}}} + A^{\paren{T_2, T_{2}}}} / d }\\
        = & ~ \exp\paren{A^{\paren{T_1, T_{1}}} / d} + \exp\paren{A^{\paren{T_2, T_{1}}} / d} + \exp\paren{A^{\paren{T_1, T_{2}}} / d} + \exp\paren{A^{\paren{T_2, T_{2}}} / d} - 3 \cdot {\bf 1}_{n \times n}\\
        = & ~ \sum_{\ell = 1}^{2} \sum_{\ell' = 1}^{2} \exp\paren{A^{\paren{T_\ell, T_{\ell'}}} / d} - \paren{2^2 - 1}\cdot \mathbf{1}_{n \times n},
    \end{align*}
    where the first step follows from the definition of $Q$ and $K$, the third step follows from the definition of $A^{\paren{T_\ell, T_{\ell'}}}$, the fourth step follows from combining Lemma~\ref{lem:support_basis} and Fact~\ref{fac:exp_split}, and the last step follows from $m = 2$.

    {\bf Inductive case.}

    Let $t$ be an arbitrary positive integer greater than or equal to $2$. Suppose for all $k \in [t]$, we have $Q = \sum_{\ell = 1}^{t} Q^{\paren{T_\ell}} \in \mathbb{R}^{n \times d} $ and $K = \sum_{\ell' = 1}^{t} K^{\paren{T_{\ell'}}} \in \mathbb{R}^{n \times d}$, and we have
    \begin{align}\label{eq:inductive_assumption}
        \exp\paren{QK^\top / d} 
        = & ~ \exp\paren{\sum_{\ell = 1}^{t} \sum_{\ell' = 1}^{t} A^{\paren{T_\ell, T_{\ell'}}}} \notag \\
        = & ~ \sum_{\ell = 1}^{t} \sum_{\ell' = 1}^{t} \exp\paren{A^{\paren{T_\ell, T_{\ell'}}} / d} - \paren{t^2 - 1}\cdot \mathbf{1}_{n \times n}.
    \end{align}

    Now, we consider the case of $t + 1$.

    We have
    \begin{align*}
        & ~ \exp\paren{QK^\top / d}\\
        = & ~ \exp\paren{\paren{\sum_{\ell = 1}^{t + 1} Q^{\paren{T_\ell}}} \paren{\sum_{\ell' = 1}^{t + 1} K^{\paren{T_{\ell'}}}}^\top / d}\\
        = & ~ \exp\left(\left(\paren{\sum_{\ell = 1}^{t} Q^{\paren{T_\ell}}} \paren{\sum_{\ell' = 1}^{t} K^{\paren{T_{\ell'}}}}^\top + \paren{\sum_{\ell = 1}^{t} Q^{\paren{T_\ell}}} \paren{K^{\paren{T_{t + 1}}}}^\top \right.\right. \\
        & ~ + \left. \left. \paren{Q^{\paren{T_{t + 1}}}} \paren{\sum_{\ell' = 1}^{t} K^{\paren{T_{\ell'}}}}^\top + \paren{Q^{\paren{T_{t + 1}}}} \paren{K^{\paren{T_{t + 1}}}}^\top\right) / d \right)\\
        = & ~ \exp\paren{\paren{\paren{\sum_{\ell = 1}^{t} \sum_{\ell' = 1}^{t} A^{\paren{T_\ell, T_{\ell'}}} } + A^{\paren{T_{t + 1}, T_{t + 1}}} + \paren{\sum_{\ell' = 1}^{t} A^{\paren{T_{t + 1}, T_{\ell'}}}} + \paren{\sum_{\ell = 1}^{t} A^{\paren{T_\ell, T_{t + 1}}}}} / d} \\
        = & ~ \exp\paren{\sum_{\ell = 1}^{t} \sum_{\ell' = 1}^{t} A^{\paren{T_\ell, T_{\ell'}}} / d} + \exp\paren{A^{\paren{T_{t + 1}, T_{t + 1}}} / d} + \exp\paren{\sum_{\ell' = 1}^{t} A^{\paren{T_{t + 1}, T_{\ell'}}} / d} \\
        & ~ + \exp\paren{\sum_{\ell = 1}^{t} A^{\paren{T_\ell, T_{t + 1}}} / d} - 3 \cdot {\bf 1}_{n \times n},
    \end{align*}
    where the first step follows from the definition of $Q$ and $K$, the third step follows from the definition of $A^{\paren{T_\ell, T_{\ell'}}}$ (as defined in Definition~\ref{def:multi_threshold_disjoint}), and the fourth step follows from combining Lemma~\ref{lem:support_basis} and Fact~\ref{fac:exp_split}.

    We note that by Lemma~\ref{lem:support_basis} and Fact~\ref{fac:exp_split}, we have
    \begin{align*}
        \exp\paren{\sum_{\ell' = 1}^{t} A^{\paren{T_{t + 1}, T_{\ell'}}}} = \sum_{\ell' = 1}^{t} \exp\paren{A^{\paren{T_{t + 1}, T_{\ell'}}}} - \paren{t - 1} \cdot {\bf 1}_{n \times n}
    \end{align*}
    and
    \begin{align*}
        \exp\paren{\sum_{\ell = 1}^{t} A^{\paren{T_\ell, T_{t + 1}}}} = \sum_{\ell = 1}^{t} \exp\paren{A^{\paren{T_\ell, T_{t + 1}}}} - \paren{t - 1} \cdot {\bf 1}_{n \times n}.
    \end{align*}

    Combining everything together, we have
    \begin{align*}
        \exp\paren{QK^\top / d} 
        = & ~ \sum_{\ell = 1}^{t + 1} \sum_{\ell' = 1}^{t + 1} \exp\paren{A^{\paren{T_\ell, T_{\ell'}}} / d} - \paren{t^2 - 1 + 2\paren{t - 1} + 3}\cdot \mathbf{1}_{n \times n} \\
        = & ~ \sum_{\ell = 1}^{t + 1} \sum_{\ell' = 1}^{t + 1} \exp\paren{A^{\paren{T_\ell, T_{\ell'}}} / d} - \paren{\paren{t + 1}^2 - 1}\cdot \mathbf{1}_{n \times n},
    \end{align*}
    which completes the proof.
\end{proof}

\ifdefined\isarxiv

\section{Threshold Values Specification and Interval-Based Factorization}

\else
\section{THRESHOLD VALUES SPECIFICATION AND INTERVAL-BASED FACTORIZATION}

\fi

\label{sec:multi_basis_threshold}

We have constructed a multiple-threshold support basis which can make the $\paren{Q, K}$-softmax-attention matrix $\exp\paren{QK^\top / d}$ be expressed as a sum of the exponential of each matrix in this support basis, namely $\sum_{\ell = 1}^{m} \sum_{\ell' = 1}^{m} \exp\paren{A^{\paren{T_\ell, T_{\ell'}}} / d} - \paren{m^2 - 1}\cdot \mathbf{1}_{n \times n}$ (see Lemma~\ref{lem:disjoint_decomposition_entrywise_exponential}). We eventually want to approximate the attention computation problem $D^{-1} \exp\paren{QK^\top / d} V$ (see Definition~\ref{def:exact_attention_computation} for details). Therefore, we have to find a bound for $m$ so that the total running time for approximating $m^2$ numbers of $D^{-1} \paren{\exp\paren{A^{\paren{T_\ell, T_{\ell'}}} / d} - \paren{m^2 - 1}\cdot \mathbf{1}_{n \times n}} V$ can be upper bounded. 

Specifically, in Appendix~\ref{sub:multi_basis_threshold:threshold}, we specify the values of the thresholds $\left\{ T_\ell \right\}_{\ell \in [m] \cup \{0\}}$ for our multiple-threshold support basis—motivated by Birge bucketing—so that $m$ can eventually be bounded. In Appendix~\ref{sub:multi_basis_threshold:fact}, we demonstrate how to tightly bound $\|A^{\circ p} - B^{\circ p}\|_\infty$ using the Mean Value Theorem. In Appendix~\ref{sub:multi_basis_threshold:poly}, we show that the $\paren{Q, K}$-softmax-attention matrix $\exp\paren{QK^\top / d}$ (Definition~\ref{def:exact_attention_computation}) can be approximated by a sum of polynomial attention matrices (see Definition~\ref{def:poly_attention_matrix}) of the form $\paren{U_1 U_2^\top / d}^{\circ p}$. In Appendix~\ref{sub:multi_basis_threshold:nobucket}, we prove that the $\ell_\infty$ error between the $\paren{Q, K}$-softmax-attention matrix and a single polynomial attention matrix is the same as that between the softmax matrix and the sum of polynomial attention matrices. Finally, in Appendix~\ref{sub:multi_basis_threshold:l1error}, we show that the $\ell_1$ error is significantly reduced when using Birge-bucketing-inspired thresholds.

\subsection{Thresholds for Support Basis}

\label{sub:multi_basis_threshold:threshold}

Birge bucketing~\citep{b87,bkr04,cgr16,bfr+00,abrr23} is a well-known technique used in distribution testing: it partitions the support of a distribution into buckets to better analyze its tail behavior and test for heavy-tailedness. In particular,~\cite{abrr23} introduces an equal-weight bucketing scheme that partitions the domain of a distribution into intervals (or buckets) such that each bucket contains the same probability mass.

Unlike the single-threshold support basis, which assumes that the entries of $Q, K \in \R^{n \times d}$ are sub-Gaussian, the multiple-threshold support basis makes no specific distributional assumptions about the entries of $Q$ and $K$. However, since their entries still follow some underlying distribution, we can apply the Birge bucketing technique to partition them. Because the exact distribution of the entries of $Q$ and $K$ is unknown\footnote{We even suppose that we do not have sample access to the entries of $Q$ and $K$.}, we cannot perform weight-based bucketing as in \cite{abrr23}. Moreover, uniform bucketing—i.e., using a fixed bucket length of $o\paren{\sqrt{\log n}}$—is also impractical, as the entries may take extremely large values, leading to an excessively large number of buckets $m$. Since we must approximate $\exp\paren{A^{\paren{T_\ell, T_{\ell'}}} / d} - \paren{m^2 - 1}\cdot \mathbf{1}_{n \times n}$ for all $\paren{\ell, \ell'} \in [m] \times [m]$, a large value of $m$ would prevent sub-quadratic runtime. 

To keep $m$ small, we define the thresholds as $T_\ell = b \paren{ 1 + \epsilon}^\ell$. We formally present the following definition:

\begin{definition}\label{def:exp_threshold_Q_K}
    Let $\epsilon < 0$ be the bucketing parameter. Let $n, m, d$ be positive integers where $2 \leq m \leq n$. Let $Q, K \in \R^{n \times d}$ be the query and key matrices, respectively. Let $b = \min_{i \in [n], j \in [d]}\left \{|Q_{i, j}|, |K_{i, j}|\right \}$ and $B = \max_{i \in [n], j \in [d]}\left \{|Q_{i, j}|, |K_{i, j}|\right \}$. For all $\ell \in [m] \cup \left \{0\right \}$, we let $T_\ell = b \paren{ 1 + \epsilon}^\ell$.

    We define
    \[
    Q^{\paren{ T_\ell}}_{i,*} := 
    \begin{cases}
        Q_{i,*} & \text{if } T_{\ell - 1} \leq \max_{j \in [d]} |Q_{i,j}| < T_\ell \\ 
        {\bf 0}_{1 \times d} & \text{otherwise},
    \end{cases}
    \]
    and similarly for $K^{\paren{ T_\ell}}_{i,j}$.
\end{definition}

Using $T_\ell = b \paren{ 1 + \epsilon}^\ell$, we can eventually show that the number of buckets $m$ can be bounded. It suffices to have $\left \lfloor \log_{1+\epsilon}\paren{ B/b} \right \rfloor + 1$ numbers of buckets.

\begin{fact}\label{fac:m}
    Let $\epsilon < 0$ be the bucketing parameter. Let $n, m, d$ be positive integers where $2 \leq m \leq n$. Let $Q, K \in \R^{n \times d}$ be the query and key matrices, respectively. Let $b = \min_{i \in [n], j \in [d]}\left \{|Q_{i, j}|, |K_{i, j}|\right \}$ and $B = \max_{i \in [n], j \in [d]}\left \{|Q_{i, j}|, |K_{i, j}|\right \}$. For all $\ell \in [m] \cup \left \{0\right \}$, we let $T_\ell = b \paren{ 1 + \epsilon}^\ell$. Let 
    \begin{align*}
        M = \left \{m_i \mid  [b, B] \subseteq \bigcup_{\ell = 0}^{m_i - 1} [T_\ell, T_{\ell + 1}) \right \}
    \end{align*}
    be a sequence of positive integers.

    Then, we have
    \begin{align*}
        \min_{i} M = \left \lfloor \log_{1+\epsilon}\paren{B/b} \right \rfloor + 1.
    \end{align*}
\end{fact}
\begin{proof}
    We note that $\bigcup_{\ell = 0}^{m - 1} [T_\ell, T_{\ell + 1})$ is an interval and, by definition, $T_0 = b \paren{ 1 + \epsilon}^0 = b$. Therefore, it suffices to find the smallest $m$ such that
    \begin{align*}
        T_m > B.
    \end{align*}

    We have
    \begin{align*}
        T_m = b \paren{ 1 + \epsilon}^m &> B\\
        \paren{ 1 + \epsilon}^m &> B/b\\
        m &> \log_{1 + \epsilon}\paren{ B/b}.
    \end{align*}

    Therefore, the smallest possible $m$ is $\left \lfloor \log_{1+\epsilon}\paren{ B/b} \right \rfloor + 1$.
\end{proof}

We need to ensure that within each bucket $\paren{\ell, \ell'} \in [m] \times [m]$, we can make $\left \|Q^{\paren{ T_\ell}}\right\|_\infty, \left \|K^{\paren{ T_\ell}}\right\|_\infty < o\paren{\sqrt{\log n}}$. Although $\left \|Q^{\paren{ T_\ell}}\right\|_\infty, \left \|K^{\paren{ T_\ell}}\right\|_\infty \in \R$, without imposing any assumption, they can be much larger than $o\paren{\sqrt{\log n}}$. Therefore, we take out a large constant $C^{\paren{ T_\ell}}$ from $Q^{\paren{ T_\ell}}$ and $K^{\paren{ T_{\ell'}}}$ to make $Q^{\paren{ T_\ell}} = C^{\paren{ T_\ell}} Q^{\paren{ \ell}}$ and $K^{\paren{ T_{\ell'}}} = C^{\paren{ T_{\ell'}}} K^{\paren{ \ell'}}$, where $\left \|Q^{\paren{ \ell}}\right\|_\infty, \left \|K^{\paren{ \ell'}}\right\|_\infty < o\paren{\sqrt{\log n}}$. Furthermore, we show that $C^{\paren{ T_\ell, T_{\ell'}}} = C^{\paren{ T_\ell}} C^{\paren{ T_{\ell'}}} \geq \frac{b^2 \paren{ 1 + \epsilon}^{\ell + \ell'}}{\log n}$, where $b = \min_{i \in [n], j \in [d]}\left \{|Q_{i, j}|, |K_{i, j}|\right \}$ and $\epsilon < 0$ is the bucketing parameter.

\begin{lemma}[Normalized block decomposition]\label{lem:normalized_decomposition}

Let $n, m, d$ be positive integers where $2 \leq m \leq n$. Let $\epsilon < 0$ be the bucketing parameter.
Let $Q = \sum_{\ell = 1}^{m} Q^{\paren{ T_\ell}} \in \mathbb{R}^{n \times d} $ and $K = \sum_{\ell' = 1}^{m} K^{\paren{ T_{\ell'}}} \in \mathbb{R}^{n \times d}$ be defined as in Definition~\ref{def:exp_threshold_Q_K}, with $b = \min_{i \in [n], j \in [d]}\left \{|Q_{i, j}|, |K_{i, j}|\right \}$, $B = \max_{i \in [n], j \in [d]}\left \{|Q_{i, j}|, |K_{i, j}|\right \}$, and for all $\ell \in [m] \cup \left \{0\right \}$, we let $T_\ell = b \paren{ 1 + \epsilon}^\ell$.
Let $A^{\paren{ T_\ell, T_{\ell'}}} = Q^{\paren{ T_\ell}} \paren{ K^{\paren{ T_{\ell'}}}}^\top \in \R^{n \times n}$.

Let $m := \left \lfloor \log_{1+\epsilon}\paren{ B/b} \right \rfloor + 1$ and $C^{\paren{ T_\ell, T_{\ell'}}} \geq \frac{b^2 \paren{ 1 + \epsilon}^{\ell + \ell'}}{\log n}$.
By Lemma~\ref{lem:disjoint_decomposition_entrywise_exponential}, we have
\begin{align*}
    \exp\paren{ QK^\top} = \sum_{\ell = 1}^{m} \sum_{\ell' = 1}^{m} \exp\paren{ A^{\paren{ T_\ell, T_{\ell'}}}} - \paren{ m^2 - 1}\cdot \mathbf{1}_{n \times n}.
\end{align*}

Then, for all $A^{\paren{ T_\ell, T_{\ell'}}}$, there exists $Q^{\paren{ \ell}}, K^{\paren{ \ell'}} \in \R^{n \times d}$ such that
\begin{align*}
    A^{\paren{ T_\ell, T_{\ell'}}} = C^{\paren{ T_\ell, T_{\ell'}}} \cdot \underbrace{Q^{\paren{ \ell}} \paren{ K^{\paren{ \ell'}}}^\top}_{:= A^{\paren{ \ell, \ell'}}},
\end{align*}
and $\left \|Q^{\paren{ \ell}}\right \|_\infty, \left \|K^{\paren{ \ell'}}\right \|_\infty \leq o\paren{ \sqrt{\log n}}$.
\end{lemma}

\begin{proof}
    By the definition of $A^{\paren{ T_\ell, T_{\ell'}}}$, we have for all $p, q \in [n]$,
    \begin{align*}
        \paren{ A^{\paren{ T_\ell, T_{\ell'}}}}_{p, q}
        = & ~ \sum_{k = 1}^d \paren{ Q^{\paren{ T_\ell}}}_{p, k} \paren{ K^{\paren{ T_{\ell'}}}}_{q, k}.
    \end{align*}

    By Definition~\ref{def:exp_threshold_Q_K}, we notice that $\left \|Q^{\paren{ T_\ell}}\right \|_\infty \leq b \paren{ 1 + \epsilon}^\ell$ and $\left \|K^{\paren{ T_\ell'}}\right \|_\infty \leq b \paren{ 1 + \epsilon}^{\ell'}$.

    By letting $C^{\paren{ T_\ell}} = \frac{b \paren{ 1 + \epsilon}^{\ell}}{\sqrt{\log n}}$, $K^{\paren{ T_{\ell'}}} = C^{\paren{ T_{\ell'}}} K^{\paren{ \ell'}}$, and $Q^{\paren{ T_\ell}} = C^{\paren{ T_\ell}} Q^{\paren{ \ell}}$, we have
    \begin{align*}
        \left \|Q^{\paren{ \ell}}\right \|_\infty, \left \|K^{\paren{ \ell'}}\right \|_\infty \leq o\paren{ \sqrt{\log n}}.
    \end{align*}

    Additionally, we have
    \begin{align*}
        \paren{ A^{\paren{ T_\ell, T_{\ell'}}}}_{p, q}
        = & ~ \sum_{k = 1}^d \paren{ Q^{\paren{ T_\ell}}}_{p, k} \paren{ K^{\paren{ T_{\ell'}}}}_{q, k}\\
        = & ~ \sum_{k = 1}^d \paren{ C^{\paren{ T_\ell}} Q^{\paren{ \ell}}}_{p, k} \paren{ C^{\paren{ T_{\ell'}}} K^{\paren{ \ell'}}}_{q, k}\\
        = & ~ C^{\paren{ T_\ell}} C^{\paren{ T_{\ell'}}} \sum_{k = 1}^d \paren{ Q^{\paren{ \ell}}}_{p, k} \paren{ K^{\paren{ \ell'}}}_{q, k}\\
        = & ~ C^{\paren{ T_\ell}} C^{\paren{ T_{\ell'}}} \cdot Q^{\paren{ \ell}} \paren{ K^{\paren{ \ell'}}}^\top.
    \end{align*}

    Defining $C^{\paren{ T_\ell, T_{\ell'}}} := C^{\paren{ T_\ell}} C^{\paren{ T_{\ell'}}}$, we have
    \begin{align*}
        C^{\paren{ T_\ell}} C^{\paren{ T_{\ell'}}} 
        = & ~ \frac{b \paren{ 1 + \epsilon}^{\ell}}{\sqrt{\log n}} \frac{b \paren{ 1 + \epsilon}^{\ell'}}{\sqrt{\log n}} \\
        = & ~ \frac{b^2 \paren{ 1 + \epsilon}^{\ell + \ell'}}{\log n}.
    \end{align*}

\end{proof}

\subsection{Basic Facts}
\label{sub:multi_basis_threshold:fact}

Now, we use the Mean Value Theorem to show that for all arbitrary matrices $A, B$, positive integers $p$, $\beta = \max\left \{\left \|A\right \|_\infty, \left \|B\right \|_\infty\right \}$, and $\epsilon > 0$, we can get $\left \|A^{\circ p} - B^{\circ p}\right \|_\infty \leq p \cdot \beta^{p-1} \cdot \epsilon$.

\begin{fact}\label{fac:poly_l_infty_bound}
Let $ A, B \in \mathbb{R}^{n \times n} $, and let $ p \in \mathbb{N} $ with $ p \geq 1 $. Let $\beta := \max\left \{\left \|A\right \|_\infty, \left \|B\right \|_\infty\right \}$.
Suppose $ \left \|A - B\right \|_\infty < \epsilon $. 

Then, we have:
\[
\left \|A^{\circ p} - B^{\circ p}\right \|_\infty \leq p \cdot \beta^{p-1} \cdot \epsilon.
\]
\end{fact}

\begin{proof}
Let $ f\paren{ x} = x^p $, which is differentiable on $ \mathbb{R} $. By the Mean Value Theorem, there exists a point $ c $ between $ a $ and $ b $ such that:
\[
f\paren{ a} - f\paren{ b} = f'\paren{ c}\paren{ a - b}.
\]
Since $ f'\paren{ x} = p x^{p-1} $, we obtain:
\[
a^p - b^p = p c^{p-1}\paren{ a - b}.
\]
Taking absolute values on both sides gives:
\[
|a^p - b^p| = |p c^{p-1}||a - b| = p |c|^{p-1} |a - b|.
\]
Because $ c $ lies between $ a $ and $ b $, we have $ |c| \leq \max\left \{|a|, |b|\right \} $. Therefore, we have
\begin{align*}
    |a^p - b^p| \leq p \cdot \max\left \{|a|, |b|\right \}^{p-1} \cdot |a - b| < p \cdot \max\left \{|a|, |b|\right \}^{p-1} \cdot \epsilon. 
\end{align*}
\end{proof}

\begin{remark}\label{rem:abuse_notation}
By Lemma~\ref{lem:disjoint_decomposition_entrywise_exponential}, we have 
\begin{align*}
    \exp\paren{QK^\top / d} = \sum_{\ell = 1}^{m} \sum_{\ell' = 1}^{m} \exp\paren{A^{\paren{T_\ell, T_{\ell'}}} / d} - \paren{m^2 - 1}\cdot \mathbf{1}_{n \times n}.
\end{align*}
With some abuse of notation, we may say 
\begin{align*}
    \exp\paren{QK^\top / d} = \sum_{\ell = 1}^{m} \sum_{\ell' = 1}^{m} \exp\paren{A^{\paren{T_\ell, T_{\ell'}}} / d}.
\end{align*}

This does not affect the correctness and running time of our algorithm. We note that since the collection $\{ A^{\paren{T_l, T_{l'}}}  \}_{T_l, T_{l'} \in  \{T_\ell \}_{\ell = 1}^{m}}$ forms a support basis (as defined in Definition~\ref{def:support_basis}) of $QK^\top$, $\exp\paren{A^{\paren{T_\ell, T_{\ell'}}} / d}$ transforms the zero entries to $\exp(0) = 1$. Subtracting $\paren{m^2 - 1}\cdot \mathbf{1}_{n \times n}$ is equivalent to say that we redefine ``$\exp(0) = 1$'' to ``$\exp(0) = 0$'' in $\sum_{\ell = 1}^{m} \sum_{\ell' = 1}^{m} \exp\paren{A^{\paren{T_\ell, T_{\ell'}}} / d}$.
\end{remark}

We have shown that $Q^{\paren{ T_\ell}} = C^{\paren{ T_\ell}} Q^{\paren{ \ell}}$ and $K^{\paren{ T_{\ell'}}} = C^{\paren{ T_{\ell'}}} K^{\paren{ \ell'}}$ in Lemma~\ref{lem:normalized_decomposition}. Since in this paper, we apply $\exp(A)$ entry-wisely to a matrix, we can get for all $c \in \R$, $\exp(c A) = \exp(A)^{\circ c}$.

\begin{fact}\label{fac:multi_threshold_basic_fact}

Let $n, m, d$ be positive integers where $2 \leq m \leq n$. Let $\epsilon \in \paren{ 0,1}$.
Let $Q = \sum_{\ell = 1}^{m} Q^{\paren{ T_\ell}} \in \mathbb{R}^{n \times d} $ and $K = \sum_{\ell' = 1}^{m} K^{\paren{ T_{\ell'}}} \in \mathbb{R}^{n \times d}$ be defined as in Definition~\ref{def:exp_threshold_Q_K}, with $b = \min_{i \in [n], j \in [d]}\left \{|Q_{i, j}|, |K_{i, j}|\right \}$, $B = \max_{i \in [n], j \in [d]}\left \{|Q_{i, j}|, |K_{i, j}|\right \}$, and for all $\ell \in [m] \cup \left \{0\right \}$, we let $T_\ell = b \paren{ 1 + \epsilon}^\ell$. 
Let $A^{\paren{ T_\ell, T_{\ell'}}} = Q^{\paren{ T_\ell}} \paren{ K^{\paren{ T_{\ell'}}}}^\top =  C^{\paren{ T_\ell, T_{\ell'}}} \cdot Q^{\paren{ \ell}} \paren{ K^{\paren{ \ell'}}}^\top \in \R^{n \times n}$. Let $m := \left \lfloor \log_{1+\epsilon}\paren{ B/b} \right \rfloor + 1$ and $C^{\paren{ T_\ell, T_{\ell'}}} =  \frac{b^2 \paren{ 1 + \epsilon}^{\ell + \ell'}}{\log n}$.

Then, we have
\begin{itemize}
    \item {\bf Part 1.}
    \begin{align*}
        \exp\paren{QK^\top / d} = \sum_{\ell = 1}^{m} \sum_{\ell' = 1}^{m} \exp\paren{ Q^{\paren{ \ell}} \paren{ K^{\paren{ \ell'}}}^\top /d }^{\circ C^{\paren{ T_\ell, T_{\ell'}}}},
    \end{align*}
    \item {\bf Part 2.} and for all $\paren{\ell, \ell'} \in [m] \times [m]$,
    \begin{align*}
        \left \|Q^{\paren{ \ell}} \paren{ K^{\paren{ \ell'}}}^\top /d \right\|_\infty \leq o\paren{\log n}
    \end{align*}
\end{itemize}

\end{fact}

\begin{proof}
{\bf Proof of Part 1.}

    We have
\begin{align*}
    \exp\paren{QK^\top / d}
    = & ~ \sum_{\ell = 1}^{m} \sum_{\ell' = 1}^{m} \exp\paren{A^{\paren{T_\ell, T_{\ell'}}} / d} \\
    = & ~ \sum_{\ell = 1}^{m} \sum_{\ell' = 1}^{m} \exp\paren{C^{\paren{ T_\ell, T_{\ell'}}} \cdot Q^{\paren{ \ell}} \paren{ K^{\paren{ \ell'}}}^\top / d} \\
    = & ~ \sum_{\ell = 1}^{m} \sum_{\ell' = 1}^{m} \exp\paren{Q^{\paren{ \ell}} \paren{ K^{\paren{ \ell'}}}^\top / d}^{\circ C^{\paren{ T_\ell, T_{\ell'}}}},
\end{align*}
where the first step follows from Remark~\ref{rem:abuse_notation}, the second step follows from the lemma statement, and the last step follows from $\exp(cA) = \exp(A)^{\circ c}$ with the degree $c$ is applied entry-wisely to the matrix $\exp(A)$.

{\bf Proof of Part 2.}

Similar as Lemma~\ref{lem:bounded_entry}, for all arbitrary $\paren{\ell, \ell'} \in [m] \times [m]$, we have
\begin{align*}
    \left \|Q^{\paren{ \ell}} \paren{ K^{\paren{ \ell'}}}^\top \right \|_\infty
    = & ~ \max_{\paren{i, j} \in [n] \times [n]} \left|\paren{Q^{\paren{ \ell}} \paren{ K^{\paren{ \ell'}}}^\top}_{i, j} \right|\\
    = & ~ \max_{\paren{i, j} \in [n] \times [n]} \left|\sum_{l = 1}^d Q^{\paren{ \ell}}_{i, l} K^{\paren{ \ell'}}_{j, l} \right|\\
    \leq & ~ \sum_{l = 1}^d \left |\max_{i \in [n]} Q^{\paren{ \ell}}_{i, l} \right | \cdot \left |\max_{j \in [n]} K^{\paren{ \ell'}}_{j, l} \right |\\
    \leq & ~ \sum_{l = 1}^d o\paren{\sqrt{\log n}}^2\\
    = & ~ o\paren{d \log n},
\end{align*}
where the first step follows from the definition of the $\ell_\infty$ norm, the third step follows from the triangle inequality (see Fact~\ref{fac:vector_norm}), and the fourth step follows from Lemma~\ref{lem:normalized_decomposition}.
\end{proof}

\subsection{Reducing the Softmax Attention Matrix to the Polynomial Attention Matrix}
\label{sub:multi_basis_threshold:poly}

We define the polynomial attention matrix, which replaces the exponential in the standard attention formula with a polynomial of a given degree.

\begin{definition}[Polynomial attention matrix \citep{kmz24}]\label{def:poly_attention_matrix}
    Let $p$ be an arbitrary positive integer. Let $Q, K \in \R^{n \times d}$. The polynomial attention matrix $A \in \R^{n \times n}$ is defined as $A := \paren{QK^\top}^{\circ p}$. 
\end{definition}

Now, we show that the $\paren{Q, K}$-softmax-attention matrix is ``close'' to the sum of polynomial attention matrices.

\begin{lemma}\label{lem:l_infty_error_with_bucketing}
Let $n, m, d$ be positive integers where $2 \leq m \leq n$. Let $\epsilon \in \paren{ 0,1}$.
Let $Q = \sum_{\ell = 1}^{m} Q^{\paren{ T_\ell}} \in \mathbb{R}^{n \times d} $ and $K = \sum_{\ell' = 1}^{m} K^{\paren{ T_{\ell'}}} \in \mathbb{R}^{n \times d}$ be defined as in Definition~\ref{def:exp_threshold_Q_K}, with $b = \min_{i \in [n], j \in [d]}\left \{|Q_{i, j}|, |K_{i, j}|\right \}$, $B = \max_{i \in [n], j \in [d]}\left \{|Q_{i, j}|, |K_{i, j}|\right \}$, and for all $\ell \in [m] \cup \left \{0\right \}$, we let $T_\ell = b \paren{ 1 + \epsilon}^\ell$.
Let $A^{\paren{ T_\ell, T_{\ell'}}} = Q^{\paren{ T_\ell}} \paren{ K^{\paren{ T_{\ell'}}}}^\top =  C^{\paren{ T_\ell, T_{\ell'}}} \cdot Q^{\paren{ \ell}} \paren{ K^{\paren{ \ell'}}}^\top \in \R^{n \times n}$.

Let $m := \left \lfloor \log_{1+\epsilon}\paren{ B/b} \right \rfloor + 1$ and $C^{\paren{ T_\ell, T_{\ell'}}} =  \frac{b^2 \paren{ 1 + \epsilon}^{\ell + \ell'}}{\log n}$. Let $\beta = \max \left \{\left \|Q\right \|_\infty, \left \|K\right \|_\infty\right \}$.

Suppose by Lemma~\ref{lem:wt_A_small_rank}, there exists $U_1^{\paren{ \ell, \ell'}}, U_2^{\paren{ \ell, \ell'}} \in \R^{n \times r}$ such that $U_1^{\paren{ \ell, \ell'}} \paren{ U_2^{\paren{ \ell, \ell'}}}^\top$ is the $\paren{ \epsilon_0, r}$-approximation of $\exp\paren{ Q^{\paren{ \ell}} \paren{ K^{\paren{ \ell'}}}^\top}$.

Then, we have
\begin{align*}
    \left \|\exp\paren{QK^\top / d} - \sum_{\ell = 1}^{m} \sum_{\ell' = 1}^{m} \paren{ U_1^{\paren{ \ell, \ell'}} \paren{ U_2^{\paren{ \ell, \ell'}}}^\top}^{\circ C^{\paren{ T_\ell, T_{\ell'}}}}\right \|_\infty
        \leq O\paren{\frac{B^2}{\log n} e^{o\paren{B^2}} \epsilon_0}.
\end{align*}
\end{lemma}

\begin{proof}

    By {\bf Part 1} of Fact~\ref{fac:multi_threshold_basic_fact}, we have
    \begin{align*}
        \exp\paren{QK^\top / d} = \sum_{\ell = 1}^{m} \sum_{\ell' = 1}^{m} \exp\paren{ Q^{\paren{ \ell}} \paren{ K^{\paren{ \ell'}}}^\top /d }^{\circ C^{\paren{ T_\ell, T_{\ell'}}}}.
    \end{align*}

    By Lemma~\ref{lem:wt_A_small_rank}, since for all $\paren{\ell, \ell'} \in [m] \times [m]$, we have $\left \|Q^{\paren{ \ell}} \paren{ K^{\paren{ \ell'}}}^\top /d \right \|_\infty \leq o\paren{\log n}$ (see {\bf Part 2} of Fact~\ref{fac:multi_threshold_basic_fact}), we can construct $U_1^{\paren{ \ell, \ell'}}, U_2^{\paren{ \ell, \ell'}} \in \R^{n \times r}$ such that
    \begin{align}\label{eq:wt_A_small_rank_in_proof}
        \left \|\exp\paren{ Q^{\paren{ \ell}} \paren{ K^{\paren{ \ell'}}}^\top / d} - U_1^{\paren{ \ell, \ell'}} \paren{ U_2^{\paren{ \ell, \ell'}}}^\top\right \|_\infty < \epsilon_0.
    \end{align}

    By the triangle inequality (see Fact~\ref{fac:vector_norm}), we have
    \begin{align}\label{eq:exp_circ_C}
        & ~ \left \|\sum_{\ell = 1}^{m} \sum_{\ell' = 1}^{m} \exp\paren{ Q^{\paren{ \ell}} \paren{ K^{\paren{ \ell'}}}^\top /d }^{\circ C^{\paren{ T_\ell, T_{\ell'}}}} - \sum_{\ell = 1}^{m} \sum_{\ell' = 1}^{m} \paren{ U_1^{\paren{ \ell, \ell'}} \paren{ U_2^{\paren{ \ell, \ell'}}}^\top}^{\circ C^{\paren{ T_\ell, T_{\ell'}}}}\right \|_\infty \notag\\
        \leq & ~ \sum_{\ell = 1}^{m} \sum_{\ell' = 1}^{m} \left \|\exp\paren{ Q^{\paren{ \ell}} \paren{ K^{\paren{ \ell'}}}^\top /d }^{\circ C^{\paren{ T_\ell, T_{\ell'}}}} - \paren{ U_1^{\paren{ \ell, \ell'}} \paren{ U_2^{\paren{ \ell, \ell'}}}^\top}^{\circ C^{\paren{ T_\ell, T_{\ell'}}}}\right \|_\infty \notag\\
        \leq & ~ \sum_{\ell = 1}^{m} \sum_{\ell' = 1}^{m} C^{\paren{ T_\ell, T_{\ell'}}} \beta^{C^{\paren{ T_\ell, T_{\ell'}}} - 1} \left \|\exp\paren{ Q^{\paren{ \ell}} \paren{ K^{\paren{ \ell'}}}^\top /d  } - U_1^{\paren{ \ell, \ell'}} \paren{ U_2^{\paren{ \ell, \ell'}}}^\top\right \|_\infty,
    \end{align}
    where the second step follows from Fact~\ref{fac:poly_l_infty_bound} with
    \begin{align*}
        \beta = \max \left\{\left\|\exp\paren{ Q^{\paren{ \ell}} \paren{ K^{\paren{ \ell'}}}^\top /d } \right\|_\infty, \left\|U_1^{\paren{ \ell, \ell'}} \paren{ U_2^{\paren{ \ell, \ell'}}}^\top \right\|_\infty \right\}.
    \end{align*}

    Considering $C^{\paren{ T_\ell, T_{\ell'}}}$, we have 
    \begin{align*}
        \max_{T_\ell, T_{\ell'}} C^{\paren{ T_\ell, T_{\ell'}}} 
        = & ~ C^{\paren{ T_m, T_{m}}} \\
        = & ~ \frac{b^2 \paren{ 1 + \epsilon}^{2m}}{\log n} \\
        = & ~ \frac{B^2}{\log n},
    \end{align*}
    where the first and the second step follows from the fact that $C^{\paren{ T_\ell, T_{\ell'}}} =  \frac{b^2 \paren{ 1 + \epsilon}^{\ell + \ell'}}{\log n}$ is strictly increasing (see from the lemma statement), the third step follows from $B = b \paren{ 1 + \epsilon}^{m}$.

    Considering $\beta$, by Fact~\ref{fac:poly_l_infty_bound}, we have
    \begin{align*}
        \beta 
        = & ~ \max \left\{\left\|\exp\paren{ Q^{\paren{ \ell}} \paren{ K^{\paren{ \ell'}}}^\top /d } \right\|_\infty, \left\|U_1^{\paren{ \ell, \ell'}} \paren{ U_2^{\paren{ \ell, \ell'}}}^\top \right\|_\infty \right\}\\
        \leq & ~ \left\|\exp\paren{ Q^{\paren{ \ell}} \paren{ K^{\paren{ \ell'}}}^\top /d} \right\|_\infty + \epsilon_0\\
        = & ~ \exp\paren{\left\| Q^{\paren{ \ell}} \paren{ K^{\paren{ \ell'}}}^\top  /d  \right \|_\infty }  + \epsilon_0\\
        = & ~ \exp\paren{o\paren{\log n}}  + \epsilon_0\\
        = & ~ n^{o(1)},
    \end{align*}
    where the first step follows from Fact~\ref{fac:poly_l_infty_bound}, the second step follows from Eq.~\eqref{eq:wt_A_small_rank_in_proof}, and the fourth step follows from {\bf Part 2} of Fact~\ref{fac:multi_threshold_basic_fact}.

    Combining everything together, we have
    \begin{align*}
        & ~ \left \|\sum_{\ell = 1}^{m} \sum_{\ell' = 1}^{m} \exp\paren{ Q^{\paren{ \ell}} \paren{ K^{\paren{ \ell'}}}^\top /d }^{\circ C^{\paren{ T_\ell, T_{\ell'}}}} - \sum_{\ell = 1}^{m} \sum_{\ell' = 1}^{m} \paren{ U_1^{\paren{ \ell, \ell'}} \paren{ U_2^{\paren{ \ell, \ell'}}}^\top}^{\circ C^{\paren{ T_\ell, T_{\ell'}}}}\right \|_\infty\\
        \leq & ~ O\paren{\frac{B^2}{\log n} n^\frac{o(1) \cdot B^2}{\log n} \epsilon_0}
    \end{align*}
\end{proof}

\subsection{Reducing to Polynomial Attention Without Bucketing}
\label{sub:multi_basis_threshold:nobucket}

Additionally, we note that the bucketing technique does not improve the $\ell_\infty$-closeness between the polynomial attention matrix and the $\paren{Q, K}$-softmax attention matrix. Without using this bucketing technique, we obtain only a single polynomial attention matrix, which is precisely the case studied in \cite{kmz24}.

\begin{lemma}\label{lem:without_bucketing}

Let $n, m, d$ be positive integers. Let $\epsilon > 0$.
Let $Q = \sum_{\ell = 1}^{m} Q^{\paren{ T_\ell}} \in \mathbb{R}^{n \times d} $ and $K = \sum_{\ell' = 1}^{m} K^{\paren{ T_{\ell'}}} \in \mathbb{R}^{n \times d}$ be defined as in Definition~\ref{def:exp_threshold_Q_K}, with $b = \min_{i \in [n], j \in [d]}\left \{|Q_{i, j}|, |K_{i, j}|\right \}$, $B = \max_{i \in [n], j \in [d]}\left \{|Q_{i, j}|, |K_{i, j}|\right \}$, and for all $\ell \in [m] \cup \left \{0\right \}$, we let $T_\ell = b \paren{ 1 + \epsilon}^\ell$.
Let $A^{\paren{ T_\ell, T_{\ell'}}} = Q^{\paren{ T_\ell}} \paren{ K^{\paren{ T_{\ell'}}}}^\top = C^{\paren{ T_\ell, T_{\ell'}}} \cdot Q^{\paren{ \ell}} \paren{ K^{\paren{ \ell'}}}^\top \in \R^{n \times n}$.
Let $\beta = \max \left \{\left \|Q\right \|_\infty, \left \|K\right \|_\infty\right \}$.

Suppose $m = 1$ and by Lemma~\ref{lem:wt_A_small_rank}, there exists $U_1^{\paren{ 1, 1}}, U_2^{\paren{ 1, 1}} \in \R^{n \times r}$ such that $U_1^{\paren{ 1, 1}} \paren{ U_2^{\paren{ 1, 1}}}^\top$ is the $\paren{ \epsilon_0, r}$-approximation of $\exp\paren{ Q^{\paren{1}} \paren{ K^{\paren{1}}}^\top /d }$.

Then, we have
\begin{align*}
    \left \|\exp\paren{QK^\top / d} - \paren{ U_1^{\paren{ 1, 1}} \paren{ U_2^{\paren{ 1, 1}}}^\top}^{\circ C^{\paren{ T_1, T_{1}}}}\right \|_\infty \leq O\paren{\frac{B^2}{\log n} e^{o\paren{B^2}} \epsilon_0}.
\end{align*}
    
\end{lemma}

\begin{proof}

This proof is similar to that of Lemma~\ref{lem:l_infty_error_with_bucketing}. By {\bf Part 1} of Fact~\ref{fac:multi_threshold_basic_fact}, with $m = 1$, we have
    \begin{align*}
    \exp\paren{QK^\top / d} = \exp\paren{ Q^{\paren{1}} \paren{ K^{\paren{1}}}^\top /d }^{\circ C^{\paren{ T_1, T_{1}}}},
\end{align*}
with 
\begin{align*}
    C^{\paren{ T_1, T_{1}}} 
    = & ~ \frac{b^2 \paren{ 1 + \epsilon}^{1 + 1}}{\log n}\\
    = & ~ \frac{b^2 \paren{ 1 + \epsilon}^{2m}}{\log n}\\
    = & ~ B^2 / \log n,
\end{align*}
where the first step follows from the definition of $C^{\paren{ T_\ell, T_{\ell'}}}$, the second step follows from $m = 1$, and the last step follows from $B = b \paren{ 1 + \epsilon}^{m}$.

    By Lemma~\ref{lem:wt_A_small_rank}, since we have $\left \|Q^{\paren{ 1}} \paren{ K^{\paren{ 1}}}^\top /d \right \|_\infty \leq o\paren{\log n}$ (see {\bf Part 2} of Fact~\ref{fac:multi_threshold_basic_fact}), we can construct $U_1^{\paren{ 1, 1}}, U_2^{\paren{ 1, 1}} \in \R^{n \times r}$ such that
    \begin{align}\label{eq:wt_A_small_rank_in_proof_2}
        \left \|\exp\paren{ Q^{\paren{ 1}} \paren{ K^{\paren{ 1}}}^\top / d} - U_1^{\paren{ 1, 1}} \paren{ U_2^{\paren{ 1, 1}}}^\top\right \|_\infty < \epsilon_0.
    \end{align}

We have
    \begin{align*}
        & ~ \left \|\exp\paren{ Q^{\paren{1}} \paren{ K^{\paren{1}}}^\top /d }^{\circ C^{\paren{ T_1, T_{1}}}} - \paren{ U_1^{\paren{ 1, 1}} \paren{ U_2^{\paren{ 1, 1}}}^\top}^{\circ C^{\paren{ T_1, T_{1}}}}\right \|_\infty\\
        \leq & ~ C^{\paren{ T_1, T_{1}}} \beta^{C^{\paren{ T_1, T_{1}}} - 1} \left \|\exp\paren{ Q^{\paren{ 1}} \paren{ K^{\paren{ 1}}}^\top /d  } - U_1^{\paren{ 1, 1}} \paren{ U_2^{\paren{ 1, 1}}}^\top\right \|_\infty\\
        = & ~ C^{\paren{ T_1, T_{1}}} \beta^{C^{\paren{ T_1, T_{1}}} - 1} \epsilon_0,
    \end{align*}
    where the first step follows from Fact~\ref{fac:poly_l_infty_bound} and the second step follows from Eq.~\eqref{eq:wt_A_small_rank_in_proof_2}.

    Considering $\beta$, we have
    \begin{align*}
        \beta 
        = & ~ \max \left\{\left\|\exp\paren{ Q^{\paren{ 1}} \paren{ K^{\paren{ 1}}}^\top /d } \right\|_\infty, \left\|U_1^{\paren{ 1, 1}} \paren{ U_2^{\paren{ 1, 1}}}^\top \right\|_\infty \right\}\\
        \leq & ~ \left\|\exp\paren{ Q^{\paren{ 1}} \paren{ K^{\paren{ 1}}}^\top /d} \right\|_\infty + \epsilon_0\\
        = & ~ \exp\paren{\left\| Q^{\paren{ 1}} \paren{ K^{\paren{ 1}}}^\top  /d  \right \|_\infty }  + \epsilon_0\\
        = & ~ \exp\paren{o\paren{\log n}}  + \epsilon_0\\
        = & ~ n^{o(1)},
    \end{align*}
     where the first step follows from Fact~\ref{fac:poly_l_infty_bound}, the second step follows from Eq.~\eqref{eq:wt_A_small_rank_in_proof_2}, and the fourth step follows from {\bf Part 2} of Fact~\ref{fac:multi_threshold_basic_fact}.

    Combining everything together, we have
    \begin{align*}
        \left \|\exp\paren{ Q^{\paren{1}} \paren{ K^{\paren{1}}}^\top /d }^{\circ C^{\paren{ T_1, T_{1}}}} - \paren{ U_1^{\paren{ 1, 1}} \paren{ U_2^{\paren{ 1, 1}}}^\top}^{\circ C^{\paren{ T_1, T_{1}}}}\right \|_\infty
        \leq O\paren{\frac{B^2}{\log n} n^\frac{o(1) \cdot B^2}{\log n} \epsilon_0}.
    \end{align*}

\end{proof}

\subsection{Bucketing Reduces the Error in \texorpdfstring{$\ell_1$}{} Norm}
\label{sub:multi_basis_threshold:l1error}

In the following lemma, we analyze how the $\ell_1$ error between the $\paren{Q, K}$-softmax-attention matrix and its polynomial approximation is affected by the bucketing technique. While prior sections mainly focus on $\ell_\infty$ approximations, the $\ell_1$ norm offers a complementary view by aggregating errors across entries. We show that the sum of multiple polynomial attention matrices generated by the bucketing technique significantly reduces the $\ell_1$ error, thereby providing a stronger global approximation guarantee.

First, we analyze the case where we use the bucketing technique.

\begin{lemma}

Let $n, m, d$ be positive integers where $2 \leq m \leq n$. Let $\epsilon \in \paren{ 0,1}$.
Let $Q = \sum_{\ell = 1}^{m} Q^{\paren{ T_\ell}} \in \mathbb{R}^{n \times d} $ and $K = \sum_{\ell' = 1}^{m} K^{\paren{ T_{\ell'}}} \in \mathbb{R}^{n \times d}$ be defined as in Definition~\ref{def:exp_threshold_Q_K}, with $b = \min_{i \in [n], j \in [d]}\left \{|Q_{i, j}|, |K_{i, j}|\right \}$, $B = \max_{i \in [n], j \in [d]}\left \{|Q_{i, j}|, |K_{i, j}|\right \}$, and for all $\ell \in [m] \cup \left \{0\right \}$, we let $T_\ell = b \paren{ 1 + \epsilon}^\ell$.
Let $A^{\paren{ T_\ell, T_{\ell'}}} = Q^{\paren{ T_\ell}} \paren{ K^{\paren{ T_{\ell'}}}}^\top = C^{\paren{ T_\ell, T_{\ell'}}} \cdot Q^{\paren{ \ell}} \paren{ K^{\paren{ \ell'}}}^\top \in \R^{n \times n}$.

Let $m := \left \lfloor \log_{1+\epsilon}\paren{ B/b} \right \rfloor + 1$ and $C^{\paren{ T_\ell, T_{\ell'}}} =  \frac{b^2 \paren{ 1 + \epsilon}^{\ell + \ell'}}{\log n}$. Let $\beta = \max \left \{\left \|Q\right \|_\infty, \left \|K\right \|_\infty\right \}$.

Suppose by Lemma~\ref{lem:wt_A_small_rank}, there exists $U_1^{\paren{ \ell, \ell'}}, U_2^{\paren{ \ell, \ell'}} \in \R^{n \times r}$ such that $U_1^{\paren{ \ell, \ell'}} \paren{ U_2^{\paren{ \ell, \ell'}}}^\top$ is the $\paren{ \epsilon_0, r}$-approximation of $\exp\paren{ Q^{\paren{ \ell}} \paren{ K^{\paren{ \ell'}}}^\top}$.

Then, we have
        \begin{align*}
            & ~ \left \|\sum_{\ell = 1}^{m} \sum_{\ell' = 1}^{m} \exp\paren{ Q^{\paren{ \ell}} \paren{ K^{\paren{ \ell'}}}^\top /d }^{\circ C^{\paren{ T_\ell, T_{\ell'}}}} - \sum_{\ell = 1}^{m} \sum_{\ell' = 1}^{m} \paren{ U_1^{\paren{ \ell, \ell'}} \paren{ U_2^{\paren{ \ell, \ell'}}}^\top}^{\circ C^{\paren{ T_\ell, T_{\ell'}}}}\right \|_1 \\
            \leq & ~ O\paren{\frac{n^2 \epsilon_0 e^{o\paren{B^2}}}{ \log\left(n\right) \log\left({\epsilon} + 1\right)}}.
        \end{align*}
\end{lemma}
\begin{proof}

    Now, we consider the $\ell_1$ error. We have
    \begin{align*}
        & ~ \left \|\sum_{\ell = 1}^{m} \sum_{\ell' = 1}^{m} \exp\paren{ Q^{\paren{ \ell}} \paren{ K^{\paren{ \ell'}}}^\top /d }^{\circ C^{\paren{ T_\ell, T_{\ell'}}}} - \sum_{\ell = 1}^{m} \sum_{\ell' = 1}^{m} \paren{ U_1^{\paren{ \ell, \ell'}} \paren{ U_2^{\paren{ \ell, \ell'}}}^\top}^{\circ C^{\paren{ T_\ell, T_{\ell'}}}}\right \|_1 \\
        = & ~ \sum_{i = 1}^n \sum_{j = 1}^n \paren{\sum_{\ell = 1}^{m} \sum_{\ell' = 1}^{m} \paren{\exp\paren{ Q^{\paren{ \ell}} \paren{ K^{\paren{ \ell'}}}^\top /d }^{\circ C^{\paren{ T_\ell, T_{\ell'}}}} - \paren{ U_1^{\paren{ \ell, \ell'}} \paren{ U_2^{\paren{ \ell, \ell'}}}^\top}^{\circ C^{\paren{ T_\ell, T_{\ell'}}}}} }_{i, j}\\
        = & ~ \sum_{i = 1}^n \sum_{j = 1}^n \sum_{\ell = 1}^{m} \sum_{\ell' = 1}^{m} \paren{\exp\paren{ Q^{\paren{ \ell}} \paren{ K^{\paren{ \ell'}}}^\top /d }_{i, j}^{C^{\paren{ T_\ell, T_{\ell'}}}} - \paren{ U_1^{\paren{ \ell, \ell'}} \paren{ U_2^{\paren{ \ell, \ell'}}}^\top}_{i, j}^{C^{\paren{ T_\ell, T_{\ell'}}}}} \\
        = & ~ \sum_{\ell = 1}^{m} \sum_{\ell' = 1}^{m} \paren{\sum_{i = 1}^n \sum_{j = 1}^n \exp\paren{ Q^{\paren{ \ell}} \paren{ K^{\paren{ \ell'}}}^\top /d }_{i, j}^{C^{\paren{ T_\ell, T_{\ell'}}}} - \paren{ U_1^{\paren{ \ell, \ell'}} \paren{ U_2^{\paren{ \ell, \ell'}}}^\top}_{i, j}^{C^{\paren{ T_\ell, T_{\ell'}}}}} \\
        \leq & ~ \sum_{\ell = 1}^{m} \sum_{\ell' = 1}^{m} \paren{ \left |\mathrm{supp}\paren{ \exp\paren{ Q^{\paren{ \ell}} \paren{ K^{\paren{ \ell'}}}^\top /d }^{\circ C^{\paren{ T_\ell, T_{\ell'}}}}} \right | C^{\paren{ T_\ell, T_{\ell'}}} \beta^{C^{\paren{ T_\ell, T_{\ell'}}} - 1} \epsilon_0} \\
        = & ~ \epsilon_0 \sum_{\ell = 1}^{m} \sum_{\ell' = 1}^{m} \paren{ \left|\supp\paren{A^{\paren{\ell, \ell'}}}\right|
        \frac{b^2 \paren{ 1 + \epsilon}^{\ell + \ell'}}{\log n} n^{o\paren{\frac{b^2 \paren{ 1 + \epsilon}^{\ell + \ell'}}{\log n}}}},
        \end{align*}
        where the first step follows from definition of $\ell_1$ norm, the third step follows from exchanging the order of the summation, the fourth step follows from the mean value theorem (see Fact~\ref{fac:poly_l_infty_bound}) and the fact that there are 
        \begin{align*}
            \left |\mathrm{supp}\paren{ \exp\paren{ Q^{\paren{ \ell}} \paren{ K^{\paren{ \ell'}}}^\top /d }^{\circ C^{\paren{ T_\ell, T_{\ell'}}}}} \right |
        \end{align*}
        amount of non-zero entries to approximate, and the last step follows from the definition of $A^{\paren{\ell, \ell'}}$, $\beta$, and $C^{\paren{ T_\ell, T_{\ell'}}}$ (see from the lemma statement).

        Furthermore, by using the Cauchy--Schwarz inequality (see Fact~\ref{fac:vector_norm}), we can get:
        \begin{align}\label{eq:ell_1_error}
        & ~ \left \|\sum_{\ell = 1}^{m} \sum_{\ell' = 1}^{m} \exp\paren{ Q^{\paren{ \ell}} \paren{ K^{\paren{ \ell'}}}^\top /d }^{\circ C^{\paren{ T_\ell, T_{\ell'}}}} - \sum_{\ell = 1}^{m} \sum_{\ell' = 1}^{m} \paren{ U_1^{\paren{ \ell, \ell'}} \paren{ U_2^{\paren{ \ell, \ell'}}}^\top}^{\circ C^{\paren{ T_\ell, T_{\ell'}}}}\right \|_1 \notag\\
        = & ~ \epsilon_0 \paren{ \sum_{\ell = 1}^{m} \sum_{\ell' = 1}^{m} \left|\supp\paren{A^{\paren{\ell, \ell'}}}\right|^2}^{1/2} \cdot \paren{\sum_{\ell = 1}^{m} \sum_{\ell' = 1}^{m}
        \frac{b^4 \paren{ 1 + \epsilon}^{2\ell + 2\ell'}}{\log^2 n} n^{o\paren{\frac{b^2 \paren{ 1 + \epsilon}^{\ell + \ell'}}{\log n}}}}^{1/2} \notag\\
        \leq & ~ \epsilon_0 \paren{ \sum_{\ell = 1}^{m} \sum_{\ell' = 1}^{m} \left|\supp\paren{A^{\paren{\ell, \ell'}}}\right|} \cdot \paren{\sum_{\ell = 1}^{m} \sum_{\ell' = 1}^{m}
        \frac{b^4 \paren{ 1 + \epsilon}^{2\ell + 2\ell'}}{\log^2 n} n^{o\paren{\frac{b^2 \paren{ 1 + \epsilon}^{\ell + \ell'}}{\log n}}}}^{1/2} \notag\\
        \leq & ~ \epsilon_0 \paren{ \sum_{\ell = 1}^{m} \sum_{\ell' = 1}^{m} \left|\supp\paren{A^{\paren{\ell, \ell'}}}\right|} \cdot \paren{\int_1^{m + 1} \int_1^{m + 1}
        \frac{b^4 \paren{ 1 + \epsilon}^{2x + 2y}}{\log^2 n} n^{o\paren{\frac{b^2 \paren{ 1 + \epsilon}^{x + y}}{\log n}}} \, \mathrm{d} x \, \mathrm{d} y}^{1/2},
    \end{align}
    where the second step follows from Fact~\ref{fac:vector_norm} and the third step follows from the definition of Riemann integral.

    Since $\{ A^{(\ell, \ell')} \}_{\ell, \ell' \in [m]}$ forms a support basis of $QK^\top$ (see Lemma~\ref{lem:support_basis}), by applying Fact~\ref{fac:support_of_A}, we have
    \begin{align}\label{eq:supp_bound}
        \sum_{\ell = 1}^{m} \sum_{\ell' = 1}^{m} \left|\supp\paren{A^{\paren{\ell, \ell'}}}\right| \leq n^2.
    \end{align}

    Additionally, for all $c > 0$, we have
    \begin{align}\label{eq:first_integral_bound}
        & ~ \int_1^{m + 1} \frac{b^4 \paren{ 1 + \epsilon}^{2x + 2y}}{\log^2 n} n^{o\paren{\frac{b^2 \paren{ 1 + \epsilon}^{x + y}}{\log n}}} \, \mathrm{d} x \notag\\
        \leq & ~ \frac{cb^2 \paren{ 1 + \epsilon}^{y + m + 1} e^{b^2 c (\epsilon + 1)^{y + m + 1}} - cb^2 \paren{ 1 + \epsilon}^{y + 1} e^{b^2 c (\epsilon + 1)^{y + 1}} - e^{b^2 c (\epsilon + 1)^{y + m + 1}} + e^{b^2 c (\epsilon + 1)^{y + 1}}}{c^2 \log^2 n \log(\epsilon + 1)} \notag\\
        = & ~ O\paren{\frac{b^2 \paren{ 1 + \epsilon}^{y + m + 1} e^{b^2 c (\epsilon + 1)^{y + m + 1}}}{c \log^{2}\left(n\right) \log\left({\epsilon} + 1\right)}},
    \end{align}
    where the first step follows from {\bf Part 1} of Fact~\ref{fac:integral} and the second step follows from the fact that $cb^2 \paren{ 1 + \epsilon}^{y + m + 1} e^{b^2 c (\epsilon + 1)^{y + m + 1}}$ is the dominating term in the numerator. 

    We can further get
    \begin{align}\label{eq:integral_bound}
        \int_1^{m + 1} \int_1^{m + 1}
        \frac{b^4 \paren{ 1 + \epsilon}^{2x + 2y}}{\log^2 n} n^{o\paren{\frac{b^2 \paren{ 1 + \epsilon}^{x + y}}{\log n}}} \, \mathrm{d} x \, \mathrm{d} y
        \leq & ~ \int_1^{m + 1} O\paren{\frac{b^2 \paren{ 1 + \epsilon}^{y + m + 1} e^{b^2 c (\epsilon + 1)^{y + m + 1}}}{c \log^{2}\left(n\right) \log\left({\epsilon} + 1\right)}} \, \mathrm{d} y \notag \\
        = & ~ O\paren{\int_1^{m + 1} \frac{b^2 \paren{ 1 + \epsilon}^{y + m + 1} e^{b^2 c (\epsilon + 1)^{y + m + 1}}}{c \log^{2}\left(n\right) \log\left({\epsilon} + 1\right)} \, \mathrm{d} y } \notag\\
        = & ~ O\paren{\frac{e^{b^2 c (\epsilon + 1)^{2m + 2}} - e^{b^2 c (\epsilon + 1)^{m + 2}}}{c^2 \log^2(n) \log^2(\epsilon + 1)}} \notag\\
        = & ~ O\paren{\frac{e^{B^2 c (\epsilon + 1)^{2}}}{c^2 \log^2(n) \log^2(\epsilon + 1)}},
    \end{align}
    where the first step follows from Eq.~\eqref{eq:first_integral_bound}, the third step follows from {\bf Part 2} of Fact~\ref{fac:integral}, and the last step follows from the fact that $e^{b^2 c (\epsilon + 1)^{2m + 2}}$ is the dominating term in numerator.

    Combining Eq.~\eqref{eq:ell_1_error}, Eq.~\eqref{eq:supp_bound}, and Eq.~\eqref{eq:integral_bound} together, we have
    \begin{align*}
        & ~ \left \|\sum_{\ell = 1}^{m} \sum_{\ell' = 1}^{m} \exp\paren{ Q^{\paren{ \ell}} \paren{ K^{\paren{ \ell'}}}^\top /d }^{\circ C^{\paren{ T_\ell, T_{\ell'}}}} - \sum_{\ell = 1}^{m} \sum_{\ell' = 1}^{m} \paren{ U_1^{\paren{ \ell, \ell'}} \paren{ U_2^{\paren{ \ell, \ell'}}}^\top}^{\circ C^{\paren{ T_\ell, T_{\ell'}}}}\right \|_1 \\
        \leq & ~ O\paren{\frac{n^2 \epsilon_0 e^{o\paren{B^2}}}{ \log\left(n\right) \log\left({\epsilon} + 1\right)}}
    \end{align*}
\end{proof}

Then, we analyze the case where we do not use the bucketing technique.

\begin{lemma}

Let $n, m, d$ be positive integers. Let $\epsilon \in \paren{ 0,1}$.
Let $Q = \sum_{\ell = 1}^{m} Q^{\paren{ T_\ell}} \in \mathbb{R}^{n \times d} $ and $K = \sum_{\ell' = 1}^{m} K^{\paren{ T_{\ell'}}} \in \mathbb{R}^{n \times d}$ be defined as in Definition~\ref{def:exp_threshold_Q_K}, with $b = \min_{i \in [n], j \in [d]}\left \{|Q_{i, j}|, |K_{i, j}|\right \}$, $B = \max_{i \in [n], j \in [d]}\left \{|Q_{i, j}|, |K_{i, j}|\right \}$, and for all $\ell \in [m] \cup \left \{0\right \}$, we let $T_\ell = b \paren{ 1 + \epsilon}^\ell$.
Let $A^{\paren{ T_\ell, T_{\ell'}}} = Q^{\paren{ T_\ell}} \paren{ K^{\paren{ T_{\ell'}}}}^\top = C^{\paren{ T_\ell, T_{\ell'}}} \cdot Q^{\paren{ \ell}} \paren{ K^{\paren{ \ell'}}}^\top \in \R^{n \times n}$.
Let $\beta = \max \left \{\left \|Q\right \|_\infty, \left \|K\right \|_\infty\right \}$.

Suppose $m = 1$, which implies 
\begin{align*}
    \exp\paren{QK^\top / d} = \exp\paren{ Q^{\paren{m}} \paren{ K^{\paren{m}}}^\top /d }^{\circ C^{\paren{ T_m, T_{m}}}},
\end{align*}
where $C^{\paren{ T_m, T_{m}}} = B^2 / \log n$.

Suppose by Lemma~\ref{lem:wt_A_small_rank}, there exists $U_1^{\paren{ m, m}}, U_2^{\paren{ m, m}} \in \R^{n \times r}$ such that $U_1^{\paren{ m, m}} \paren{ U_2^{\paren{ m, m}}}^\top$ is the $\paren{ \epsilon_0, r}$-approximation of $\exp\paren{ Q^{\paren{m}} \paren{ K^{\paren{m}}}^\top /d }$.

Then, we have
\begin{align*}
    \left \|\exp\paren{QK^\top / d} - \paren{ U_1^{\paren{ m, m}} \paren{ U_2^{\paren{ m, m}}}^\top}^{\circ C^{\paren{ T_m, T_{m}}}}\right \|_1 \leq O\paren{\frac{n^2B^2}{\log n} e^{o\paren{B^2}} \epsilon_0}.
\end{align*}
    
\end{lemma}

\begin{proof}

We have
    \begin{align*}
    & ~ \left \|\exp\paren{QK^\top / d} - \paren{ U_1^{\paren{ m, m}} \paren{ U_2^{\paren{ m, m}}}^\top}^{\circ C^{\paren{ T_m, T_{m}}}} \right \|_1 \\
    = & ~ \sum_{i = 1}^n \sum_{j = 1}^n \paren{\exp\paren{QK^\top / d}_{i, j} - \paren{ U_1^{\paren{ m, m}} \paren{ U_2^{\paren{ m, m}}}^\top}_{i, j}^{C^{\paren{ T_m, T_{m}}}}} \\
    \leq & ~ \sum_{i = 1}^n \sum_{j = 1}^n O\paren{\frac{B^2}{\log n} e^{o\paren{B^2}} \epsilon_0} \\
    = & ~ O\paren{\frac{n^2B^2}{\log n} e^{o\paren{B^2}} \epsilon_0},
\end{align*}
where the first step follows from the definition of $\ell_1$ norm and the second step follows from Lemma~\ref{lem:without_bucketing}.
\end{proof}

\subsection{Bucketing Reduces the Error in \texorpdfstring{$\ell_p$}{} Norm}
\label{sub:multi_basis_threshold:lperror}

\begin{lemma}

Let $n, m, d$ be positive integers where $2 \leq m \leq n$. Let $\epsilon \in \paren{ 0,1}$.
Let $Q = \sum_{\ell = 1}^{m} Q^{\paren{ T_\ell}} \in \mathbb{R}^{n \times d} $ and $K = \sum_{\ell' = 1}^{m} K^{\paren{ T_{\ell'}}} \in \mathbb{R}^{n \times d}$ be defined as in Definition~\ref{def:exp_threshold_Q_K}, with $b = \min_{i \in [n], j \in [d]}\left \{|Q_{i, j}|, |K_{i, j}|\right \}$, $B = \max_{i \in [n], j \in [d]}\left \{|Q_{i, j}|, |K_{i, j}|\right \}$, and for all $\ell \in [m] \cup \left \{0\right \}$, we let $T_\ell = b \paren{ 1 + \epsilon}^\ell$.
Let $A^{\paren{ T_\ell, T_{\ell'}}} = Q^{\paren{ T_\ell}} \paren{ K^{\paren{ T_{\ell'}}}}^\top = C^{\paren{ T_\ell, T_{\ell'}}} \cdot Q^{\paren{ \ell}} \paren{ K^{\paren{ \ell'}}}^\top \in \R^{n \times n}$.

Let $m := \left \lfloor \log_{1+\epsilon}\paren{ B/b} \right \rfloor + 1$ and $C^{\paren{ T_\ell, T_{\ell'}}} =  \frac{b^2 \paren{ 1 + \epsilon}^{\ell + \ell'}}{\log n}$. Let $\beta = \max \left \{\left \|Q\right \|_\infty, \left \|K\right \|_\infty\right \}$.

Suppose by Lemma~\ref{lem:wt_A_small_rank}, there exists $U_1^{\paren{ \ell, \ell'}}, U_2^{\paren{ \ell, \ell'}} \in \R^{n \times r}$ such that $U_1^{\paren{ \ell, \ell'}} \paren{ U_2^{\paren{ \ell, \ell'}}}^\top$ is the $\paren{ \epsilon_0, r}$-approximation of $\exp\paren{ Q^{\paren{ \ell}} \paren{ K^{\paren{ \ell'}}}^\top}$.

Then, we can get a tighter bound for
        \begin{align*}
            & ~ \left \|\sum_{\ell = 1}^{m} \sum_{\ell' = 1}^{m} \exp\paren{ Q^{\paren{ \ell}} \paren{ K^{\paren{ \ell'}}}^\top /d }^{\circ C^{\paren{ T_\ell, T_{\ell'}}}} - \sum_{\ell = 1}^{m} \sum_{\ell' = 1}^{m} \paren{ U_1^{\paren{ \ell, \ell'}} \paren{ U_2^{\paren{ \ell, \ell'}}}^\top}^{\circ C^{\paren{ T_\ell, T_{\ell'}}}}\right \|_p 
        \end{align*}
        compared with
        \begin{align*}
    \left \|\exp\paren{QK^\top / d} - \paren{ U_1^{\paren{ m, m}} \paren{ U_2^{\paren{ m, m}}}^\top}^{\circ C^{\paren{ T_m, T_{m}}}}\right \|_p.
\end{align*}
\end{lemma}
\begin{proof}

    Now, we consider the $\ell_p$ error. 
    
    Using {\bf Part 2} of Fact~\ref{fac:more_disjoint_prop}, we have
    \begin{align}\label{eq:ell_p_error}
        & ~ \left \|\sum_{\ell = 1}^{m} \sum_{\ell' = 1}^{m} \exp\paren{ Q^{\paren{ \ell}} \paren{ K^{\paren{ \ell'}}}^\top /d }^{\circ C^{\paren{ T_\ell, T_{\ell'}}}} - \sum_{\ell = 1}^{m} \sum_{\ell' = 1}^{m} \paren{ U_1^{\paren{ \ell, \ell'}} \paren{ U_2^{\paren{ \ell, \ell'}}}^\top}^{\circ C^{\paren{ T_\ell, T_{\ell'}}}}\right \|_p^p \notag\\
        = & ~ \sum_{\ell = 1}^{m} \sum_{\ell' = 1}^{m}  \left \|\exp\paren{ Q^{\paren{ \ell}} \paren{ K^{\paren{ \ell'}}}^\top /d }^{\circ C^{\paren{ T_\ell, T_{\ell'}}}} - \paren{ U_1^{\paren{ \ell, \ell'}} \paren{ U_2^{\paren{ \ell, \ell'}}}^\top}^{\circ C^{\paren{ T_\ell, T_{\ell'}}}} \right \|_p^p.
        \end{align}

        Below, we apply the same multi-threshold support basis for $\exp\paren{QK^\top / d}$, but for each matrix, we take out the constant $C^{\paren{ T_m, T_{m}}}$ to simulate the case where we do not use bucketing.

        Therefore, we can get
        \begin{align*}
            & ~ \left \|\exp\paren{QK^\top / d} - \paren{ U_1^{\paren{\ell, \ell'}} \paren{ U_2^{\paren{\ell, \ell'}}}^\top}^{\circ C^{\paren{ T_m, T_{m}}}} \right \|_p^p \\
            = & ~ \left \|\sum_{\ell = 1}^{m} \sum_{\ell' = 1}^{m} \exp\paren{ Q^{\paren{ \ell}} \paren{ K^{\paren{ \ell'}}}^\top /d }^{\circ C^{\paren{ T_m, T_{m}}}} - \sum_{\ell = 1}^{m} \sum_{\ell' = 1}^{m} \paren{ U_1^{\paren{\ell, \ell'}} \paren{ U_2^{\paren{\ell, \ell'}}}^\top}^{\circ C^{\paren{ T_m, T_{m}}}} \right \|_p^p \\
            = & ~ \left \|\sum_{\ell = 1}^{m} \sum_{\ell' = 1}^{m} \paren{\exp\paren{ Q^{\paren{ \ell}} \paren{ K^{\paren{ \ell'}}}^\top /d }^{\circ C^{\paren{ T_m, T_{m}}}} - \paren{ U_1^{\paren{\ell, \ell'}} \paren{ U_2^{\paren{\ell, \ell'}}}^\top}^{\circ C^{\paren{ T_m, T_{m}}}}} \right \|_p^p \\
            \leq & ~ \sum_{\ell = 1}^{m} \sum_{\ell' = 1}^{m} \left \|\exp\paren{ Q^{\paren{ \ell}} \paren{ K^{\paren{ \ell'}}}^\top /d }^{\circ C^{\paren{ T_m, T_{m}}}} - \paren{ U_1^{\paren{\ell, \ell'}} \paren{ U_2^{\paren{\ell, \ell'}}}^\top}^{\circ C^{\paren{ T_m, T_{m}}}} \right \|_p^p,
        \end{align*}
        where the second step follows from the distributive law and the third step follows from {\bf Part 2} of Fact~\ref{fac:more_disjoint_prop}.

        Applying {\bf Part 1} and {\bf Part 3} of Fact~\ref{fac:more_disjoint_prop}, we can get
        \begin{align*}
            \left \| A^{\circ k} + B^{\circ k} \right\|_p^p
            = & ~ \left \| \paren{A + B}^{\circ k} \right\|_p^p
            \leq \left \| A + B \right\|_p^{p k},
        \end{align*}
        and for each $\ell$ and $\ell'$, we can use it to show:
        \begin{align*}
            & ~ \left \|\exp\paren{ Q^{\paren{ \ell}} \paren{ K^{\paren{ \ell'}}}^\top /d }^{\circ C^{\paren{ T_\ell, T_{\ell'}}}} - \paren{ U_1^{\paren{ \ell, \ell'}} \paren{ U_2^{\paren{ \ell, \ell'}}}^\top}^{\circ C^{\paren{ T_\ell, T_{\ell'}}}} \right \|_p^p\\
            \leq & ~ \left \|\exp\paren{ Q^{\paren{ \ell}} \paren{ K^{\paren{ \ell'}}}^\top /d } - \paren{ U_1^{\paren{ \ell, \ell'}} \paren{ U_2^{\paren{ \ell, \ell'}}}^\top} \right \|_p^{p \cdot C^{\paren{ T_\ell, T_{\ell'}}}}
        \end{align*}
        which is tighter compared with
        \begin{align*}
            & ~ \left \|\exp\paren{ Q^{\paren{ \ell}} \paren{ K^{\paren{ \ell'}}}^\top /d }^{\circ C^{\paren{ T_m, T_{m}}}} - \paren{ U_1^{\paren{\ell, \ell'}} \paren{ U_2^{\paren{\ell, \ell'}}}^\top}^{\circ C^{\paren{ T_m, T_{m}}}} \right \|_p^p\\
            \leq & ~ \left \|\exp\paren{ Q^{\paren{ \ell}} \paren{ K^{\paren{ \ell'}}}^\top /d } - \paren{ U_1^{\paren{\ell, \ell'}} \paren{ U_2^{\paren{\ell, \ell'}}}^\top} \right \|_p^{p \cdot C^{\paren{ T_m, T_{m}}}}.
        \end{align*}
\end{proof}

Then, we analyze the case where we do not use the bucketing technique.

\begin{lemma}

Let $n, m, d$ be positive integers. Let $\epsilon \in \paren{ 0,1}$.
Let $Q = \sum_{\ell = 1}^{m} Q^{\paren{ T_\ell}} \in \mathbb{R}^{n \times d} $ and $K = \sum_{\ell' = 1}^{m} K^{\paren{ T_{\ell'}}} \in \mathbb{R}^{n \times d}$ be defined as in Definition~\ref{def:exp_threshold_Q_K}, with $b = \min_{i \in [n], j \in [d]}\left \{|Q_{i, j}|, |K_{i, j}|\right \}$, $B = \max_{i \in [n], j \in [d]}\left \{|Q_{i, j}|, |K_{i, j}|\right \}$, and for all $\ell \in [m] \cup \left \{0\right \}$, we let $T_\ell = b \paren{ 1 + \epsilon}^\ell$.
Let $A^{\paren{ T_\ell, T_{\ell'}}} = Q^{\paren{ T_\ell}} \paren{ K^{\paren{ T_{\ell'}}}}^\top = C^{\paren{ T_\ell, T_{\ell'}}} \cdot Q^{\paren{ \ell}} \paren{ K^{\paren{ \ell'}}}^\top \in \R^{n \times n}$.
Let $\beta = \max \left \{\left \|Q\right \|_\infty, \left \|K\right \|_\infty\right \}$.

Suppose $m = 1$, which implies 
\begin{align*}
    \exp\paren{QK^\top / d} = \exp\paren{ Q^{\paren{m}} \paren{ K^{\paren{m}}}^\top /d }^{\circ C^{\paren{ T_m, T_{m}}}},
\end{align*}
where $C^{\paren{ T_m, T_{m}}} = B^2 / \log n$.

Suppose by Lemma~\ref{lem:wt_A_small_rank}, there exists $U_1^{\paren{ m, m}}, U_2^{\paren{ m, m}} \in \R^{n \times r}$ such that $U_1^{\paren{ m, m}} \paren{ U_2^{\paren{ m, m}}}^\top$ is the $\paren{ \epsilon_0, r}$-approximation of $\exp\paren{ Q^{\paren{m}} \paren{ K^{\paren{m}}}^\top /d }$.

Then, we have
\begin{align*}
    \left \|\exp\paren{QK^\top / d} - \paren{ U_1^{\paren{ m, m}} \paren{ U_2^{\paren{ m, m}}}^\top}^{\circ C^{\paren{ T_m, T_{m}}}}\right \|_p \leq O\paren{\frac{n^2B^2}{\log n} e^{o\paren{B^2}} \epsilon_0}.
\end{align*}
    
\end{lemma}

\begin{proof}

We have
    \begin{align*}
    & ~ \left \|\exp\paren{QK^\top / d} - \paren{ U_1^{\paren{ m, m}} \paren{ U_2^{\paren{ m, m}}}^\top}^{\circ C^{\paren{ T_m, T_{m}}}} \right \|_p^p \\
    \leq & ~ \left | \supp\paren{A} \right | \cdot \left \|\exp\paren{QK^\top / d} - \paren{ U_1^{\paren{ m, m}} \paren{ U_2^{\paren{ m, m}}}^\top}^{\circ C^{\paren{ T_m, T_{m}}}} \right \|_\infty^p \\
    \leq & ~ \left | \supp\paren{A} \right | \cdot  O\paren{\frac{B^2}{\log n} e^{o\paren{B^2}} \epsilon_0}^p,
\end{align*}
where the first step follows from the definition of $\ell_1$ norm and the second step follows from Lemma~\ref{lem:without_bucketing}.

Therefore, we can get
\begin{align*}
    \left \|\exp\paren{QK^\top / d} - \paren{ U_1^{\paren{ m, m}} \paren{ U_2^{\paren{ m, m}}}^\top}^{\circ C^{\paren{ T_m, T_{m}}}} \right \|_p \leq O\paren{\frac{n^{2 / p} B^2}{\log n} e^{o\paren{B^2}} \epsilon_0}
\end{align*}
\end{proof}

\ifdefined\isarxiv

\section{Sketching Reduces the Time Complexity of Polynomial Attention}

\else
\section{SKETCHING REDUCES THE TIME COMPLEXITY OF POLYNOMIAL ATTENTION}

\fi

\label{sec:multi_basis_sketching}

We note that \cite{kmz24} only analyzes the Frobenius norm error of the sketching matrix. In this section, we find the $\ell_\infty$ error guarantee. 

In Appendix~\ref{sub:multi_basis_sketching:prob}, we express the two-vector JL moment property from \cite{akk+20} into the probabilistic statement. In Appendix~\ref{sub:multi_basis_sketching:point}, we use the sketching technique to show that for each $i, j$, the inner product of the sketched vectors $\left \langle \phi'\paren{\paren{U_1}_{i, *}}, \phi'\paren{\paren{U_2}_{j, *}} \right \rangle$ is close to the original un-sketched polynomial kernel $\left \langle \paren{U_1}_{i, *}, \paren{U_2}_{j, *} \right \rangle^p$. In Appendix~\ref{sub:multi_basis_sketching:union}, we use the union bound over all $i$ and $j$ to show that the $\ell_\infty$ norm of $\paren{U_1 U_2^\top}^{\circ p} - \phi'(U_1) \phi'(U_2)^\top$ is upper bounded, where the randomized feature mapping $\phi'$ is applied row-wisely to matrices $U_1$ and $U_2$.

\subsection{Probabilistic Statement of Two Vector JL Moment Property}
\label{sub:multi_basis_sketching:prob}

We formalize the guarantee provided by random projections. This guarantees approximately preserving the inner products between vectors with high probability. This result is a direct consequence of the JL moment property (Definition~\ref{def:jl_moment}) and plays a crucial role in ensuring that geometric relationships between vectors are maintained under dimensionality reduction:

\begin{lemma}[Inner Product Preservation]\label{lem:probability_jl}
Let \( S \in \mathbb{R}^{m \times d} \) be a random matrix satisfying the \((\epsilon, \delta, t)\)-JL moment property for some \( \epsilon, \delta > 0 \) and integer \( t \geq 1 \). Then for all vectors \( x, y \in \mathbb{R}^d \), with probability at least \( 1 - \delta \), we have:
\[
\left| (Sx)^\top (Sy) - x^\top y \right| \leq \epsilon \|x\|_2 \|y\|_2.
\]
\end{lemma}

\begin{proof}

By Lemma~\ref{lem:jl_moment}, we can get
\begin{align}\label{eq:lt_bound}
    \left\| (Sx)^\top (Sy) - x^\top y \right\|_{L_t} \leq \epsilon \delta^{1/t} \|x\|_2 \|y\|_2.
\end{align}

By Definition~\ref{def:jl_moment}, we can get
\begin{align}\label{eq:lt_expectation}
    \left\| (Sx)^\top (Sy) - x^\top y \right\|_{L_t}
    = \E\left[\left|(Sx)^\top (Sy) - x^\top y \right|^t \right]^{1 / t}.
\end{align}

Combining Eq.~\eqref{eq:lt_bound} and Eq.~\eqref{eq:lt_expectation}, we can get
\begin{align*}
    \E\left[\left|(Sx)^\top (Sy) - x^\top y \right|^t \right] \leq \epsilon^t \delta \|x\|_2^t \|y\|_2^t.
\end{align*}

Using Markov’s inequality (see Fact~\ref{fac:markov}), we have
\begin{align*}
    \Pr\left[ \left|(Sx)^\top (Sy) - x^\top y \right| > \epsilon \|x\|_2 \|y\|_2 \right] 
    = & ~ \Pr\left[ \left|(Sx)^\top (Sy) - x^\top y \right|^t > \epsilon^t \|x\|_2^t \|y\|_2^t \right] \\
    \leq & ~ \frac{\mathbb{E}\left[\left|(Sx)^\top (Sy) - x^\top y \right|^t\right]}{\epsilon^t \|x\|_2^t \|y\|_2^t}  \\
    \leq & ~ \delta.
\end{align*}
\end{proof}

\subsection{Pointwise Approximation of Polynomial Kernel via Sketching}
\label{sub:multi_basis_sketching:point}

Now, we show that the randomized feature mapping $\phi'$ can approximate the value of a degree-$p$ polynomial kernel between any pair of vectors up to a small additive error, with high probability:

\begin{lemma}[Pointwise Approximation of Polynomial Kernel via Sketching]
\label{lem:pointwise-poly-sketch}
Let $\{\paren{U_1}_{i, *}\}_{i=1}^n, \{\paren{U_2}_{j, *}\}_{j=1}^n \subset \mathbb{R}^r$. 
Let $p$ be an even positive integer. 
Let $\epsilon \in (0, 1)$ be the accuracy parameter and $\delta \in (0, 1)$ be the failure probability.

Then, there exists a randomized feature mapping 
\[
\phi': \mathbb{R}^r \to \mathbb{R}^{z^2}, \quad \text{for } z = \Theta\left( p \epsilon^{-2} \log(1/\delta) \right)
\]
defined as $\phi'(x) := \paren{Sx^{\otimes \paren{p / 2}}}^{\otimes 2}$ where $S \in \R^{z \times r^{p / 2}}$ is a sketching matrix, such that for all $i, j \in [n]$, the following hold with probability at least $1 - \delta$:
\begin{align*}
    \left | \left \langle \phi'\paren{\paren{U_1}_{i, *}}, \phi'\paren{\paren{U_2}_{j, *}} \right \rangle - \left \langle \paren{U_1}_{i, *}, \paren{U_2}_{j, *} \right \rangle^p \right | \leq \epsilon \|\paren{U_1}_{i, *}\|_2^p \|\paren{U_2}_{j, *}\|_2^p.
\end{align*}
\end{lemma}
\begin{proof}

We have
\begin{align*}
    & ~ \left | \left \langle \phi'\paren{\paren{U_1}_{i, *}}, \phi'\paren{\paren{U_2}_{j, *}} \right \rangle - \left \langle \paren{U_1}_{i, *}, \paren{U_2}_{j, *} \right \rangle^p \right |\\
    = & ~ \left | \left \langle \paren{S \paren{U_1}_{i, *}^{\otimes \paren{p / 2}}}^{\otimes 2}, \paren{S \paren{U_2}_{j, *}^{\otimes \paren{p / 2}}}^{\otimes 2} \right \rangle - \left \langle \paren{U_1}_{i, *}, \paren{U_2}_{j, *} \right \rangle^p \right |\\
    = & ~ \left | \left \langle S \paren{U_1}_{i, *}^{\otimes \paren{p / 2}}, S \paren{U_2}_{j, *}^{\otimes \paren{p / 2}} \right \rangle^2 - \left \langle \paren{U_1}_{i, *}, \paren{U_2}_{j, *} \right \rangle^p \right |\\
    = & ~ \left | \paren{\paren{S \paren{U_1}_{i, *}^{\otimes \paren{p / 2}}}^\top \paren{S \paren{U_2}_{j, *}^{\otimes \paren{p / 2}}}}^2 - \paren{\left \langle \paren{U_1}_{i, *}, \paren{U_2}_{j, *} \right \rangle^{p / 2}}^2 \right |\\
    = & ~ \left | \paren{S \paren{U_1}_{i, *}^{\otimes \paren{p / 2}}}^\top \paren{S \paren{U_2}_{j, *}^{\otimes \paren{p / 2}}} - \left \langle \paren{U_1}_{i, *}, \paren{U_2}_{j, *} \right \rangle^{p / 2} \right | \\
    \cdot & ~ \left | \paren{S \paren{U_1}_{i, *}^{\otimes \paren{p / 2}}}^\top \paren{S \paren{U_2}_{j, *}^{\otimes \paren{p / 2}}} + \left \langle \paren{U_1}_{i, *}, \paren{U_2}_{j, *} \right \rangle^{p / 2} \right |,
\end{align*}
where the first step follows from the definition of $\phi'$ (see from the lemma statement), the second step follows from {\bf Part 1} of Fact~\ref{fac:kronecker}, the third step follows from $\left \langle a, b \right \rangle = a^\top b$ (see Fact~\ref{fac:vector_norm}), and the last step follows from $a^2 - b^2 = (a + b)(a - b)$.

Considering $\left | \paren{S \paren{U_1}_{i, *}^{\otimes \paren{p / 2}}}^\top \paren{S \paren{U_2}_{j, *}^{\otimes \paren{p / 2}}} - \left \langle \paren{U_1}_{i, *}, \paren{U_2}_{j, *} \right \rangle^{p / 2} \right |$, with probability $1 - \delta$, we have
\begin{align*}
    & ~ \left | \paren{S \paren{U_1}_{i, *}^{\otimes \paren{p / 2}}}^\top \paren{S \paren{U_2}_{j, *}^{\otimes \paren{p / 2}}} - \left \langle \paren{U_1}_{i, *}, \paren{U_2}_{j, *} \right \rangle^{p / 2} \right |\\
    = & ~ \left | \paren{S \paren{U_1}_{i, *}^{\otimes \paren{p / 2}}}^\top \paren{S \paren{U_2}_{j, *}^{\otimes \paren{p / 2}}} - \left \langle \paren{U_1}_{i, *}^{\otimes p / 2}, \paren{U_2}_{j, *}^{\otimes p / 2} \right \rangle \right |\\
    = & ~ \left | \paren{S \paren{U_1}_{i, *}^{\otimes \paren{p / 2}}}^\top \paren{S \paren{U_2}_{j, *}^{\otimes \paren{p / 2}}} - \paren{\paren{U_1}_{i, *}^{\otimes p / 2}}^\top \paren{U_2}_{j, *}^{\otimes p / 2}  \right |\\
    \leq & ~ \epsilon \left \| \paren{U_1}_{i, *}^{\otimes p / 2} \right\|_2 \left \| \paren{U_2}_{j, *}^{\otimes p / 2} \right\|_2\\
    = & ~ \epsilon \left \| \paren{U_1}_{i, *} \right\|_2^{p / 2} \left \| \paren{U_2}_{j, *} \right\|_2^{p / 2},
\end{align*}
where the first step follows from {\bf Part 2} of Fact~\ref{fac:kronecker}, the second step follows from Fact~\ref{fac:vector_norm}, the third step follows from Lemma~\ref{lem:probability_jl}, and the last step follows from {\bf Part 3} of Fact~\ref{fac:kronecker}.

Now, we consider $\left | \paren{S \paren{U_1}_{i, *}^{\otimes \paren{p / 2}}}^\top \paren{S \paren{U_2}_{j, *}^{\otimes \paren{p / 2}}} + \left \langle \paren{U_1}_{i, *}, \paren{U_2}_{j, *} \right \rangle^{p / 2} \right |$: we have with probability $1 - \delta$,
\begin{align*}
    & ~ \left | \paren{S \paren{U_1}_{i, *}^{\otimes \paren{p / 2}}}^\top \paren{S \paren{U_2}_{j, *}^{\otimes \paren{p / 2}}} + \left \langle \paren{U_1}_{i, *}, \paren{U_2}_{j, *} \right \rangle^{p / 2} \right | \\ 
    = & ~ \left | \paren{S \paren{U_1}_{i, *}^{\otimes \paren{p / 2}}}^\top \paren{S \paren{U_2}_{j, *}^{\otimes \paren{p / 2}}} - \left \langle \paren{U_1}_{i, *}, \paren{U_2}_{j, *} \right \rangle^{p / 2} + 2\left \langle \paren{U_1}_{i, *}, \paren{U_2}_{j, *} \right \rangle^{p / 2} \right |\\
    \leq & ~ \left | \paren{S \paren{U_1}_{i, *}^{\otimes \paren{p / 2}}}^\top \paren{S \paren{U_2}_{j, *}^{\otimes \paren{p / 2}}} - \left \langle \paren{U_1}_{i, *}, \paren{U_2}_{j, *} \right \rangle^{p / 2}\right | + 2\left |\left \langle \paren{U_1}_{i, *}, \paren{U_2}_{j, *} \right \rangle^{p / 2} \right |\\
    \leq & ~ \epsilon \left \| \paren{U_1}_{i, *} \right\|_2^{p / 2} \left \| \paren{U_2}_{j, *} \right\|_2^{p / 2} + 2 \left |\left \langle \paren{U_1}_{i, *}, \paren{U_2}_{j, *} \right \rangle^{p / 2} \right |\\
    \leq & ~ \epsilon \left \| \paren{U_1}_{i, *} \right\|_2^{p / 2} \left \| \paren{U_2}_{j, *} \right\|_2^{p / 2} + 2 \left \| \paren{U_1}_{i, *} \right\|_2^{p / 2} \left \| \paren{U_2}_{j, *} \right\|_2^{p / 2},
\end{align*}
where the second step follows from the triangle inequality (see Fact~\ref{fac:vector_norm}), the third step follows from Lemma~\ref{lem:probability_jl}, and the last step follows from the Cauchy-Schwarz inequality (see Fact~\ref{fac:vector_norm}).

Thus, combining everything together, we have that
\begin{align*}
    \left | \left \langle \phi'\paren{\paren{U_1}_{i, *}}, \phi'\paren{\paren{U_2}_{j, *}} \right \rangle - \left \langle \paren{U_1}_{i, *}, \paren{U_2}_{j, *} \right \rangle^p \right | \leq \epsilon \|\paren{U_1}_{i, *}\|_2^p \|\paren{U_2}_{j, *}\|_2^p
\end{align*}
holds with probability at least $1 - \delta$.
    
\end{proof}

\subsection{Union Bound to All Entries}
\label{sub:multi_basis_sketching:union}

We generalize the previous pointwise sketching result (Lemma~\ref{lem:pointwise-poly-sketch}) by providing an $\ell_\infty$ bound over the entire polynomial kernel matrix. It shows that, with high probability, the randomized sketching method can approximate the full matrix $\paren{U_1 U_2^\top}^{\circ p}$ entry-wise up to a controllable additive error, uniformly across all pairs of rows:

\begin{lemma}[$\ell_\infty$ Guarantee for Polynomial Sketching]\label{lem:poly_to_poly}
Let $\{\paren{U_1}_{i, *}\}_{i=1}^n, \{\paren{U_2}_{j, *}\}_{j=1}^n \subset \mathbb{R}^r$. 
Fix an even integer $p \geq 2$. Let $\epsilon \in (0, 1)$ be the accuracy parameter and $\delta \in (0, 1)$ be the failure probability. 
Let $\phi' : \mathbb{R}^r \to \mathbb{R}^{z^2}$ be a randomized feature mapping constructed using the sketching technique in Theorem~\ref{thm:poly_attention}, with sketch size $z = \Theta(p \epsilon^{-2} \log(n / \delta))$ and $\phi'\paren{\paren{U_1}_{i, *}} := \paren{S \paren{\paren{U_1}_{i, *}}^{\otimes \paren{p / 2}}}^{\otimes 2}$. For all matrix $A \in \R^{n \times r}$, we write $\phi'(A) \in \R^{n \times z^2}$ to denote $\phi'$ being applied to $A$ row-wisely.

Then, with probability at least $1 - \delta$, we have
\begin{align*}
    \left \|\paren{U_1 U_2^\top}^{\circ p} - \phi'(U_1) \phi'(U_2)^\top \right\|_\infty \leq \max_{i, j \in [n]} \left \{\epsilon \left \|\paren{U_1}_{i, *} \right \|_2^p \left \|\paren{U_2}_{j, *} \right\|_2^p \right \}.
\end{align*}
\end{lemma}

\begin{proof}
By Lemma~\ref{lem:pointwise-poly-sketch}, we know that for each $i, j \in [n]$, the randomized mapping $\phi'$ satisfies
\begin{align*}
    \Pr\left[ \left| \left\langle \phi'\paren{\paren{U_1}_{i, *}}, \phi'\paren{\paren{U_2}_{j, *}} \right\rangle - \left \langle \paren{U_1}_{i, *}, \paren{U_2}_{j, *} \right \rangle^p \right| > \epsilon \left\|\paren{U_1}_{i, *} \right\|_2^p \cdot \left\|\paren{U_2}_{j, *} \right\|_2^p \right] \leq \delta',
\end{align*}
for some small $\delta' \in (0, 1)$.

Since $i, j \in [n]$, we have $n^2$ total pairs $(i,j)$. Applying the union bound (see Fact~\ref{fac:union}) over all $i, j \in [n]$, we have
\begin{align*}
    & ~ \Pr\left[ \exists i, j \in [n], \left| \left\langle \phi'\paren{\paren{U_1}_{i, *}}, \phi'\paren{\paren{U_2}_{j, *}} \right\rangle - \left \langle \paren{U_1}_{i, *}, \paren{U_2}_{j, *} \right \rangle^p \right| > \epsilon \left\|\paren{U_1}_{i, *} \right\|_2^p \cdot \left\|\paren{U_2}_{j, *} \right\|_2^p \right] \\
    \leq & ~ n^2 \cdot \delta'.
\end{align*}

Setting $\delta' = \delta / n^2$, we can get that, with probability at least $1 - \delta$,
\begin{align}\label{eq:error_bound_poly}
    \max_{i, j \in [n]} \left| \left\langle \phi'\paren{\paren{U_1}_{i, *}}, \phi'\paren{\paren{U_2}_{j, *}} \right\rangle - \left \langle \paren{U_1}_{i, *}, \paren{U_2}_{j, *} \right \rangle^p \right| \leq \max_{i, j \in [n]} \left \{ \epsilon \left\|\paren{U_1}_{i, *} \right\|_2^p \cdot \left\|\paren{U_2}_{j, *} \right\|_2^p \right\}.
\end{align}

On the other hand, we have
\begin{align}\label{eq:ell_infty_equa}
    & ~ \max_{i, j \in [n]} \left| \left\langle \phi'\paren{\paren{U_1}_{i, *}}, \phi'\paren{\paren{U_2}_{j, *}} \right\rangle - \left \langle \paren{U_1}_{i, *}, \paren{U_2}_{j, *} \right \rangle^p \right| \notag\\
    = & ~ \max_{i, j \in [n]} \left| \left\langle \phi'\paren{U_1}_{i, *}, \phi'\paren{U_2}_{j, *} \right\rangle - \left \langle \paren{U_1}_{i, *}, \paren{U_2}_{j, *} \right \rangle^p \right| \notag\\
    = & ~ \max_{i, j \in [n]} \left| \paren{\phi'\paren{U_1} \phi'\paren{U_2}^\top}_{i, j} - \paren{\paren{U_1} \paren{U_2^\top}}_{i, j}^p \right| \notag\\
    = & ~ \max_{i, j \in [n]} \left| \phi'\paren{U_1} \phi'\paren{U_2}^\top - \paren{\paren{U_1} \paren{U_2^\top}}^{\circ p} \right|_{i, j} \notag\\
    = & ~ \left \|\paren{U_1 U_2^\top}^{\circ p} - \phi'(U_1) \phi'(U_2)^\top \right\|_\infty,
\end{align}
where the first step follows from the fact that $\phi'$ is applied row-wisely, the second step follows from the definition of matrix multiplication, and the last step follows from the definition of $\ell_\infty$ norm.

Finally, we consider the sketch size. By Lemma~\ref{lem:pointwise-poly-sketch}, to get pointwise approximation to the polynomial kernel, we need
\begin{align*}
    \Theta\left( p \epsilon^{-2} \log(1/\delta') \right).
\end{align*}

Now, since we use union bound and setting $\delta' = \delta / n^2$, we can get the sketch size
\begin{align}\label{eq:sketch_size}
    z 
    = & ~ \Theta\left( p \epsilon^{-2} \log(1/\paren{\delta / n^2}) \right) \notag\\
    = & ~ \Theta\left( p \epsilon^{-2} \log(n/\delta) \right).
\end{align}

Combining Eq.~\eqref{eq:error_bound_poly}, \eqref{eq:ell_infty_equa}, and \eqref{eq:sketch_size}, we finish the proof.
\end{proof}

\ifdefined\isarxiv

\section{Attention Optimization via Multi-Threshold Support Basis and Sketching}

\else
\section{ATTENTION OPTIMIZATION VIA MULTI-THRESHOLD SUPPORT BASIS AND SKETCHING}

\fi

\label{sec:multi_basis_attention}

Now, we combine everything together and present our main result using multi-threshold support basis.

In Appendix~\ref{sub:multi_basis_attention:def}, we give a basic definition. In Appendix~\ref{sub:multi_basis_attention:approx}, we approximate the $\paren{Q, K}$-softmax-attention matrix using multi-threshold support basis decomposition and sketching. In Appendix~\ref{sub:multi_basis_attention:time}, we analyze the running time for constructing $\phi'\paren{U_1^{\paren{\ell, \ell'}}}$ and $\phi'\paren{U_2^{\paren{\ell, \ell'}}}$. In Appendix~\ref{sub:multi_basis_attention:kde}, we solve the batch Gaussian kernel density estimation problem (see Definition~\ref{def:gaussian_kde}) using multi-threshold support basis decomposition. Finally, in Appendix~\ref{sub:multi_basis_attention:main}, we present our main theorem, which states the running time and correctness of approximating the attention computation problem (see Definition~\ref{def:approximate_attention_computation}) using multi-threshold support basis decomposition and sketching. 

\subsection{A Basic Definition}
\label{sub:multi_basis_attention:def}

In this section, we provide basic definitions and clarify the meaning of all notation. We note that in previous sections, some notational abuse may have occurred—for example, the symbol $\epsilon$ may have been used both as the accuracy parameter for sketching and as the bucketing parameter. Our goal here is to unify all theoretical results to derive the main result. Therefore, we redefine the relevant notations to improve clarity.

\begin{definition}\label{def:basic_definition}
        Given the query and the key matrices $Q, K \in \R^{n \times d}$, we define the $\paren{Q, K}$-softmax-attention matrix as $A = \exp(QK^\top / d)$. Let $Q = \sum_{\ell = 1}^{m} Q^{\paren{ T_\ell}} \in \mathbb{R}^{n \times d} $ and $K = \sum_{\ell' = 1}^{m} K^{\paren{ T_{\ell'}}} \in \mathbb{R}^{n \times d}$ be defined as in Definition~\ref{def:exp_threshold_Q_K}, with $b = \min_{i \in [n], j \in [d]}\left \{|Q_{i, j}|, |K_{i, j}|\right \}$, $B = \max_{i \in [n], j \in [d]}\left \{|Q_{i, j}|, |K_{i, j}|\right \}$, and for all $\ell \in [m] \cup \left \{0\right \}$, we let $T_\ell = b \paren{ 1 + \epsilon_B}^\ell$, where $\epsilon_B$ is the bucketing parameter. Let $A^{\paren{ T_\ell, T_{\ell'}}} = Q^{\paren{ T_\ell}} \paren{ K^{\paren{ T_{\ell'}}}}^\top =  C^{\paren{ T_\ell, T_{\ell'}}} \cdot Q^{\paren{ \ell}} \paren{ K^{\paren{ \ell'}}}^\top \in \R^{n \times n}$. Suppose by Lemma~\ref{lem:wt_A_small_rank}, there exists $U_1^{\paren{ \ell, \ell'}}, U_2^{\paren{ \ell, \ell'}} \in \R^{n \times r}$ such that $U_1^{\paren{ \ell, \ell'}} \paren{ U_2^{\paren{ \ell, \ell'}}}^\top$ is the $\paren{ \epsilon_0, r}$-approximation of $\exp\paren{ Q^{\paren{ \ell}} \paren{ K^{\paren{ \ell'}}}^\top}$. Let $\phi': \mathbb{R}^d \to \mathbb{R}^{r^2}$ be defined as $\phi'\paren{q_i} := \paren{S q_i^{\otimes \paren{p / 2}}}^{\otimes 2}$, where $S \in \mathbb{R}^{r \times d^{p/2}}$ with $r = \Theta\left( p \epsilon^{-2} \log(1/\delta) \right)$ is a sketching matrix. With $\epsilon, \epsilon_0 \in (0, 1)$, we let 
    \begin{align*}
        \epsilon_2 := O\paren{\frac{B^2}{\log n} e^{o\paren{B^2}} \epsilon_0 + \sum_{\ell = 1}^{m} \sum_{\ell' = 1}^{m} \max_{i, j \in [n]} \left \{\epsilon \left \|\paren{U_1}_{i, *} \right \|_2^{C^{\paren{ T_\ell, T_{\ell'}}}} \left \|\paren{U_2}_{j, *} \right\|_2^{C^{\paren{ T_\ell, T_{\ell'}}}} \right \}}.
    \end{align*}
\end{definition}

\subsection{\texorpdfstring{$\paren{Q, K}$}{}-Softmax-Attention Matrix Approximation}
\label{sub:multi_basis_attention:approx}

The $\paren{Q, K}$-softmax-attention matrix $\exp(QK^\top / d)$ is central to transformer models, but its computation is costly due to the exponential operation and quadratic complexity. To reduce computational overhead, we approximate this matrix using a combination of polynomial kernel sketching and bucketing. The following lemma shows that by decomposing the input into multiple components and applying randomized feature mappings to each, we can uniformly approximate the entire softmax-attention matrix.

\begin{lemma}[\texorpdfstring{$\paren{Q, K}$}{}-Softmax-Attention Matrix Approximation]\label{lem:attention_approx}

    Let everything be defined as in Definition~\ref{def:basic_definition}.
    Then, we can get
    \begin{align*}
        \left \|\exp(QK^\top / d) - \sum_{\ell = 1}^{m} \sum_{\ell' = 1}^{m} \phi'\paren{U_1^{\paren{\ell, \ell'}}} \phi'\paren{U_2^{\paren{\ell, \ell'}}}^\top \right\|_\infty \leq \epsilon_2.
    \end{align*}
\end{lemma}

\begin{proof}
We can get
\begin{align*}
    & ~ \left \|\exp(QK^\top / d) - \sum_{\ell = 1}^{m} \sum_{\ell' = 1}^{m} \phi'\paren{U_1^{\paren{\ell, \ell'}}} \phi'\paren{U_2^{\paren{\ell, \ell'}}}^\top \right\|_\infty\\
    = & ~ \left \|\exp(QK^\top / d) - \sum_{\ell = 1}^{m} \sum_{\ell' = 1}^{m} \paren{ U_1^{\paren{ \ell, \ell'}} \paren{ U_2^{\paren{ \ell, \ell'}}}^\top}^{\circ C^{\paren{ T_\ell, T_{\ell'}}}} \right.\\
    & ~ + \left. \sum_{\ell = 1}^{m} \sum_{\ell' = 1}^{m} \paren{ U_1^{\paren{ \ell, \ell'}} \paren{ U_2^{\paren{ \ell, \ell'}}}^\top}^{\circ C^{\paren{ T_\ell, T_{\ell'}}}} - \sum_{\ell = 1}^{m} \sum_{\ell' = 1}^{m} \phi'\paren{U_1^{\paren{\ell, \ell'}}} \phi'\paren{U_2^{\paren{\ell, \ell'}}}^\top \right\|_\infty\\
    \leq & ~ \left \|\exp(QK^\top / d) - \sum_{\ell = 1}^{m} \sum_{\ell' = 1}^{m} \paren{ U_1^{\paren{ \ell, \ell'}} \paren{ U_2^{\paren{ \ell, \ell'}}}^\top}^{\circ C^{\paren{ T_\ell, T_{\ell'}}}} \right\|_\infty\\
    & ~ + \left\| \sum_{\ell = 1}^{m} \sum_{\ell' = 1}^{m} \paren{ U_1^{\paren{ \ell, \ell'}} \paren{ U_2^{\paren{ \ell, \ell'}}}^\top}^{\circ C^{\paren{ T_\ell, T_{\ell'}}}} - \sum_{\ell = 1}^{m} \sum_{\ell' = 1}^{m} \phi'\paren{U_1^{\paren{\ell, \ell'}}} \phi'\paren{U_2^{\paren{\ell, \ell'}}}^\top \right\|_\infty\\
    \leq & ~ \left \|\exp(QK^\top / d) - \sum_{\ell = 1}^{m} \sum_{\ell' = 1}^{m} \paren{ U_1^{\paren{ \ell, \ell'}} \paren{ U_2^{\paren{ \ell, \ell'}}}^\top}^{\circ C^{\paren{ T_\ell, T_{\ell'}}}} \right\|_\infty\\
    & ~ + \sum_{\ell = 1}^{m} \sum_{\ell' = 1}^{m} \left\|\paren{ U_1^{\paren{ \ell, \ell'}} \paren{ U_2^{\paren{ \ell, \ell'}}}^\top}^{\circ C^{\paren{ T_\ell, T_{\ell'}}}} - \phi'\paren{U_1^{\paren{\ell, \ell'}}} \phi'\paren{U_2^{\paren{\ell, \ell'}}}^\top \right\|_\infty\\
    \leq & ~ O\paren{\frac{B^2}{\log n} e^{o\paren{B^2}} \epsilon_0 + \sum_{\ell = 1}^{m} \sum_{\ell' = 1}^{m} \max_{i, j \in [n]} \left \{\epsilon \left \|\paren{U_1}_{i, *} \right \|_2^{C^{\paren{ T_\ell, T_{\ell'}}}} \left \|\paren{U_2}_{j, *} \right\|_2^{C^{\paren{ T_\ell, T_{\ell'}}}} \right \}},
\end{align*}
where the second and the third step follows from the triangle inequality (see Fact~\ref{fac:vector_norm}), and the last step follows from combining Lemma~\ref{lem:l_infty_error_with_bucketing} and Lemma~\ref{lem:poly_to_poly}.
\end{proof}

\subsection{Running Time Analysis of Constructing \texorpdfstring{$\phi'\paren{U_1^{\paren{\ell, \ell'}}}$}{} and \texorpdfstring{$\phi'\paren{U_2^{\paren{\ell, \ell'}}}$}{}}
\label{sub:multi_basis_attention:time}

We analyze the runtime complexity of constructing the sketched feature representations used in approximating the softmax-attention matrix. The following lemma shows that the total time required to compute all randomized feature mappings $\phi'\paren{U_1^{\paren{\ell, \ell'}}}$ and $\phi'\paren{U_2^{\paren{\ell, \ell'}}}$, across all bucketed matrix pairs $(\ell, \ell') \in [m] \times [m]$, is almost linear in $n$:

\begin{lemma}\label{lem:construction}
    Let everything be defined as in Definition~\ref{def:basic_definition}.
    
    Then, it takes 
    \begin{align*}
        O\paren{\frac{B^6 \log^2(n/\delta)}{\epsilon^{4}}n^{1 + o(1)}}
    \end{align*}
    time to construct $\phi'\paren{U_1^{\paren{\ell, \ell'}}}$ and $\phi'\paren{U_2^{\paren{\ell, \ell'}}}$, for all $(\ell, \ell') \in [m] \times [m]$.
\end{lemma}
\begin{proof}
    By Lemma~\ref{lem:rank_is_small}, for each $\paren{\ell, \ell'} \in [m] \times [m]$, it takes $O\paren{n^{1 + o(1)}}$ time to construct $U_1^{\paren{\ell, \ell'}} \in \R^{n \times r}$ and $U_2^{\paren{\ell, \ell'}} \in \R^{n \times r}$, with $r < n^{o(1)}$.

By Theorem~\ref{thm:poly_attention}, with 
\begin{align}\label{eq:z}
    z = \Theta\left( C^{\paren{ T_\ell, T_{\ell'}}} \epsilon^{-2} \log(n/\delta) \right)
\end{align}
being the sketch size (see Lemma~\ref{lem:poly_to_poly}), where $\epsilon \in (0, 0.5)$ and $\delta \in (0, 1)$ respectively are the accuracy parameter and failure probability for sketching, since $\phi'$ is applied row-wisely to a matrix, computing $\phi'\paren{U_1^{\paren{\ell, \ell'}}}$ and $\phi'\paren{U_2^{\paren{\ell, \ell'}}}$ respectively requires $n$ times of:
\begin{enumerate}
    \item $\frac{C^{\paren{ T_\ell, T_{\ell'}}}}{2}$ matrix-vector multiplications with matrices of size $r \times z$,
    \item $\frac{C^{\paren{ T_\ell, T_{\ell'}}}}{2} - 2$ matrix-vector multiplications with matrices of size $z \times z$,
    \item $\frac{C^{\paren{ T_\ell, T_{\ell'}}}}{2} - 1$ Hadamard products of $z$-dimensional vectors,
    \item and $1$ self-Kronecker product of an $z$-dimensional vector.
\end{enumerate}

Step 1 takes 
\begin{align*}
    \frac{C^{\paren{ T_\ell, T_{\ell'}}}}{2} \cdot rz = O\paren{C^{\paren{ T_\ell, T_{\ell'}}} r z}.
\end{align*}

Step 2 takes 
\begin{align*}
    \paren{\frac{C^{\paren{ T_\ell, T_{\ell'}}}}{2} - 2} \cdot z^2 = O\paren{C^{\paren{ T_\ell, T_{\ell'}}} z^2}.
\end{align*}

Step 3 takes 
\begin{align*}
    \paren{\frac{C^{\paren{ T_\ell, T_{\ell'}}}}{2} - 1} \cdot z = O\paren{C^{\paren{ T_\ell, T_{\ell'}}} z}.
\end{align*}

Step 4 takes
\begin{align*}
    O\paren{z^2}.
\end{align*}

Therefore, for each $\paren{\ell, \ell'} \in [m] \times [m]$, it takes 
\begin{align*}
    & ~ O\paren{n^{1 + o(1)} + n \cdot \paren{C^{\paren{ T_\ell, T_{\ell'}}} r z + C^{\paren{ T_\ell, T_{\ell'}}} z^2 + C^{\paren{ T_\ell, T_{\ell'}}} z + z^2}} \\
    = & ~ O\paren{n^{1 + o(1)} + n C^{\paren{ T_\ell, T_{\ell'}}} \cdot \paren{r z + z^2}}.
\end{align*}

In total, it takes
\begin{align}\label{eq:total}
    & ~ \sum_{\ell = 1}^{m} \sum_{\ell' = 1}^{m} O\paren{n^{1 + o(1)} + n C^{\paren{ T_\ell, T_{\ell'}}} \cdot \paren{r z + z^2}} \notag \\
    = & ~ O\paren{\sum_{\ell = 1}^{m} \sum_{\ell' = 1}^{m} n^{1 + o(1)} + \sum_{\ell = 1}^{m} \sum_{\ell' = 1}^{m} n C^{\paren{ T_\ell, T_{\ell'}}} \cdot \paren{r z + z^2}} \notag \\
    = & ~ O\paren{m^2 n^{1 + o(1)} + nr \sum_{\ell = 1}^{m} \sum_{\ell' = 1}^{m} C^{\paren{ T_\ell, T_{\ell'}}} z + n\sum_{\ell = 1}^{m} \sum_{\ell' = 1}^{m} C^{\paren{ T_\ell, T_{\ell'}}} z^2}
\end{align}
time to compute $\phi'\paren{U_1^{\paren{\ell, \ell'}}}, \phi'\paren{U_2^{\paren{\ell, \ell'}}} \in \R^{n \times z^2}$ for all $\paren{\ell, \ell'} \in [m] \times [m]$.

Now, we analyze the first term of Eq.~\eqref{eq:total}: $m^2 n^{1 + o(1)}$. 

By Fact~\ref{fac:m}, we have
\begin{align}\label{eq:m}
    m = \left \lfloor \log_{1+\epsilon_B}\paren{ B/b} \right \rfloor + 1.
\end{align}

Therefore, by Eq.~\eqref{eq:m}, we have
\begin{align}\label{eq:t1}
    m^2 n^{1 + o(1)} 
    = & ~ \paren{\left \lfloor \log_{1+\epsilon_B}\paren{ B/b} \right \rfloor + 1}^2 n^{1 + o(1)} \notag\\
    = & ~ O\paren{\log_{1+\epsilon_B}^2\paren{ B/b}  n^{1 + o(1)}}.
\end{align}

Now, we analyze the second term of Eq.~\eqref{eq:total}: $nr \sum_{\ell = 1}^{m} \sum_{\ell' = 1}^{m} C^{\paren{ T_\ell, T_{\ell'}}} z$. 

We have
\begin{align}\label{eq:t2}
    nr \sum_{\ell = 1}^{m} \sum_{\ell' = 1}^{m}  C^{\paren{ T_\ell, T_{\ell'}}} \cdot z
    = & ~ n^{1 + o(1)} \sum_{\ell = 1}^{m} \sum_{\ell' = 1}^{m}  C^{\paren{ T_\ell, T_{\ell'}}} \cdot z \notag\\
    = & ~ n^{1 + o(1)} \sum_{\ell = 1}^{m} \sum_{\ell' = 1}^{m}  C^{\paren{ T_\ell, T_{\ell'}}} \cdot C^{\paren{ T_\ell, T_{\ell'}}} \epsilon^{-2} \log(n/\delta)\notag\\
    = & ~ n^{1 + o(1)} \epsilon^{-2} \log(n/\delta) \sum_{\ell = 1}^{m} \sum_{\ell' = 1}^{m}  \paren{C^{\paren{ T_\ell, T_{\ell'}}}}^2\notag\\
    = & ~ n^{1 + o(1)} \epsilon^{-2} \log(n/\delta) \sum_{\ell = 1}^{m} \sum_{\ell' = 1}^{m}  \paren{\frac{b^2 \paren{ 1 + \epsilon_B}^{\ell + \ell'}}{\log n}}^2\notag\\
    = & ~ \frac{b^4 n^{1 + o(1)} \epsilon^{-2} \log(n/\delta) }{\log^2 n} \sum_{\ell = 1}^{m} \sum_{\ell' = 1}^{m} \paren{ 1 + \epsilon_B}^{2\ell + 2\ell'} \notag\\
    = & ~ \frac{b^4 n^{1 + o(1)} \epsilon^{-2} \log(n/\delta) }{\log^2 n} \paren{\sum_{\ell = 1}^{m} \paren{ 1 + \epsilon_B}^{2\ell}}^2,
\end{align}
where the first step follows from $r = n^{o(1)}$ (see Lemma~\ref{lem:wt_A_small_rank}), the second step follows from Eq.~\eqref{eq:z}, the fourth step follows from Lemma~\ref{lem:normalized_decomposition} and $\epsilon_B > 0$ denote the bucketing parameter, and the last step follows from $\paren{ 1 + \epsilon_B}^{2\ell + 2\ell'} = \paren{ 1 + \epsilon_B}^{2\ell} \cdot \paren{ 1 + \epsilon_B}^{2\ell'}$.

Furthermore, we can see that
\begin{align}\label{eq:t2_more}
    \sum_{\ell = 1}^{m} \paren{ 1 + \epsilon_B}^{2\ell}
    = & ~ \frac{\paren{ 1 + \epsilon_B}^2\paren{1 - \paren{ 1 + \epsilon_B}^{2m}}}{1 - \paren{ 1 + \epsilon_B}^{2}} \notag  \\
    = & ~ O\paren{\paren{ 1 + \epsilon_B}^{2m}},
\end{align}
where the first step follows from the geometric series formula. 

Combining Eq.~\eqref{eq:t2} and \eqref{eq:t2_more}, we have
\begin{align}\label{eq:t2_final}
    nr \sum_{\ell = 1}^{m} \sum_{\ell' = 1}^{m}  C^{\paren{ T_\ell, T_{\ell'}}} \cdot z 
    = & ~ O\paren{\frac{b^4 \log(n/\delta) e^{4m \epsilon_B}}{\epsilon^{2} \log^2 n} n^{1 + o(1)}} \notag\\
    = & ~ O\paren{\frac{B^4 \log(n/\delta) }{\epsilon^{2} \log^2 n} n^{1 + o(1)}},
\end{align}
where the second step follows from Eq.~\eqref{eq:m}.

Now, we analyze the third term of Eq.~\eqref{eq:total}: $n\sum_{\ell = 1}^{m} \sum_{\ell' = 1}^{m} C^{\paren{ T_\ell, T_{\ell'}}} z^2$. 

We have
\begin{align}\label{eq:t3}
    n \sum_{\ell = 1}^{m} \sum_{\ell' = 1}^{m}  C^{\paren{ T_\ell, T_{\ell'}}} \cdot z^2
    = & ~ n \sum_{\ell = 1}^{m} \sum_{\ell' = 1}^{m}  C^{\paren{ T_\ell, T_{\ell'}}} \cdot \paren{C^{\paren{ T_\ell, T_{\ell'}}} \epsilon^{-2} \log(n/\delta)}^2 \notag\\
    = & ~ n \epsilon^{-4} \log^2(n/\delta) \sum_{\ell = 1}^{m} \sum_{\ell' = 1}^{m}  \paren{C^{\paren{ T_\ell, T_{\ell'}}}}^3 \notag\\
    = & ~ n \epsilon^{-4} \log^2(n/\delta) \sum_{\ell = 1}^{m} \sum_{\ell' = 1}^{m}  \paren{\frac{b^2 \paren{ 1 + \epsilon_B}^{\ell + \ell'}}{\log n}}^3 \notag\\
    = & ~ \frac{b^6 n \epsilon^{-4} \log^2(n/\delta) }{\log^3 n} \sum_{\ell = 1}^{m} \sum_{\ell' = 1}^{m} \paren{ 1 + \epsilon_B}^{3\ell + 3\ell'} \notag\\
    = & ~ \frac{b^6 n \epsilon^{-4} \log^2(n/\delta) }{\log^3 n} \paren{\sum_{\ell = 1}^{m} \paren{ 1 + \epsilon_B}^{3\ell}}^2,
\end{align}
where the first step follows from Eq.~\eqref{eq:z}, the third step follows from Lemma~\ref{lem:normalized_decomposition} and $\epsilon_B > 0$ denote the bucketing parameter, and the last step follows from $\paren{ 1 + \epsilon_B}^{3\ell + 3\ell'} = \paren{ 1 + \epsilon_B}^{3\ell} \cdot \paren{ 1 + \epsilon_B}^{3\ell'}$.

Furthermore, we can see that
\begin{align}\label{eq:t3_more}
    \sum_{\ell = 1}^{m} \paren{ 1 + \epsilon_B}^{3\ell}
    = & ~ \frac{\paren{ 1 + \epsilon_B}^3\paren{1 - \paren{ 1 + \epsilon_B}^{3m}}}{1 - \paren{ 1 + \epsilon_B}^{3}} \notag\\
    = & ~ O\paren{\paren{ 1 + \epsilon_B}^{3m}}
\end{align}
where the first step follows from the geometric series formula. 

Combining Eq.~\eqref{eq:t3} and \eqref{eq:t3_more}, we have
\begin{align}\label{eq:t3_final}
    n \sum_{\ell = 1}^{m} \sum_{\ell' = 1}^{m}  C^{\paren{ T_\ell, T_{\ell'}}} \cdot z^2
    = & ~ O\paren{\frac{b^6 \log^2(n/\delta) e^{6 m \epsilon_B} }{\epsilon^{4} \log^3 n}n} \notag\\
    = & ~ O\paren{\frac{B^6 \log^2(n/\delta)}{\epsilon^{4} \log^3 n}n},
\end{align}
where the second step follows from Eq.~\eqref{eq:m}.

Finally, combining Eq.~\eqref{eq:total}, \eqref{eq:t1}, \eqref{eq:t2_final}, and \eqref{eq:t3_final}, we can get that it takes
\begin{align*}
    O\paren{\frac{B^6 \log^2(n/\delta)}{\epsilon^{4}}n^{1 + o(1)}}
\end{align*}
time to compute $\phi'\paren{U_1^{\paren{\ell, \ell'}}}, \phi'\paren{U_2^{\paren{\ell, \ell'}}} \in \R^{n \times z^2}$ for all $\paren{\ell, \ell'} \in [m] \times [m]$.
\end{proof}

\subsection{(Batch) Gaussian Kernel Density Estimation}
\label{sub:multi_basis_attention:kde}

The following lemma shows that, by using the polynomial sketching and bucketing approach described earlier, we can approximate the batch Gaussian kernel density computation with provably small $\ell_\infty$ error. Moreover, the algorithm runs in almost linear time with respect to $n$:

\begin{lemma}\label{lem:gaussian_kde_bucketing}
    Let everything be defined as in Definition~\ref{def:basic_definition}.
    Then, with probability $1 - \delta$, we can solve the (Batch) Gaussian kernel density estimation (Definition~\ref{def:gaussian_kde}) by outputting $\mathsf{S} \in \R^{n \times d}$ satisfying
    \begin{align*}
        \left \|\mathsf{S} - A V\right \|_\infty < n \epsilon_2 \left \|V \right \|_\infty
    \end{align*}
    in $O\paren{\frac{B^6 \log^2(n/\delta)}{\epsilon^{4}}n^{1 + o(1)}}$ time.
\end{lemma}
\begin{proof}

{\bf Proof of correctness.}

We can get 
\begin{align*}
    AV = \exp\paren{QK^\top / d}V.
\end{align*}

We define $\mathsf S := \paren{\sum_{\ell = 1}^{m} \sum_{\ell' = 1}^{m} \phi'\paren{U_1^{\paren{\ell, \ell'}}} \phi'\paren{U_2^{\paren{\ell, \ell'}}}^\top} V$.

Therefore, we can get
\begin{align*}
    \left \|A V - \mathsf S \right\|_\infty    
    = & ~ \left \|\exp(QK^\top / d) V - \sum_{\ell = 1}^{m} \sum_{\ell' = 1}^{m} \phi'\paren{U_1^{\paren{\ell, \ell'}}} \phi'\paren{U_2^{\paren{\ell, \ell'}}}^\top V \right\|_\infty\\
    = & ~ \left \|\paren{\exp(QK^\top / d) - \sum_{\ell = 1}^{m} \sum_{\ell' = 1}^{m} \phi'\paren{U_1^{\paren{\ell, \ell'}}} \phi'\paren{U_2^{\paren{\ell, \ell'}}}^\top} V \right\|_\infty\\
    \leq & ~ n \left \|\exp(QK^\top / d) - \sum_{\ell = 1}^{m} \sum_{\ell' = 1}^{m} \phi'\paren{U_1^{\paren{\ell, \ell'}}} \phi'\paren{U_2^{\paren{\ell, \ell'}}}^\top\right \|_\infty \left \| V \right\|_\infty\\
    \leq & ~ n \epsilon_2 \left \| V \right\|_\infty,
\end{align*}
where the second step follows from the distributive law, the third step follows from the definition of the $\ell_\infty$ norm, and the fourth step follows from Lemma~\ref{lem:attention_approx}.

{\bf Proof of the running time.}

By Lemma~\ref{lem:construction}, we note that to obtain $\phi'\paren{U_1^{\paren{\ell, \ell'}}}, \phi'\paren{U_2^{\paren{\ell, \ell'}}} \in \R^{n \times z^2}$, it takes 
\begin{align}\label{eq:total_final}
    O\paren{\frac{B^6 \log^2(n/\delta)}{\epsilon^{4}}n^{1 + o(1)}}
\end{align}
time. 

Then, we compute 
\begin{align*}
    \underbrace{\phi'\paren{U_2^{\paren{\ell, \ell'}}}^\top}_{z^2 \times n} \underbrace{V}_{n \times d},
\end{align*}
for all $\paren{\ell, \ell'} \in [m] \times [m]$, which takes 
\begin{align}\label{eq:kde_1}
    \sum_{\ell = 1}^{m} \sum_{\ell' = 1}^{m} O\paren{z^2 n d}
    = & ~ \sum_{\ell = 1}^{m} \sum_{\ell' = 1}^{m} O\paren{\paren{C^{\paren{ T_\ell, T_{\ell'}}}}^2 \epsilon^{-4} \log^2(n/\delta) n d} \notag\\
    = & ~ O\paren{\epsilon^{-4} \log^2(n/\delta) n^{1 + o(1)} \sum_{\ell = 1}^{m} \sum_{\ell' = 1}^{m} \paren{C^{\paren{ T_\ell, T_{\ell'}}}}^2} \notag\\
    = & ~ O\paren{\epsilon^{-4} \log^2(n/\delta) n^{1 + o(1)} \sum_{\ell = 1}^{m} \sum_{\ell' = 1}^{m} \paren{\frac{b^2 \paren{ 1 + \epsilon_B}^{\ell + \ell'}}{\log n}}^2} \notag\\
    = & ~ O\paren{\frac{b^4 \log^2(n/\delta)}{\epsilon^{4}}  n^{1 + o(1)} \sum_{\ell = 1}^{m} \sum_{\ell' = 1}^{m} \paren{ 1 + \epsilon_B}^{2\ell + 2\ell'}} \notag\\
    = & ~ O\paren{\frac{b^4 \log^2(n/\delta)}{\epsilon^{4}}  n^{1 + o(1)} \paren{\sum_{\ell = 1}^{m} \paren{ 1 + \epsilon_B}^{2\ell}}^2} \notag\\
    = & ~ O\paren{\frac{b^4 \log^2(n/\delta) e^{4m\epsilon_B}}{\epsilon^{4}}  n^{1 + o(1)}} \notag\\
    = & ~ O\paren{\frac{B^4 \log^2(n/\delta)}{\epsilon^{4}}  n^{1 + o(1)}},
\end{align}
where the first step follows from Eq.~\eqref{eq:z}, the second step follows from $d = O(\log n)$, the third step follows from Lemma~\ref{lem:normalized_decomposition}, the fifth step follows from $\paren{ 1 + \epsilon_B}^{2\ell + 2\ell'} = \paren{ 1 + \epsilon_B}^{2\ell} \cdot \paren{ 1 + \epsilon_B}^{2\ell'}$, the sixth step follows from Eq.~\eqref{eq:t2_more}, and the last step follows from Eq.~\eqref{eq:m}.

Finally, for all $\paren{\ell, \ell'} \in [m] \times [m]$, we compute
\begin{align*}
    \underbrace{\phi'\paren{U_1^{\paren{\ell, \ell'}}}}_{n \times z^2} \cdot \paren{\underbrace{\phi'\paren{U_2^{\paren{\ell, \ell'}}}^\top V}_{z^2 \times d}},
\end{align*}
which, similar as in Eq.~\eqref{eq:kde_1} also takes 
\begin{align}\label{eq:kde_2}
    \sum_{\ell = 1}^{m} \sum_{\ell' = 1}^{m} O\paren{z^2 n d} = O\paren{\frac{B^4 \log^2(n/\delta)}{\epsilon^{4}}  n^{1 + o(1)}}
\end{align}
time.

In conclusion, to construct all of $\phi'\paren{U_1^{\paren{\ell, \ell'}}}, \phi'\paren{U_2^{\paren{\ell, \ell'}}} \in \R^{n \times z^2}$, for all $\paren{\ell, \ell'} \in [m] \times [m]$, we need Eq.~\eqref{eq:total_final} time; computing all of $\underbrace{\phi'\paren{U_2^{\paren{\ell, \ell'}}}^\top}_{z^2 \times n} \underbrace{V}_{n \times d}$, we need Eq.~\eqref{eq:kde_1} time; computing all of $\underbrace{\phi'\paren{U_1^{\paren{\ell, \ell'}}}}_{n \times z^2} \cdot \paren{\underbrace{\phi'\paren{U_2^{\paren{\ell, \ell'}}}^\top V}_{z^2 \times d}}$, we need Eq.~\eqref{eq:kde_2} time. Adding them all together, we need
\begin{align*}
    O\paren{\frac{B^6 \log^2(n/\delta)}{\epsilon^{4}}n^{1 + o(1)}}
\end{align*}
time.
\end{proof}

\subsection{Main Result}
\label{sub:multi_basis_attention:main}

We now state the main result of our attention approximation framework. By combining polynomial sketching, bucketing, and sparse additive decompositions, our algorithm approximates the attention computation to within a small $\ell_\infty$ error, with high probability. Importantly, it does so without requiring bounded entries or restrictive assumptions on the input matrices. The theorem below guarantees both the correctness and the runtime efficiency of our method, showing that the overall computation can be carried out in almost linear time:

\begin{theorem}\label{thm:attention_approximation_bucketing}
    Let everything be defined as in Definition~\ref{def:basic_definition}.
    
    Then, with probability $1 - \delta$, we can solve the approximate attention computation (Definition~\ref{def:approximate_attention_computation}) by outputting $P \in \R^{n \times d}$ satisfying
    \begin{align*}
        \left \|P - D^{-1} A V\right \|_\infty \lesssim \epsilon_2 \exp(3B^2) \cdot \left \| V \right \|_\infty
    \end{align*}
    in $O\paren{\frac{B^6 \log^2(n/\delta)}{\epsilon^{4}}n^{1 + o(1)}}$ time.
\end{theorem}

\begin{proof}
    {\bf Proof of correctness.}

    We define 
    \begin{align}\label{eq:P}
        P := \wt D^{-1} \wt A V = \diag\paren{\wt A {\bf 1}_n }^{-1} \wt A V,
    \end{align}
    where 
    \begin{align}\label{eq:wtA_P}
        \wt A := \sum_{\ell = 1}^{m} \sum_{\ell' = 1}^{m} \phi'\paren{U_1^{\paren{\ell, \ell'}}} \phi'\paren{U_2^{\paren{\ell, \ell'}}}^\top.
    \end{align}

    By the triangle inequality (see Fact~\ref{fac:vector_norm}), we have
    \begin{align}\label{eq:D-1AV_bucketing}
        \left \| D^{-1} AV - \wt D^{-1} \wt AV \right \|_\infty
        = & ~ \left \| D^{-1} AV - D^{-1} \wt AV + D^{-1} \wt AV -  \wt D^{-1} \wt AV \right \|_\infty \notag\\
        \leq & ~ \left \| D^{-1} AV - D^{-1} \wt AV \right \|_\infty + \left \| D^{-1} \wt AV -  \wt D^{-1} \wt AV \right \|_\infty.
    \end{align}

    Considering the first term $\left \| D^{-1} AV - D^{-1} \wt AV \right \|_\infty$, we have
    \begin{align}\label{eq:approx_atten}
        \left \| D^{-1} AV - D^{-1} \wt AV \right \|_\infty
        = & ~ \left \| D^{-1} \paren{A - \wt A} V \right \|_\infty \notag\\
        \leq & ~ n \left \| D^{-1} \paren{A - \wt A}\right \|_\infty \left \| V \right \|_\infty \notag\\
        \leq & ~ n \left \| D^{-1}\right \|_\infty \left \|A - \wt A\right \|_\infty \left \| V \right \|_\infty \notag\\
        \leq & ~ n \left \| D^{-1}\right \|_\infty \epsilon_2 \left \| V \right \|_\infty,
    \end{align}
    where the second step follows from the definition of inner product, the third step follows from the fact that $D^{-1}$ is a diagonal matrix, and the last step follows from Lemma~\ref{lem:attention_approx}.

    Now, we consider $\left \| D^{-1}\right \|_\infty$: we have
    \begin{align}\label{eq:D-1}
        \left \| D^{-1} \right \|_\infty
        = & ~ \max_i \frac{1}{D_{i,i}} \notag\\
        = & ~ \frac{1}{\min_i D_{i,i}} \notag\\
        = & ~ \frac{1}{\min_i \left( \exp(QK^\top / d) \cdot \mathbf{1}_n \right)_{i, *}} \notag\\
        = & ~ \frac{1}{\min_i \sum_{j=1}^n \exp\left( \frac{ \langle Q_{i, *}, K_{j, *} \rangle }{d} \right)} \notag\\
        \leq & ~ \frac{1}{n \cdot \min_{i,j} \exp\left( \frac{ \langle Q_{i, *}, K_{j, *} \rangle }{d} \right)} \notag\\
        \leq & ~ \frac{\exp(B^2)}{n},
    \end{align}
    where the first step follows from the fact that $D^{-1}$ is a diagonal matrix, the third and fourth steps follow from Definition~\ref{def:exact_attention_computation}, and the last step follows from Definition~\ref{def:basic_definition}.
    
    Combining Eq.~\eqref{eq:approx_atten} and Eq.~\eqref{eq:D-1}, we can get
    \begin{align}\label{eq:approx_atten_final}
        \left \| D^{-1} AV - D^{-1} \wt AV \right \|_\infty \leq \epsilon_2 \exp\paren{B^2} \left \| V \right \|_\infty.
    \end{align}

    Considering the second term $\left \| D^{-1} \wt AV -  \wt D^{-1} \wt AV \right \|_\infty$: we have
    \begin{align}\label{eq:approx_atten_2}
        \left \| D^{-1} \wt{A} V - \wt{D}^{-1} \wt{A} V \right \|_\infty
        = & ~ \left \| \left( D^{-1} - \wt{D}^{-1} \right) \wt{A} V \right \|_\infty \notag\\
        \leq & ~ n \cdot \left \| D^{-1} - \wt{D}^{-1} \right \|_\infty \cdot \left \| \wt{A} \right \|_\infty \cdot \left \| V \right \|_\infty \notag\\
        = & ~ n \cdot \max_i \left| \frac{1}{D_{i,i}} - \frac{1}{\wt{D}_{i,i}} \right| \cdot \left \| \wt{A} \right \|_\infty \cdot \left \| V \right \|_\infty \notag\\
        = & ~ n \cdot \max_i \left| \frac{D_{i,i} - \wt{D}_{i,i}}{D_{i,i} \cdot \wt{D}_{i,i}} \right| \cdot \left \| \wt{A} \right \|_\infty \cdot \left \| V \right \|_\infty \notag\\
        \leq & ~ \frac{n \cdot \| D - \wt{D} \|_\infty}{\min_i D_{i,i} \cdot \min_i \wt{D}_{i,i}} \cdot \left \| \wt{A} \right \|_\infty \cdot \left \| V \right \|_\infty \notag\\
        \leq & ~ \frac{\exp(2B^2) \cdot \| D - \wt{D} \|_\infty}{n} \cdot \left \| \wt{A} \right \|_\infty \cdot \left \| V \right \|_\infty,
    \end{align}
    where the second step follows from the definition of inner product and $D^{-1}$ is a diagonal matrix, the third step follows from the definition of $\ell_\infty$ norm, and the sixth step follows from Eq.~\eqref{eq:D-1}.

    Note that
    \begin{align}\label{eq:D-wtD}
        \left \|D - \wt D \right \|_\infty 
        = & ~ \left \|\diag(A {\bf 1}_n) - \diag(\wt A {\bf 1}_n) \right \|_\infty \notag\\
        = & ~ \left \|\paren{A - \wt A} {\bf 1}_n \right \|_\infty \notag\\
        \leq & ~ n \left \|A - \wt A \right \|_\infty \left \| {\bf 1}_n \right \|_\infty \notag\\
        = & ~ n \left \|A - \wt A \right \|_\infty.
    \end{align}

    Combining Eq.~\eqref{eq:D-wtD} and Eq.~\eqref{eq:approx_atten_2}, we have
    \begin{align}\label{eq:approx_atten_22}
        \left \| D^{-1} \wt{A} V - \wt{D}^{-1} \wt{A} V \right \|_\infty
        \leq \exp(2B^2) \cdot \| A - \wt{A} \|_\infty \cdot \left \| \wt{A} \right \|_\infty \cdot \left \| V \right \|_\infty .
    \end{align}

    Additionally, combining the reverse triangle inequality (see Fact~\ref{fac:vector_norm}) and Lemma~\ref{lem:attention_approx}, we have
    \begin{align}\label{eq:wtA_infty}
        \|\wt{A} \|_\infty \leq \| A \|_\infty + \epsilon_2.
    \end{align}

    In particular, considering $\| A \|_\infty$, we note that
    \begin{align}\label{eq:A_infty}
        \| A \|_\infty
        = & ~ \| \exp(QK^\top / d) \|_\infty \notag\\
        = & ~  \exp(\| QK^\top \|_\infty / d) \notag\\
        \leq & ~ \exp(B^2),
    \end{align}
    where the first step follows from Definition~\ref{def:exact_attention_computation} and the second step follows from the definition of inner product.

    Combining Eq.~\eqref{eq:approx_atten_22}, Eq.~\eqref{eq:wtA_infty}, Eq.~\eqref{eq:A_infty}, and Lemma~\ref{lem:attention_approx}, we have
    \begin{align}\label{eq:approx_atten_2_final}
        \left \| D^{-1} \wt{A} V - \wt{D}^{-1} \wt{A} V \right \|_\infty \lesssim \epsilon_2 \exp(3B^2) \cdot \left \| V \right \|_\infty.
    \end{align}

    Combining Eq.~\eqref{eq:D-1AV_bucketing}, Eq.~\eqref{eq:approx_atten_final}, and Eq.~\eqref{eq:approx_atten_2_final}, we have
    \begin{align*}
        \left \| D^{-1} AV - \wt D^{-1} \wt AV \right \|_\infty \lesssim \epsilon_2 \exp(3B^2) \cdot \left \| V \right \|_\infty.
    \end{align*}

{\bf Proof of Running time.}

Our goal is to compute $P$ (see Eq.~\eqref{eq:P} and Eq.~\eqref{eq:wtA_P}). 

    By Lemma~\ref{lem:construction}, we have shown that it takes 
    \begin{align*}
        O\paren{\frac{B^6 \log^2(n/\delta)}{\epsilon^{4}}n^{1 + o(1)}}
    \end{align*}
    time to construct $\phi'\paren{U_1^{\paren{\ell, \ell'}}}$ and $\phi'\paren{U_2^{\paren{\ell, \ell'}}}$, for all $(\ell, \ell') \in [m] \times [m]$.

    By Lemma~\ref{lem:gaussian_kde_bucketing}, we have shown that it takes 
    \begin{align*}
        O\paren{\frac{B^6 \log^2(n/\delta)}{\epsilon^{4}}n^{1 + o(1)}}
    \end{align*}
    time to compute 
    \begin{align*}
        \underbrace{\phi'\paren{U_1^{\paren{\ell, \ell'}}}}_{n \times z^2} \cdot \paren{\underbrace{\phi'\paren{U_2^{\paren{\ell, \ell'}}}^\top V}_{z^2 \times d}},
    \end{align*}
    and 
    \begin{align*}
        \underbrace{\phi'\paren{U_2^{\paren{\ell, \ell'}}}^\top}_{z^2 \times n} \underbrace{V}_{n \times d},
    \end{align*}
    for all $\paren{\ell, \ell'} \in [m] \times [m]$.

    Similarly, it also takes $O\paren{\frac{B^6 \log^2(n/\delta)}{\epsilon^{4}}n^{1 + o(1)}}$ time to compute 
    \begin{align*}
        \underbrace{\phi'\paren{U_1^{\paren{\ell, \ell'}}}}_{n \times z^2} \cdot \paren{\underbrace{\phi'\paren{U_2^{\paren{\ell, \ell'}}}^\top {\bf 1}_n}_{z^2 \times 1}},
    \end{align*}
    and 
    \begin{align*}
        \underbrace{\phi'\paren{U_2^{\paren{\ell, \ell'}}}^\top}_{z^2 \times n} \underbrace{{\bf 1}_n}_{n \times 1},
    \end{align*}
    for all $\paren{\ell, \ell'} \in [m] \times [m]$.

    This implies that we can compute 
    \begin{align*}
        \wt D^{-1} 
        = & ~ \diag\paren{\wt A {\bf 1}_n }^{-1}\\
        = & ~ \diag\paren{\sum_{\ell = 1}^{m} \sum_{\ell' = 1}^{m} \phi'\paren{U_1^{\paren{\ell, \ell'}}} \phi'\paren{U_2^{\paren{\ell, \ell'}}}^\top {\bf 1}_n }^{-1}
    \end{align*}
    in $O\paren{\frac{B^6 \log^2(n/\delta)}{\epsilon^{4}}n^{1 + o(1)}}$ time.

    Finally, as $\wt D^{-1}$ is a diagonal matrix, computing
    \begin{align*}
        P = \wt D^{-1} \wt A V = \sum_{\ell = 1}^{m} \sum_{\ell' = 1}^{m} \paren{\underbrace{\wt D^{-1}}_{n \times n} \underbrace{\phi'\paren{U_1^{\paren{\ell, \ell'}}}}_{n \times z^2}} \cdot \paren{\underbrace{\phi'\paren{U_2^{\paren{\ell, \ell'}}}^\top}_{z^2 \times n} \underbrace{V}_{n \times d}}
    \end{align*}
    takes $O\paren{\frac{B^6 \log^2(n/\delta)}{\epsilon^{4}}n^{1 + o(1)}}$ time.

\end{proof}

\ifdefined\isarxiv

\section{More Facts}

\else
\section{MORE FACTS}

\fi

\label{sec:facts}

In this section, we present basic mathematical properties that support our theoretical proofs.

\begin{fact}\label{fac:support_inequality}
    Let $A, B \in \R^{n \times n}$. 

    Then, we have
    \begin{align*}
        |\mathrm{supp}\paren{A + B}| \leq |\mathrm{supp}\paren{A}| + |\mathrm{supp}\paren{B}|.
    \end{align*}
\end{fact}
\begin{proof}
    It suffices to show that 
    \begin{align}\label{eq:supp_subset}
        \mathrm{supp}\paren{A + B} \subseteq \mathrm{supp}\paren{A} \cup \mathrm{supp}\paren{B}.
    \end{align}

    Suppose $\paren{i, j}$ is an arbitrary element in $\mathrm{supp}\paren{A + B}$.

    Then, we have $\paren{A + B}_{i, j} \neq 0$.

    Note that by the basic definition of matrix addition, we have
    \begin{align*}
        \paren{A + B}_{i, j}
        = & ~ A_{i, j} + B_{i, j}\\
        \neq & ~ 0,
    \end{align*}
    which implies that either $A_{i, j} \neq 0$ or $B_{i, j} \neq 0$.

    Therefore, $\paren{i, j} \in \mathrm{supp}\paren{A}$ or $\paren{i, j} \in \mathrm{supp}\paren{B}$, which completes the proof of Eq.~\eqref{eq:supp_subset}.
\end{proof}

\begin{fact}\label{fac:support_of_A}

Let $\{ A^{(\ell, \ell')} \}_{\ell, \ell' \in [m]}$ be a collection of $n \times n$ matrices forming a support basis of $A \in \mathbb{R}^{n \times n}$.

Then, we have
\[
\sum_{\ell = 1}^{m} \sum_{\ell' = 1}^{m} \left| \supp\left(A^{(\ell, \ell')}\right) \right| = \left| \supp(A) \right| \leq n^2.
\]
\end{fact}
\begin{proof}

By the definition of support basis (see Definition~\ref{def:support_basis}), we know that $A^{(\ell, \ell')}$'s are disjoint matrices (Definition~\ref{def:disjoint}).

Therefore, we have
\begin{align*}
    \supp\paren{A}
    = & ~ \supp \paren{\sum_{\ell, \ell'} A^{(\ell, \ell')}}\\
    = & ~ \bigcup_{\ell, \ell'} \supp\left( A^{(\ell, \ell')} \right).
\end{align*}

Thus, we can further get
\begin{align*}
    \left | \supp\paren{A} \right |
    = & ~ \left | \bigcup_{\ell, \ell'} \supp\left( A^{(\ell, \ell')} \right) \right |\\
    = & ~ \sum_{\ell, \ell'} \left| \supp\left( A^{(\ell, \ell')} \right) \right|.
\end{align*}

Finally, since $A \in \R^{n \times n}$, we can get that $\left| \supp(A) \right| \leq n^2$.
\end{proof}

\begin{fact}\label{fac:vector_norm}
Let $u, v \in \R^n$. 

Then, we have
\begin{itemize}
    \item $\left| \langle u, v \rangle \right| \leq \|u\|_2 \cdot \|v\|_2$ (Cauchy--Schwarz inequality).
    \item $\| u \|_2 \leq \| u \|_1$.
    \item $\langle u, v \rangle = u^\top v$.
    \item $\langle u, u \rangle = u^\top u = \|u\|_2^2$.
    \item $\|u + v\|_2 \leq \|u\|_2 + \|v\|_2$ (triangle inequality).
    \item $\left |\|u\|_2 - \|v\|_2 \right| \leq \|u - v\|_2$ (reverse triangle inequality).
\end{itemize}
The (reverse) triangle inequality holds in all metric spaces. In particular, it applies to $(\R, |\cdot|)$, $(\R^{n \times d}, \|\cdot\|_p)$, and $(\R^{n}, \|\cdot\|_p)$, for all positive integers $n, d$, and $p \in \{1, 2, \dots, \infty\}$.
\end{fact}

\begin{fact}[Integral analysis]\label{fac:integral}
    We have
    \begin{itemize}
        \item {\bf Part 1.} if $x$ is the variable, then
        \begin{align*}
            & ~ \int_1^{m + 1} \frac{b^4 \paren{ 1 + \epsilon}^{2x + 2y}}{\log^2 n} n^{\frac{c b^2 \paren{ 1 + \epsilon}^{x + y}}{\log n}} \, \mathrm{d} x\\
            = & ~ \frac{cb^2 \paren{ 1 + \epsilon}^{y + m + 1} e^{b^2 c (\epsilon + 1)^{y + m + 1}} - cb^2 \paren{ 1 + \epsilon}^{y + 1} e^{b^2 c (\epsilon + 1)^{y + 1}} - e^{b^2 c (\epsilon + 1)^{y + m + 1}} + e^{b^2 c (\epsilon + 1)^{y + 1}}}{c^2 \log^2 n \log(\epsilon + 1)}
        \end{align*}
        \item {\bf Part 2.} if $y$ is the variable, then
        \begin{align*}
            \int_1^{m + 1} \frac{b^2 \paren{ 1 + \epsilon}^{y + m + 1} e^{b^2 c (\epsilon + 1)^{y + m + 1}}}{c \log^{2}\left(n\right) \log\left({\epsilon} + 1\right)} \, \mathrm{d} y 
            = \frac{e^{b^2 c (\epsilon + 1)^{2m + 2}} - e^{b^2 c (\epsilon + 1)^{m + 2}}}{c^2 \log^2(n) \log^2(\epsilon + 1)}.
        \end{align*}
    \end{itemize}
\end{fact}
\begin{proof}
    {\bf Proof of Part 1.}

    By the linearity of the integral, we have
    \begin{align}\label{eq:integral_part1_1}
        \int_1^{m + 1} \frac{b^4 \paren{ 1 + \epsilon}^{2x + 2y}}{\log^2 n} n^{\frac{cb^2 \paren{ 1 + \epsilon}^{x + y}}{\log n}} \, \mathrm{d} x
        = \frac{b^4 \paren{ 1 + \epsilon}^{2y}}{\log^2 n}\int_1^{m + 1} \paren{ 1 + \epsilon}^{2x} n^{\frac{cb^2 \paren{ 1 + \epsilon}^{x + y}}{\log n}} \, \mathrm{d} x.
    \end{align}

    We can further get
    \begin{align*}
        \int_1^{m + 1} (\epsilon + 1)^{2x} n^{\frac{b^2 c (\epsilon + 1)^{x + y}}{\log(n)}} \, \d x
        = & ~ \int_1^{m + 1} (\epsilon + 1)^x \log(\epsilon + 1) \cdot \frac{(\epsilon + 1)^x e^{b^2 c (\epsilon + 1)^{x + y}}}{\log(\epsilon + 1)} \, \d x
    \end{align*}

    Now, we use $u$-substitution by letting $u = (\epsilon + 1)^x$, which implies that
    \begin{align*}
        x = \frac{\log(u)}{\log(\epsilon + 1)}
    \end{align*}
    and
    \begin{align*}
        \d x = \frac{1}{(\epsilon + 1)^x \log(\epsilon + 1)} \d u
    \end{align*}

    Therefore, we have
    \begin{align}\label{eq:integral_part1_3}
        \int_1^{m + 1} (\epsilon + 1)^{2x} n^{\frac{b^2 c (\epsilon + 1)^{x + y}}{\log(n)}} \, \d x
        = & ~ \frac{1}{\log(\epsilon + 1)} \int_{1 + \epsilon}^{\paren{1 + \epsilon}^{m + 1}} u e^{b^2 c (\epsilon + 1)^y u} \, \d u
    \end{align}

Now, we consider
\[
\int_{1 + \epsilon}^{\paren{1 + \epsilon}^{m + 1}} u e^{b^2 c (\epsilon + 1)^y u} \, \d u
\]
We use integration by parts, namely $\int f g' = fg - \int f' g$.

With
\[
f = u, \quad g' = e^{b^2 c (\epsilon + 1)^y u}
\]
\[
f' = 1, \quad g = \frac{e^{b^2 c (\epsilon + 1)^y u}}{b^2 c (\epsilon + 1)^y},
\]
we have
\begin{align}\label{eq:integral_part1_4}
    & ~ \int_{1 + \epsilon}^{\paren{1 + \epsilon}^{m + 1}} u e^{b^2 c (\epsilon + 1)^y u} \, \d u \notag \\
    = & ~ \left. \frac{u e^{b^2 c (\epsilon + 1)^y u}}{b^2 c (\epsilon + 1)^y}\right|_{1 + \epsilon}^{\paren{1 + \epsilon}^{m + 1}} - \int_{1 + \epsilon}^{\paren{1 + \epsilon}^{m + 1}} \frac{e^{b^2 c (\epsilon + 1)^y u}}{b^2 c (\epsilon + 1)^y} \, \d u \notag\\
    = & ~ \paren{\frac{\paren{1 + \epsilon}^{m + 1} e^{b^2 c (\epsilon + 1)^{y + m + 1}} - \paren{1 + \epsilon} e^{b^2 c (\epsilon + 1)^{y + 1}}}{b^2 c (\epsilon + 1)^y}} - \int_{1 + \epsilon}^{\paren{1 + \epsilon}^{m + 1}} \frac{e^{b^2 c (\epsilon + 1)^y u}}{b^2 c (\epsilon + 1)^y} \, \d u.
\end{align}

We again use $u$-substitution by letting $v = b^2 c (\epsilon + 1)^y u$ so that 
\begin{align*}
    \d u = \frac{1}{b^2 c (\epsilon + 1)^y} \d v
\end{align*}

Therefore, we get
\begin{align}\label{eq:integral_part1_5}
    \int_{1 + \epsilon}^{\paren{1 + \epsilon}^{m + 1}} \frac{e^{b^2 c (\epsilon + 1)^y u}}{b^2 c (\epsilon + 1)^y} \, \d u 
    = & ~ \frac{1}{b^4 c^2 (\epsilon + 1)^{2y}} \int_{b^2 c (\epsilon + 1)^y \paren{1 + \epsilon}}^{b^2 c (\epsilon + 1)^y \paren{1 + \epsilon}^{m + 1}} e^{v} \, \d v \notag\\
    = & ~ \frac{e^{b^2 c (\epsilon + 1)^y \paren{1 + \epsilon}^{m + 1}} - e^{b^2 c (\epsilon + 1)^y \paren{1 + \epsilon}}}{b^4 c^2 (\epsilon + 1)^{2y}}.
\end{align}

Combining Eq.~\eqref{eq:integral_part1_1}, \eqref{eq:integral_part1_3}, \eqref{eq:integral_part1_4}, and \eqref{eq:integral_part1_5}, we have
\begin{align*}
    & ~ \int_1^{m + 1} \frac{b^4 \paren{ 1 + \epsilon}^{2x + 2y}}{\log^2 n} n^{\frac{c b^2 \paren{ 1 + \epsilon}^{x + y}}{\log n}} \, \mathrm{d} x\\
    = & ~ \paren{\frac{b^4 \paren{ 1 + \epsilon}^{2y}}{\log^2 n}} \cdot \paren{\frac{\frac{\paren{1 + \epsilon}^{m + 1} e^{b^2 c (\epsilon + 1)^{y + m + 1}} - \paren{1 + \epsilon} e^{b^2 c (\epsilon + 1)^{y + 1}}}{b^2 c (\epsilon + 1)^y} - \frac{e^{b^2 c (\epsilon + 1)^y \paren{1 + \epsilon}^{m + 1}} - e^{b^2 c (\epsilon + 1)^y \paren{1 + \epsilon}}}{b^4 c^2 (\epsilon + 1)^{2y}}}{\log(\epsilon + 1)}}\\
    = & ~ \frac{\frac{b^2 \paren{ 1 + \epsilon}^{y} \paren{1 + \epsilon}^{m + 1} e^{b^2 c (\epsilon + 1)^{y + m + 1}} - b^2 \paren{ 1 + \epsilon}^{y} \paren{1 + \epsilon} e^{b^2 c (\epsilon + 1)^{y + 1}}}{c} - \frac{e^{b^2 c (\epsilon + 1)^y \paren{1 + \epsilon}^{m + 1}} - e^{b^2 c (\epsilon + 1)^y \paren{1 + \epsilon}}}{c^2}}{\log^2 n \log(\epsilon + 1)}\\
    = & ~ \frac{cb^2 \paren{ 1 + \epsilon}^{y + m + 1} e^{b^2 c (\epsilon + 1)^{y + m + 1}} - cb^2 \paren{ 1 + \epsilon}^{y + 1} e^{b^2 c (\epsilon + 1)^{y + 1}} - e^{b^2 c (\epsilon + 1)^{y + m + 1}} + e^{b^2 c (\epsilon + 1)^{y + 1}}}{c^2 \log^2 n \log(\epsilon + 1)}.
\end{align*}

{\bf Proof of Part 2.}

Now, we consider
\begin{align*}
    \int_1^{m + 1} \frac{b^2(\epsilon + 1)^{y + m + 1} e^{b^2 c (\epsilon + 1)^{y + m + 1}}}{c \log^2(n) \log(\epsilon + 1)} \, \d y
\end{align*}

We use $u$-substitution by defining $u = b^2 c (\epsilon + 1)^{y + m + 1}$, which implies
\begin{align*}
    \d y = \frac{1}{b^2 c (\epsilon + 1)^{y + m + 1} \log(\epsilon + 1)} \d u.
\end{align*}

Therefore, we have
\begin{align*}
    & ~ \int_1^{m + 1} \frac{b^2(\epsilon + 1)^{y + m + 1} e^{b^2 c (\epsilon + 1)^{y + m + 1}}}{c \log^2(n) \log(\epsilon + 1)} \, \d y \\
    = & ~ \frac{1}{c \log^2(n) \log(\epsilon + 1)} \int_1^{m + 1} b^2(\epsilon + 1)^{y + m + 1} e^{b^2 c (\epsilon + 1)^{y + m + 1}}  \, \d y\\
    = & ~ \frac{1}{c^2 \log^2(n) \log^2(\epsilon + 1)} \int_{b^2 c (\epsilon + 1)^{m + 2}}^{b^2 c (\epsilon + 1)^{2m + 2}} e^{u}  \, \d u\\
    = & ~ \frac{e^{b^2 c (\epsilon + 1)^{2m + 2}} - e^{b^2 c (\epsilon + 1)^{m + 2}}}{c^2 \log^2(n) \log^2(\epsilon + 1)}.
\end{align*}
\end{proof}

\begin{fact}[Multiplicative Chernoff Bound]\label{fac:chernoff}
Let $X_1, X_2, \dots, X_n$ be independent random variables taking values in $[0,1]$, and let $X = \sum_{i=1}^n X_i$ with $\mu = \mathbb{E}[X]$. Then for all $\delta > 0$, the following holds:
\begin{align*}
    \Pr[X \geq (1+\delta)\mu] &\leq \exp\left( -\frac{\delta^2 \mu}{3} \right) \quad \text{for } 0 < \delta \leq 1, \\
    \Pr[X \geq (1+\delta)\mu] &\leq \exp\left( -\frac{\delta \mu}{3} \right) \quad \text{for } \delta > 1.
\end{align*}
\end{fact}

\begin{fact}\label{fac:kronecker}
    Let $a, b \in \R^d$. Let $p$ be an arbitrary positive integer.
    
    Then, we have
    \begin{itemize}
        \item {\bf Part 1.} $\langle a^{\otimes 2}, b^{\otimes 2} \rangle = \langle a, b \rangle^2$.
        \item {\bf Part 2.} $\langle a^{\otimes p}, b^{\otimes p} \rangle = \langle a, b \rangle^p$ (generalizing {\bf Part 1} to arbitrary $p$).
        \item {\bf Part 3.} $\|a^{\otimes p}\|_2 = \|a\|_2^p$.
    \end{itemize}
\end{fact}

\begin{proof}

{\bf Proof of Part 1.}

We have
\begin{align*}
    \langle a^{\otimes 2}, b^{\otimes 2} \rangle
    = & ~ \langle a \otimes a, b \otimes b \rangle \\
    = & ~ \sum_{i=1}^d \sum_{j=1}^d (a_i a_j)(b_i b_j) \\
    = & ~ \sum_{i=1}^d \sum_{j=1}^d a_i b_i a_j b_j \\
    = & ~ \left( \sum_{i=1}^d a_i b_i \right) \left( \sum_{j=1}^d a_j b_j \right) \\
    = & ~ \left( \sum_{i=1}^d a_i b_i \right)^2 \\
    = & ~ \langle a, b \rangle^2,
\end{align*}
where the first step follows from the definition of $a^{\otimes 2}$ and $b^{\otimes 2}$, the second step follows from the definition of inner product and Kronecker product, the third step follows from the associative law of multiplication, and the last step follows from the definition of inner product.

{\bf Proof of Part 2}

We prove this part using mathematical induction.

We treat {\bf Part 1} as the base case, and below, we prove the inductive case.

Assume for all given positive integer $k$, we have
\begin{align}\label{eq:induction_hypothesis}
    \langle a^{\otimes k}, b^{\otimes k} \rangle = \langle a, b \rangle^k.
\end{align}

Similar as what we have in {\bf Part 1}, we can get
\begin{align*}
    \langle a^{\otimes (k+1)}, b^{\otimes (k+1)} \rangle 
    = & ~ \langle a^{\otimes k} \otimes a, b^{\otimes k} \otimes b \rangle \\
    = & ~ \sum_{j=1}^d \sum_{i=1}^{d^k} \paren{\paren{a^{\otimes k}}_i a_j}\paren{\paren{b^{\otimes k}}_i b_j} \\
    = & ~ \sum_{j=1}^d \sum_{i=1}^{d^k} \paren{a^{\otimes k}}_i a_j\paren{b^{\otimes k}}_i b_j \\
    = & ~ \left( \sum_{i=1}^{d^k} \paren{a^{\otimes k}}_i \paren{b^{\otimes k}}_i \right) \left( \sum_{j=1}^d a_j b_j \right) \\
    = & ~ \langle a^{\otimes k}, b^{\otimes k} \rangle \cdot \langle a, b \rangle,
\end{align*}
where the first step follows from the definition of $a^{\otimes (k + 1)}$ and $b^{\otimes (k + 1)}$, the second step follows from the definition of inner product and Kronecker product, the third step follows from the associative law of multiplication, and the last step follows from the definition of inner product.

By the inductive hypothesis (Eq.~\eqref{eq:induction_hypothesis}), we have
\begin{align*}
    \langle a^{\otimes (k+1)}, b^{\otimes (k+1)} \rangle 
    = & ~ \langle a, b \rangle^k \cdot \langle a, b \rangle\\
    = & ~ \langle a, b \rangle^{k + 1}.
\end{align*}

{\bf Proof of Part 3.}

We can get
\begin{align*}
    \|a^{\otimes p}\|_2^2 
    = & ~ \sum_{i_1=1}^d \cdots \sum_{i_p=1}^d (a_{i_1} a_{i_2} \cdots a_{i_p})^2 \\
    = & ~ \sum_{i_1=1}^d \cdots \sum_{i_p=1}^d a_{i_1}^2 a_{i_2}^2 \cdots a_{i_p}^2 \\
    = & ~ \left( \sum_{i=1}^d a_i^2 \right)^p \\
    = & ~ \|a\|_2^{2p},
\end{align*}
where the first step follows from the definition of the $\ell_2$ norm and the Kronecker product, and the last step follows from the definition of the $\ell_2$ norm.

Taking square roots of both sides, we have
\[
\|a^{\otimes p}\|_2 = \|a\|_2^p
\]
\end{proof}

\begin{fact}[Markov's Inequality]\label{fac:markov}
Let $X$ be a non-negative random variable and let $a > 0$. Then
\[
\Pr[X \geq a] \leq \frac{\mathbb{E}[X]}{a}.
\]
\end{fact}

\begin{fact}[Union Bound]\label{fac:union}
Let $A_1, A_2, \dots, A_n$ be events in a probability space. Then
\[
\Pr\left[\bigcup_{i=1}^n A_i\right] \leq \sum_{i=1}^n \Pr[A_i].
\]
\end{fact}

\begin{fact}\label{fac:more_disjoint_prop}
    Let $A \in \R^{n \times n}$ and $B \in \R^{n \times n}$ be disjoint matrices. Let $p$ and $k$ be arbitrary positive integers. 
    
    Then, we can get
    \begin{itemize}
        \item {\bf Part 1.} $\paren{A + B}^{\circ p} = \paren{A}^{\circ p} + \paren{B}^{\circ p}$,
        \item {\bf Part 2} $\left \|A + B \right \|_p^p = \left \|A \right \|_p^p + \left \| B \right \|_p^p$,
        \item {\bf Part 3} $\left \| A^{\circ k} \right\|_p^p \leq \left \| A \right\|_p^{p k}$.
    \end{itemize}
\end{fact}
\begin{proof}
    {\bf Proof of Part 1.}

    Since $A \in \R^{n \times n}$ and $B \in \R^{n \times n}$ are disjoint matrices, for all arbitrary $\paren{i, j} \in [n] \times [n]$, we can either have
    \begin{align}\label{eq:bij_0}
        A_{i, j} \neq 0 \text{ and } B_{i, j} = 0
    \end{align}
    or
    \begin{align}\label{eq:aij_0}
        B_{i, j} \neq 0 \text{ and } A_{i, j} = 0.
    \end{align}
    
    If Eq.~\eqref{eq:bij_0} holds, then we can get
    \begin{align*}
        \paren{\paren{A + B}^{\circ p}}_{i, j}
        = & ~ \paren{A + B}_{i, j}^p \\
        = & ~ A_{i, j}^p.
    \end{align*}

    If Eq.~\eqref{eq:aij_0} holds, then we can get
    \begin{align*}
        \paren{\paren{A + B}^{\circ p}}_{i, j}
        = & ~ \paren{A + B}_{i, j}^p \\
        = & ~ B_{i, j}^p.
    \end{align*}

    {\bf Proof of Part 2.}

    We can get
    \begin{align*}
        \left \|A + B \right \|_p^p
        = & ~ \sum_{i, j} (A + B)_{i, j}^p \\
        = & ~ \sum_{i, j} (A_{i, j}^p + B_{i, j}^p) \\
        = & ~ \sum_{i, j} A_{i, j}^p + \sum_{i, j} B_{i, j}^p \\
        = & ~ \left \|A \right \|_p^p + \left \| B \right \|_p^p,
    \end{align*}
    where the first step follows from the definition of $\ell_p$ norm and the second step follows from {\bf Part 1}.

    {\bf Proof of Part 3.}

    We have
    \begin{align*}
        \left \| A^{\circ k} \right\|_p^p
        = & ~ \sum_{i, j} (A^{\circ k})_{i, j}^p \\
        = & ~ \sum_{i, j} (A_{i, j}^k)^p \\
        = & ~ \sum_{i, j} (A_{i, j}^p)^k \\
        \leq & ~ \paren{\sum_{i, j} A_{i, j}^p}^k\\
        = & ~ \left \| A \right\|_p^{p k}.
    \end{align*}
\end{proof}

\ifdefined\isarxiv

\section{More Related Work}

\else
\section{MORE RELATED WORK}

\fi

\label{sec:more_related}

\paragraph{Attention regression problems}

Another line of work that attempts to reduce the computational complexity of attention approximation transforms the attention computation into regression problems and applies the approximate Newton method to solve them. These methods use sketching matrices to reduce the dimensionality of the Hessian in the Newton method, thereby accelerating the algorithm with provable guarantees. However, most of these works simplify the attention computation problem (Definition~\ref{def:exact_attention_computation}). 

For example, \cite{lssw25} proposes a softmax regression formulation, 
\begin{align*}
    \min_{x \in \R^d} \left \| \langle \exp\paren{ Ax} , {\bf 1}_n \rangle^{-1} \exp\paren{ Ax} - c \right \|_2^2
\end{align*}
where the value matrix is completely ignored. \cite{gsy23_hyper} studies a rescaled version of softmax regression, 
\begin{align*}
    \min_{x \in \R^d} \left \| \exp\paren{ Ax} - \langle \exp\paren{ Ax} , {\bf 1}_n \rangle c \right \|_2^2
\end{align*}
\cite{syz23} analyzes an exponential regression problem, 
\begin{align*}
    \min_{x \in \R^n} \left \| \exp\paren{ AA^\top} x - b \right \|_2^2,
\end{align*}
where both $D^{-1}$ and $V$ are omitted. \cite{gswy23} is the only work that does not make any such simplifications, studying the attention regression problem 
\begin{align*}
    \min_{X,Y \in \R^{d \times d}}  \left \| D\paren{ X}^{-1} \exp\paren{ A_1 X A_2^\top} A_3 Y - B \right \|_F^2.
\end{align*}
However, the approximate Newton method can only solve convex problems, whereas attention computation is inherently non-convex. To apply the approximate Newton method, \cite{gswy23} still relies on regularization terms and the choice of a good initialization point for the iterative procedure—assumptions that limit the applicability of the proposed algorithm.

\paragraph{Sketching}

For a large data matrix $A \in \R^{n \times d}$, earlier works have centered on designing a sketching matrix $S \in \R^{m \times n}$ with $m \ll n$ such that, for every $x \in \R^d$, $\|SAx\|_2^2 = (1 \pm \epsilon)\|Ax\|_2^2$, where $\epsilon \in (0,1)$ denotes the approximation error. A matrix $S$ with this property is known as a subspace embedding for $A$. This idea has become a standard tool across many machine learning tasks, including linear regression \citep{psw17,syyz23_linf,syz23_quantum}, (weighted) low-rank approximation \citep{cw17,syyz23_weight}, tensor power method \citep{dsy23}, online weighted matching problem \citep{swyy25}, low-rank matrix completion \citep{gsyz24}, $k$-means clustering \citep{lss+22}, and attention mechanisms \citep{kmz24}.

\paragraph{Large language models}

Besides the line of works that focus on the computational efficiency of LLMs, such as \cite{lss+24,cls+24,lss+25}, there is another line of work on memory efficiency, such as LoRA \citep{hsw22}, GaLore \citep{zzc24}, SARA \citep{zyw25}, KV-Cache compression \citep{cll+25_kv}, and CoVE \citep{zzy+25}. Another line of work focuses on analyzing the softmax unit in attention, such as binary hypothesis testing \citep{gsy25}, softmax regression \citep{lssw25,swy23,lswy23}. Other works focus on the circuit complexity of LLMs \citep{kll+25,cll+24}, limitations of LLMs \citep{cll+25}, and hallucination \citep{lls+25}. In reinforcement learning \citep{zcz+25,zcy23}, \cite{wpx+24} design reinforcement learning targeted attack on LLMs, and \cite{phh+24} propose the reinforcement learning framework of LLM agents. Furthermore, the application of large transformers with long-context capabilities is not limited to language processing; broader domains such as time-series prediction \citep{zwz23,zpt2022,jin2023time,liu2024autotimes,gruver2023large} are also of practical importance. Extending and validating our method to preserve the accuracy of large transformer models on these tasks is, therefore, a worthwhile direction for future work.

\ifdefined\isarxiv

\section{Query and Key Entries Distribution}

\else
\section{QUERY AND KEY ENTRIES DISTRIBUTION}

\fi

\label{sec:distribution}

In this section, we empirically validate our assumption that the entries of query and key matrices resemble sub-Gaussian distributions on transformer architectures such as TinyLlama-1.1B \citep{zzwl24}, LLaDA-8B-Base \citep{nie2025large}, OPT-1.3B \citep{zrg+22}, and Phi-2 \citep{jba23} across multiple layers. Their model IDs are GSAI-ML/LLaDA-8B-Base (LLaDA-8B-Base), facebook/opt-1.3b (OPT-1.3B), TinyLlama/TinyLlama-1.1B-Chat-v1.0 (TinyLlama-1.1B), and microsoft/phi-2 (Phi-2).  

\begin{figure}[H]
    \centering
    \includegraphics[width=\linewidth]{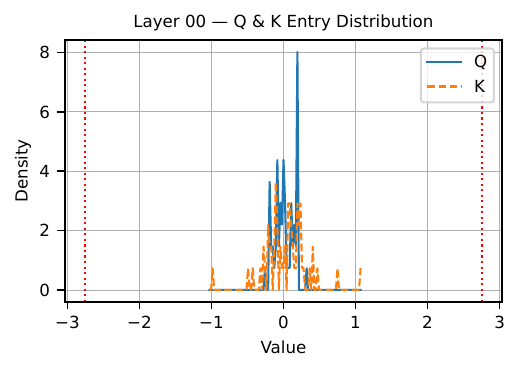}
    \caption{Distribution of entries in the query and key matrices of Layer~0 in TinyLlama-1.1B.
    The red dashed lines mark the thresholds $\pm\sqrt{\log n}$.}
    \label{fig:QK_0}
\end{figure}

\begin{figure}[H]
    \centering
    \includegraphics[width=\linewidth]{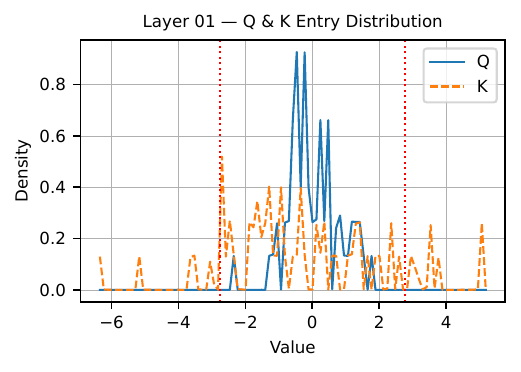}
    \caption{Distribution of entries in the query and key matrices of Layer~1 in TinyLlama-1.1B.
    The red dashed lines mark the thresholds $\pm\sqrt{\log n}$.}
    \label{fig:QK_1}
\end{figure}

\begin{figure}[H]
    \centering
    \includegraphics[width=\linewidth]{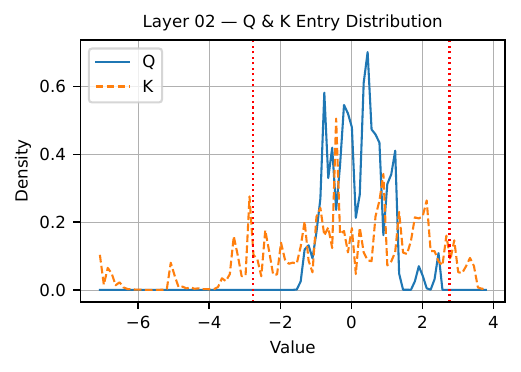}
    \caption{Distribution of entries in the query and key matrices of Layer~2 in TinyLlama-1.1B.
    The red dashed lines mark the thresholds $\pm\sqrt{\log n}$.}
    \label{fig:QK_2}
\end{figure}

\begin{figure}[H]
    \centering
    \includegraphics[width=\linewidth]{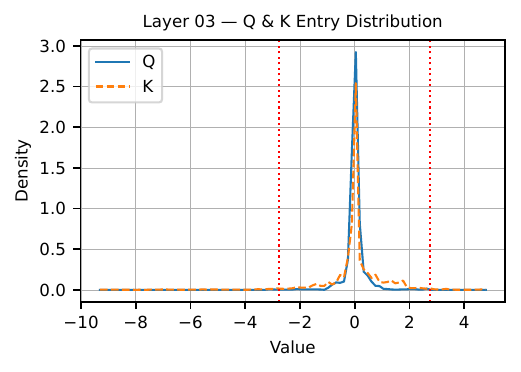}
    \caption{Distribution of entries in the query and key matrices of Layer~3 in TinyLlama-1.1B.
    The red dashed lines mark the thresholds $\pm\sqrt{\log n}$.}
    \label{fig:QK_3}
\end{figure}

\begin{figure}[H]
    \centering
    \includegraphics[width=\linewidth]{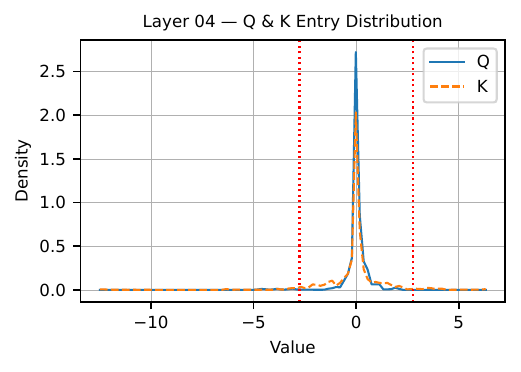}
    \caption{Distribution of entries in the query and key matrices of Layer~4 in TinyLlama-1.1B.
    The red dashed lines mark the thresholds $\pm\sqrt{\log n}$.}
    \label{fig:QK_4}
\end{figure}

\begin{figure}[H]
    \centering
    \includegraphics[width=\linewidth]{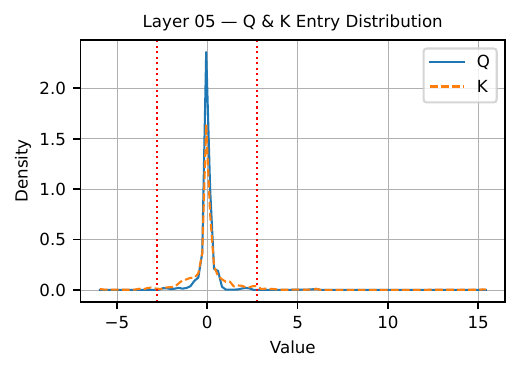}
    \caption{Distribution of entries in the query and key matrices of Layer~5 in TinyLlama-1.1B.
    The red dashed lines mark the thresholds $\pm\sqrt{\log n}$.}
    \label{fig:QK_5}
\end{figure}

\begin{figure}[H]
    \centering
    \includegraphics[width=\linewidth]{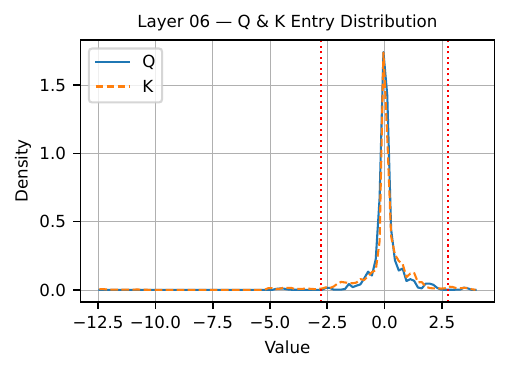}
    \caption{Distribution of entries in the query and key matrices of Layer~6 in TinyLlama-1.1B.
    The red dashed lines mark the thresholds $\pm\sqrt{\log n}$.}
    \label{fig:QK_6}
\end{figure}

\begin{figure}[H]
    \centering
    \includegraphics[width=\linewidth]{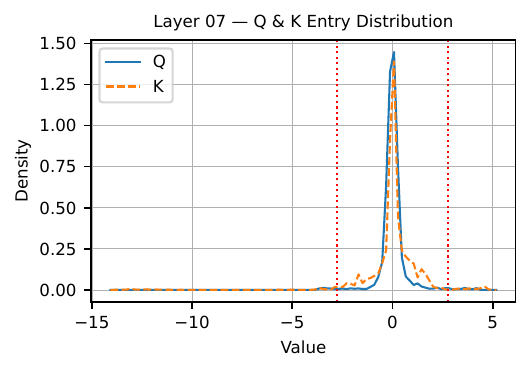}
    \caption{Distribution of entries in the query and key matrices of Layer~7 in TinyLlama-1.1B.
    The red dashed lines mark the thresholds $\pm\sqrt{\log n}$.}
    \label{fig:QK_7}
\end{figure}

\begin{figure}[H]
    \centering
    \includegraphics[width=\linewidth]{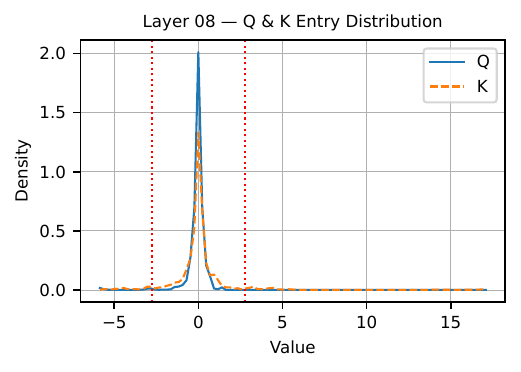}
    \caption{Distribution of entries in the query and key matrices of Layer~8 in TinyLlama-1.1B.
    The red dashed lines mark the thresholds $\pm\sqrt{\log n}$.}
    \label{fig:QK_8}
\end{figure}

\begin{figure}[H]
    \centering
    \includegraphics[width=\linewidth]{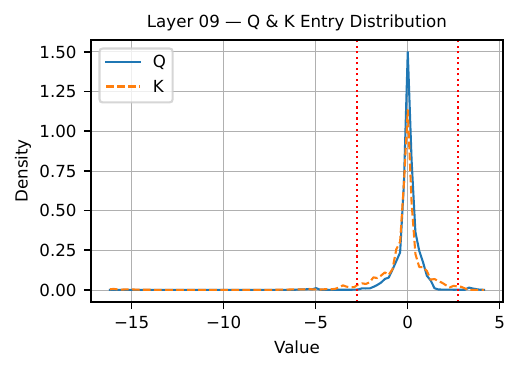}
    \caption{Distribution of entries in the query and key matrices of Layer~9 in TinyLlama-1.1B.
    The red dashed lines mark the thresholds $\pm\sqrt{\log n}$.}
    \label{fig:QK_9}
\end{figure}

\begin{figure}[H]
    \centering
    \includegraphics[width=\linewidth]{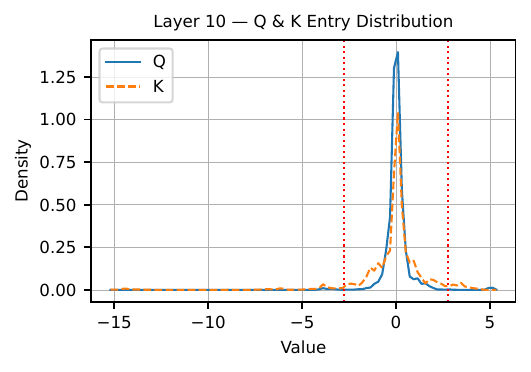}
    \caption{Distribution of entries in the query and key matrices of Layer~10 in TinyLlama-1.1B.
    The red dashed lines mark the thresholds $\pm\sqrt{\log n}$.}
    \label{fig:QK_10}
\end{figure}

\begin{figure}[H]
    \centering
    \includegraphics[width=\linewidth]{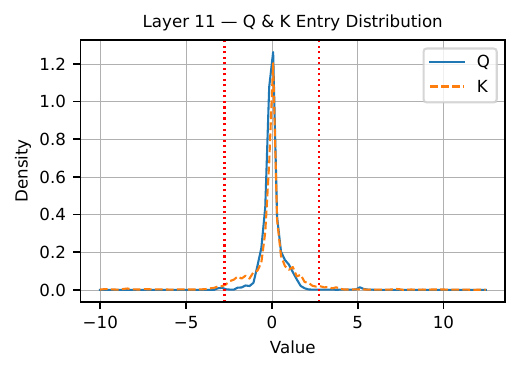}
    \caption{Distribution of entries in the query and key matrices of Layer~11 in TinyLlama-1.1B.
    The red dashed lines mark the thresholds $\pm\sqrt{\log n}$.}
    \label{fig:QK_11}
\end{figure}

\begin{figure}[H]
    \centering
    \includegraphics[width=\linewidth]{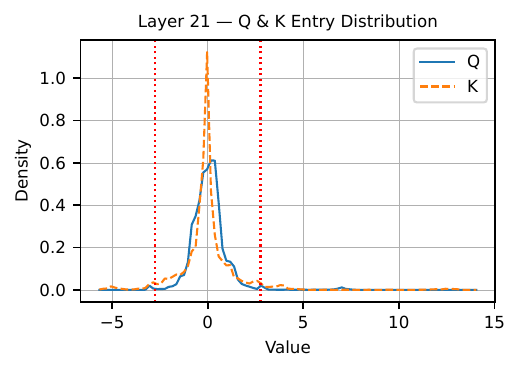}
    \caption{Distribution of entries in the query and key matrices of Layer~21 in TinyLlama-1.1B.
    The red dashed lines mark the thresholds $\pm\sqrt{\log n}$.}
    \label{fig:QK_21}
\end{figure}

To empirically validate our assumption, we visualize the distribution of the entries of $Q$ and $K$ across multiple layers, generated by natural human-like tokens (Appendix~\ref{sub:distribution:llama}, \ref{sub:distribution:gpt}, \ref{sub:distribution:llada}, and \ref{sub:distribution:phi2}). We observe that for all of the layers of attention in these transformer architectures (TinyLlama-1.1B \citep{zzwl24}, LLaDA-8B-Base \citep{nie2025large}, OPT-1.3B \citep{zrg+22}, and Phi-2 \citep{jba23}), the entry distributions are sub-Gaussian-like, which satisfies the “practical assumption” made in this paper. 
In Appendix~\ref{sub:distribution:llama_hhh} we present the entry distribution of $Q$ and $K$ of TinyLlama-1.1B \citep{zzwl24} by deliberately constructing a non-human-like token sequence consisting of the character “h” repeated 2048 times. For these tokens, the distributions of the entries of $Q$ and $K$ are more dispersed in the early layers (layers 0–2, Figures~\ref{fig:QK_0}–\ref{fig:QK_2}). However, starting from layer 3 onward, all entries of $Q$ and $K$ become increasingly concentrated (Figures~\ref{fig:QK_3}–\ref{fig:QK_21}). LLMs are trained on natural human-like tokens, so it is expected that the entries of $Q$ and $K$ remain tightly concentrated across all layers. Our result in Appendix~\ref{sub:distribution:llama_hhh} can be viewed as a worst-case scenario: even if the entries are not well centered around the mean in the first few iterations, they become more tightly concentrated as $\ell$ increases.

In Appendix~\ref{sub:distribution:llama}, we show that in TinyLlama-1.1B \citep{zzwl24}, after the first few layers, the entries of $Q$ and $K$ become tightly concentrated with light tails. We present the entry distribution of $Q$ and $K$ of TinyLlama-1.1B \citep{zzwl24} from layer 0 to layer 21. In Appendix~\ref{sub:distribution:gpt}, we present the entry distribution of $Q$ and $K$ of OPT-1.3B \citep{zrg+22} from layer 0 to layer 23. 
In Appendix~\ref{sub:distribution:llada}, we present the entry distribution of $Q$ and $K$ of LLaDA-8B-Base \citep{nie2025large} from layer 0 to layer 31. In Appendix~\ref{sub:distribution:phi2}, we present the entry distribution of $Q$ and $K$ of Phi-2 \citep{jba23} from layer 0 to layer 31.

\subsection{TinyLlama-1.1B With Repeated Token}

\label{sub:distribution:llama_hhh}

In this section, we present the entry distribution of $Q$ and $K$ in TinyLlama-1.1B \citep{zzwl24} from layer 0 to layer 21. The token sequence consists of the character “h” repeated 2048 times.

\begin{figure}[H]
    \centering
    \includegraphics[width=\linewidth]{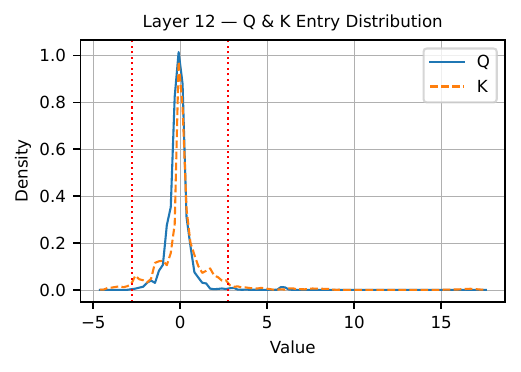}
    \caption{Distribution of entries in the query and key matrices of Layer~12 in TinyLlama-1.1B.
    The red dashed lines mark the thresholds $\pm\sqrt{\log n}$.}
    
\end{figure}

\begin{figure}[H]
    \centering
    \includegraphics[width=\linewidth]{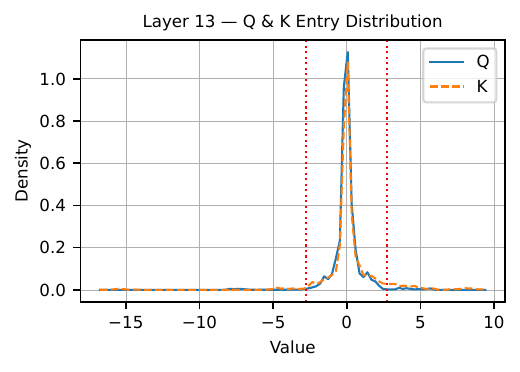}
    \caption{Distribution of entries in the query and key matrices of Layer~13 in TinyLlama-1.1B.
    The red dashed lines mark the thresholds $\pm\sqrt{\log n}$.}
    
\end{figure}

\begin{figure}[H]
    \centering
    \includegraphics[width=\linewidth]{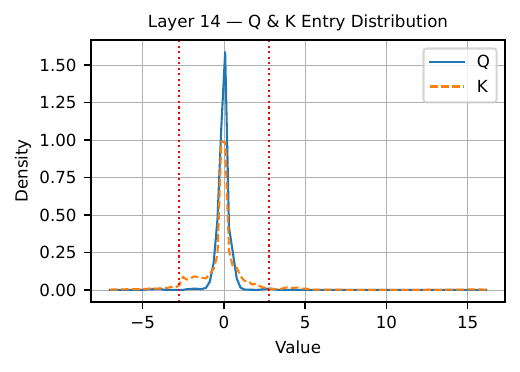}
    \caption{Distribution of entries in the query and key matrices of Layer~14 in TinyLlama-1.1B.
    The red dashed lines mark the thresholds $\pm\sqrt{\log n}$.}
    
\end{figure}

\begin{figure}[H]
    \centering
    \includegraphics[width=\linewidth]{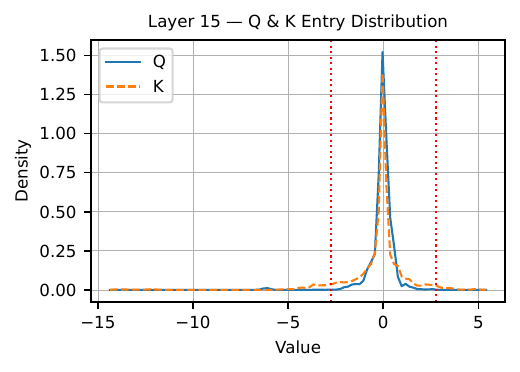}
    \caption{Distribution of entries in the query and key matrices of Layer~15 in TinyLlama-1.1B.
    The red dashed lines mark the thresholds $\pm\sqrt{\log n}$.}
    
\end{figure}

\begin{figure}[H]
    \centering
    \includegraphics[width=\linewidth]{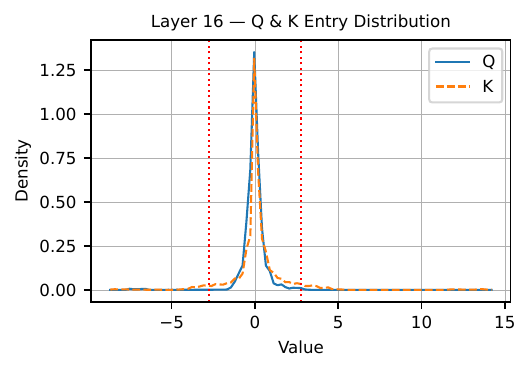}
    \caption{Distribution of entries in the query and key matrices of Layer~16 in TinyLlama-1.1B.
    The red dashed lines mark the thresholds $\pm\sqrt{\log n}$.}
    
\end{figure}

\begin{figure}[H]
    \centering
    \includegraphics[width=\linewidth]{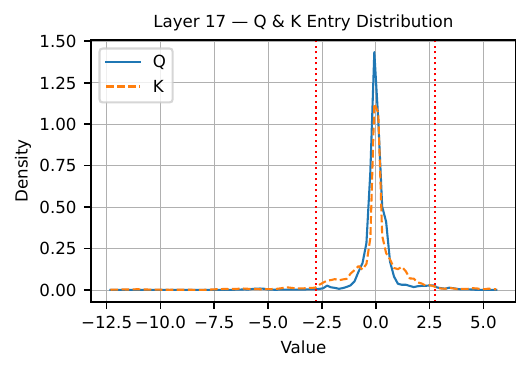}
    \caption{Distribution of entries in the query and key matrices of Layer~17 in TinyLlama-1.1B.
    The red dashed lines mark the thresholds $\pm\sqrt{\log n}$.}
    
\end{figure}

\begin{figure}[H]
    \centering
    \includegraphics[width=\linewidth]{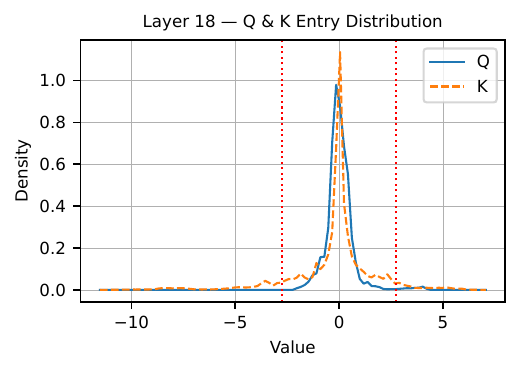}
    \caption{Distribution of entries in the query and key matrices of Layer~18 in TinyLlama-1.1B.
    The red dashed lines mark the thresholds $\pm\sqrt{\log n}$.}
    
\end{figure}

\begin{figure}[H]
    \centering
    \includegraphics[width=\linewidth]{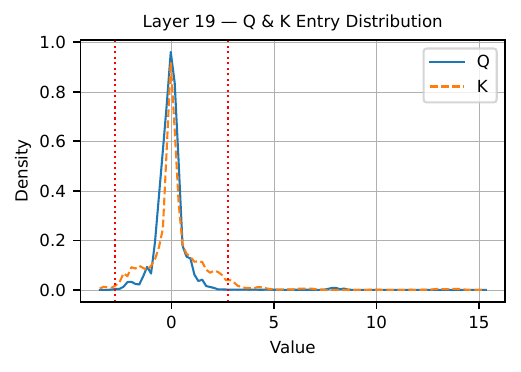}
    \caption{Distribution of entries in the query and key matrices of Layer~19 in TinyLlama-1.1B.
    The red dashed lines mark the thresholds $\pm\sqrt{\log n}$.}
    
\end{figure}

\begin{figure}[H]
    \centering
    \includegraphics[width=\linewidth]{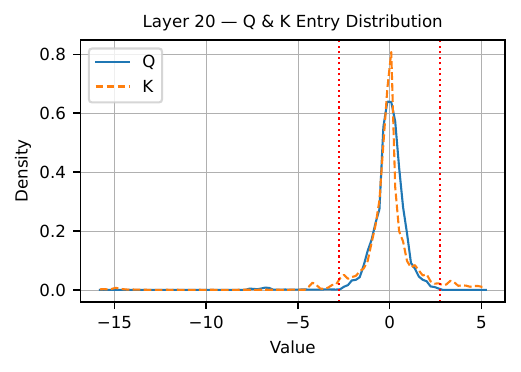}
    \caption{Distribution of entries in the query and key matrices of Layer~20 in TinyLlama-1.1B.
    The red dashed lines mark the thresholds $\pm\sqrt{\log n}$.}
    
\end{figure}

\subsection{TinyLlama-1.1B}

\label{sub:distribution:llama}

In this section, we present the entry distribution of $Q$ and $K$ in TinyLlama-1.1B \citep{zzwl24} from layer 0 to layer 21. The token sequence consists of the following natural human-like tokens.

\begin{AIbox}{Natural human-like tokens}
In the fast-paced world of modern research and technology, the ability to adapt quickly to new knowledge, integrate interdisciplinary ideas, and communicate them clearly has become a defining factor for success, not only for academic researchers but also for professionals working in industry, policy, or creative sectors. The challenge is no longer simply about having access to information—since the digital age has democratized knowledge to an unprecedented degree—but rather about cultivating the skills necessary to filter, evaluate, and synthesize the vast amount of data that is constantly flowing around us. A researcher today may begin the morning reading about advances in large language models, spend the afternoon designing experiments to validate theoretical insights, and finish the day by considering applications in medicine, finance, or education. This fluid movement between levels of abstraction demands both intellectual flexibility and a strong methodological foundation. At the same time, collaboration has emerged as a core feature of progress: breakthroughs increasingly arise not from the lone genius archetype, but from teams that combine different strengths, whether it is the theoretical rigor of mathematicians, the practical engineering sense of computer scientists, the domain knowledge of biologists, or the design intuition of human-computer interaction experts. Alongside collaboration, communication is equally essential. A brilliant idea poorly explained is often a wasted opportunity, while even a moderately novel insight presented with clarity and precision can influence the trajectory of a field. In this context, the role of writing, speaking, and visualizing cannot be underestimated. Writing a paper or report is not just a matter of documenting results but of shaping the interpretation and framing of those results, guiding how others will build upon them. Similarly, presenting at conferences, creating effective figures, and explaining technical ideas to broader audiences are all skills that expand the impact of one’s work beyond immediate circles. Another layer to this landscape is the increasing pressure to balance depth with breadth. Specialization is necessary to push the frontier of a subfield, yet the most impactful research often arises from unexpected connections, such as applying methods from signal processing to neuroscience or adapting optimization techniques from physics to machine learning. To navigate this dual demand, researchers must cultivate a meta-skill: the ability to learn how to learn efficiently, to enter new fields without being overwhelmed, and to identify the key assumptions, tools, and open questions that define them. Underlying all of this is resilience and persistence, since research inevitably involves setbacks, failed experiments, and long periods of uncertainty. The process is rarely linear; rather, it is iterative and recursive, resembling more a spiral of refinement than a straight path toward discovery. In the end, success in modern research and professional life lies in the interplay of curiosity, rigor, creativity, and communication—qualities that allow individuals and teams not only to generate knowledge but to ensure that this knowledge becomes meaningful, usable, and transformative in the broader world.
\end{AIbox}

\begin{figure}[H]
    \centering
    \includegraphics[width=\linewidth]{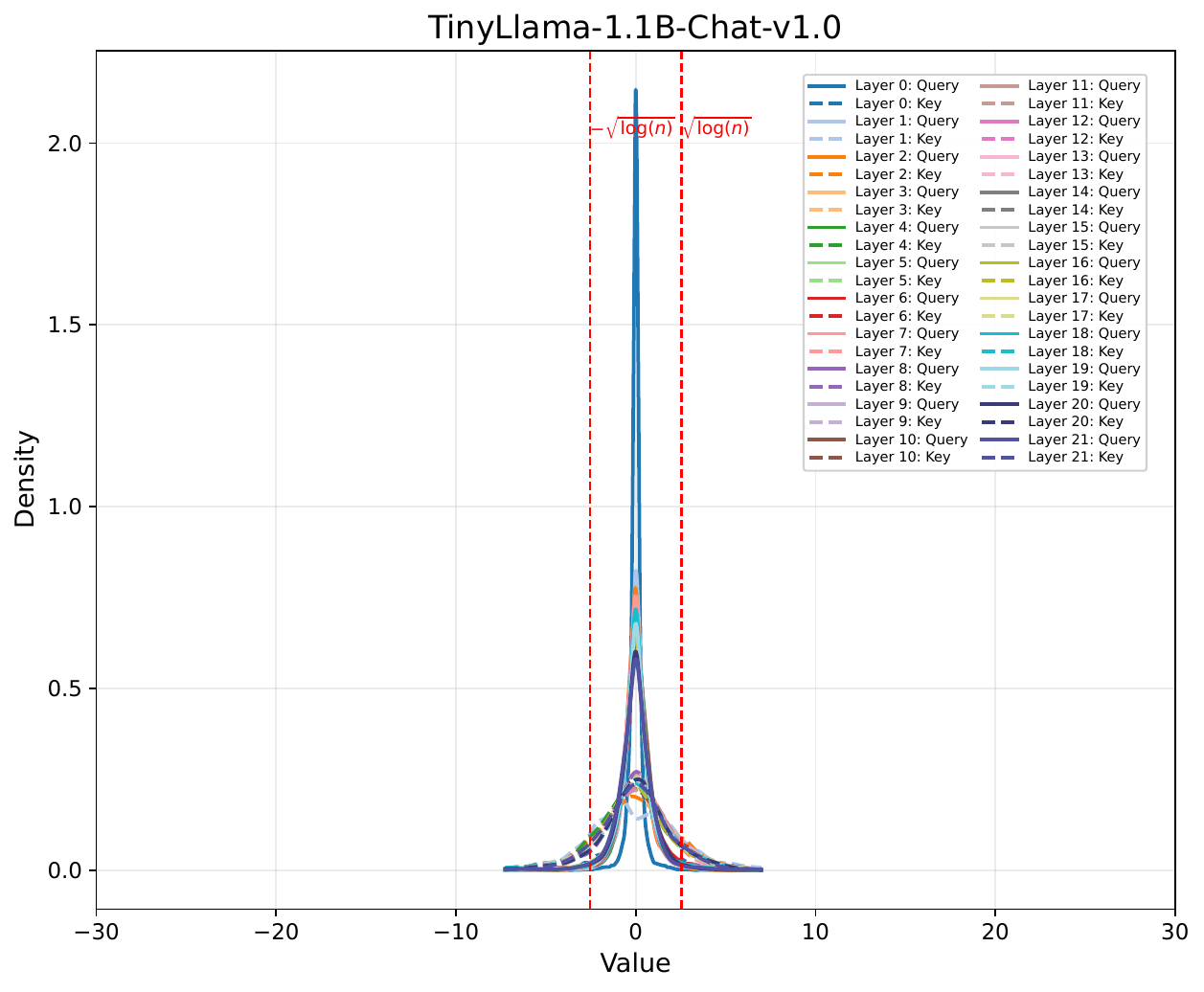}
    \caption{Distribution of entries in the query, key, and value matrices of TinyLlama-1.1B from layer 0 to layer 21.
    The red dashed lines mark the thresholds $\pm\sqrt{\log n}$.}
    
\end{figure}

\subsection{OPT-1.3B}
\label{sub:distribution:gpt}

In this section, we present the entry distribution of $Q$ and $K$ in OPT-1.3B \citep{zrg+22} from layer 0 to layer 23. The token sequence consists of the natural human-like tokens.

\begin{figure}[H]
    \centering
    \includegraphics[width=\linewidth]{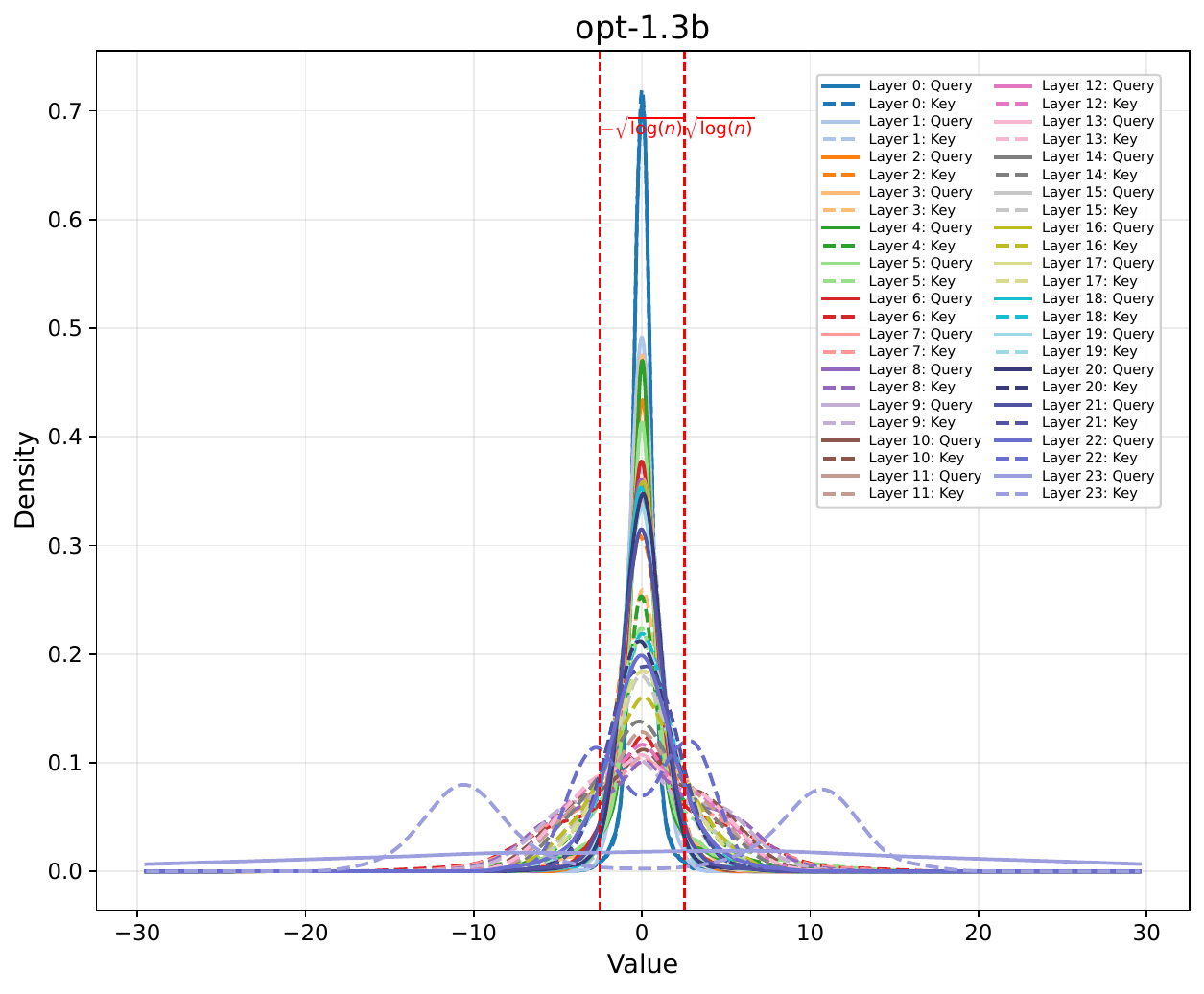}
    \caption{Distribution of entries in the query, key, and value matrices of OPT-1.3B from layer 0 to layer 23.
    The red dashed lines mark the thresholds $\pm\sqrt{\log n}$.}
    
\end{figure}

\subsection{LLaDA-8B-Base}
\label{sub:distribution:llada}

In this section, we present the entry distribution of $Q$ and $K$ in LLaDA-8B-Base \citep{nie2025large} from layer 0 to layer 31.  The token sequence consists of the natural human-like tokens.

\begin{figure}[H]
    \centering
    \includegraphics[width=\linewidth]{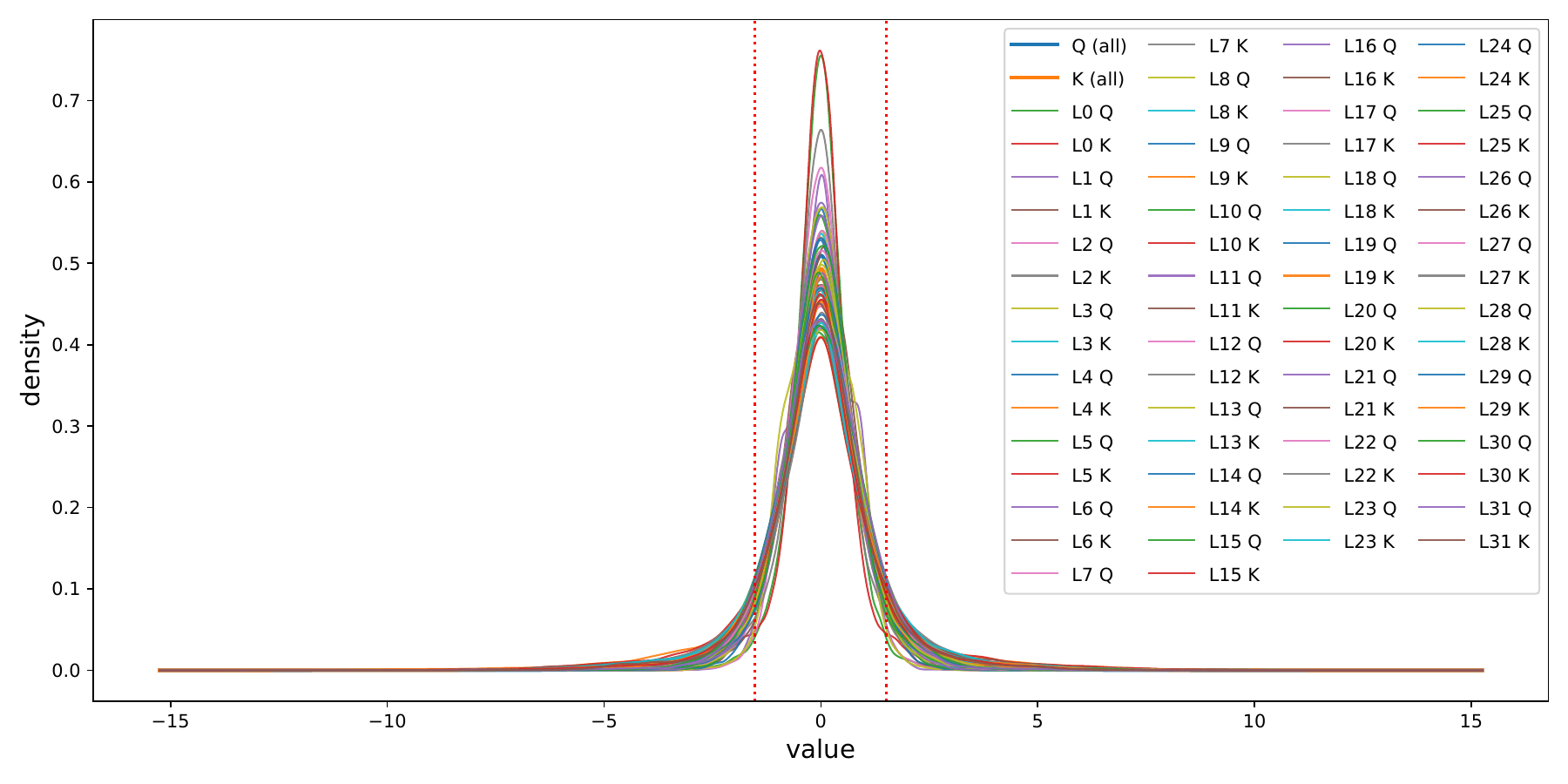}
    \caption{Distribution of entries in the query and key matrices of LLaDA-8B-Base from layer 0 to layer 31.
    The red dashed lines mark the thresholds $\pm\sqrt{\log n}$.}
    
\end{figure}

\subsection{Phi-2}
\label{sub:distribution:phi2}

In this section, we present the entry distribution of $Q$ and $K$ in Phi-2 \citep{jba23} from layer 0 to layer 31. The token sequence consists of the natural human-like tokens.

\begin{figure}[H]
    \centering
    \includegraphics[width=\linewidth]{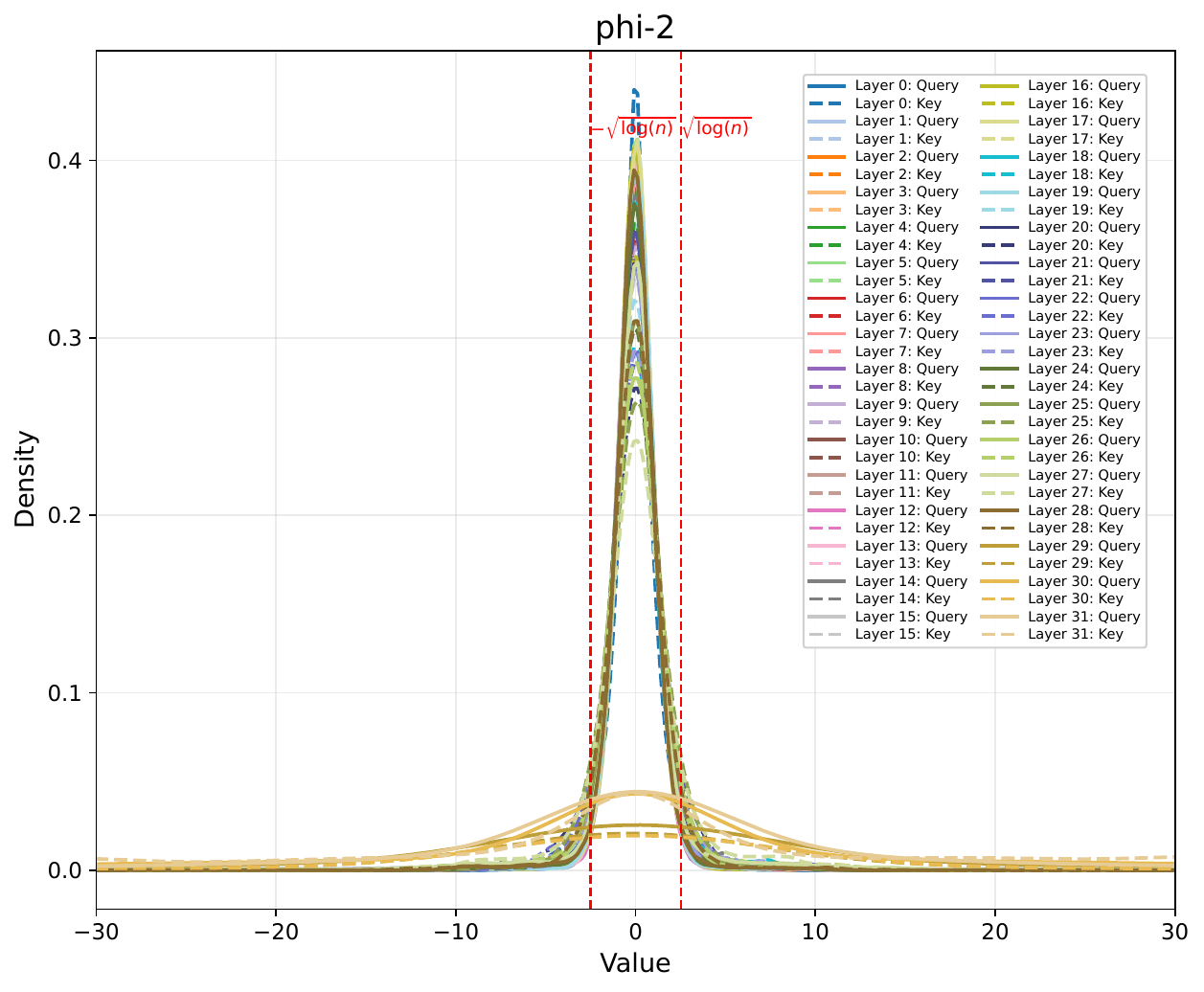}
    \caption{Distribution of entries in the query and key matrices of Phi-2 from layer 0 to layer 31.
    The red dashed lines mark the thresholds $\pm\sqrt{\log n}$.}
    
\end{figure}

\ifdefined\isarxiv

\bibliographystyle{alpha}
\bibliography{ref}
\else

\fi




\end{document}